\documentclass[oneside,12pt]{book}

\usepackage[utf8]{inputenc}

\usepackage{graphicx}
\setkeys{Gin}{width=\linewidth,totalheight=\textheight,keepaspectratio}

\usepackage{xcolor}
\definecolor{darkblue}{HTML}{00416A}

\usepackage{longtable}
\usepackage{booktabs}
\usepackage{amssymb}
\usepackage{amsmath}
\usepackage{amsthm}
\usepackage{boxedminipage}
\usepackage{microtype}
\usepackage{makeidx}
\usepackage{hyperref}
\usepackage{ifthen}

\usepackage{titlesec,titletoc}

\usepackage[letterpaper,textwidth=5.5in,textheight=8.5in]{geometry}

\usepackage[osf,sc]{mathpazo}
\RequirePackage[scaled=0.90]{helvet}
\RequirePackage[scaled=0.85]{beramono}
\RequirePackage[T1]{fontenc}
\RequirePackage{textcomp}

\usepackage[square,sort,comma,numbers]{natbib}
\usepackage[superscript,biblabel]{cite}
\renewcommand{\citep}[1]{\cite{#1}}

\titleformat{\chapter}%
  [display]
  {\relax}
  {\itshape\huge\thechapter}
  {0pt}
  {\huge\rmfamily\itshape}
  []

\titleformat{\section}%
  [hang]
  {\normalfont\Large\itshape}
  {\thesection}
  {1em}
  {}
  []

\titleformat{\subsection}%
  [hang]
  {\normalfont\large\itshape}
  {\thesubsection}
  {1em}
  {}
  []

\titleformat{\paragraph}%
  [runin]
  {\normalfont\itshape}
  {\theparagraph}
  {1em}
  {}
  []

\titlespacing*{\chapter}{0pt}{50pt}{40pt}
\titlespacing*{\section}{0pt}{3.5ex plus 1ex minus .2ex}{2.3ex plus .2ex}
\titlespacing*{\subsection}{0pt}{3.25ex plus 1ex minus .2ex}{1.5ex plus.2ex}

\makeatother

\newtheorem{Definition}{Definition}
\newtheorem{Theorem}{Theorem}
\newtheorem{Lemma}{Lemma}

\newtheorem{Proposition}{Proposition}

\newenvironment{Algorithm}{\begin{center}\begin{boxedminipage}{0.92\textwidth}}{\end{boxedminipage}\end{center}}
\newenvironment{Proof}{\begin{proof}}{\end{proof}}
\newtheorem*{summary*}{Summary}

\providecommand{\tightlist}{%
  \setlength{\itemsep}{0pt}\setlength{\parskip}{0pt}}

\renewcommand{\bot}{\perp}
\renewcommand{\hat}{\widehat}

\usepackage{wrapfig}
\usepackage{marginfix}
\usepackage[morefloats=100]{morefloats}

\newcommand{\marginnote}[1]{
\begin{wrapfigure}{o}{0.4\textwidth}
\footnotesize #1
\end{wrapfigure}
}

\renewcommand{\footnote}[1]{\marginnote{#1}}

\makeatletter
\renewcommand{\@seccntformat}[1]{}
\makeatother

\newcommand{\gengap}{\Delta_{\mathrm{gen}}}
\newcommand{\E}{\mathop\mathbb{E}}
\renewcommand{\Pr}{\mathop\mathbb{P}}
\newcommand{\R}{\mathbb{R}}
\newcommand{\cA}{\mathcal{A}}
\newcommand{\cT}{\mathcal{T}}
\newcommand{\mper}{\,.}

\newcommand{\doop}{\mathrm{do}}
\newcommand{\Cov}{\mathop\mathrm{Cov}}
\newcommand{\Var}{\mathop\mathrm{Var}}
\newcommand{\loss}{\mathit{loss}}
\newcommand{\jac}{\mathsf{D}}
\newcommand{\Id}{I}
\newcommand{\dif}{\mathrm{d}}
\newcommand{\FPR}{\mathrm{FPR}}
\newcommand{\TPR}{\mathrm{TPR}}

\usepackage{ifthen}
\usepackage{chngcntr}
\counterwithout{figure}{chapter}
\counterwithout{table}{chapter}

\hypersetup{
  colorlinks,
  linkcolor = black,
  urlcolor = black,
  pdftitle = {Patterns, Predictions, and Actions},
  pdfauthor = {Moritz Hardt and Benjamin Recht},
  hyperindex = true
}

\usepackage{eso-pic}
\newcommand\BackgroundPic{%
\put(0,0){%
\parbox[b][\paperheight]{\paperwidth}{%
\vfill
\centering
\includegraphics[width=\paperwidth,height=\paperheight,%
keepaspectratio]{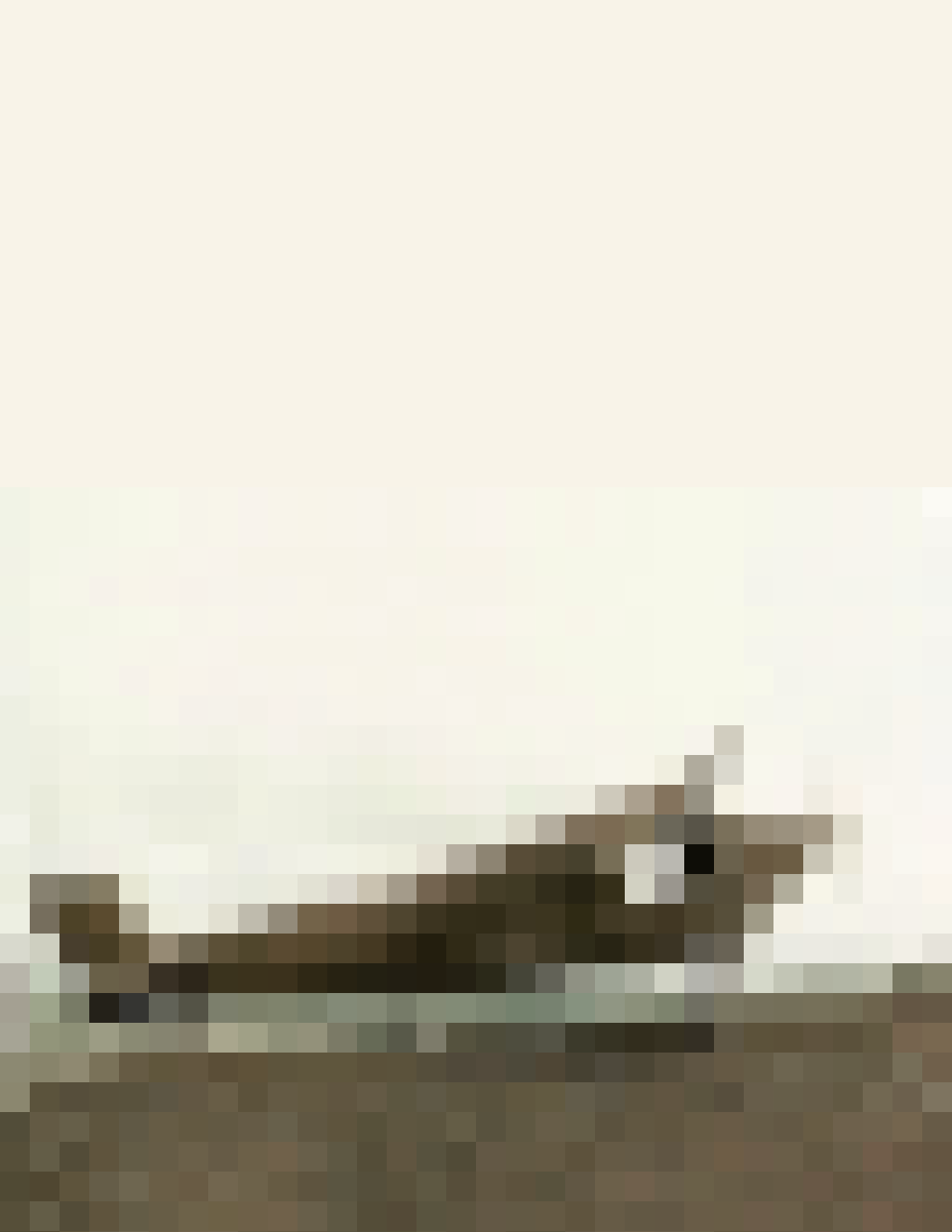}%
\vfill
}}}
\title{Patterns, Predictions, and Actions}
\author{Moritz Hardt and Benjamin Recht}
\makeindex

\newif\ifshowchapternumber

\let\OLDthebibliography\thebibliography
\renewcommand\thebibliography[1]{
  \OLDthebibliography{#1}
  \setlength{\parskip}{0pt}
  \setlength{\itemsep}{0pt plus 0.5ex}
}

\begin{document}
\AddToShipoutPicture*{\BackgroundPic}

\pagenumbering{roman}

\definecolor{browntext}{RGB}{85,70,44}
\definecolor{authortext}{RGB}{118,16,0}

\noindent{\Huge\tt\color{browntext}\uppercase{Patterns, Predictions, and Actions}}

\vspace{1cm}

\noindent{\huge\tt\color{browntext} A story about machine learning}

\vspace{3cm}

\noindent{\LARGE \tt \color{authortext} Moritz Hardt and Benjamin Recht}

\thispagestyle{empty}

\pagebreak

\vfill

\noindent Licensed under the \href{https://creativecommons.org/licenses/by-nc-nd/4.0/}{Creative Commons BY-NC-ND 4.0} license.

\vspace{1cm}

\noindent Compiled on Tue 26 Oct 2021 08:50:30 AM PDT.

\noindent Latest version available at \url{https://mlstory.org}.

\thispagestyle{empty}

\pagebreak

\vspace{2cm}

For Isaac, Leonora, and Quentin

\thispagestyle{empty}

\pagebreak

\tableofcontents

\chapter*{Preface}

In its conception, our book is both an old take on something new and a
new take on something old.

Looking at it one way, we return to the roots with our emphasis on
pattern classification. We believe that the practice of machine learning
today is surprisingly similar to pattern classification of the 1960s,
with a few notable innovations from more recent decades.

This is not to understate recent progress. Like many, we are amazed by
the advances that have happened in recent years. Image recognition has
improved dramatically. Even small devices can now reliably recognize
speech. Natural language processing and machine translation have made
massive leaps forward. Machine learning has even been helpful in some
difficult scientific problems, such as protein folding.

However, we think that it would be a mistake not to recognize pattern
classification as a driving force behind these improvements. The
ingenuity behind many advances in machine learning so far lies not in a
fundamental departure from pattern classification, but rather in finding
new ways to make problems amenable to the model fitting techniques of
pattern classification.

Consequently, the first few chapters of this book follow relatively
closely the excellent text ``Pattern Classification and Scene Analysis''
by Duda and Hart, particularly, its first edition from 1973, which
remains relevant today. Indeed, Duda and Hart summarize the state of
pattern classification in 1973, and it bears a striking resemblance to
the core of what we consider today to be machine learning. We add new
developments on representations, optimization, and generalization, all
of which remain topics of evolving, active research.

Looking at it differently, our book departs in some considerable ways
from the way machine learning is commonly taught.

First, our text emphasizes the role that datasets play in machine
learning. A full chapter explores the histories, significance, and
scientific basis of machine learning benchmarks. Although ubiquitous and
taken for granted today, the datasets-as-benchmarks paradigm was a
relatively recent development of the 1980s. Detailed consideration of
datasets, the collection and construction of data, as well as the
training and testing paradigm, tend to be lacking from theoretical
courses on machine learning.

Second, the book includes a modern introduction to causality and the
practice of causal inference that lays to rest dated controversies in
the field. The introduction is self-contained, starts from first
principles, and requires no prior commitment intellectually or
ideologically to the field of causality. Our treatment of causality
includes the conceptual foundations, as well as some of the practical
tools of causal inference increasingly applied in numerous applications.
It's interesting to note that many recent causal estimators reduce the
problem of causal inference in clever ways to pattern classification.
Hence, this material fits quite well with the rest of the book.

Third, our book covers sequential and dynamic models thoroughly. Though
such material could easily fill a semester course on its own, we wanted
to provide the basic elements required to think about making decisions
in dynamic contexts. In particular, given so much recent interest in
reinforcement learning, we hope to provide a self-contained short
introduction to the concepts underpinning this field. Our approach here
follows our approach to supervised learning: we focus on how we would
make decisions given a probabilistic model of our environment, and then
turn to how to take action when the model is unknown. Hence, we begin
with a focus on optimal sequential decision making and dynamic
programming. We describe some of the basic solution approaches to such
problems, and discuss some of the complications that arise as our
measurement quality deteriorates. We then turn to making decisions when
our models are unknown, providing a survey of bandit optimization and
reinforcement learning. Our focus here is to again highlight the power
of prediction. We show that for most problems, pattern recognition can
be seen as a complement to feedback control, and we highlight how
``certainty equivalent'' decision making---where we first use data to
estimate a model and then use feedback control acting as if this model
were true---is optimal or near optimal in a surprising number of
scenarios.

Finally, we attempt to highlight in a few different places throughout
the potential harms, limitations, and social consequences of machine
learning. From its roots in World War II, machine learning has always
been political. Advances in artificial intelligence feed into a global
industrial military complex, and are funded by it. As useful as machine
learning is for some unequivocally positive applications such as
assistive devices, it is also used to great effect for tracking,
surveillance, and warfare. Commercially its most successful use cases to
date are targeted advertising and digital content recommendation, both
of questionable value to society. Several scholars have explained how
the use of machine learning can perpetuate inequity through the ways
that it can put additional burden on already marginalized, oppressed,
and disadvantaged communities. Narratives of artificial intelligence
also shape policy in several high stakes debates about the replacement
of human judgment in favor of statistical models in the criminal justice
system, health care, education, and social services.

There are some notable topics we left out. Some might find that the most
glaring omission is the lack of material on unsupervised learning.
Indeed, there has been a significant amount of work on unsupervised
learning in recent years. Thankfully, some of the most successful
approaches to learning without labels could be described as
\emph{reductions to pattern recognition}. For example, researchers have
found ingenious ways of procuring labels from unlabeled data points, an
approach called self supervision. We believe that the contents of this
book will prepare students interested in these topics well.

The material we cover supports a one semester graduate introduction to
machine learning. We invite readers from all backgrounds. However,
mathematical maturity with probability, calculus, and linear algebra is
required. We provide a chapter on mathematical background for review.
Necessarily, this chapter cannot replace prerequesite coursework.

In writing this book, our goal was to balance mathematical rigor against
presenting insights we have found useful in the most direct way
possible. In contemporary learning theory important results often have
short sketches, yet making these arguments rigorous and precise may
require dozens of pages of technical calculations. Such proofs are
critical to the community's scientific activities but often make
important insights hard to access for those not yet versed in the
appropriate techniques. On the other hand, many machine learning courses
drop proofs altogether, thereby losing the important foundational ideas
that they contain. We aim to strike a balance, including full details
for as many arguments as possible, but frequently referring readers to
the relevant literature for full details.

\chapter*{Acknowledgments}

We are indebted to Alexander Rakhlin, who pointed us to the early
generalization bound for the Perceptron algorithm. This result both in
its substance and historical position shaped our understanding of
machine learning. Kevin Jamieson was the first to point out to us the
similarity between the structure of our course and the text by Duda and
Hart. Peter Bartlett provided many helpful pointers to the literature
and historical context about generalization theory. Jordan Ellenberg
helped us improve the presentation of algorithmic stability. Dimitri
Bertsekas pointed us to an elegant proof of the Neyman-Pearson Lemma. We
are grateful to Rediet Abebe and Ludwig Schmidt for discussions relating
to the chapter on datasets. We also are grateful to David Aha, Thomas
Dietterich, Michael I. Jordan, Pat Langley, John Platt, and Csaba
Szepesvari for giving us additional context about the state of machine
learning in the 1980s. Finally, we are indebted to Boaz Barak, David
Blei, Adam Klivans, Csaba Szepesvari, and Chris Wiggins for detailed
feedback and suggestions on an early draft of this text. We're also
grateful to Chris Wiggins for pointing us to Highleyman's data.

We thank all students of UC Berkeley's CS 281a in the Fall of 2019,
2020, and 2021, who worked through various iterations of the material in
this book. Special thanks to our graduate student instructors Mihaela
Curmei, Sarah Dean, Frances Ding, Sara Fridovich-Keil, Wenshuo Guo,
Chloe Hsu, Meena Jagadeesan, John Miller, Robert Netzorg, Juan C.
Perdomo, and Vickie Ye, who spotted and corrected many mistakes we made.

\mainmatter

\sloppy

\setcounter{page}{1}
\pagenumbering{arabic}

\chapter{Introduction}

``Reflections on life and death of those who in Breslau lived and died''
is the title of a manuscript that Protestant pastor Caspar Neumann sent
to mathematician Gottfried Wilhelm Leibniz in the late 17th century.
Neumann had spent years keeping track of births and deaths in his Polish
hometown now called Wrocław. Unlike sprawling cities like London or
Paris, Breslau had a rather small and stable population with limited
migration in and out. The parishes in town took due record of the newly
born and deceased.

Neumann's goal was to find patterns in the occurrence of births and
deaths. He thereby sought to dispel a persisting superstition that
ascribed critical importance to certain climacteric years of age. Some
believed it was age 63, others held it was either the 49th or the 81st
year, that particularly critical events threatened to end the journey of
life. Neumann recognized that his data defied the existence of such
climacteric years.

Leibniz must have informed the Royal Society of Neumann's work. In turn,
the Society invited Neumann in 1691 to provide the Society with the data
he had collected. It was through the Royal Society that British
astronomer Edmund Halley became aware of Neumann's work. A friend of
Isaac Newton's, Halley had spent years predicting the trajectories of
celestial bodies, but not those of human lives.

After a few weeks of processing the raw data through smoothing and
interpolation, it was in the Spring of 1693 that Halley arrived at what
became known as Halley's life table.

\begin{figure}
\centering
\includegraphics[width=0.75\textwidth,height=\textheight]{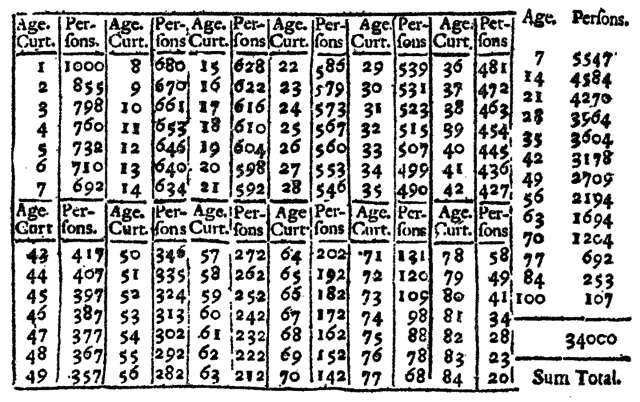}
\caption{Halley's life table}
\end{figure}

At the outset, Halley's table displayed for each year of age, the number
of people of that age alive in Breslau at the time. Halley estimated
that a total of approximately 34000 people were alive, of which
approximately 1000 were between the ages zero and one, 855 were between
age one and two, and so forth.

Halley saw multiple applications of his table. One of them was to
estimate the proportion of men in a population that could bear arms. To
estimate this proportion he computed the number of people between age 18
and 56, and divided by two. The result suggested that 26\% of the
population were men neither too old nor too young to go to war.

At the same time, King William III of England needed to raise money for
his country's continued involvement in the Nine Years War raging from
1688 to 1697. In 1692, William turned to a financial innovation imported
from Holland, the public sale of life annuities. A life annuity is a
financial product that pays out a predetermined annual amount of money
while the purchaser of the annuity is alive. The king had offered
annuities at fourteen times the annual payout, a price too low for the
young and too high for the old.

Halley recognized that his table could be used to estimate the odds that
a person of a certain age would die within the next year. Based on this
observation, he described a formula for pricing an annuity that,
expressed in modern language, computes the sum of expected discounted
payouts over the course of a person's life starting from their current
age.

\hypertarget{ambitions-of-the-20th-century}{%
\section{Ambitions of the 20th
century}\label{ambitions-of-the-20th-century}}

Halley had stumbled upon the fact that prediction requires no physics.
Unknown outcomes, be they future or unobserved, often follow patterns
found in past observations. This empirical law would become the basis of
consequential decision making for centuries to come.

On the heels of Halley and his contemporaries, the 18th century saw the
steady growth of the life insurance industry. The industrial revolution
fueled other forms of insurance sold to a population seeking safety in
tumultuous times. Corporations and governments developed risk models of
increasing complexity with varying degrees of rigor. Actuarial science
and financial risk assessment became major fields of study built on the
empirical law.

Modern statistics and decision theory emerged in the late 19th and early
20th century. Statisticians recognized that the scope of the empirical
law extended far beyond insurance pricing, that it could be a method for
both scientific discovery and decision making writ large.

Emboldened by advances in probability theory, statisticians modeled
populations as probability distributions. Attention turned to what a
scientist could say about a population by looking at a random draw from
its probability distribution. From this perspective, it made sense to
study how to decide between one of two plausible probability models for
a population in light of available data. The resulting concepts, such as
true positive and false positive, as well as the resulting technical
repertoire, are in broad use today as the basis of hypothesis testing
and binary classification.

As statistics flourished, two other developments around the middle of
the 20th century turned out to be transformational. The works of Turing,
Gödel, and von Neumann, alongside dramatic improvements in hardware,
marked the beginning of the computing revolution. Computer science
emerged as a scientific discipline. General purpose programmable
computers promised a new era of automation with untold possibilities.

World War II spending fueled massive research and development programs
on radar, electronics, and servomechanisms. Established in 1940, the
United States National Defense Research Committee, included a division
devoted to control systems. The division developed a broad range of
control systems, including gun directors, target predictors, and
radar-controlled devices. The agency also funded theoretical work by
mathematician Norbert Wiener, including plans for an ambitious
anti-aircraft missile system that used statistical methods for
predicting the motion of enemy aircraft.

In 1948, Wiener released his influential book \emph{Cybernetics} at the
same time as Shannon released \emph{A Mathematical Theory of
Communication}. Both proposed theories of information and communication,
but their goals were different. Wiener's ambition was to create a new
science, called cybernetics, that unified communications and control in
one conceptual framework. Wiener believed that there was a close analogy
between the human nervous system and digital computers. He argued that
the principles of control, communication, and feedback could be a way
not only to create mind-like machines, but to understand the interaction
of machines and humans. Wiener even went so far as to posit that the
dynamics of entire social systems and civilizations could be understood
and steered through the organizing principles of cybernetics.

The zeitgeist that animated cybernetics also drove ambitions to create
artificial neural networks, capable of carrying out basic cognitive
tasks. Cognitive concepts such as learning and intelligence had entered
research conversations about computing machines and with it came the
quest for machines that learn from experience.

The 1940s were a decade of active research on artificial neural
networks, often called connectionism. A 1943 paper by McCulloch and
Pitts formalized artificial neurons and provided theoretical results
about the universality of artificial neural networks as computing
devices. A 1949 book by Donald Hebb pursued the central idea that neural
networks might learn by constructing internal representations of
concepts.

\hypertarget{pattern-classification}{%
\section{Pattern classification}\label{pattern-classification}}

Around the mid 1950s, it seemed that progress on connectionism had
started to slow and would have perhaps tapered off had psychologist
Frank Rosenblatt not made a striking discovery.

Rosenblatt had devised a machine for image classification. Equipped with
400 photosensors the machine could read an image composed of 20 by 20
pixels and sort it into one of two possible classes. Mathematically, the
Perceptron computes a linear function of its input pixels. If the value
of the linear function applied to the input image is positive, the
Perceptron decides that its input belongs to class 1, otherwise class
-1. What made the Perceptron so successful was the way it could learn
from examples. Whenever it misclassified an image, it would adjust the
coefficients of its linear function via a local correction.

Rosenblatt observed in experiments what would soon be a theorem. If a
sequence of images could at all be perfectly classified by a linear
function, the Perceptron would only make so many mistakes on the
sequence before it correctly classified all images it encountered.

Rosenblatt developed the Perceptron in 1957 and continued to publish on
the topic in the years that followed. The Perceptron project was funded
by the US Office of Naval Research, who jointly announced the project
with Rosenblatt in a press conference in 1958, that led to the New York
Times to exclaim:

\begin{quote}
The Navy revealed the embryo of an electronic computer that it expects
will be able to walk, talk, see, write, reproduce itself and be
conscious of its existence.\citep{nytimes1958new}
\end{quote}

This development sparked significant interest in perceptrons and
reinvigorated neural networks research throughout the 1960s. By all
accounts, the research in the decade that followed Rosenblatt's work had
essentially all the ingredients of what is now called machine learning,
specifically, supervised learning.

Practitioners experimented with a range of different features and model
architectures, moving from linear functions to Perceptrons with multiple
layers, the equivalent of today's deep neural networks. A range of
variations to the optimization method and different ways of propagating
errors came and went.

Theory followed closely behind. Not long after the invention came a
theorem, called mistake bound, that gave an upper bound on the number of
mistakes the Perceptron would make in the worst case on any sequence of
labeled data points that can be fit perfectly with a linear separator.

Today, we recognize the Perceptron as an instance of the stochastic
gradient method applied to a suitable objective function. The stochastic
gradient method remains the optimization workhorse of modern machine
learning applications.

Shortly after the well-known mistake bound came a lesser known theorem.
The result showed that when the Perceptron succeeded in fitting training
data, it would also succeed in classifying unseen examples correctly
provided that these were drawn from the same distribution as the
training data. We call this \emph{generalization}: Finding rules
consistent with available data that apply to instances we have yet to
encounter.

By the late 1960s, these ideas from perceptrons had solidified into a
broader subject called \emph{pattern recognition} that knew most of the
concepts we consider core to machine learning today. In 1939, Wald
formalized the basic problem of classification as one of optimal
decision making when the data is generated by a known probabilistic
model. Researchers soon realized that pattern classification could be
achieved using data alone to guide prediction methods such as
perceptrons, nearest neighbor classifiers, or density estimators. The
connections with mathematical optimization including gradient descent
and linear programming also took shape during the 1960s.

Pattern classification---today more popularly known as supervised
learning---built on statistical tradition in how it formalized the idea
of generalization. We assume observations come from a fixed data
generating process, such as, samples drawn from a fixed distribution. In
a first optimization step, called training, we fit a model to a set of
data points labeled by class membership. In a second step, called
testing, we judge the model by how well it performs on newly generated
data from the very same process.

This notion of generalization as performance on fresh data can seem
mundane. After all, it simply requires the classifier to do, in a sense,
more of the same. We require consistent success on the same data
generating process as encountered during training. Yet the seemingly
simple question of what theory underwrites the generalization ability of
a model has occupied the machine learning research community for
decades.

\hypertarget{pattern-classification-once-again}{%
\subsection{Pattern classification, once
again}\label{pattern-classification-once-again}}

Machine learning as a field, however, is not a straightforward evolution
of the pattern recognition of the 1960s, at least not culturally and not
historically.

After a decade of perceptrons research, a group of influential
researchers, including McCarthy, Minsky, Newell, and Simon put forward a
research program by the name of artificial intelligence. The goal was to
create human-like intelligence in a machine. Although the goal itself
was in many ways not far from the ambitions of connectionists, the group
around McCarthy fancied entirely different formal techniques. Rejecting
the numerical pattern fitting of the connectionist era, the proponents
of this new discipline saw the future in symbolic and logical
manipulation of knowledge represented in formal languages.

Artificial intelligence became the dominant academic discipline to deal
with cognitive capacities of machines within the computer science
community. Pattern recognition and neural networks research continued,
albeit largely outside artificial intelligence. Indeed, journals on
pattern recognition flourished during the 1970s.

During this time, artificial intelligence research led to a revolution
in \emph{expert systems}, logic and rule based models that had
significant industrial impact. Expert systems were hard coded and left
little room for adapting to new information. AI researchers interested
in such adaptation and improvement---learning, if you will---formed
their own subcommunity, beginning in 1981 with the first International
Workshop on Machine Learning. The early work from this community
reflects the logic-based research that dominated artificial intelligence
at the time; the papers read as if of a different field than what we now
recognize as machine learning research. It was not until the late 1980s
that machine learning began to look more like pattern recognition, once
again.

Personal computers had made their way from research labs into home
offices across wealthy nations. Internet access, if slow, made email a
popular form of communication among researchers. File transfer over the
internet allowed researchers to share code and datasets more easily.

Machine learning researchers recognized that in order for the discipline
to thrive it needed a way to more rigorously evaluate progress on
concrete tasks. Whereas in the 1950s it had seemed miraculous enough if
training errors decreased over time on any non-trivial task, it was
clear now that machine learning needed better benchmarks.

In the late 1980s, the first widely used benchmarks emerged. Then
graduate student David Aha created the UCI machine learning repository
that made several datasets widely available via FTP. Aiming to better
quantify the performance of AI systems, the Defense Advanced Research
Projects Agency (DARPA) funded a research program on speech recognition
that led to the creation of the influential TIMIT speech recognition
benchmark.

These benchmarks had the data split into two parts, one called training
data, one called testing data. This split elicits the promise that the
learning algorithm must only access the training data when it fits the
model. The testing data is reserved for evaluating the trained model.
The research community can then rank learning algorithms by how well the
trained models perform on the testing data.

Splitting data into training and testing sets was an old practice, but
the idea of reusing such datasets as benchmarks was novel and
transformed machine learning. The \emph{dataset-as-benchmark paradigm}
caught on and became core to applied machine learning research for
decades to come. Indeed, machine learning benchmarks were at the center
of the most recent wave of progress on deep learning. Chief among them
was ImageNet, a large repository of images, labeled by nouns of objects
displayed in the images. A subset of roughly 1 million images belonging
to 1000 different object classes was the basis of the ImageNet Large
Scale Visual Recognition Challenge. Organized from 2010 until 2017, the
competition became a striking showcase for performance of deep learning
methods for image classification.

Increases in computing power and volume of available data were a key
driving factor for progress in the field. But machine learning
benchmarks did more than to provide data. Benchmarks gave researchers a
way to compare results, share ideas, and organize communities. They
implicitly specified a problem description and a minimal interface
contract for code. Benchmarks also became a means of knowledge transfer
between industry and academia.

The most recent wave of machine learning as pattern classification was
so successful, in fact, that it became the new artificial intelligence
in the public narrative of popular media. The technology reached
entirely new levels of commercial significance with companies competing
fiercely over advances in the space.

This new artificial intelligence had done away with the symbolic
reasoning of the McCarthy era. Instead, the central drivers of progress
were widely regarded as growing datasets, increasing compute resources,
and more benchmarks along with publicly available code to start from.
Are those then the only ingredients needed to secure the sustained
success of machine learning in the real world?

\hypertarget{prediction-and-action}{%
\section{Prediction and action}\label{prediction-and-action}}

Unknown outcomes often follow patterns found in past observations. But
what do we do with the patterns we find and the predictions we make?
Like Halley proposing his life table for annuity pricing, predictions
only become useful when they are acted upon. But going from patterns and
predictions to successful actions is a delicate task. How can we even
anticipate the effect of a hypothetical action when our actions now
influence the data we observe and value we accrue in the future?

One way to determine the effect of an action is experimentation: try it
out and see what happens. But there's a lot more we can do if we can
model the situation more carefully. A model of the environment specifies
how an action changes the state of the world, and how in turn this state
results in a gain or loss of utility. We include some aspects of the
environment explicitly as variables in our model. Others we declare
\emph{exogenous} and model as noise in our system.

The solution of how to take such models and turn them into plans of
actions that maximize expected utility is a mathematical achievement of
the 20th century. By and large, such problems can be solved by
\emph{dynamic programming}. Initially formulated by Bellman in 1954,
dynamic programming poses optimization problems where at every time
step, we observe data, take an action, and pay a cost. By chaining these
together in time, elaborate plans can be made that remain optimal under
considerable stochastic uncertainty. These ideas revolutionized
aerospace in the 1960s, and are still deployed in infrastructure
planning, supply chain management, and the landing of SpaceX rockets.
Dynamic programming remains one of the most important algorithmic
building blocks in the computer science toolkit.

Planning actions under uncertainty has also always been core to
artificial intelligence research, though initial proposals for
sequential decision making in AI were more inspired by neuroscience than
operations research. In 1950-era AI, the main motivating concept was one
of \emph{reinforcement learning}, which posited that one should
encourage taking actions that were successful in the past. This
reinforcement strategy led to impressive game-playing algorithms like
Samuel's Checkers Agent circa 1959. Surprisingly, it wasn't until the
1990s that researchers realized that reinforcement learning methods were
approximation schemes for dynamic programming. Powered by this
connection, a mix of researchers from AI and operations research applied
neural nets and function approximation to simplify the approximate
solution of dynamic programming problems. The subsequent 30 years have
led to impressive advances in reinforcement learning and approximate
dynamic programming techniques for playing games, such as Go, and in
powering dexterous manipulation in robotic systems.

Central to the reinforcement learning paradigm is understanding how to
balance learning about an environment and acting on it. This balance is
a non-trivial problem even in the case where actions do not lead to a
change in state. In the context of machine learning, experimentation in
the form of taking an action and observing its effect often goes by the
name \emph{exploration}. Exploration reveals the payoff of an action,
but it comes at the expense of not taking an action that we already knew
had a decent payoff. Thus, there is an inherent tradeoff between
exploration and \emph{exploitation} of previous actions. Though in
theory, the optimal balance can be computed by dynamic programming, it
is more common to employ techniques from \emph{bandit optimization} that
are simple and effective strategies to balance exploration and
exploitation.

Not limited to experimentation, causality is a comprehensive conceptual
framework to reason about the effect of actions. Causal inference, in
principle, allows us to estimate the effect of hypothetical actions from
observational data. A growing technical repertoire of causal inference
is taking various sciences by storm as witnessed in epidemiology,
political science, policy, climate, and development economics.

There are good reasons that many see causality as a promising avenue for
making machine learning methods more robust and reliable. Current
state-of-the-art predictive models remain surprisingly fragile to
changes in the data. Even small natural variations in a data-generating
process can significantly deteriorate performance. There is hope that
tools from causality could lead to machine learning methods that perform
better under changing conditions.

However, causal inference is no panacea. There are no causal insights
without making substantive judgments about the problem that are not
verifiable from data alone. The reliance on hard earned substantive
domain knowledge stands in contrast with the nature of recent advances
in machine learning that largely did without---and that was the point.

\hypertarget{chapter-notes}{%
\section{Chapter notes}\label{chapter-notes}}

Halley's life table has been studied and discussed extensively; for an
entry point, see recent articles by Bellhouse\citep{bellhouse2011new}
and Ciecka\citep{ciecka2008edmond}, or the article by Pearson and
Pearson.\citep{pearson1981history}

Halley was not the first to create a life table. In fact, what Halley
created is more accurately called a population table. Instead, John
Grount deserves credit for the first life table in 1662 based on
mortality records from London. Considered to be the founder of
demography and an early epidemiologist, Grount's work was in many ways
more detailed than Halley's fleeting engagement with Breslau's
population. However, to Grount's disadvantage the mortality records
released in London at the time did not include the age of the deceased,
thus complicating the work significantly.

Mathematician de Moivre picked up Halley's life table in 1725 and
sharpened the mathematical rigor of Halley's idea. A few years earlier,
de Moivre had published the first textbook on probability theory called
``The Doctrine of Chances: A Method of Calculating the Probability of
Events in Play''. Although de Moivre lacked the notion of a probability
distribution, his book introduced an expression resembling the normal
distribution as an approximation to the Binomial distribution, what was
in effect the first central limit theorem. The time of Halley coincides
with the emergence of probability. Hacking's book provides much
additional context, particularly relevant are Chapter 12 and
13.\citep{hacking2006emergence}

For the history of feedback, control, and computing before cybernetics,
see the excellent text by Mindell.\citep{mindell2002between} For more on
the cybernetics era itself, see the books by
Kline\citep{kline2015cybernetics} and Heims\citep{heims1991cybernetics}.
See Beninger\citep{beniger2009control} for how the concepts of control
and communication and the technology from that era lead to the modern
information society.

The prologue from the 1988 edition of \emph{Perceptrons} by Minsky and
Papert presents a helpful historical perspective. The recent 2017
reprint of the same book contains additional context and commentary in a
foreword by Léon Bottou.

Much of the first International Workshop on Machine Learning was
compiled in an edited volume, which summarizes the motivations and
perspectives that seeded the field.\citep{Michalski83} Langley's article
provides helpful context on the state of evaluation in machine learning
in the 1980s and how the desire for better metrics led to a renewed
emphasis on pattern recognition.\citep{langley2011changing} Similar
calls for better evaluation motivated the speech transcription program
at DARPA, leading to the TIMIT dataset, arguably the first machine
learning benchmark
dataset.\citep{liberman10obituary, Church18, Liberman20}

It is worth noting that the Parallel Distributed Processing Research
Group led by Rummelhart and McLeland actively worked on neural networks
during the 1980s and made extensive use of the rediscovered
back-propagation algorithm, an efficient algorithm for computing partial
derivatives of a circuit.\citep{mcclelland1986parallel}

A recent article by Jordan provides an insightful perspective on how the
field came about and what challenges it still
faces.\citep{jordan2019artificial}

\chapter{Fundamentals of prediction}

\emph{Prediction}\index{prediction} is the art and science of leveraging
patterns found in natural and social processes to conjecture about
uncertain events. We use the word \emph{prediction} broadly to refer to
statements about things we don't know for sure \emph{yet}, including but
not limited to the outcome of future events.

Machine learning is to a large extent the study of algorithmic
prediction. Before we can dive into machine learning, we should
familiarize ourselves with prediction. Starting from first principles,
we will motivate the goals of prediction before building up to a
statistical theory of prediction.

We can formalize the goal of prediction problems by assuming a
population of~\(N\) instances with a variety of attributes. We associate
with each instance two variables, denoted~\(X\) and~\(Y\). The goal of
prediction is to conjecture a plausible value for~\(Y\) after observing
\(X\) alone. But when is a prediction good? For that, we must quantify
some notion of the quality of prediction and aim to optimize that
quantity.

To start, suppose that for each variable~\(X\) we make a deterministic
prediction~\(f(X)\) by means of some prediction function~\(f\). A
natural goal is to find a function~\(f\) that makes the fewest number of
incorrect predictions, where~\(f(X)\ne Y\), across the population. We
can think of this function as a computer program that reads~\(X\) as
input and outputs a prediction~\(f(X)\) that we hope matches the
value~\(Y\). For a fixed prediction function,~\(f\), we can sum up all
of the errors made on the population. Dividing by the size of the
population, we observe the average (or mean) error rate of the function.

\hypertarget{minimizing-errors}{%
\subsection{Minimizing errors}\label{minimizing-errors}}

Let's understand how we can find a prediction function that makes as few
errors as possible on a given population in the case of binary
prediction, where the variable~\(Y\) has only two values.

Consider a population of Abalone, a type of marine snail with colorful
shells featuring a varying number of rings. Our goal is to predict the
sex, male or female, of the Abalone from the number of rings on the
shell. We can tabulate the population of Abalone by counting for each
possible number of rings, the number of male and female instances in the
population.

\begin{figure}
\centering
\includegraphics{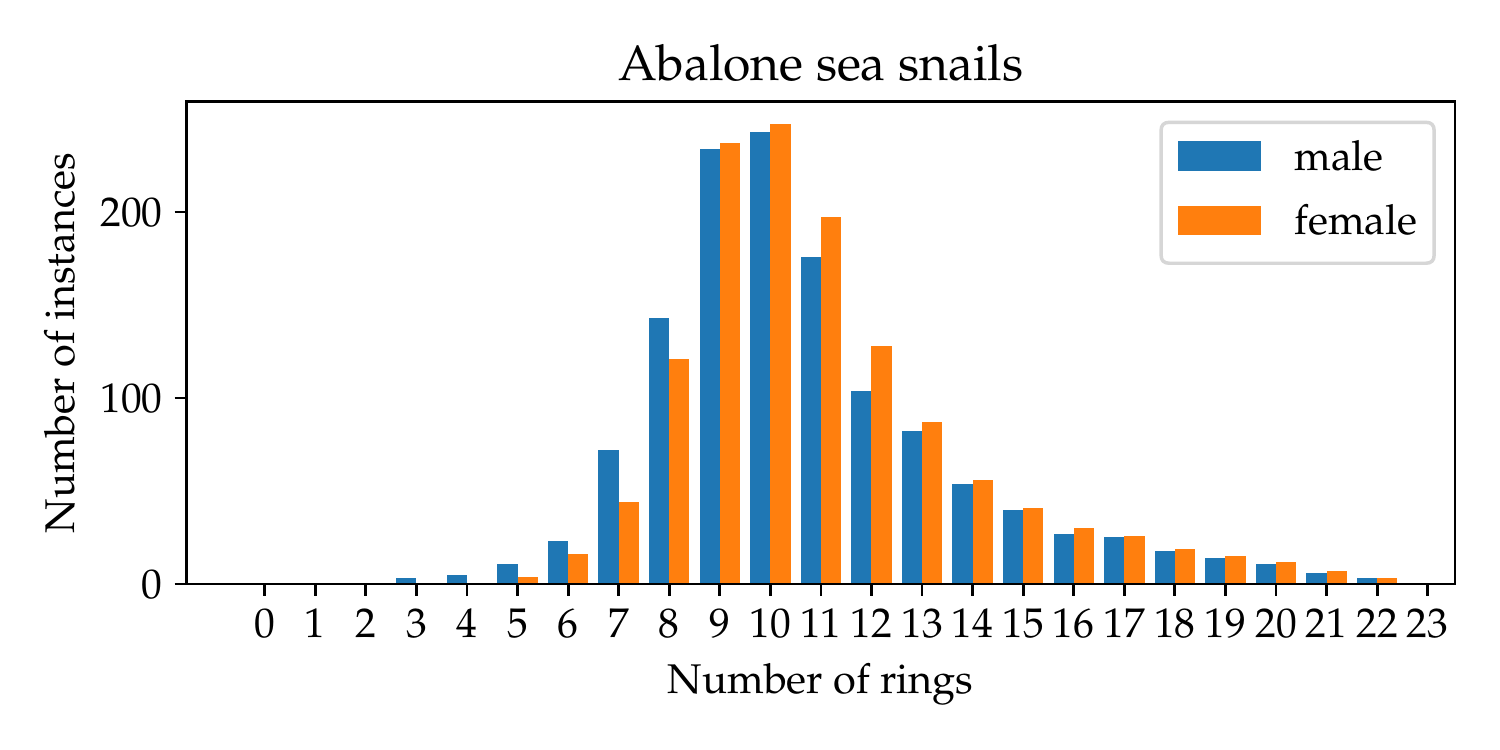}
\caption{Predicting the sex of Abalone sea snails}
\end{figure}

From this way of presenting the population, it is not hard to compute
the predictor that makes the fewest mistakes. For each value on the
X-axis, we predict ``female'' if the number of female instances with
this X-value is larger than the number of male instances. Otherwise, we
predict ``male'' for the given X-value. For example, there's a majority
of male Abalone with seven rings on the shell. Hence, it makes sense to
predict ``male'' when we see seven rings on a shell. Scrutinizing the
figure a bit further, we can see that the best possible predictor is a
\emph{threshold function} that returns ``male'' whenever the number of
rings is at most 8, and ``female'' whenever the number of rings is
greater or equal to 9.

The number of mistakes our predictor makes is still significant. After
all, most counts are pretty close to each other. But it's better than
random guessing. It uses whatever there is that we can say from the
number of rings about the sex of an Abalone snail, which is just not
much.

What we constructed here is called the \emph{minimum error
rule}.\index{minimum error rule} It generalizes to multiple attributes.
If we had measured not only the number of rings, but also the length of
the shell, we would repeat the analogous counting exercise over the
two-dimensional space of all possible values of the two attributes.

The minimum error rule is intuitive and simple, but computing the rule
exactly requires examining the entire population. Tracking down every
instance of a population is not only intractable. It also defeats the
purpose of prediction in almost any practical scenario. If we had a way
of enumerating the~\(X\) and~\(Y\) value of all instances in a
population, the prediction problem would be solved. Given an
instance~\(X\) we could simply look up the corresponding value of~\(Y\)
from our records.

What's missing so far is a way of doing prediction that does not require
us to enumerate the entire population of interest.

\hypertarget{modeling-knowledge}{%
\section{Modeling knowledge}\label{modeling-knowledge}}

Fundamentally, what makes prediction without enumeration possible is
\emph{knowledge}\index{knowledge} about the population. Human beings
organize and represent knowledge in different ways. In this chapter, we
will explore in depth the consequences of one particular way to
represent populations, specifically, as \emph{probability
distributions}.

The assumption we make is that we have knowledge of a probability
distribution~\(p(x,y)\) over pairs of~\(X\) and~\(Y\) values. We assume
that this distribution conceptualizes the ``typical instance'' in a
population. If we were to select an instance uniformly at random from
the population, what relations between its attributes might we expect?
We expect that a uniform sample from our population would be the same as
a sample from~\(p(x,y)\). We call such a distribution a
\emph{statistical model} or simply \emph{model} of a population. The
word \emph{model} emphasizes that the distribution isn't the population
itself. It is, in a sense, a sketch of a population that we use to make
predictions.

Let's revisit our Abalone example in probabilistic form. Assume we know
the distribution of the number of rings of male and female Abalone, as
illustrated in the figure.

\begin{figure}
\centering
\includegraphics{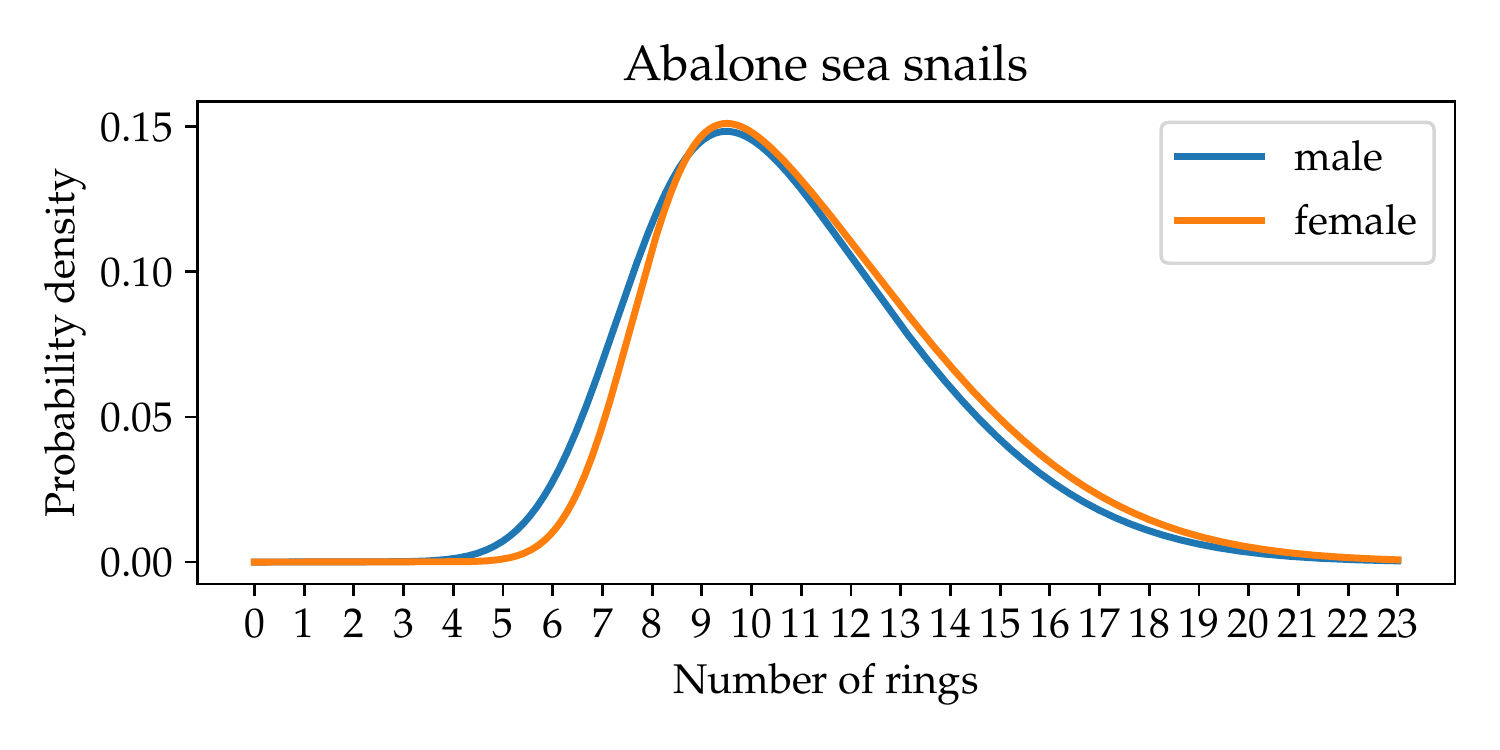}
\caption{Representing Abalone population as a distribution}
\end{figure}

Both follow a skewed normal distribution described by three parameters
each, a location, a scale, and a skew parameter. Knowing the
distribution is to assume that we know these parameters. Although the
specific numbers won't matter for our example, let's spell them out for
concreteness. The distribution for male Abalone has location~\(7.4\),
scale~\(4.48\), and skew~\(3.12\), whereas the distribution for female
Abalone has location~\(7.63\), scale~\(4.67\), and skew~\(4.34\). To
complete the specification of the joint distribution over~\(X\)
and~\(Y\), we need to determine the relative proportion of males and
females. Assume for this example that male and female Abalone are
equally likely.

Representing the population this way, it makes sense to predict ``male''
whenever the probability density for male Abalone is larger than that
for female Abalone. By inspecting the plot we can see that the density
is higher for male snails up until~\(8\) rings at which point it is
larger for female instances. We can see that the predictor we derive
from this representation is the same threshold rule that we had before.

We arrived at the same result without the need to enumerate and count
all possible instances in the population. Instead, we recovered the
minimum error rule from knowing only~\(7\) parameters, three for each
conditional distribution, and one for the balance of the two classes.

Modeling populations as probability distributions is an important step
in making prediction algorithmic. It allows us to represent populations
succinctly, and gives us the means to make predictions about instances
we haven't encountered.

Subsequent chapters extend these fundamentals of prediction to the case
where we don't know the exact probability distribution, but only have a
random sample drawn from the distribution. It is tempting to think about
machine learning as being all about \emph{that}, namely what we do with
a sample of data drawn from a distribution. However, as we learn in this
chapter, many fundamentally important questions arise even if we have
full knowledge of the population.

\hypertarget{prediction-from-statistical-models}{%
\subsection{Prediction from statistical
models}\label{prediction-from-statistical-models}}

Let's proceed to formalize prediction assuming we have full knowledge of
a statistical model of the population. Our first goal is to formally
develop the minimum error rule in greater generality.

We begin with binary prediction where we suppose~\(Y\) has two
alternative values,~\(0\) and~\(1\). Given some measured
information~\(X\), our goal is to conjecture whether~\(Y\) equals zero
or one.

Throughout we assume that~\(X\) and~\(Y\) are random variables drawn
from a joint probability distribution. It is convenient both
mathematically and conceptually to specify the joint distribution as
follows. We assume that~\(Y\) has \emph{a priori} (or \emph{prior})
probabilities: \[
p_0 = \Pr[Y=0]\,,
\qquad p_1 = \Pr[Y=1]
\] That is, the we assume we know the proportion of instances
with~\(Y=1\) and~\(Y=0\) in the population. We'll always model available
information as being a random vector~\(X\) with support in~\(\R^d\). Its
distribution depends on whether~\(Y\) is equal to zero or one. In other
words, there are two different statistical models for the data, one for
each value of \(Y\). These models are the conditional probability
densities of~\(X\) given a value~\(y\) for~\(Y\),
denoted~\(p(x \mid Y=y)\). This density function is often called a
\emph{generative model} or \emph{likelihood function} for each
scenario.\index{generative model}\index{likelihood function}

\hypertarget{example-signal-versus-noise}{%
\subsection{Example: signal versus
noise}\label{example-signal-versus-noise}}

For a simple example with more mathematical formalism, suppose that when
\(Y=0\) we observe a scalar~\(X=\omega\) where~\(\omega\) is
unit-variance, zero mean Gaussian noise~\(\omega\sim \mathcal{N}(0,1)\).
Recall that the Gaussian distribution of mean~\(\mu\) and
variance~\(\sigma^2\) is given by the density
\(\frac{1}{\sigma\sqrt{2\pi}}e^{-\frac12\left(\frac{x-\mu}{\sigma}\right)^2}.\)

Suppose when~\(Y=1\), we would observe~\(X=s+\omega\) for some
scalar~\(s\). That is, the conditional densities are \[
\begin{aligned}
    p(x\mid Y=0) &= \mathcal{N}(0,1) \,, \\
    p(x\mid Y=1) &= \mathcal{N}(s,1)\,.
\end{aligned}
\] The larger the shift~\(s\) is, the easier it is to predict whether
\(Y=0\) or~\(Y=1\). For example, suppose~\(s=10\) and we
observed~\(X=11\). If we had~\(Y=0\), the probability that the
observation is greater than~\(10\) is on the order of~\(10^{-23}\), and
hence we'd likely think we're in the alternative scenario where~\(Y=1\).
However, if~\(s\) were very close to zero, distinguishing between the
two alternatives is rather challenging. We can think of a small
difference~\(s\) that we're trying to detect as a \emph{needle in a
haystack}.

\begin{figure}
\centering
\includegraphics{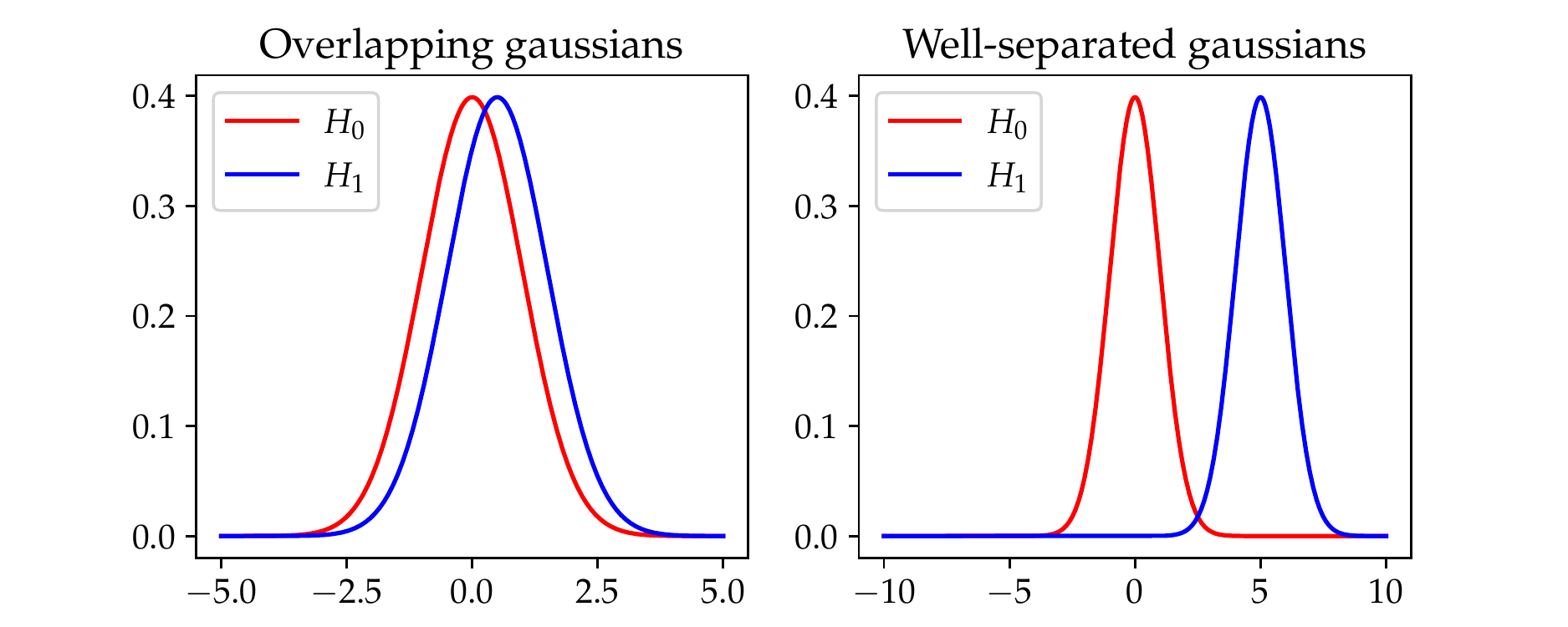}
\caption{Illustration of shifted Gaussians}
\end{figure}

\hypertarget{prediction-via-optimization}{%
\section{Prediction via
optimization}\label{prediction-via-optimization}}

Our core approach to all statistical decision making will be to
formulate an appropriate optimization problem for which the decision
rule is the optimal solution. That is, we will optimize over
\emph{algorithms}, searching for functions that map data to decisions
and predictions. We will define an appropriate notion of the cost
associated to each decision, and attempt to construct decision rules
that minimize the expected value of this cost. As we will see, choosing
this optimization framework has many immediate consequences.

\hypertarget{predictors-and-labels}{%
\subsection{Predictors and labels}\label{predictors-and-labels}}

A \emph{predictor} is a function~\(\hat Y(x)\) that maps an input~\(x\)
to a prediction~\(\hat y=\hat Y(x).\) The prediction~\(\hat y\) is also
called a \emph{label} for the point~\(x\).\index{predictor}\index{label}
The target variable~\(Y\) can be both real valued or discrete.
When~\(Y\) is a discrete random variable, each different value it can
take on is called a \emph{class} of the prediction problem.\index{class}

To ease notation, we take the liberty to write~\(\hat Y\) as a shorthand
for the random variable~\(\hat Y(X)\) that we get by applying the
prediction function~\(\hat Y\) to the random variable~\(X\).

The most common case we consider through the book is binary prediction,
where we have two classes,~\(0\) and~\(1\). Sometimes it's
mathematically convenient to instead work with the numbers~\(-1\)
and~\(1\) for the two classes.

In most cases we consider, labels are scalars that are either discrete
or real-valued. Sometimes it also makes sense to consider vector-valued
predictions and target variables.

The creation and encoding of suitable labels for a prediction problem is
an important step in applying machine learning to real world problems.
We will return to it multiple times.

\hypertarget{loss-functions-and-risk}{%
\subsection{Loss functions and risk}\label{loss-functions-and-risk}}

The final ingredient in our formal setup is a \emph{loss function} which
generalizes the notion of an error that we defined as a mismatch between
prediction and target value.

A \emph{loss function} takes two inputs,\index{loss function} \(\hat y\)
and~\(y\), and returns a real number~\(\loss(\hat y, y)\) that we
interpret as a quantified loss for predicting~\(\hat y\) when the target
is~\(y\). A loss could be negative in which case we think of it as a
reward.

A prediction error corresponds to the loss function
\(\loss(\hat y, y)=\mathbb{1}\{\hat y \ne y\}\) that indicates
disagreement between its two inputs. Loss functions give us modeling
flexibility that will become crucial as we apply this formal setup
throughout this book.

An important notion is the expected loss of a predictor taken over a
population. This construct is called \emph{risk}.

\begin{Definition}

We define the \emph{risk} associated with~\(\hat{Y}\) to be \[
    R[\hat{Y}] := \E[\loss(\hat{Y}(X),Y)]\,.
\] Here, the expectation is taken jointly over~\(X\) and~\(Y.\)

\end{Definition}

Now that we defined risk, our goal is to determine which decision rule
minimizes risk.\index{risk} Let's get a sense for how we might go about
this.

In order to minimize risk, theoretically speaking, we need to solve an
\emph{infinite dimensional} optimization problem over binary-valued
functions. That is, \emph{for every} \(x\), we need to find a binary
assignment. Fortunately, the infinite dimension here turns out to not be
a problem analytically once we make use of the law of iterated
expectation.

\begin{Lemma}

We claim that the optimal predictor is given by \[
\hat {Y}(x) = \mathbb{1}\left\{\Pr[Y=1\mid X=x] \geq  \frac{\loss(1,0)-\loss(0,0)}{\loss(0,1)-\loss(1,1)} \,\, \Pr[Y=0\mid X=x]\right\}\,.
\]

\end{Lemma}

This rule corresponds to the intuitive rule we derived when thinking
about how to make predictions over the population. For a fixed value of
the data~\(X=x\), we compare the frequency of which~\(Y=1\) occurs to
which \(Y=0\) occurs. If this frequency exceeds some threshold that is
defined by our loss function, then we set~\(\hat{Y}(x)=1\). Otherwise,
we set \(\hat{Y}(x)=0\).

\begin{Proof}

To see why this is rule is optimal, we make use of the law of iterated
expectation: \[
\begin{aligned}
    \E[\loss(\hat{Y}(X),Y)] &= \E\left[  \E\left[\loss(\hat{Y}(X),Y)\mid X\right] \right]\,.
\end{aligned}
\] Here, the outer expectation is over a random draw of~\(X\) and the
inner expectation samples~\(Y\) conditional on~\(X\). Since there are no
constraints on the predictor~\(\hat Y\), we can minimize the expression
by minimizing the inner expectation independently for each possible
setting that~\(X\) can assume.

Indeed, for a fixed value~\(x\), we can expand the expected loss for
each of the two possible predictions: \[
\begin{aligned}
\E[\loss(0,Y)\mid X=x]  &= \loss(0,0) \Pr[Y=0\mid X=x] + \loss(0,1) \Pr[Y=1\mid X=x]\\
\E[\loss(1,Y)\mid X=x]  &= \loss(1,0) \Pr[Y=0\mid X=x] + \loss(1,1) \Pr[Y=1\mid X=x]\,.
\end{aligned}
\] The optimal assignment for this~\(x\) is to set~\(\hat Y(x)=1\)
whenever the second expression is smaller than the first. Writing out
this inequality and rearranging gives us the rule specified in the
lemma.

\end{Proof}

Probabilities of the form~\(\Pr[Y=y\mid X=x]\), as they appeared in the
lemma, are called \emph{posterior} probability.\index{posterior}

We can relate them to the likelihood function via Bayes rule: \[
    \Pr[Y=y\mid X=x] = \frac{p(x\mid Y=y) p_y }{p(x)}\,,
\] where~\(p(x)\) is a density function for the marginal distribution of
\(X\).

When we use posterior probabilities, we can rewrite the optimal
predictor as \[
\hat {Y}(x) = \mathbb{1}\left\{
    \frac{p(x\mid Y=1)}{p(x\mid Y=0)} \geq  \frac{p_0(\loss(1,0)-\loss(0,0))}{p_1(\loss(0,1)-\loss(1,1))}\right\}\,.
\] This rule is an example of a likelihood ratio test.

\begin{Definition}

The \emph{likelihood ratio} is the ratio of the likelihood
functions:\index{likelihood ratio} \[
    \mathcal{L}(x) := \frac{p(x\mid Y=1)}{p(x\mid Y=0)}
\] A \emph{likelihood ratio test} (LRT) is a predictor of the
form\index{likelihood ratio test} \[
\hat{Y}(x) = \mathbb{1}\{ \mathcal{L}(x) \geq \eta\}
\] for some scalar threshold~\(\eta>0\).

\end{Definition}

If we denote the optimal threshold value \[
\eta =  \frac{p_0(\loss(1,0)-\loss(0,0))}{p_1(\loss(0,1)-\loss(1,1))}\,,\qquad(1)
\] then the predictor that minimizes the risk is the likelihood ratio
test \[
    \hat{Y}(x) = \mathbb{1}\{ \mathcal{L}(x)\geq \eta \}\,.
\]

A LRT naturally partitions the sample space in two regions: \[
\begin{aligned}
\mathcal{X}_0 & = \left\{ x \in \mathcal{X} \colon \mathcal{L}(x) \leq \eta  \right\}\\
\mathcal{X}_1 &=  \left\{ x \in \mathcal{X} \colon \mathcal{L}(x) > \eta  \right\}\,.
\end{aligned}
\] The sample space~\(\mathcal{X}\) then becomes the disjoint union of
\(\mathcal{X}_0\) and~\(\mathcal{X}_1\). Since we only need to identify
which set~\(x\) belongs to, we can use any function
\(h:\mathcal{X} \rightarrow \R\) which gives rise to the same threshold
rule. As long as~\(h(x) \leq t\) whenever~\(\mathcal{L}(x) \leq \eta\)
and vice versa, these functions give rise to the same partition into
\(\mathcal{X}_0\) and~\(\mathcal{X}_1\). So, for example, if~\(g\) is
any monotonically increasing function, then the predictor \[
        \hat{Y}_g(x) = \mathbb{1} \{ g(\mathcal{L}(x))\geq g(\eta) \}
\] is equivalent to using~\(\hat{Y}(x)\). In particular, it's popular to
use the logarithmic predictor \[
    \hat{Y}_{\mathrm{log}}(x)  =  \mathbb{1} \{\log p(x\mid Y=1)- \log p(x\mid Y=0)   \geq \log(\eta)\}\,,
\] as it is often more convenient or numerically stable to work with
logarithms of likelihoods.

This discussion shows that there are an \emph{infinite number of
functions} which give rise to the same binary predictor. Hence, we don't
need to know the conditional densities exactly and can still compute the
optimal predictor. For example, suppose the true partitioning of the
real line under an LRT is \[
  \mathcal{X}_0 = \{x\colon x\geq 0\}\quad\text{and}\quad\mathcal{X}_1 = \{x\colon x< 0\}\,.
\] Setting the threshold to~\(t=0\), the functions~\(h(x) = x\) or
\(h(x)=x^3\) give the same predictor, as does any odd function which is
positive on the right half line.

\hypertarget{example-needle-in-a-haystack-revisited}{%
\subsection{Example: needle in a haystack
revisited}\label{example-needle-in-a-haystack-revisited}}

Let's return to our needle in a haystack example with
\begin{align*}     p(X\mid Y=0) &= \mathcal{N}(0,1)\,,\\     p(X\mid Y=1) &= \mathcal{N}(s,1)\,, \end{align*}
and assume that the prior probability of~\(Y=1\) is very small,
say,~\(p_1 = 10^{-6}\). Suppose that if we declare \(\hat{Y}=0\), we do
not pay a cost. If we declare~\(\hat{Y}=1\) but are wrong, we incur a
cost of~\(100\). But if we guess~\(\hat{Y}=1\) and it is actually true
that~\(Y=1\), we actually gain a reward of~\(1,000,000\). That
is~\(\loss(0,0) = 0,\) \(\loss(0,1)=0,\) \(\loss(1,0) = 100,\) and
\(\loss(1,1)=-1,000,000\,.\)

What is the LRT for this problem? Here, it's considerably easier to work
with logarithms: \[
    \log(\eta) = \log\left( \frac{(1-10^{-6}) \cdot 100}{10^{-6} \cdot 10^{6}}\right) \approx 4.61
\] Now, \[
    \log p(x\mid Y=1)- \log p(x\mid Y=0) =  -\frac{1}{2} (x-s)^2 + \frac{1}{2} x^2 = sx-\frac{1}{2}s^2
\] Hence, the optimal predictor is to declare \[
\hat{Y}= \mathbb{1}\left\{ sx > \tfrac{1}{2}s^2+\log(\eta) \right\}\,.
\] The optimal rule here is \emph{linear}. Moreover, the rule divides
the space into two open intervals. While the entire real line lies in
the union of these two intervals, it is exceptionally unlikely to ever
see an~\(x\) larger than~\(|s|+5\). Hence, even if our predictor were
incorrect in these regions, the risk would still be nearly optimal as
these terms have almost no bearing on our expected risk!

\hypertarget{maximum-a-posteriori-and-maximum-likelihood}{%
\subsection{Maximum a posteriori and maximum
likelihood}\label{maximum-a-posteriori-and-maximum-likelihood}}

A folk theorem of statistical decision theory states that essentially
all optimal rules are equivalent to likelihood ratio tests. While this
isn't \emph{always} true, many important prediction rules end up being
equivalent to LRTs. Shortly, we'll see an optimization problem that
speaks to the power of LRTs. But before that, we can already show that
the well known \emph{maximum likelihood} and \emph{maximum a posteriori}
predictors are both LRTs.

The expected error of a predictor is the expected number of times we
declare~\(\hat{Y}=0\) (resp.~\(\hat{Y}=1\)) when~\(\hat{Y}=1\) (resp.
\(\hat{Y}=0\)) is true. Minimizing the error is equivalent to minimizing
the risk with
cost~\(\loss(0,0)=\loss(1,1)=0\),~\(\loss(1,0)=\loss(0,1)=1\). The
optimum predictor is hence a likelihood ratio test. In particular, \[
    \hat{Y}(x) =\mathbb{1}\left\{ \mathcal{L}(x) \geq \tfrac{p_0}{p_1}\right\}\,.
\] Using Bayes rule, one can see that this rule is equivalent to \[
    \hat{Y}(x) = \arg \max_{y\in\{0,1\}}\Pr[Y=y\mid X=x]\,.
\] Recall that the expression~\(\Pr[Y=y\mid X=x]\) is called the
posterior probability\index{posterior} of~\(Y=y\) given~\(X=x\). And
this rule is hence referred to as the \emph{maximum a posteriori} (MAP)
rule.\index{MAP}\index{maximum a posteriori}

As we discussed above, the expression~\(p(x\mid Y=y)\) is called the
\emph{likelihood} of the point~\(x\) given the class~\(Y=y\). A maximum
likelihood rule would set \[
    \hat{Y}(x) = \arg \max_y p(x\mid Y=y)\,.
\] This is completely equivalent to the LRT when~\(p_0=p_1\) and the
costs are~\(\loss(0,0)=\loss(1,1)=0\),~\(\loss(1,0)=\loss(0,1)=1\).
Hence, the maximum likelihood rule is equivalent to the MAP rule with a
uniform prior on the labels.

That both of these popular rules ended up reducing to LRTs is no
accident. In what follows, we will show that LRTs are almost always the
optimal solution of optimization-driven decision theory.

\hypertarget{types-of-errors-and-successes}{%
\section{Types of errors and
successes}\label{types-of-errors-and-successes}}

Let~\(\hat{Y}(x)\) denote any predictor mapping into~\(\{0,1\}\). Binary
predictions can be right or wrong in four different ways summarized by
the \emph{confusion table}.\index{confusion table}

\begin{longtable}[]{@{}lcc@{}}
\caption{Confusion table}\tabularnewline
\toprule
& \(Y=0\) & \(Y=1\) \\
\midrule
\endfirsthead
\toprule
& \(Y=0\) & \(Y=1\) \\
\midrule
\endhead
\(\hat Y = 0\) & true negative & false negative \\
\(\hat Y = 1\) & false positive & true positive \\
\bottomrule
\end{longtable}

Taking expected values over the populations give us four corresponding
\emph{rates} that are characteristics of a predictor.

\begin{enumerate}
\def\labelenumi{\arabic{enumi}.}
\tightlist
\item
  \textbf{True Positive Rate:}
  \(\mathrm{TPR} = \Pr[\hat{Y}(X)=1\mid Y=1]\). Also known as
  \emph{power}, \emph{sensitivity}, \emph{probability of detection}, or
  \emph{recall}.
\item
  \textbf{False Negative Rate:} \(\mathrm{FNR} = 1-\mathrm{TPR}\). Also
  known as \emph{type II error} or \emph{probability of missed
  detection}.
\item
  \textbf{False Positive Rate:}
  \(\mathrm{FPR} = \Pr[\hat{Y}(X)=1\mid Y=0]\). Also known as
  \emph{size} or \emph{type I error} or \emph{probability of false
  alarm}.
\item
  \textbf{True Negative Rate} \(\mathrm{TNR}=1-\mathrm{FPR}\), the
  probability of declaring \(\hat{Y}=0\) given \(Y=0\). This is also
  known as
  \emph{specificity}.\index{true positive rate}\index{false positive rate}\index{false negative rate}\index{true negative rate}
\end{enumerate}

There are other quantities that are also of interest in statistics and
machine learning:

\begin{enumerate}
\def\labelenumi{\arabic{enumi}.}
\tightlist
\item
  \textbf{Precision:} \(P[Y=1 \mid \hat{Y}(X)=1]\). This is equal to
  \((p_1 \mathrm{TPR})/(p_0 \mathrm{FPR}+p_1 \mathrm{TPR})\).
\item
  \textbf{F1-score:} \(F_1\) is the harmonic mean of precision and
  recall. We can write this as \[
  F_1 = \frac{2 \mathrm{TPR}}{1+\mathrm{TPR}+\tfrac{p_0}{p_1} \mathrm{FPR}}
  \]
\item
  \textbf{False discovery rate:} False discovery rate (FDR) is equal to
  the expected ratio of the number of false positives to the total
  number of positives.
\end{enumerate}

\index{precision}\index{false discovery rate}\index{F1-score}

In the case where both labels are equally likely, precision,~\(F_1\),
and \(FDR\) are also only functions of~\(\mathrm{FPR}\)
and~\(\mathrm{TPR}\). However, these quantities explicitly account for
\emph{class imbalances}: when there is a significant skew
between~\(p_0\) and~\(p_1\), such measures are often preferred.

\(\mathrm{TPR}\) and~\(\mathrm{FPR}\) are competing objectives. We'd
like \(\mathrm{TPR}\) as large as possible and~\(\mathrm{FPR}\) as small
as possible.

We can think of risk minimization as optimizing a balance between
\(\mathrm{TPR}\) and~\(\mathrm{FPR}\): \[
        R[\hat{Y}] := \E[\loss(\hat{Y}(X),Y)] = \alpha \mathrm{FPR} - \beta \mathrm{TPR} + \gamma\,,
\] where~\(\alpha\) and~\(\beta\) are nonnegative and~\(\gamma\) is some
constant. For all such~\(\alpha\),~\(\beta\), and~\(\gamma\), the
risk-minimizing predictor is an LRT.

Other cost functions might try to balance~\(\mathrm{TPR}\) versus
\(\mathrm{FPR}\) in other ways. Which pairs of
\((\mathrm{FPR},\mathrm{TPR})\) are achievable?

\hypertarget{roc-curves}{%
\subsection{ROC curves}\label{roc-curves}}

True and false positive rate lead to another fundamental notion, called
the the \emph{receiver operating characteristic (ROC) curve}.
\index{ROC}\index{receiver operating characteristic}

The ROC curve is a property of the joint distribution~\((X, Y)\) and
shows for every possible value~\(\alpha=[0, 1]\) the best possible true
positive rate that we can hope to achieve with any predictor that has
false positive rate~\(\alpha\). As a result the ROC curve is a curve in
the FPR-TPR plane. It traces out the maximal TPR for any given FPR.
Clearly the ROC curve contains values~\((0,0)\) and~\((1,1)\), which are
achieved by constant predictors that either reject or accept all inputs.

\begin{figure}
\centering
\includegraphics[width=0.5\textwidth,height=\textheight]{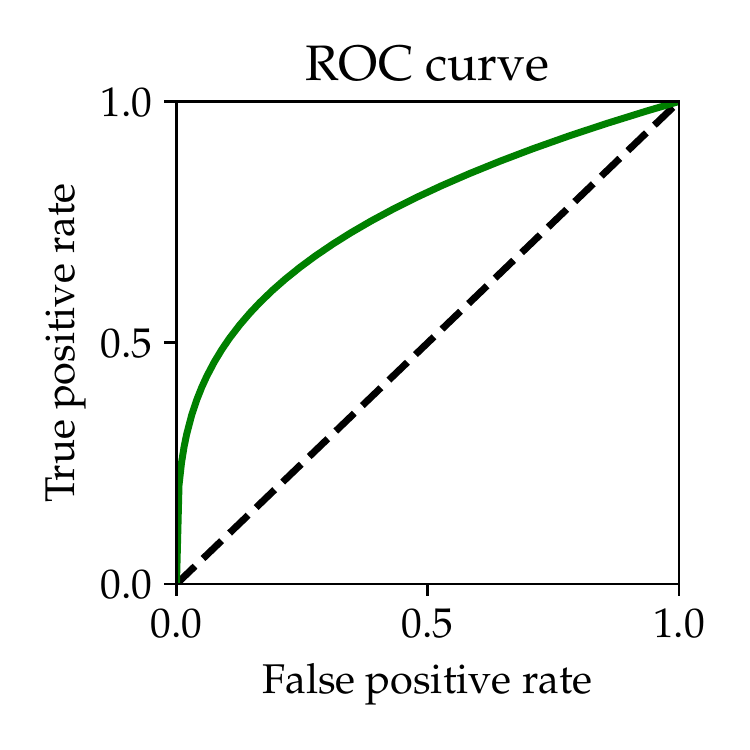}
\caption{Example of an ROC curve}
\end{figure}

We will now show, in a celebrated result by Neyman and Pearson, that the
ROC curve is given by varying the threshold in the likelihood ratio test
from negative to positive infinity.

\hypertarget{the-neyman-pearson-lemma}{%
\section{The Neyman-Pearson Lemma}\label{the-neyman-pearson-lemma}}

The Neyman-Pearson Lemma, a fundamental lemma of decision theory, will
be an important tool for us to establish three important facts. First,
it will be a useful tool for understanding the geometric properties of
ROC curves. Second, it will demonstrate another important instance where
an optimal predictor is a likelihood ratio test. Third, it introduces
the notion of probabilistic predictors.\index{Neyman-Pearson}

Suppose we want to maximize true positive rate subject to an upper bound
on the false positive rate. That is, we aim to solve the optimization
problem: \[
\begin{array}{ll}
    \text{maximize} & \mathrm{TPR}\\
    \text{subject to} & \mathrm{FPR} \leq \alpha
\end{array}
\]

Let's optimize over \emph{probabilistic predictors}. A probabilistic
predictor~\(Q\) returns~\(1\) with probability~\(Q(x)\) and~\(0\) with
probability~\(1-Q(x)\). With such rules, we can rewrite our optimization
problem as: \[
\begin{array}{ll}
    \text{maximize}_Q & \E[ Q(X) \mid Y=1 ] \\
    \text{subject to} & \E[ Q(X) \mid Y=0 ] \leq \alpha\\
       & \forall x\colon Q(x) \in [0,1]
\end{array}
\]

\begin{Lemma}

\textbf{Neyman-Pearson Lemma.} Suppose the likelihood
functions~\(p(x|y)\) are continuous. Then the optimal probabilistic
predictor that maximizes \(\mathrm{TPR}\) with an upper bound
on~\(\mathrm{FPR}\) is a deterministic likelihood ratio test.

\end{Lemma}

Even in this constrained setup, allowing for more powerful probabilistic
rules, we can't escape likelihood ratio tests. The Neyman-Pearson Lemma
has many interesting consequences in its own right that we will discuss
momentarily. But first, let's see why the lemma is true.

The key insight is that for any LRT, we can find a loss function for
which it is optimal. We will prove the lemma by constructing such a
problem, and using the associated condition of optimality.

\begin{Proof}

Let~\(\eta\) be the threshold for an LRT such that the predictor \[
 Q_\eta(x)= \mathbb{1}\{ \mathcal{L}(x) > \eta\}
\] has~\(\mathrm{FPR}=\alpha\). Such an LRT exists because we assumed
our likelihoods were continuous. Let~\(\beta\) denote
the~\(\mathrm{TPR}\) of \(Q_\eta\).

We claim that~\(Q_\eta\) is optimal for the risk minimization problem
corresponding to the loss function \[
\loss(1,0) = \tfrac{\eta p_1}{p_0},~\loss(0,1)=1,~\loss(1,1)=0,~\loss(0,0)=0\,.
\] Indeed, recalling Equation~\(1\), the risk minimizer for this loss
function corresponds to a likelihood ratio test with threshold value \[
\frac{p_0( \loss(1,0) - \loss(0,0) )}
{p_1(\loss(0, 1) - \loss(1, 1) )}
=\frac{p_0\loss(1,0)}
{p_1\loss(0, 1)}
= \eta\,.
\] Moreover, under this loss function, the risk of a predictor~\(Q\)
equals \[
\begin{aligned}
R[Q] &= p_0 \mathrm{FPR}(Q) \loss(1,0) + p_1 (1-\mathrm{TPR}(Q)) \loss(0,1)\\
&= p_1 \eta \mathrm{FPR}(Q) + p_1 (1-\mathrm{TPR}(Q))\,.
\end{aligned}
\]

Now let~\(Q\) be any other predictor with~\(\FPR(Q)\leq \alpha\). We
have by the optimality of~\(Q_\eta\) that \[
\begin{aligned}
    p_1 \eta \alpha + p_1 (1-\beta)
    &\leq p_1 \eta \FPR(Q) + p_1 (1-\TPR(Q))  \\
    &\leq  p_1 \eta \alpha + p_1 (1-\TPR(Q)) \,,
\end{aligned}
\] which implies~\(\TPR(Q) \leq \beta\). This in turn means
that~\(Q_\eta\) maximizes~\(\TPR\) for all rules
with~\(\FPR\leq \alpha\), proving the lemma.

\end{Proof}

\hypertarget{properties-of-roc-curves}{%
\subsection{Properties of ROC curves}\label{properties-of-roc-curves}}

A specific randomized predictor that is useful for analysis combines two
other rules. Suppose predictor one yields
\((\mathrm{FPR}^{(1)},\mathrm{TPR}^{(1)})\) and the second rule achieves
\((\mathrm{FPR}^{(2)},\mathrm{TPR}^{(2)})\). If we flip a biased coin
and use rule one with probability~\(p\) and rule 2 with
probability~\(1-p\), then this yields a randomized predictor with
\((\mathrm{FPR},\mathrm{TPR}) = (p\mathrm{FPR}^{(1)}+(1-p)\mathrm{FPR}^{(2)} ,p \mathrm{TPR}^{(1)}+(1-p)\mathrm{TPR}^{(2)})\).
Using this rule lets us prove several properties of ROC curves.

\begin{Proposition}

The points~\((0,0)\) and~\((1,1)\) are on the ROC curve.

\end{Proposition}

\begin{Proof}

This proposition follows because the point~\((0,0)\) is achieved when
the threshold~\(\eta = \infty\) in the likelihood ratio test,
corresponding to the constant~\(0\) predictor. The point~\((1,1)\) is
achieved when \(\eta = 0\), corresponding to the constant~\(1\)
predictor.

\end{Proof}

The Neyman-Pearson Lemma gives us a few more useful properties.

\begin{Proposition}

The ROC must lie above the main diagonal.

\end{Proposition}

\begin{Proof}

To see why this proposition is true, fix some~\(\alpha>0\). Using a
randomized rule, we can achieve a predictor with
\(\mathrm{TPR}=\mathrm{FPR}=\alpha\). But the Neyman-Pearson LRT with
\(\mathrm{FPR}\) constrained to be less than or equal to~\(\alpha\)
achieves true positive rate greater than or equal to the randomized
rule.

\end{Proof}

\begin{Proposition}

The ROC curve is concave.

\end{Proposition}

\begin{Proof}

Suppose~\((\mathrm{FPR}(\eta_1),\mathrm{TPR}(\eta_1))\) and
\((\mathrm{FPR}(\eta_2),\mathrm{TPR}(\eta_2))\) are achievable. Then \[
    (t\mathrm{FPR}(\eta_1)+(1-t)\mathrm{FPR}(\eta_2) ,t \mathrm{TPR}(\eta_1)+(1-t)\mathrm{TPR}(\eta_2))
\] is achievable by a randomized test. Fixing
\(\mathrm{FPR} \leq t\mathrm{FPR}(\eta_1)+(1-t)\mathrm{FPR}(\eta_2)\),
we see that the optimal Neyman-Pearson LRT achieves
\(\mathrm{TPR} \geq \mathrm{TPR}(\eta_1)+(1-t)\mathrm{TPR}(\eta_2)\).

\end{Proof}

\hypertarget{example-the-needle-one-more-time}{%
\subsection{Example: the needle one more
time}\label{example-the-needle-one-more-time}}

Consider again the \emph{needle in a haystack} example, where
\(p(x\mid Y=0) = \mathcal{N}(0,\sigma^2)\) and
\(p(x\mid Y=1) = \mathcal{N}(s,\sigma^2)\) with~\(s\) a positive scalar.
The optimal predictor is to declare~\(\hat{Y}=1\) when~\(X\) is greater
than \(\gamma:= \frac{s}{2} + \frac{\sigma^2 \log \eta}{s}\). Hence we
have \[
\begin{aligned}
\mathrm{TPR} &= \int_{\gamma}^{\infty} p(x\mid Y=1) \,\dif x = \tfrac{1}{2} \operatorname{erfc} \left(\frac{\gamma-s}{\sqrt{2}\sigma}\right)\\
\mathrm{FPR} &= \int_{\gamma}^{\infty} p(x\mid Y=0) \,\dif x = \tfrac{1}{2} \operatorname{erfc} \left(\frac{\gamma}{\sqrt{2}\sigma}\right)\,.
\end{aligned}
\]

For fixed~\(s\) and~\(\sigma\), the ROC curve
\((\mathrm{FPR}(\gamma),\mathrm{TPR}(\gamma))\) only depends on the
\emph{signal to noise ratio} (SNR),~\(s/\sigma\). For small SNR, the ROC
curve is close to the~\(\mathrm{FPR}=\mathrm{TPR}\) line. For large SNR,
\(\mathrm{TPR}\) approaches~\(1\) for all values of~\(\mathrm{FPR}\).

\begin{figure}
\centering
\includegraphics[width=0.75\textwidth,height=\textheight]{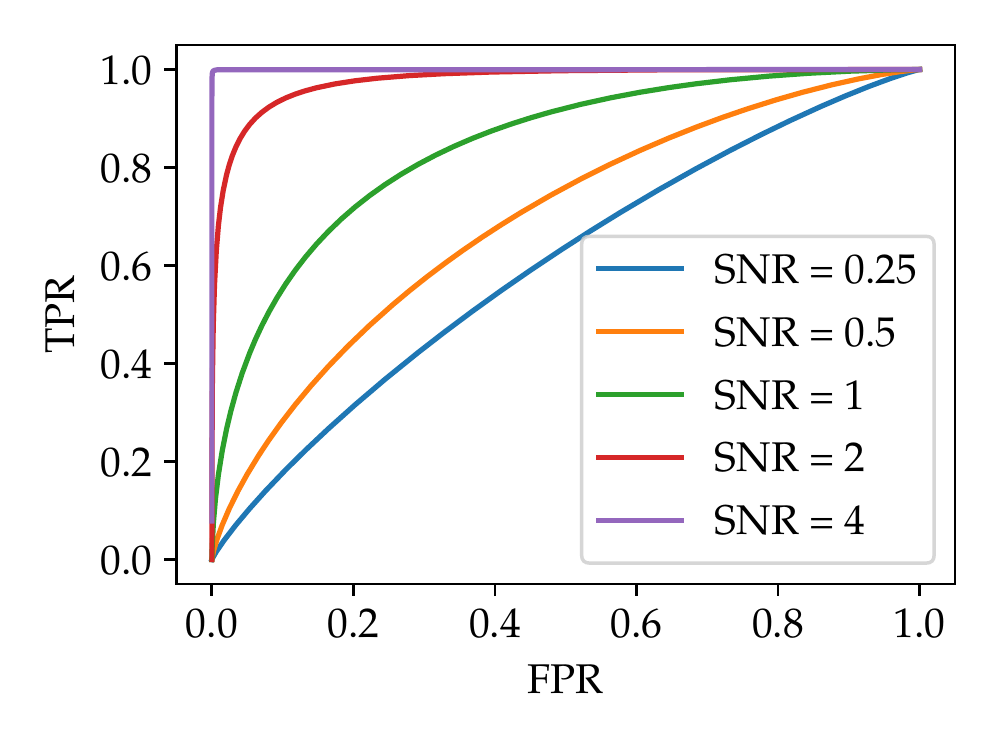}
\caption{The ROC curves for various signal to noise ratios in the needle
in the haystack problem.}
\end{figure}

\hypertarget{area-under-the-roc-curve}{%
\subsection{Area under the ROC curve}\label{area-under-the-roc-curve}}

Oftentimes in information retrieval and machine learning, the term ROC
curve is overloaded to describe the achievable FPR-TPR pairs that we get
by varying the threshold~\(t\) in any predictor
\(\hat Y(x) = \mathbb{1}\{ R(x) > t\}.\) Note such curves must lie below
the ROC curves that are traced out by the optimal likelihood ratio test,
but may approximate the true ROC curves in many cases.

A popular summary statistic for evaluating the quality of a decision
function is the area under its associated ROC curve. This is commonly
abbreviated as AUC. In the ROC curve plotted in the previous section, as
the SNR increases, the AUC increases. However, AUC does not tell the
entire story. Here we plot two ROC curves with the same AUC.

\begin{figure}
\centering
\includegraphics[width=0.75\textwidth,height=\textheight]{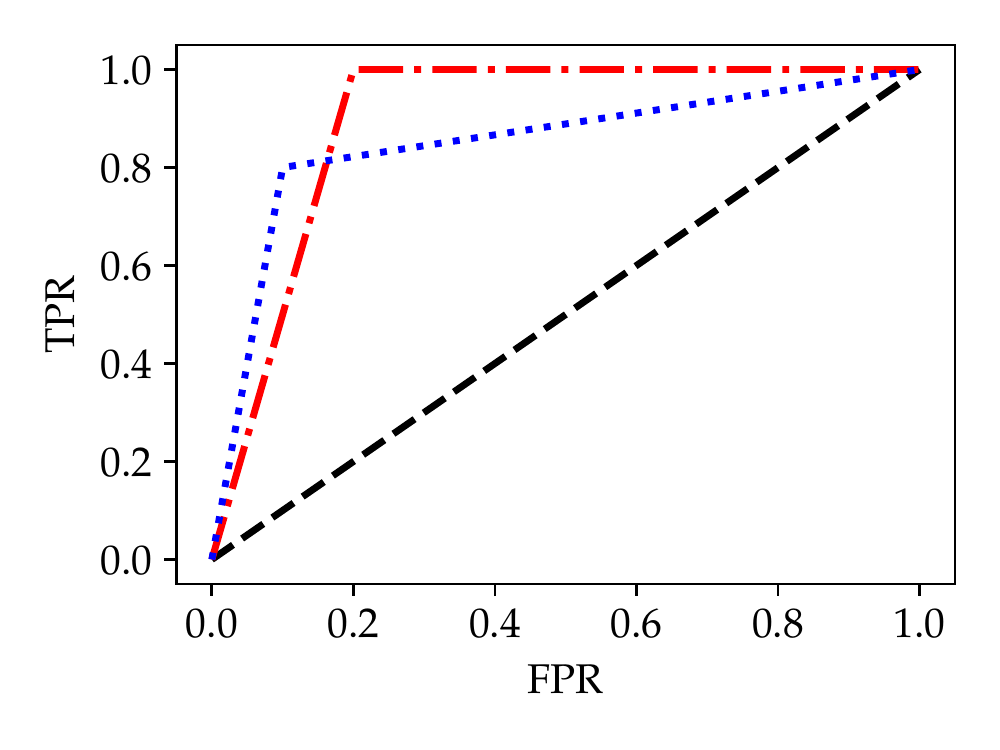}
\caption{Two ROC curves with the same AUC. Note that if we constrain
\(\mathrm{FPR}\) to be less than 10\%, for the blue curve,
\(\mathrm{TPR}\) can be as high as 80\% whereas it can only reach 50\%
for the red.}
\end{figure}

If we constrain~\(\mathrm{FPR}\) to be less than 10\%, for the blue
curve, \(\mathrm{TPR}\) can be as high as 80\% whereas it can only reach
50\% for the red. AUC should be always viewed skeptically: the shape of
an ROC curve is always more informative than any individual number.

\hypertarget{decisions-that-discriminate}{%
\section{Decisions that
discriminate}\label{decisions-that-discriminate}}

The purpose of prediction is almost always decision making. We build
predictors to guide our decision making by acting on our predictions.
Many decisions entail a life changing event for the individual. The
decision could grant access to a major opportunity, such as college
admission, or deny access to a vital resource, such as a social
benefit.\index{decision making}

Binary decision rules always draw a boundary between one group in the
population and its complement. Some are labeled \emph{accept}, others
are labeled \emph{reject}. When decisions have serious consequences for
the individual, however, this decision boundary is not just a technical
artifact. Rather it has moral and legal significance.

The decision maker often has access to data that encode an individual's
status in socially salient groups relating to race, ethnicity, gender,
religion, or disability status. These and other categories that have
been used as the basis of adverse treatment, oppression, and denial of
opportunity in the past and in many cases to this day.

Some see formal or algorithmic decision making as a neutral mathematical
tool. However, numerous scholars have shown how formal models can
perpetuate existing inequities and cause harm. In her book on this
topic, Ruha Benjamin warns of

\begin{quote}
the employment of new technologies that reflect and reproduce existing
inequities but that are promoted and perceived as more objective or
progressive than the discriminatory systems of a previous
era.\citep{benjamin2019race}
\end{quote}

Even though the problems of inequality and injustice are much broader
than one of formal decisions, we already encounter an important and
challenging facet within the narrow formal setup of this chapter.
Specifically, we are concerned with decision rules that
\emph{discriminate} in the sense of creating an unjustified basis of
differentiation between individuals.\index{discrimination}

A concrete example is helpful. Suppose we want to accept or reject
individuals for a job. Suppose we have a perfect estimate of the number
of hours an individual is going to work in the next 5 years. We decide
that this a reasonable measure of productivity and so we accept every
applicant where this number exceeds a certain threshold. On the face of
it, our rule might seem neutral. However, on closer reflection, we
realize that this decision rule systematically disadvantages individuals
who are more likely than others to make use of their parental leave
employment benefit that our hypothetical company offers. We are faced
with a conundrum. On the one hand, we trust our estimate of
productivity. On the other hand, we consider taking parental leave
\emph{morally irrelevant} to the decision we're making. It should not be
a disadvantage to the applicant. After all that is precisely the reason
why the company is offering a parental leave benefit in the first place.

The simple example shows that statistical accuracy alone is no safeguard
against discriminatory decisions. It also shows that ignoring
\emph{sensitive attributes} is no safeguard either. So what then is
\emph{discrimination} and how can we avoid it? This question has
occupied scholars from numerous disciplines for decades. There is no
simple answer. Before we go into attempts to formalize discrimination in
our statistical decision making setting, it is helpful to take a step
back and reflect on what the law says.

\hypertarget{legal-background-in-the-united-states}{%
\subsection{Legal background in the United
States}\label{legal-background-in-the-united-states}}

The legal frameworks governing decision making differ from country to
country, and from one domain to another. We take a glimpse at the
situation in the United States, bearing in mind that our description is
incomplete and does not transfer to other countries.

Discrimination is not a general concept. It is concerned with socially
salient categories that have served as the basis for unjustified and
systematically adverse treatment in the past. United States law
recognizes certain \emph{protected categories} including race, sex
(which extends to sexual orientation), religion, disability status, and
place of birth.

Further, discrimination is a domain specific concept concerned with
important opportunities that affect people's lives. Regulated domains
include credit (Equal Credit Opportunity Act), education (Civil Rights
Act of 1964; Education Amendments of 1972), employment (Civil Rights Act
of 1964), housing (Fair Housing Act), and \emph{public accommodation}
(Civil Rights Act of 1964). Particularly relevant to machine learning
practitioners is the fact that the scope of these regulations extends to
marketing and advertising within these domains. An ad for a credit card,
for example, allocates access to credit and would therefore fall into
the credit domain.

There are different legal frameworks available to a plaintiff that
brings forward a case of discrimination. One is called \emph{disparate
treatment}, the other is \emph{disparate impact}. Both capture different
forms of discrimination. Disparate treatment is about purposeful
consideration of group membership with the intention of discrimination.
Disparate impact is about unjustified harm, possibly through indirect
mechanisms. Whereas disparate treatment is about \emph{procedural
fairness}, disparate impact is more about \emph{distributive
justice}.\index{disparate impact}\index{disparate treatment}

It's worth noting that anti-discrimination law does not reflect one
overarching moral theory. Pieces of legislation often came in response
to civil rights movements, each hard fought through decades of activism.

Unfortunately, these legal frameworks don't give us a formal definition
that we could directly apply. In fact, there is some well-recognized
tension between the two doctrines.

\hypertarget{formal-non-discrimination-criteria}{%
\subsection{Formal non-discrimination
criteria}\label{formal-non-discrimination-criteria}}

The idea of formal non-discrimination (or \emph{fairness}) criteria goes
back to pioneering work of Anne Cleary and other researchers in the
educational testing community of the
1960s.\citep{hutchinson201950}\index{fairness}

The main idea is to introduce a discrete random variable~\(A\) that
encodes membership status in one or multiple protected classes.
Formally, this random variable lives in the same probability space as
the other covariates~\(X\), the decision~\(\hat Y=\mathbb{1}\{ R > t\}\)
in terms of a score~\(R\), and the outcome~\(Y\). The random
variable~\(A\) might coincide with one of the features in~\(X\) or
correlate strongly with some combination of them.

Broadly speaking, different statistical fairness criteria all equalize
some group-dependent statistical quantity across groups defined by the
different settings of~\(A\). For example, we could ask to equalize
acceptance rates across all groups. This corresponds to imposing the
constraint for all groups~\(a\) and~\(b\):

\[
\Pr[\hat Y = 1 \mid A=a] = \Pr[\hat Y = 1 \mid A=b]
\]

Researchers have proposed dozens of different criteria, each trying to
capture different intuitions about what is \emph{fair}. Simplifying the
landscape of fairness criteria, we can say that there are essentially
three fundamentally different ones of particular significance:

\begin{itemize}
\tightlist
\item
  Acceptance rate \(\Pr[\hat Y = 1]\)
\item
  Error rates \(\Pr[\hat Y = 0 \mid Y = 1]\) and
  \(\Pr[\hat Y = 1 \mid Y =0]\)
\item
  Outcome frequency given score value \(\Pr[Y = 1 \mid R = r ]\)
\end{itemize}

The meaning of the first two as a formal matter is clear given what we
already covered. The third criterion needs a bit more motivation. A
useful property of score functions is \emph{calibration} which asserts
that \(\Pr[Y = 1\mid R=r]=r\) for all score values~\(r\). In words, we
can interpret a score value~\(r\) as the propensity of positive outcomes
among instances assigned the score value~\(r\). What the third criterion
says is closely related. We ask that the score values have the same
meaning in each group. That is, instances labeled~\(r\) in one group are
equally likely to be positive instances as those scored~\(r\) in any
other group.

The three criteria can be generalized and simplified using three
different conditional independence statements.

\begin{longtable}[]{@{}ccc@{}}
\caption{Non-discrimination criteria}\tabularnewline
\toprule
Independence & Separation & Sufficiency \\
\midrule
\endfirsthead
\toprule
Independence & Separation & Sufficiency \\
\midrule
\endhead
\(R\bot A\) & \(R\bot A \mid Y\) & \(Y\bot A\mid R\) \\
\bottomrule
\end{longtable}

Each of these applies not only to binary prediction, but any set of
random variables where the independence statement holds. It's not hard
to see that independence implies equality of acceptance rates across
groups. Separation implies equality of error rates across groups. And
sufficiency implies that all groups have the same rate of positive
outcomes given a score value.\citep{barocas-hardt-narayanan}

Researchers have shown that any two of the three criteria are
\emph{mutually exclusive} except in special cases. That means, generally
speaking, imposing one criterion forgoes the other
two.\citep{kleinberg2017inherent, chouldechova2017fair}

Although these formal criteria are easy to state and arguably natural in
the language of decision theory, their merit as measures of
discrimination has been subject of an ongoing debate.

\hypertarget{merits-and-limitations-of-a-narrow-statistical-perspective}{%
\subsection{Merits and limitations of a narrow statistical
perspective}\label{merits-and-limitations-of-a-narrow-statistical-perspective}}

The tension between these criteria played out in a public debate around
the use of risk scores to predict \emph{recidivism} in pre-trial
detention decisions.

There's a risk score, called COMPAS, used by many jurisdictions in the
United States to assess \emph{risk of recidivism} in pre-trial bail
decisions. Recidivism refers to a person's relapse into criminal
behavior. In the United States, a defendant may either be detained or
released on bail prior to the trial in court depending on various
factors. Judges may detain defendants in part based on this
score.\index{COMPAS}

Investigative journalists at ProPublica found that Black defendants face
a higher false positive rate, i.e., more Black defendants labeled
\emph{high risk} end up not committing a crime upon release than among
White defendants labeled \emph{high risk}.\citep{angwin2016machine} In
other words, the COMPAS score fails the separation criterion.

\index{ProPublica}

A company called Northpointe, which sells the proprietary COMPAS risk
model, pointed out in return that Black and White defendants have equal
recidivism rates \emph{given} a particular score value. That is
defendants labeled, say, an `8' for \emph{high risk} would go on to
recidivate at a roughly equal rate in either group. Northpointe claimed
that this property is desirable so that a judge can interpret scores
equally in both groups.\citep{dieterich16compas}

The COMPAS debate illustrates both the merits and limitations of the
narrow framing of discrimination as a classification criterion.

On the hand, the error rate disparity gave ProPublica a tangible and
concrete way to put pressure on Northpointe. The narrow framing of
decision making identifies the decision maker as responsible for their
decisions. As such, it can be used to interrogate and possibly intervene
in the practices of an entity.

On the other hand, decisions are always part of a broader system that
embeds structural patterns of discrimination. For example, a measure of
recidivism hinges crucially on existing policing patterns. Crime is only
found where policing activity happens. However, the allocation and
severity of police force itself has racial bias. Some scholars therefore
find an emphasis on statistical criteria rather than structural
determinants of discrimination to be limited.

\hypertarget{chapter-notes-1}{%
\section{Chapter notes}\label{chapter-notes-1}}

The theory we covered in this chapter is also called \emph{detection
theory} and \emph{decision theory}.\index{detection theory} Similarly,
what we call a predictor throughout has various different names, such as
\emph{decision rule} or
\emph{classifier}.\index{decision!rule}\index{classifier}\index{detection theory}\index{decision}

The elementary detection theory covered in this chapter has not changed
much at all since the 1950s and is essentially considered a ``solved
problem''. Neyman and Pearson invented the likelihood ratio
test\citep{neyman1928use} and later proved their lemma showing it to be
optimal for maximizing true positive rates while controlling false
positive rates.\citep{neyman1933} Wald followed this work by inventing
general Bayes risk minimization in 1939.\citep{wald1939} Wald's ideas
were widely adopted during World War II for the purpose of interpreting
RADAR signals which were often very noisy. Much work was done to improve
RADAR operations, and this led to the formalization that the output of a
RADAR system (the receiver) should be a likelihood ratio, and a decision
should be made based on an LRT. Our proof of Neyman-Pearson's lemma came
later, and is due to Bertsekas and Tsitsiklis (See Section 9.3 of
\emph{Introduction to Probability}\citep{bt-probability-book}).

Our current theory of detection was fully developed by Peterson,
Birdsall, and Fox in their report on optimal signal
detectability.\citep{peterson1954} Peterson, Birdsall, and Fox may have
been the first to propose Receiver Operating Characteristics as the
means to characterize the performance of a detection system, but these
ideas were contemporaneously being applied to better understand
psychology and psychophysics as well.\citep{tanner1954}

Statistical Signal Detection theory was adopted in the pattern
recognition community at a very early stage. Chow proposed using optimal
detection theory,\citep{chow1957optimum} and this led to a proposal by
Highleyman to approximate the risk by its sample
average.\citep{highleyman1962linear} This transition from population
risk to ``empirical'' risk gave rise to what we know today as machine
learning.

Of course, how decisions and predictions are applied and interpreted
remains an active research topic. There is a large amount of literature
now on the topic of fairness and machine learning. For a general
introduction to the problem and dangers associated with algorithmic
decision making not limited to discrimination, see the books by
Benjamin\citep{benjamin2019race},
Broussard\citep{broussard2018artificial},
Eubanks\citep{eubanks2018automating}, Noble\citep{noble2018algorithms},
and O'Neil\citep{oneil2016weapons}. The technical material in our
section on discrimination follows Chapter 2 in the textbook by Barocas,
Hardt, and Narayanan.\citep{barocas-hardt-narayanan}

The abalone example was derived from data available at the UCI Machine
Learning Repository, which we will discuss in more detail in Chapter 8.
We modified the data to ease exposition. The actual data does not have
an equal number of male and female instances, and the optimal predictor
is not exactly a threshold function.

\chapter{Supervised learning}

Previously, we talked about the fundamentals of prediction and
statistical modeling of populations. Our goal was, broadly speaking, to
use available information described by a random variable~\(X\) to
conjecture about an unknown outcome~\(Y\).

In the important special case of a binary outcome~\(Y\), we saw that we
can write an optimal predictor~\(\hat Y\) as a threshold of some
function \(f\): \[
\hat Y(x) = \mathbb{1}\{f(x) > t\}
\] We saw that in many cases the optimal function is a ratio of two
likelihood functions.

This optimal predictor has a serious limitation in practice, however. To
be able to compute the prediction for a given input, we need to know a
probability density function for the positive instances in our problem
and also one for the negative instances. But we are often unable to
construct or unwilling to assume a particular density function.

As a thought experiment, attempt to imagine what a probability density
function over images labeled \emph{cat} might look like. Coming up with
such a density function appears to be a formidable task, one that's not
intuitively any easier than merely classifying whether an image contains
a cat or not.

In this chapter, we transition from a purely mathematical
characterization of optimal predictors to an algorithmic framework. This
framework has two components. One is the idea of working with finite
samples from a population. The other is the theory of supervised
learning and it tells us how to use finite samples to build predictors
algorithmically.

\hypertarget{sample-versus-population}{%
\section{Sample versus population}\label{sample-versus-population}}

Let's take a step back to reflect on the interpretation of the pair of
random variables~\((X, Y)\) that we've worked with so far. We think of
the random variables~\((X, Y)\) as modeling a population of instances in
our prediction problem. From this pair of random variables, we can
derive other random variables such as a predictor
\(\hat Y = \mathbb{1}\{f(X) > t\}\). All of these are random variables
in the same probability space. When we talk about, say, the true
positive rate of the predictor~\(\hat Y\), we therefore make a statement
about the joint distribution of~\((X, Y)\).

In almost all prediction problems, however, we do not have access to the
entire population of instances that we will encounter. Neither do we
have a probability model for the joint distribution of the random
variables~\((X, Y)\). The joint distribution is a theoretical construct
that we can reason about, but it doesn't readily tell us what to do when
we don't have precise knowledge of the joint distribution.

What knowledge then do we typically have about the underlying population
and how can we use it algorithmically to find good predictors? In this
chapter we will begin to answer both questions.

First we assume that from past experience we have observed~\(n\) labeled
instances~\((x_1, y_1),...,(x_n, y_n)\). We assume that each data point
\((x_i, y_i)\) is a draw from the same underlying
distribution~\((X, Y)\). Moreover, we will often assume that the data
points are drawn independently. This pair of assumptions is often called
the ``i.i.d. assumption'', a shorthand for \emph{independent and
identically
distributed}.\index{i.i.d.}\index{independent and identically distributed}

To give an example, consider a population consisting of all currently
eligible voters in the United States and some of their features, such
as, age, income, state of residence etc. An \emph{i.i.d. sample} from
this population would correspond to a repeated sampling process that
selects a uniformly random voter from the entire reference population.

Sampling is a difficult problem with numerous pitfalls that can strongly
affect the performance of statistical estimators and the validity of
what we learn from data. In the voting example, individuals might be
unreachable or decline to respond. Even defining a good population for
the problem we're trying to solve is often tricky. Populations can
change over time. They may depend on a particular social context,
geography, or may not be neatly characterized by formal criteria. Task
yourself with the idea of taking a random sample of spoken sentences in
the English language, for example, and you will quickly run into these
issues.

In this chapter, as is common in learning theory, we largely ignore
these important issues. We instead focus on the significant challenges
that remain even if we have a well-defined population and an unbiased
sample from it.

\hypertarget{supervised-learning}{%
\section{Supervised learning}\label{supervised-learning}}

\emph{Supervised learning} is the prevalent method for constructing
predictors from data. The essential idea is very simple. We assume we
have labeled data, in this context also called \emph{training examples},
of the form~\((x_1,y_1), ..., (x_n, y_n),\) where each \emph{example} is
a pair \((x_i,y_i)\) of an \emph{instance} \(x_i\) and a corresponding
\emph{label} \(y_i.\) The notion of \emph{supervision} refers to the
availability of these labels.

Given such a collection of labeled data points, supervised learning
turns the task of finding a good predictor into an optimization problem
involving these data points. This optimization problem is called
\emph{empirical risk minimization}. \index{empirical risk!minimization}

Recall, in the last chapter we assumed full knowledge of the joint
distribution of~\((X,Y)\) and analytically found predictors that
minimize risk. The risk is equal to the expected value of a \emph{loss
function} that quantifies the cost of each possible prediction for a
given true outcome. For binary prediction problems, there are four
possible pairs of labels corresponding to true positives, false
positives, true negatives, and false negatives. In this case, the loss
function boils down to specifying a cost to each of the four
possibilities.

More generally, a loss function is a function
\(\loss\colon\mathcal{Y}\times\mathcal{Y}\to\mathbb{R}\,,\) where
\(\mathcal{Y}\) is the set of values that~\(Y\) can assume. Whereas
previously we focused on the predictor~\(\hat Y\) as a random variable,
in this chapter our focus shifts to the functional form that the
predictor has. By convention, we write~\(\hat Y=f(X),\) where
\(f\colon\mathcal{X}\to\mathcal{Y}\) is a function that maps from the
sample space~\(\mathcal{X}\) into the label space~\(\mathcal{Y}.\)

Although the random variable~\(\hat Y\) and the function~\(f\) are
mathematically not the same objects, we will call both a predictor and
extend our risk definition to apply the function as
well:\index{predictor} \[
R[f] = \mathop\mathbb{E}\left[ \loss (f(X), Y) \right]\,.
\] The main new definition in this chapter is a finite sample analog of
the risk, called empirical risk.

\begin{Definition}

Given a set of labeled data points~\(S=((x_1,y_1),...,(x_n, y_n))\), the
\emph{empirical risk}\index{empirical risk}\index{risk!empirical} of a
predictor~\(f\colon \mathcal{X}\to\mathcal{Y}\) with respect to the
sample \(S\) is defined as \[
R_S[f] = \frac1n \sum_{i=1}^n \mathbb{\loss}( f(x_i), y_i )\,.
\]

\end{Definition}

Rather than taking expectation over the population, the empirical risk
averages the loss function over a finite sample. Conceptually, we think
of the finite sample as something that is in our possession, e.g.,
stored on our hard disk.

Empirical risk serves as a proxy for the risk. Whereas the risk~\(R[f]\)
is a population quantity---that is, a property of the joint distribution
\((X, Y)\) and our predictor~\(f\)---the empirical risk is a
\emph{sample quantity}.

We can think of the empirical risk as the sample average estimator of
the risk. When samples are drawn i.i.d., the empirical risk is a random
variable that equals the sum of~\(n\) independent random variables. If
losses are bounded, the central limit theorem therefore suggests that
the empirical risk approximates the risk for a fixed predictor~\(f\).

Regardless of the distribution of~\(S\), however, note that we can
always compute the empirical risk~\(R_S[f]\) entirely from the
sample~\(S\) and the predictor~\(f\). Since empirical risk a quantity we
can compute from samples alone, it makes sense to turn it into an
objective function that we can try to minimize numerically.

\emph{Empirical risk minimization} is the optimization problem of
finding a predictor in a given function family that minimizes the
empirical risk.\index{empirical risk!minimization}

\begin{Definition}

Given a function
class~\(\mathcal{F}\subseteq \mathcal{X}\to\mathcal{Y},\)
\emph{empirical risk minimization} on a set of labeled data points~\(S\)
corresponds to the objective: \[
\min_{f\in\mathcal{F}} R_S[f]
\] A solution to the optimization problem is called \emph{empirical risk
minimizer}.

\end{Definition}

There is a tautology relating risk and empirical risk that is good to
keep in mind: \[
R[f] = R_S[f] + (R[f] - R_S[f])
\] Although mathematically trivial, the tautology reveals an important
insight. To minimize risk, we can first attempt to minimize empirical
risk. If we successfully find a predictor~\(f\) that achieves small
empirical risk~\(R_S[f]\), we're left worrying about the term
\(R[f]-R_S[f]\). This term quantifies how much the empirical risk
of~\(f\) underestimates its risk. We call this difference
\emph{generalization gap} and it is of fundamental importance to machine
learning. Intuitively speaking, it tells us how well the performance of
our predictor transfers from seen examples (the training examples) to
unseen examples (a fresh example from the population) drawn from the
same distribution. This process is called
\emph{generalization}.\index{generalization}

Generalization is not the only goal of supervised learning. A constant
predictor that always outputs~\(0\) generalizes perfectly well, but is
almost always entirely useless. What we also need is that the predictor
achieves small empirical risk~\(R_S[f]\). Making the empirical risk
small is fundamentally about \emph{optimization}. As a consequence, a
large part of supervised learning deals with optimization. For us to be
able to talk about optimization, we need to commit to a
\emph{representation} of the function class~\(\mathcal{F}\) that appears
in the empirical risk minimization problem. The representation of the
function class, as well as the choice of a suitable loss function,
determines whether or not we can efficiently find an empirical risk
minimizer.

To summarize, introducing empirical risk minimization directly leads to
three important questions that we will work through in turn.

\begin{itemize}
\tightlist
\item
  \textbf{Representation:} What is the class of functions
  \(\mathcal{F}\) we should choose?
\item
  \textbf{Optimization:} How can we efficiently solve the resulting
  optimization problem?
\item
  \textbf{Generalization:} Will the performance of predictor transfer
  gracefully from seen training examples to unseen instances of our
  problem?
\end{itemize}

These three questions are intertwined. Machine learning is not so much
about studying these questions in isolation as it is about the often
delicate interplay between them. Our choice of representation influences
both the difficulty of optimization and our generalization performance.
Improvements in optimization may not help, or could even hurt,
generalization. Moreover, there are aspects of the problem that don't
neatly fall into only one of these categories. The choice of the loss
function, for example, affects all of the three questions above.

There are important differences between the three questions. Results in
optimization, for example, tend to be independent of the statistical
assumptions about the data generating process. We will see a number of
different optimization methods that under certain circumstances find
either a global or local minimum of the empirical risk objective. In
contrast, to reason about generalization, we need some assumptions about
the data generating process. The most common one is the
i.i.d.-assumption we discussed earlier. We will also see several
mathematical frameworks for reasoning about the gap between risk and
empirical risk.

Let's start with a foundational example that illustrates these core
concepts and their interplay.

\hypertarget{a-first-learning-algorithm-the-perceptron}{%
\section{A first learning algorithm: The
perceptron}\label{a-first-learning-algorithm-the-perceptron}}

As we discussed in the introduction, in 1958 the
\href{https://www.nytimes.com/1958/07/08/archives/new-navy-device-learns-by-doing-psychologist-shows-embryo-of.html}{New
York Times} reported the Office of Naval Research claiming the
perceptron algorithm \citep{rosenblatt58theperceptron} would ``be able
to walk, talk, see, write, reproduce itself and be conscious of its
existence.'' Let's now dive into this algorithm that seemed to have such
unbounded potential.

Toward introducing this algorithm, let's assume we're in a binary
prediction problem with labels in~\(\{-1,1\}\) for notational
convenience. The perceptron algorithm aims to find a \emph{linear
separator}\index{linear separator} of the data, that is, a hyperplane
specified by coefficients~\(w\in\mathbb{R}^d\) that so that all positive
examples lie on one side of the hyperplane and all negative ones on the
other.

\begin{figure}
\centering
\includegraphics[width=0.4\textwidth,height=\textheight]{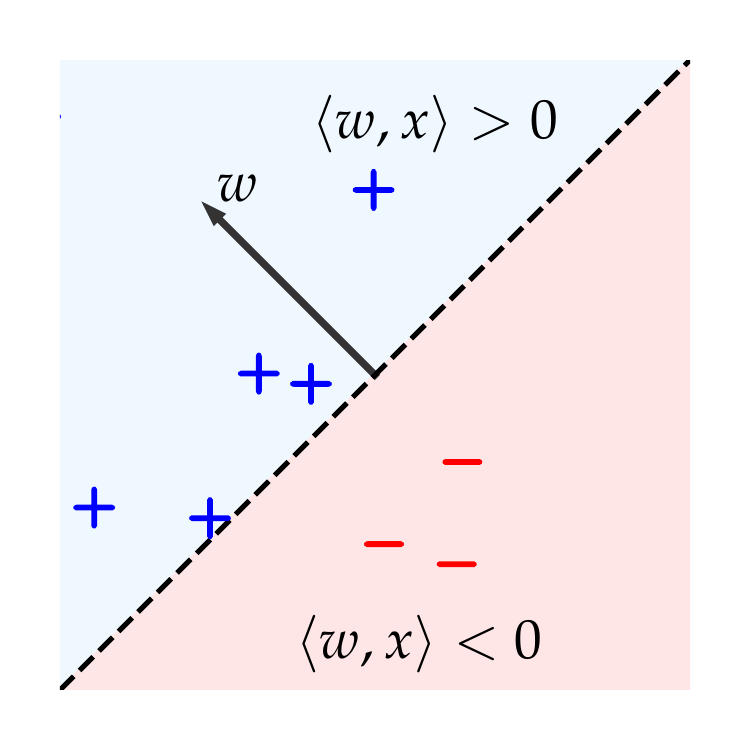}
\caption{Illustration of a linear separator}
\end{figure}

Formally, we can express this as~\(y_i\langle w, x_i\rangle > 0.\) In
other words, the linear function~\(f(x)=\langle w, x\rangle\) agrees in
sign with the labels on all training instances~\((x_i, y_i)\). In fact,
the perceptron algorithm will give us a bit more. Specifically, we
require that the sign agreement has some \emph{margin}
\(y_i\langle w, x_i\rangle \ge 1.\) That is, when~\(y=1,\) the linear
function must take on a value of at least~\(1\) and when~\(y=-1\), the
linear function must be at most~\(-1\). Once we find such a linear
function, our prediction~\(\hat Y(x)\) on a data point~\(x\) is
\(\hat{Y}(x) = 1\) if~\(\langle w, x \rangle \geq 0\)
and~\(\hat Y(x) = -1\) otherwise.

The algorithm goes about finding a linear separator~\(w\) incrementally
in a sequence of update steps.

\begin{Algorithm}

\textbf{Perceptron}

\begin{itemize}
\tightlist
\item
  Start from the initial solution \(w_0=0\)
\item
  At each step \(t=0,1,2,...\):

  \begin{itemize}
  \tightlist
  \item
    Select a random index \(i\in\{1,...,n\}\)
  \item
    Case 1: If \(y_i\langle w_t, x_i\rangle < 1\), put \[
    w_{t+1} = w_t + y_i x_i  
    \]
  \item
    Case 2: Otherwise put \(w_{t+1} = w_t\).
  \end{itemize}
\end{itemize}

\end{Algorithm}

Case 1 corresponds to what's called a \emph{margin mistake}. The sign of
the linear function may not disagree with the label, but it doesn't have
the required margin that we asked for.

\begin{figure}
\centering
\includegraphics{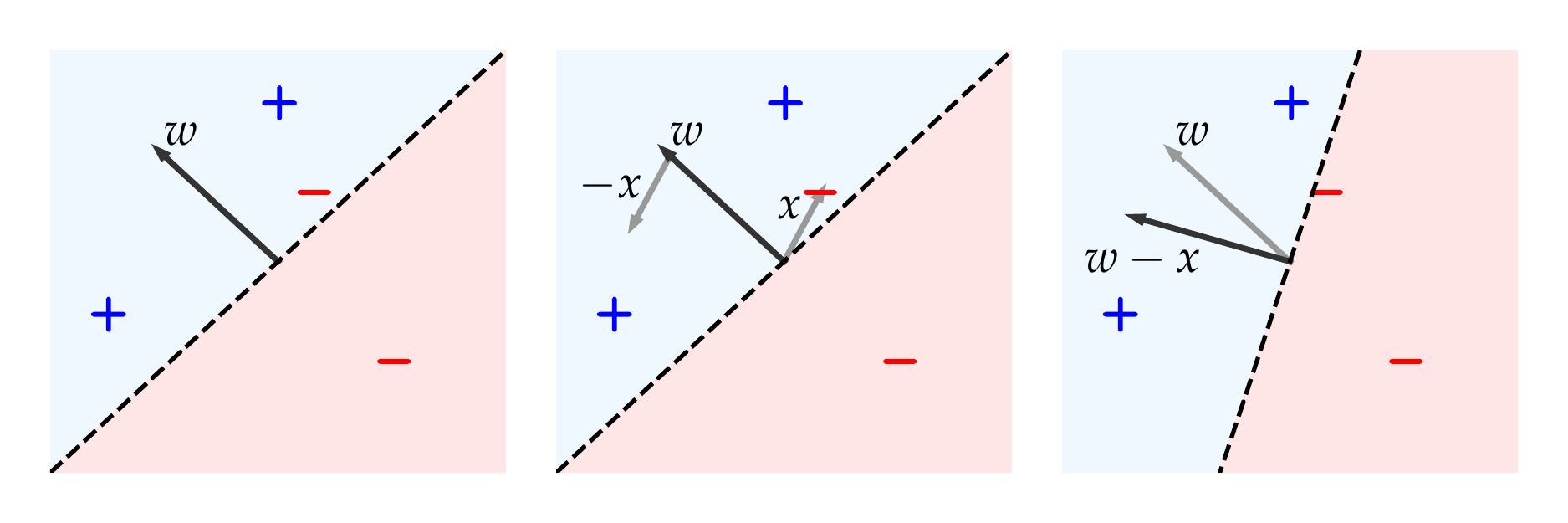}
\caption{Illustration of the perceptron update. Left: One misclassified
example \(x.\) Right: After update.}
\end{figure}

When an update occurs, we have \[
\langle w_{t+1}, x_i\rangle = \langle w_t, x_i\rangle + y_i\|x_i\|^2\,.
\] In this sense, the algorithm is nudging the hyperplane to be less
wrong on example~\(x_i\). However, in doing so it could introduce errors
on other examples. It is not yet clear that the algorithm converges to a
linear separator when this is possible.

\hypertarget{connection-to-empirical-risk-minimization}{%
\section{Connection to empirical risk
minimization}\label{connection-to-empirical-risk-minimization}}

Before we turn to the formal guarantees of the perceptron, it is
instructive to see how to relate it to empirical risk minimization. In
order to do so, it's helpful to introduce two \emph{hyperparameters} to
the algorithm by considering the alternative update rule: \[
w_{t+1} = \gamma w_t + \eta y_i x_i
\] Here~\(\eta\) is a positive scalar called a \emph{learning rate} and
\(\gamma\in [0,1]\) is called the \emph{forgetting rate}.

First, it's clear from the description that we're looking for a linear
separator. Hence, our function class is the set of linear functions
\(f_w(x)=\langle w, x\rangle,\) where~\(w\in\mathbb{R}^d\). We will
sometimes call the vector~\(w\) the \emph{weight vector} or vector of
\emph{model parameters}.

An optimization method that picks a random example at each step and
makes a local improvement to the model parameters is the
\emph{stochastic gradient method}.\index{stochastic gradient method}
This method will figure prominently in our chapter on optimization as it
is the workhorse of many machine learning applications today. The local
improvement the method picks at each step is given by a local linear
approximation of the loss function around the current model parameters.
This linear approximation can be written neatly in terms of the vector
of first derivatives, called \emph{gradient}, of the loss function with
respect to the current model parameters.\index{gradient}

The formal update rule reads \[
w_{t+1} = w_t - \eta\nabla_{w_t} \loss(f_{w_t}(x_i), y_i)
\] Here, the example~\((x_i,y_i)\) is randomly chosen and the expression
\(\nabla_{w_t} \loss(f_{w_t}(x_i), y_i)\) is the gradient of the loss
function with respect to the model parameters~\(w_t\) on the example
\((x_i, y_i)\). We will typically drop the vector~\(w_t\) from the
subscript of the gradient when it's clear from the context. The
scalar~\(\eta>0\) is a step size parameter that we will discuss more
carefully later. For now, think of it as a small constant.

It turns out that we can connect this update rule with the perceptron
algorithm by choosing a suitable loss function. Consider the loss
function\index{loss!hinge} \[
\loss(\langle w, x\rangle, y)
= \max\big\{1-y\langle w, x\rangle,\, 0\big\}\,.
\] This loss function is called \emph{hinge loss}. Note that its
gradient is \(-yx\) when~\(y\langle w, x\rangle < 1\) and~\(0\) when
\(y\langle w, x\rangle >1\).\index{loss!hinge}

\begin{figure}
\centering
\includegraphics[width=0.5\textwidth,height=\textheight]{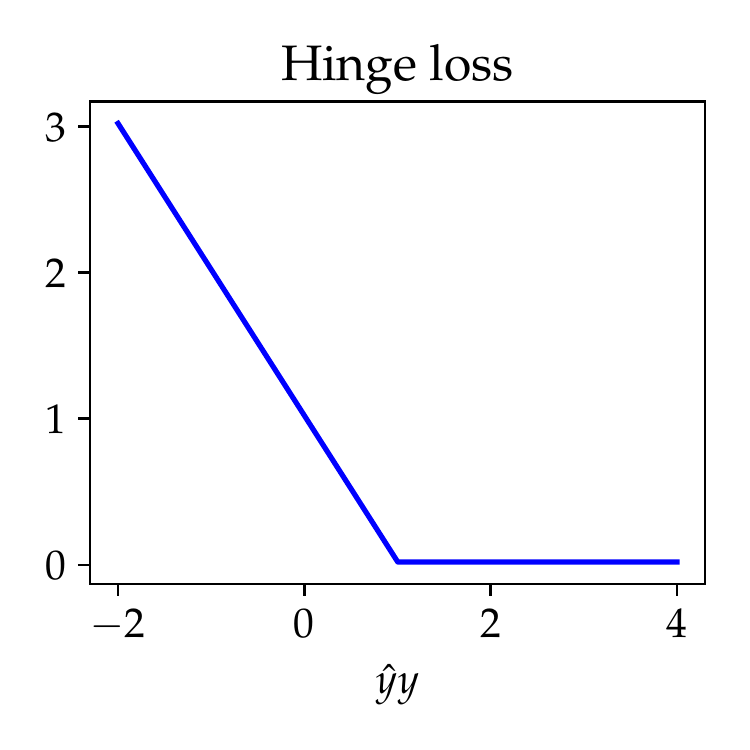}
\caption{Hinge loss}
\end{figure}

The gradient of the hinge loss is not defined when
\(y\langle w, x\rangle=1.\) In other words, the loss function is not
differentiable everywhere. This is why technically speaking the
stochastic gradient method operates with what is called a
\emph{subgradient}. The mathematical theory of subgradient optimization
rigorously justifies calling the gradient~\(0\)
when~\(y\langle w, x\rangle=1.\) We will ignore this technicality
throughout the book.\index{subgradient}

We can see that the hinge loss gives us part of the update rule in the
perceptron algorithm. The other part comes from adding a weight penalty
\(\frac\lambda 2\Vert w\Vert^2\) to the loss function that discourages
the weights from growing out of bounds. This weight penalty is called
\emph{\(\ell_2\)-regularization}, \emph{weight decay}, or \emph{Tikhonov
regularization} depending on which field you work in. The purpose of
regularization is to promote generalization. We will therefore return to
regularization in detail when we discuss generalization in more depth.
For now, note that the margin constraint we introduced is
inconsequential unless we penalize large vectors. Without the weight
penalty we could simply scale up any linear separator until it separates
the points with the desired
margin.\index{regularization}\index{weight decay}

Putting the two loss functions together, we get
the~\(\ell_2\)-regularized empirical risk minimization problem for the
hinge loss: \[
\frac{1}{n}\sum_{i=1}^n \max\big\{1-y_i\langle w, x_i\rangle,\, 0\big\} + \frac\lambda 2 \| w \|_2^2
\] The perceptron algorithm corresponds to solving this empirical risk
objective with the stochastic gradient method. The constant~\(\eta\),
which we dubbed the learning rate, is the step size of the stochastic
gradient methods. The forgetting rate constant~\(\gamma\) is equal to
\((1-\eta \lambda)\). The optimization problem is also known as
\emph{support vector machine} and we will return to it later on.

\hypertarget{a-word-about-surrogate-losses}{%
\subsection{A word about surrogate
losses}\label{a-word-about-surrogate-losses}}

\index{loss!zero-one}\index{loss!surrogate}

When the goal was to maximize the accuracy of a predictor, we
mathematically solved the risk minimization problem with respect to the
\emph{zero-one loss} \[
\loss(\hat y, y)=\mathbb{1}\{\hat  y\ne y\}
\] that gives us penalty~\(1\) if our label is incorrect, and
penalty~\(0\) if our predicted label~\(\hat y\) matches the true
label~\(y\). We saw that the optimal predictor in this case was a
\emph{maximum a posteriori} rule, where we selected the label with
higher posterior probability.

Why don't we directly solve empirical risk minimization with respect to
the zero-one loss? The reason is that the empirical risk with the
zero-one loss is computationally difficult to optimize directly. In
fact, this optimization problem is NP-hard even for linear prediction
rules.\citep{kearns1994toward} To get a better sense of the difficulty,
convince yourself that the stochastic gradient method, for example,
fails entirely on the zero-one loss objective. Of course, the stochastic
gradient method is not the only learning algorithm.

The hinge loss therefore serves as a \emph{surrogate loss} for the
zero-one loss. We hope that by optimizing the hinge loss, we end up
optimizing the zero-one loss as well. The hinge loss is not the only
reasonable choice. There are numerous loss functions that approximate
the zero-one loss in different ways.\index{loss!surrogate}

\begin{itemize}
\tightlist
\item
  The \emph{hinge loss} is \(\max\{1-y\hat y, 0\}\) and \emph{support
  vector machine} refers to empirical risk minimization with the hinge
  loss and \(\ell_2\)-regularization. This is what the perceptron is
  optimizing.
\item
  The \emph{squared loss} is given by \(\frac12(y-\hat y)^2\). Linear
  least squares regression corresponds to empirical risk minimization
  with the squared loss. \index{loss!squared}
\item
  The \emph{logistic loss} is \(-\log(\sigma(\hat y))\) when \(y=1\) and
  \(-\log(1-\sigma(\hat y))\) when \(y=-1\), where
  \(\sigma(z) = 1/(1+\exp(-z))\) is the logistic function.
  \emph{Logistic regression} corresponds to empirical risk minimization
  with the logistic loss and linear functions.
\end{itemize}

\begin{figure}
\centering
\includegraphics{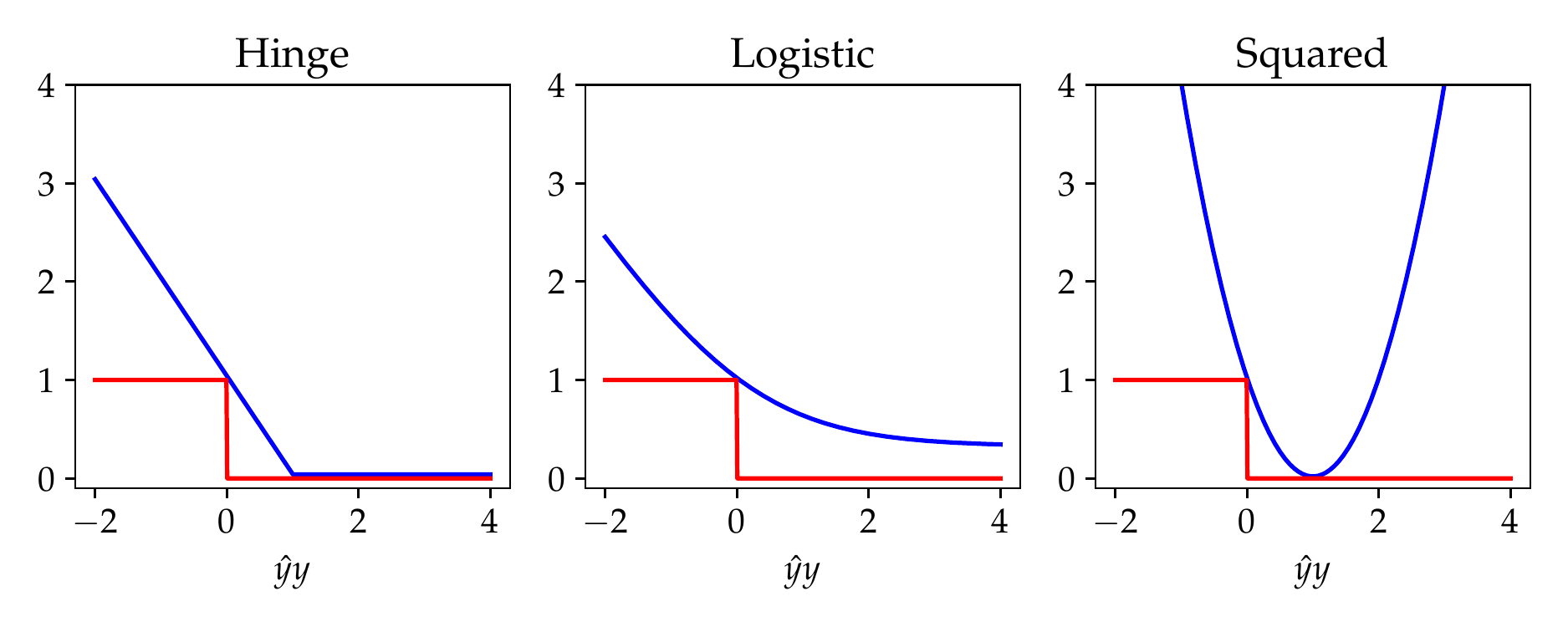}
\caption{Hinge, squared, logistic loss compared with the zero-one loss.}
\end{figure}

Sometimes we can theoretically relate empirical risk minimization under
a surrogate loss to the zero-one loss. In general, however, these loss
functions are used heuristically and practitioners evaluate performance
by trial-and-error.

\hypertarget{formal-guarantees-for-the-perceptron}{%
\section{Formal guarantees for the
perceptron}\label{formal-guarantees-for-the-perceptron}}

We saw that the perceptron corresponds to finding a linear predictor
using the stochastic gradient method. What we haven't seen yet is a
proof that the perceptron method works and under what conditions. Recall
that there are two questions we need to address. The first is why the
perceptron method successfully fits the training data, a question about
\emph{optimization}. The second is why the solution should also
correctly classify unseen examples drawn from the same distribution, a
question about \emph{generalization}. We will address each in turn with
results from the 1960s. Even though the analysis here is over 50 years
old, it has all of the essential parts of more recent theoretical
arguments in machine learning.

\hypertarget{mistake-bound}{%
\subsection{Mistake bound}\label{mistake-bound}}

\index{mistake bound}

To see why we perform well on the training data, we use a \emph{mistake
bound} due to Novikoff.\citep{Novikoff1962} The bound shows that if
there exists a linear separator of the training data, then the
perceptron will find it quickly provided that the \emph{margin} of the
separating hyperplane isn't too small.

Margin is a simple way to evaluate how well a predictor separates data.
Any vector~\(w\in \R^d\) defines a hyperplane
\(\mathcal{H}_w = \{x~:~w^Tx=0\}\). Suppose that the hyperplane
\(\mathcal{H}_w\) corresponding to the vector~\(w\) perfectly separates
the data in~\(S\). Then we define the margin of such a vector~\(w\) as
the smallest distance of our data points to this hyperplane: \[
  \gamma(S,w) = \min_{1\le i\le n} \operatorname{dist}(x_i,\mathcal{H}_w )\,.
\] Here, \[
\operatorname{dist}(x, \mathcal{H}_w) = \min\{ \|x - x'\| \colon x'\in \mathcal{H}_w\}
= \frac{|\langle x, w\rangle|}{\|w\|}.
\]

Overloading terminology, we define the margin of a \emph{dataset} to be
the maximum margin achievable by any predictor~\(w\): \[
  \gamma(S) = \max_{\Vert w \Vert =1} \gamma(S,w)\,.
\] We will now show that when a dataset has a large margin, the
perceptron algorithm will find a separating hyperplane quickly.

Let's consider the simplest form of the perceptron algorithm. We
initialize the algorithm with~\(w_0=0\). The algorithm proceeds by
selecting a random index~\(i_t\) at step~\(t\) checking whether
\(y_{i_t} w_t^Tx_{i_t} < 1\). We call this condition a margin mistake,
i.e., the prediction~\(w_t^T x_{i_t}\) is either wrong or too close to
the hyperplane. If a margin mistake occurs, the perceptron performs the
update \[
    w_{t+1} = w_t + y_{i_t} x_{i_t}\,.
\] That is, we rejigger the hyperplane to be more aligned with the
signed direction of the mistake. If no margin mistake occurs, then
\(w_{t+1}=w_t\).

To analyze the perceptron we need one additional definition. Define the
diameter of a data~\(S\) to be \[
D(S) = \max_{(x,y)\in S}\|x\|\,.
\] We can now summarize a worst case analysis of the perceptron
algorithm with the following theorem.

\begin{Theorem}

The perceptron algorithm makes at most~\((2+D(S)^2)\gamma(S)^{-2}\)
margin mistakes on any sequence of examples~\(S\) that can be perfectly
classified by a linear separator.

\end{Theorem}

\begin{Proof}

The main idea behind the proof of this theorem is that since~\(w\) only
changes when you make a mistake, we can upper bound and lower
bound~\(w\) at each time a mistake is made, and then, by comparing these
two bounds, compute an inequality on the total number of mistakes.

To find an upper bound, suppose that at step~\(t\) the algorithm makes a
margin mistake. We then have the inequality: \[
\begin{aligned}
    \Vert w_{t+1} \Vert^2
    &= \Vert w_t +y_{i_t} x_{i_t}\Vert^2 \\
    & = \Vert w_t \Vert^2 + 2 y_{i_t} \langle w_t, x_{i_t}\rangle + \Vert x_{i_t} \Vert^2 \\
    &\leq \Vert w_t\Vert^2 + 2 +  D(S)^2 \,.
\end{aligned}
\] The final inequality uses the fact that
\(y_{i_t} \langle w_t, x_{i_t}\rangle<1\). Now, let~\(m_t\) denote the
total number of mistakes made by the perceptron in the first~\(t\)
iterations. Summing up the above inequality over all the mistakes we
make and using the fact that~\(\|w_0\|=0\), we get our upper bound on
the norm of~\(w_t\): \[
    \Vert w_{t} \Vert \leq \sqrt{m_t(2+ D(S)^2)}\,.
\]

Working toward a lower bound on the norm of~\(w_t,\) we will use the
following argument. Let~\(w\) be any unit vector that correctly
classifies all points in~\(S\). If we make a mistake at iteration~\(t\),
we have \[
  \langle w, w_{t+1}-w_t\rangle
  = \langle w, y_{i_t}x_{i_t}\rangle
  = \frac{|\langle w, x_{i_t}\rangle|}{\|w\|}
  \geq \gamma(S, w)\,.
\] Note that the second equality here holds because~\(w\) correctly
classifies the point~\((x_{i_t}, y_{i_t})\). This is where we use that
the data are linearly separable. The inequality follows from the
definition of margin.

Now, let~\(w_\star\) denote the hyperplane that achieves the maximum
margin~\(\gamma(S)\). Instantiating the previous argument
with~\(w_\star\), we find that \[
    \Vert w_t \Vert \geq \langle w_\star,  w_t\rangle = \sum_{k=1}^t  w_\star^T (w_{k}-w_{k-1}) \geq m_t\gamma(S)\,,
\] where the equality follows from a telescoping sum argument.

This yields the desired lower bound on the norm of~\(w_t\). Combined
with the upper bound we already derived, it follows that the total
number of mistakes has the bound \[
m_t \leq \frac{2+D(S)^2}{\gamma(S)^2}\,.\qedhere
\]

\end{Proof}

The proof we saw has some ingredients we'll encounter again. Telescoping
sums, for example, are a powerful trick used throughout the analysis of
optimization algorithms. A telescoping sum lets us understand the
behavior of the final iterate by decomposing it into the incremental
updates of the individual iterations.

The mistake bound does not depend on the dimension of the data. This is
appealing since the requirement of linear separability and high margin,
intuitively speaking, become less taxing the larger the dimension is.

An interesting consequence of this theorem is that if we run the
perceptron repeatedly over the same dataset, we will eventually end up
with a separating hyperplane. To see this, imagine repeatedly running
over the dataset until no mistake occurred on a full pass over the data.
The mistake bound gives a bound on the number of passes required before
the algorithm terminates.

\hypertarget{from-mistake-bounds-to-generalization}{%
\subsection{From mistake bounds to
generalization}\label{from-mistake-bounds-to-generalization}}

The previous analysis shows that the perceptron finds a good predictor
on the training data. What can we say about new data that we have not
yet seen?

To talk about generalization, we need to make a statistical assumption
about the data generating process. Specifically we assume that the data
points in the training set~\(S=\{(x_1,y_1)\ldots, (x_n,y_n) \}\) where
each drawn i.i.d. from a fixed underlying distribution~\(\mathcal{D}\)
with the labels taking values~\(\{ -1,1 \}\) and each~\(x_i \in \R^d\).

We know that the perceptron finds a good linear predictor for the
training data (if it exists). What we now show is that this predictor
also works on new data drawn from the same distribution.

To analyze what happens on new data, we will employ a powerful
\emph{stability} argument. Put simply, an algorithm is stable if the
effect of removing or replacing a single data point is small. We will do
a deep dive on stability in our chapter on generalization, but we will
have a first encounter with the idea here.\index{stability}

The perceptron is stable because it makes a bounded number of mistakes.
If we remove a data point where no mistake is made, the model doesn't
change at all. In fact, it's as if we had \emph{never seen} the data
point. This lets us relate the performance on seen examples to the
performance on examples in the training data on which the algorithm
never updated.

Vapnik and Chervonenkis presented the following stability argument in
their classic text from 1974, though the original argument is likely a
decade older.\citep{VapnikChervonenkis1974Book} Their main idea was to
leverage our assumption that the data are i.i.d., so we can swap the
roles of training and test examples in the analysis.

\begin{Theorem}

Let~\(S_n\) denote a training set of~\(n\) i.i.d. samples from a
distribution~\(\mathcal{D}\) that we assume has a perfect linear
separator. Let~\(w(S)\) be the output of the perceptron on a
dataset~\(S\) after running until the hyperplane makes no more margin
mistakes on~\(S\). Let~\(Z=(X,Y)\) be an additional independent sample
from~\(\mathcal{D}\). Then, the probability of making a margin mistake
on~\((X,Y)\) satisfies the upper bound \[
    \Pr[Y w(S_n)^T X < 1] \leq \frac{1}{n+1} {\E}_{S_{n+1}}\left[ \frac{2+D(S_{n+1})^2}{\gamma(S_{n+1})^2} \right]\,.
\]

\end{Theorem}

\begin{Proof}

First note that \[
\Pr[Y w^T X < 1] = \E[\mathbb{1}\{Yw^T X < 1\}]\,.
\]

Let~\(S_n=(Z_1, ... , Z_n)\) and with~\(Z_k=(X_k, Y_k)\) and put
\(Z_{n+1}=Z=(X,Y)\). Note that these~\(n+1\) random variables are i.i.d.
draws from~\(\mathcal{D}.\) As a purely analytical device, consider the
``leave-one-out set'' \[
S^{-k}=\{Z_1,\dots,Z_{k-1},Z_{k+1},...,Z_{n+1}\}\,.
\] Since the data are drawn i.i.d., running the algorithm on~\(S^{-k}\)
and evaluating it on~\(Z_k=(X_k,Y_k)\) is equivalent to running the
algorithm on~\(S_n\) and evaluating it on~\(Z_{n+1}\). These all
correspond to the same random experiment and differ only in naming. In
particular, we have \[
\Pr[Y w(S_n)^T X < 1]
= \frac1{n+1}\sum_{k=1}^{n+1} \E[\mathbb{1}\{Y_k w(S^{-k})^T X_k < 1\}]\,.
\] Indeed, we're averaging quantities that are each identical to the
left hand side. But recall from our previous result that the perceptron
makes at most \[
m=\frac{2+D((Z_1,\dots,Z_{n+1}))^2}{\gamma((Z_1,\dots,Z_{n+1}))^2}
\] margin mistakes when run on the entire sequence
\((Z_1,\dots,Z_{n+1})\). Let~\(i_1,\dots,i_m\) denote the indices on
which the algorithm makes a mistake in any of its cycles over the data.
If \(k\not\in\{i_1,\dots,i_m\}\), the output of the algorithm remains
the same after we remove the~\(k\)-th sample from the sequence. It
follows that such~\(k\) satisfy~\(Y_k w(S^{-k})X_k \geq 1\) and
therefore~\(k\) does not contribute to the summation above. The other
terms can at most contribute~\(1\) to the summation. Hence, \[
\sum_{k=1}^{n+1}\mathbb{1}\{Y_k w(S^{-k})^T X_k < 1\} \le m\,,
\] and by linearity of expectation, as we hoped to show, \[
\Pr[Y w(S_n)^T X < 1] \le \frac{\E[m]}{n+1}\,.\qedhere
\]

\end{Proof}

We can turn our mistake bounds into bounds on the empirical risk and
risk achieved by the perceptron algorithm by choosing the loss function
\(\loss(\langle w, x\rangle, y)= \mathbb{1}\{ \langle w, x\rangle y < 1\}\).
These bounds also imply bounds on the (empirical) risk with respect to
the zero-one loss, since the prediction error is bounded by the number
of margin mistakes.

\hypertarget{chapter-notes-2}{%
\section{Chapter notes}\label{chapter-notes-2}}

Rosenblatt developed the perceptron in 1957 and continued to publish on
the topic in the years that
followed.\citep{rosenblatt1958two, rosenblatt1962principles} The
perceptron project was funded by the US Office of Naval Research (ONR),
who jointly announced the project with Rosenblatt in a press conference
in 1958, that lead to the New York Times article we quoted earlier. This
development sparked significant interest in perceptrons research
throughout the 1960s.

The simple proof the mistake bound we saw is due to
Novikoff.\citep{Novikoff1962} Block is credited with a more complicated
contemporaneous proof.\citep{block1962perceptron} Minsky and Papert
attribute a simple analysis of the convergence guarantees for the
perceptron to a 1961 paper by Papert.\citep{papert1961some}

Following these developments Vapnik and Chervonenkis proved the
generalization bound for the perceptron method that we saw earlier,
relying on the kind of stability argument that we will return to in our
chapter on generalization. The proof of Theorem 2 is available in their
1974 book.\citep{VapnikChervonenkis1974Book} Interestingly, by the
1970s, Vapnik and Chervonenkis must have abandoned the stability
argument in favor of the VC-dimension.

In 1969, Minksy and Papert published their influential book
``Perceptrons: An introduction to computational
geometry''.\citep{minsky2017perceptrons} Among other results, it showed
that perceptrons fundamentally could not learn certain concepts, like,
an XOR of its input bits. In modern language, linear predictors cannot
learn parity functions. The results remain relevant in the statistical
learning community and have been extended in numerous ways. On the other
hand, pragmatic researchers realized one could just add the XOR to the
feature vector and continue to use linear methods. We will discuss such
feature engineering in the next chapter.

The dominant narrative in the field has it that Minsky and Papert's book
curbed enthusiasm for perceptron research and their multilayer
extensions, now better known as deep neural networks. In an updated
edition of their book from 1988, Minsky and Papert argue that work on
perceptrons had already slowed significantly by the time their book was
published for a lack of new results:

\begin{quote}
One popular version is that the publication of our book so discouraged
research on learning in network machines that a promising line of
research was interrupted. Our version is that progress had already come
to a virtual halt because of the lack of adequate basic theories,
{[}\ldots{]}.
\end{quote}

On the other hand, the pattern recognition community had realized that
perceptrons were just one way to implement linear predictors. Highleyman
was arguably the first to propose empirical risk minimization and
applied this technique to optical character
recognition.\citep{highleyman1962linear} Active research in the 1960s
showed how to find linear rules using linear programming
techniques.\citep{mangasarian1965linear} Work by Aizerman, Braverman and
Rozonoer developed iterative methods to fit nonlinear rules to
data.\citep{Aizerman65} All of this work was covered in depth in the
first edition of Duda and Hart, which appeared five years after
\emph{Perceptrons}.

It was at this point that the artificial intelligence community first
split from the pattern recognition community. While the artificial
intelligence community turned towards more \emph{symbolic} techniques in
1970s, work on statistical learning continued in Soviet and IEEE
journals. The modern view of empirical risk minimization, of which we
began this chapter, came out of this work and was codified by Vapnik and
Chervonenkis in the 1970s.

It wasn't until the 1980s that work on pattern recognition, and with it
the tools of the 1960s and earlier, took a stronger foothold in the
machine learning community again.\citep{langley2011changing} We will
continue this discussion in our chapter on datasets and machine learning
benchmarks, which were pivotal in the return of pattern recognition to
the forefront of machine learning research.

\chapter{Representations and features}

The starting point for prediction is the existence of a vector~\(x\)
from which we can predict the value of~\(y\). In machine learning, each
component of this vector is called a \emph{feature}. We would like to
find a set of features that are good for prediction. Where do features
come from in the first place?

In much of machine learning, the feature vector~\(x\) is considered to
be given. However, features are not handed down from first principles.
They had to be constructed somehow, often based on models that
incorporate assumptions, design choices, and human judgments. The
construction of features often follows human intuition and domain
specific expertise. Nonetheless, there are several principles and
recurring practices we will highlight in this chapter.

Feature representations must balance many demands. First, at a
population level, they must admit decision boundaries with low error
rates. Second, we must be able to optimize the empirical risk
efficiently given then current set of features. Third, the choice of
features also influences the generalization ability of the resulting
model.

There are a few core patterns in feature engineering that are used to
meet this set of requirements. First, there is the process of turning a
measurement into a vector on a computer, which is accomplished by
\emph{quantization and embedding}. Second, in an attempt to focus on the
most discriminative directions, the binned vectors are sorted relative
to their similarity to a set of likely patterns through a process of
\emph{template matching}. Third, as a way to introduce robustness to
noise or reduce and standardize the dimension of data, feature vectors
are compressed into a low, fixed dimension via \emph{histograms and
counts}. Finally, \emph{nonlinear liftings} are used to enable
predictors to approximate complex, nonlinear decision boundaries. These
processes are often iterated, and often times the feature generation
process itself is tuned on the collected data.

\hypertarget{measurement}{%
\section{Measurement}\label{measurement}}

\index{measurement}

Before we go into specific techniques and tricks of trade, it's
important to recognize the problem we're dealing with in full
generality. Broadly speaking, the first step in any machine learning
process is to numerically represent objects in the real world and their
relationships in a way that can be processed by computers.

There is an entire scientific discipline, called measurement theory,
devoted to this subject. The field of measurement theory distinguishes
between a measurement procedure and the target \emph{construct} that we
wish to
measure.\citep{hand2010measurement, hand2016measurement, bandalos2018measurement}
Physical temperature, for example, is a construct and a thermometer is a
measurement device. Mathematical ability is another example of a
construct; a math exam can be thought of as a procedure for measuring
this construct. While we take reliable measurement of certain physical
quantities for granted today, other measurement problems remain
difficult.\index{measurement}\index{construct}

It is helpful to frame feature creation as measurement. All data stems
from some measurement process that embeds and operationalizes numerous
important choices.\citep{gitelman2013raw} Measurement theory has
developed a range of techniques over the decades. In fact, many
measurement procedures themselves involve statistical models and
approximations that blur the line between what is a feature and what is
a model. Practitioners of machine learning should consult with experts
on measurement within specific domains before creating ad-hoc
measurement procedures. More often than not there is much relevant
scholarship on how to measure various constructs of interest. When in
doubt it's best to choose constructs with an established theory.

\hypertarget{human-subjects}{%
\subsection{Human subjects}\label{human-subjects}}

\index{human subjects}

The complexities of measurement are particularly apparent when our
features are about human subjects. Machine learning problems relating to
human subjects inevitably involve features that aim to quantify a
person's traits, inclinations, abilities, and qualities. Researchers
often try to get at these constructs by designing surveys, tests, or
questionnaires. However, much data about humans is collected in an
ad-hoc manner, lacking clear measurement principles. This is especially
true in a machine learning context.

Featurization of human subjects can have serious consequences. Recall
the example of using prediction in the context of the criminal justice
system. The figure below shows sample questions that were used to create
the data that the COMPAS recidivism risk score is trained on. These
questions were designed using psychometric models to capture archetypes
of people that might indicate future criminal behavior. Though COMPAS
features have been used to predict recidivism, they have been shown to
be no more predictive than using age, gender, and past criminal
activity.\citep{Angelino2017}

\begin{figure}
\centering
\includegraphics[width=1\textwidth,height=\textheight]{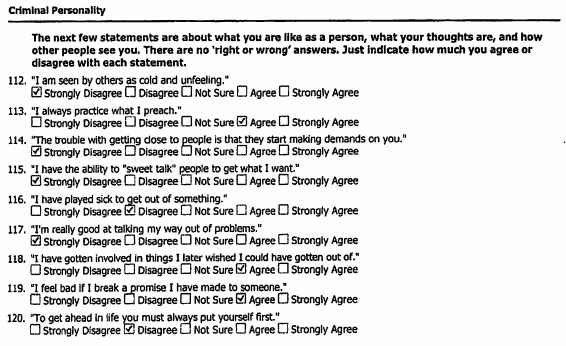}
\caption{Sample questions from the COMPAS questionnaire.}
\end{figure}

Machine learning and data creation involving human subjects should be
ethically evaluated in the same manner as any other scientific
investigation with humans.

Depending on context, different ethical guidelines and regulations exist
that aim at protecting human research subjects. The 1979 Belmont Report
is one ethical framework, commonly applied in the United States. It
rests on the three principles of respect for persons, beneficence, and
justice. Individuals should be treated as autonomous agents. Harm should
be minimized, while benefits should be maximized. Inclusion and
exclusion should not unduly burden specific individuals, as well as
marginalized and vulnerable groups.

Universities typically require obtaining institutional approval and
detailed training before conducting human subject research. These rules
apply also when data is collected from and about humans online.

We advise any reader to familiarize themselves with all applicable rules
and regulations regarding human subject research at their institution.

\hypertarget{quantization}{%
\section{Quantization}\label{quantization}}

\index{quantization}

Signals in the real world are often continuous and we have to choose how
to discretize them for use in a machine learning pipeline. Broadly
speaking, such practices fall under the rubric of \emph{quantization}.
In many cases, our goal is to quantize signals so that we can
reconstruct them almost perfectly. This is the case of high resolution
photography, high fidelity analog-to-digital conversion of audio, or
perfect sequencing of proteins. In other cases, we may want to only
record skeletons of signals that are useful for particular tasks. This
is the case for almost all quantizations of human beings. While we do
not aim to do a thorough coverage of this subject, we note quantization
is an essential preprocessing step in any machine learning pipeline.
Improved quantization schemes may very well translate into improved
machine learning applications. Let us briefly explore a few canonical
examples and issues of quantization in contemporary data science.

\hypertarget{images}{%
\subsection{Images}\label{images}}

Consider the raw bitmap formatting of an image. A bitmap file is an
array indexed by three coordinates. Mathematically, this corresponds to
a \emph{tensor} of order~\(3\). The first two coordinates index space
and the third indexes a color channel. That is,~\(x_{ijk}\) denotes the
intensity at row~\(i\), column~\(j\), and color channel~\(k\). This
representation summarizes an image by dividing two dimensional space
into a regular grid, and then counting the quantity of each of three
primary colors at each grid location.

This pixel representation suffices to render an image on a computer
screen. However, it might not be useful for prediction. Images of the
same scene with different lighting or photographic processes might end
up being quite dissimilar in a pixel representation. Even small
translations of two images might be far apart from each other in a pixel
representation. Moreover, from the vantage point of linear
classification, we could train a linear predictor on any ordering of the
pixels, but scrambling the pixels in an image certainly makes it
unrecognizable. We will describe transformations that address such
shortcomings of pixel representations in the sequel.

\hypertarget{text}{%
\subsection{Text}\label{text}}

Consider a corpus of~\(n\) documents. These documents will typically
have varying length and vocabulary. To embed a document as a vector, we
can create a giant vector for every word in the document where each
component of the vector corresponds to one dictionary word. The
dimension of the vector is therefore the size of the dictionary. For
each word that occurs in the document we set the corresponding
coordinate of the vector to~\(1\). All other coordinates corresponding
to words not in the document we set to~\(0\).

This is called a \emph{one-hot encoding} of the words in the
dictionary.\index{one-hot encoding} The one-hot encoding might seem both
wasteful due to the high dimension and lossy since we don't encode the
order of the words in the document. Nonetheless, it turns out to be
useful in a number of applications. Since typically the language has
more dictionary words than the length of the document, the encoding maps
a document to a very sparse vector. A corpus of documents would map to a
set of sparse vectors.

\hypertarget{template-matching}{%
\section{Template matching}\label{template-matching}}

\index{template matching}

While quantization can often suffice for prediction problems, we
highlighted above how this might be too fine a representation to encode
when data points are similar or dissimilar. Often times there are higher
level patterns that might be more representative for discriminative
tasks. A popular way to extract these patterns is \emph{template
matching}, where we extract the correlation of a feature vector~\(x\)
with a known pattern~\(v\), called \emph{template}.

At a high level, a template match creates a feature~\(x'\) from the
feature~\(x\) by binning the correlation with a template. A simple
example would be to fix a template~\(v\) and compute \[
    x' = \max\big\{v^T x,\,0\big\}\,.
\] We now describe some more specific examples that are ubiquitous in
pattern classification.

\hypertarget{fourier-cosine-and-wavelet-transforms}{%
\subsection{Fourier, cosine, and wavelet
transforms}\label{fourier-cosine-and-wavelet-transforms}}

One of the foundational patterns that we match to spatial or temporal
data is sinusoids. Consider a vector in~\(\R^d\) and the transformation
with~\(k\)-th component \[
    x_k' = |v_k^T x| \,.
\] Here the~\(\ell\)-th component of~\(v_k\) is given by
\(v_{k\ell} = \exp(2\pi i k \ell/d)\). In this case we are computing the
magnitude of the \emph{Fourier transform} of the feature vector. This
transformation measures the amount of oscillation in a vector. The
magnitude of the Fourier transform has the following powerful property:
Suppose~\(z\) is a translated version of~\(x\) so that \[
    z_k = x_{(k + s) \operatorname{mod} d}
\] for some shift~\(s\). Then one can check that for any~\(v_k\), \[
    |v_k^T x|  = |v_k^T z| \,.
\] That is, the magnitude of the Fourier transform is \emph{translation
invariant}. There are a variety of other transforms that fall into this
category of capturing the transformation invariant content of signals
including cosine and wavelet transforms.\index{Fourier transform}

\hypertarget{convolution}{%
\subsection{Convolution}\label{convolution}}

\index{convolution}

For spatial or temporal data, we often consider two data points to be
similar if we can translate one to align with another. For example,
small translations of digits are certainly the same digit. Similarly,
two sound utterances delayed by a few milliseconds are the same for most
applications. \emph{Convolutions} are small templates that are
translated over a feature figure to count the number of occurrences of
some pattern. The output of a convolution will have the same spatial
extent as the input, but will be a ``heat map'' denoting the amount of
correlation with the template at each location in the vector. Multiple
convolutions can be concatenated to add discriminative power. For
example, if one wants to design a system to detect animals in images,
one might design a template for heads, legs, and tails, and then linear
combinations of these appearances might indicate the existence of an
animal.

\hypertarget{summarization-and-histograms}{%
\section{Summarization and
histograms}\label{summarization-and-histograms}}

\index{summarization}\index{histogram}

Histograms summarize statistics about counts in data. These serve as a
method for both reducing the dimensionality of an input and removing
noise in the data. For instance, if a feature vector was the temperature
in a location over a year, this could be converted into a histogram of
temperatures which might better discriminate between locations. As
another example, we could downsample an image by making a histogram of
the amount of certain colors in local regions of the image.

\hypertarget{bag-of-words}{%
\subsection{Bag of words}\label{bag-of-words}}

\index{bag of words}

We could summarize a piece of text by summing up the one-hot encoding of
each word that occurs in the text. The resulting vector would have
entries where each component is the number of times the associated word
appears in the document. This is a \emph{bag of words} representation of
the document. A related representation that might take the structure of
text better into account might have a bag of words for every paragraph
or some shorter-length contiguous context.

Bag of words representations are surprisingly powerful. For example,
documents about sports tend to have a different vocabulary than
documents about fashion, and hence are far away from each other in such
an embedding. Since the number of unique words in any given document is
much less than all possible words, bag-of-words representations can be
reasonably compact and sparse. The representations of text as
large-dimensional sparse vectors can be deployed for predicting topics
and sentiment.

\hypertarget{downsamplingpooling}{%
\subsection{Downsampling/pooling}\label{downsamplingpooling}}

Another way to summarize and reduce dimension is to \emph{locally}
average a signal or image. This is called downsampling. For example, we
could break an image into 2x2 grids, and take the average or maximum
value in each grid. This would reduce the image size by a factor of 4,
and would summarize local variability in the image.\index{pooling}

\hypertarget{nonlinear-predictors}{%
\section{Nonlinear predictors}\label{nonlinear-predictors}}

Once we have a feature vector~\(x\) that we feel adequately compresses
and summarizes our data, the next step is building a prediction function
\(f(x)\). The simplest such predictors are linear functions of~\(x\),
and linear functions are quite powerful: all of the transformations we
have thus far discussed in this chapter often suffice to arrange data
such that linear decision rules have high accuracy.

However, we oftentimes desire further expressivity brought by more
complex decision rules. We now discuss many techniques that can be used
to build such nonlinear rules. Our emphasis will highlight how most of
these operations can be understood as embedding data in spaces where
linear separation is possible. That is: we seek a nonlinear
transformation of the feature vector so that linear prediction works
well on the transformed features.

\hypertarget{polynomials}{%
\subsection{Polynomials}\label{polynomials}}

\index{polynomials}

Polynomials are simple and natural nonlinear predictors. Consider the
dataset in the figure below. Here the data clearly can't be separated by
a linear function, but a quadratic function would suffice. Indeed, we'd
just use the rule that if~\((x_1-c_1)^2+(x_2-c_2)^2 \leq c_3\) then
\((x_1,x_2)\) would have label~\(y=1\). This rule is a quadratic
function of \((x_1,x_2)\).

\begin{figure}
\centering
\includegraphics[width=0.5\textwidth,height=\textheight]{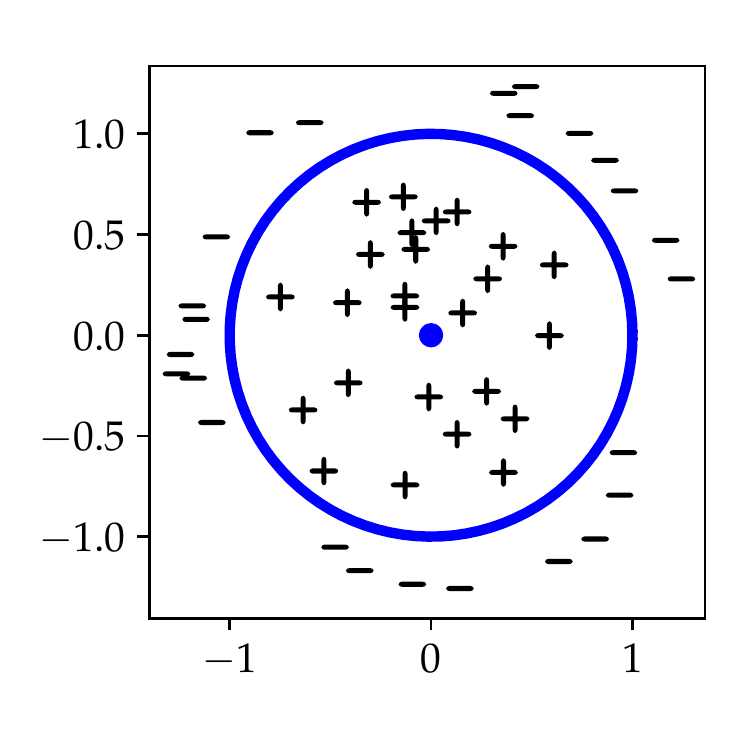}
\caption{A cartoon classification problem for polynomial classification.
Here, the blue dot denotes the center of the displayed circle.}
\end{figure}

To fit quadratic functions, we only need to fit the \emph{coefficients}
of the function. Every quadratic function can be written as a sum of
quadratic monomials. This means that we can write quadratic function
estimation as fitting a \emph{linear} function to the feature vector: \[
\Phi_2^{\text{poly}}(x) =   \begin{bmatrix} 1& x_1 &x_2 & x_1^2 & x_1 x_2 & x_2^2
    \end{bmatrix}^T
\] Any quadratic function can be written as
\(w^T \Phi_2^{\text{poly}}(x)\) for some~\(w\). The map
\(\Phi_2^{\text{poly}}\) is a \emph{lifting function} that transforms a
set of features into a more expressive set of features.

The features associated with the quadratic polynomial lifting function
have an intuitive interpretation as crossproducts of existing features.
The resulting prediction function is a linear combination of pairwise
products of features. Hence, these features capture co-occurrence and
correlation of a set of features.

The number of coefficients of a generic quadratic function in~\(d\)
dimensions is \[
{{d+2} \choose 2}\,,
\] which grows quadratically with dimension. For general degree~\(p\)
polynomials, we could construct a lifting function
\(\Phi_p^{\text{poly}}(x)\) by listing all of the monomials with degree
at most~\(p\). Again, any polynomial of degree~\(p\) can be written as
\(w^T \Phi_p^{\text{poly}}(x)\) for some~\(w\). In this case,
\(\Phi_p^{\text{poly}}(x)\) would have \[
{{d+p} \choose p}
\] terms, growing roughly as~\(d^p\). It shouldn't be too surprising to
see that as the degree of the polynomial grows, increasingly complex
behavior can be approximated.

\hypertarget{how-many-features-do-you-need}{%
\subsection{How many features do you
need?}\label{how-many-features-do-you-need}}

Our discussion of polynomials led with the motivation of creating
nonlinear decision boundaries. But we saw that we could also view
polynomial boundaries as taking an existing feature set and performing a
nonlinear transformation to embed that set in a higher dimensional space
where we could then search for a linear decision boundary. This is why
we refer to nonlinear feature maps as \emph{lifts}.

Given expressive enough functions, we can always find a lift such that a
particular dataset can be mapped to any desired set of labels. How high
of a dimension is necessary? To gain insights into this question, let us
stack all of the data points~\(x_1,\ldots,x_n \in\mathbb{R}^d\) into a
matrix~\(X\) with~\(n\) rows and~\(d\) columns. The predictions across
the entire dataset can now be written as \[
 \hat{y} = Xw\,.
\] If the~\(x_i\) are linearly independent, then as long
as~\(d \geq n\), we can make \emph{any} vector of predictions~\(y\) by
finding a corresponding vector~\(w\). For the sake of
\emph{expressivity}, the goal in feature design will be to find lifts
into high dimensional space such that our data matrix~\(X\) has linearly
independent columns. This is one reason why machine learning
practitioners lean towards models with more parameters than data points.
Models with more parameters than data points are called
\emph{overparameterized}.

\index{overparameterized}

As we saw in the analysis of the perceptron, the key quantities that
governed the number of mistakes in the perceptron algorithm were the
maximum norm of~\(x_k\) and the norm of the optimal~\(w\). Importantly,
the dimension of the data played no role. Designing features where~\(w\)
has controlled norm is a domain specific challenge, but has nothing to
do with dimension. As we will see in the remainder of this book, high
dimensional models have many advantages and few disadvantages in the
context of prediction problems.

\hypertarget{basis-functions}{%
\subsection{Basis functions}\label{basis-functions}}

\index{basis functions}

Polynomials are an example of \emph{basis functions}. More generally, we
can write prediction functions as linear combinations of~\(B\) general
nonlinear functions~\(\{b_k\}\): \[
    f(x) = \sum_{k=1}^{B} w_k b_k(x)
\] In this case, there is again a lifting function
\(\Phi_{\text{basis}}(x)\) into~\(B\) dimensions where the~\(k\)th
component is equal to~\(b_k(x)\)
and~\(f(x) =w^T \Phi_{\text{basis}}(x)\). There are a variety of basis
functions used in numerical analysis including trigonometric
polynomials, spherical harmonics, and splines. The basis function most
suitable for a given task is often dictated by prior knowledge in the
particular application domain.

A particularly useful set in pattern classification are the \emph{radial
basis functions}. A radial basis function has the form \[
    b_z(x) = \phi( \Vert x-z \Vert)
\] where~\(z \in \mathbb{R}^d\) and
\(\phi:\mathbb{R}\rightarrow\mathbb{R}\). Most commonly, \[
    \phi(t) = e^{-\gamma t^2}
\] for some~\(\gamma>0\). In this case, given~\(z_1,\ldots, z_k\), our
functions take the form \[
    f_k(x) = \sum_{j=1}^k w_j  e^{-\gamma \Vert x-z_j \Vert^2}\,.
\] Around each anchor point~\(z_k\), we place a small Gaussian bump.
Combining these bumps with different weights allows us to construct
arbitrary functions.

\begin{figure}
\centering
\includegraphics[width=0.75\textwidth,height=\textheight]{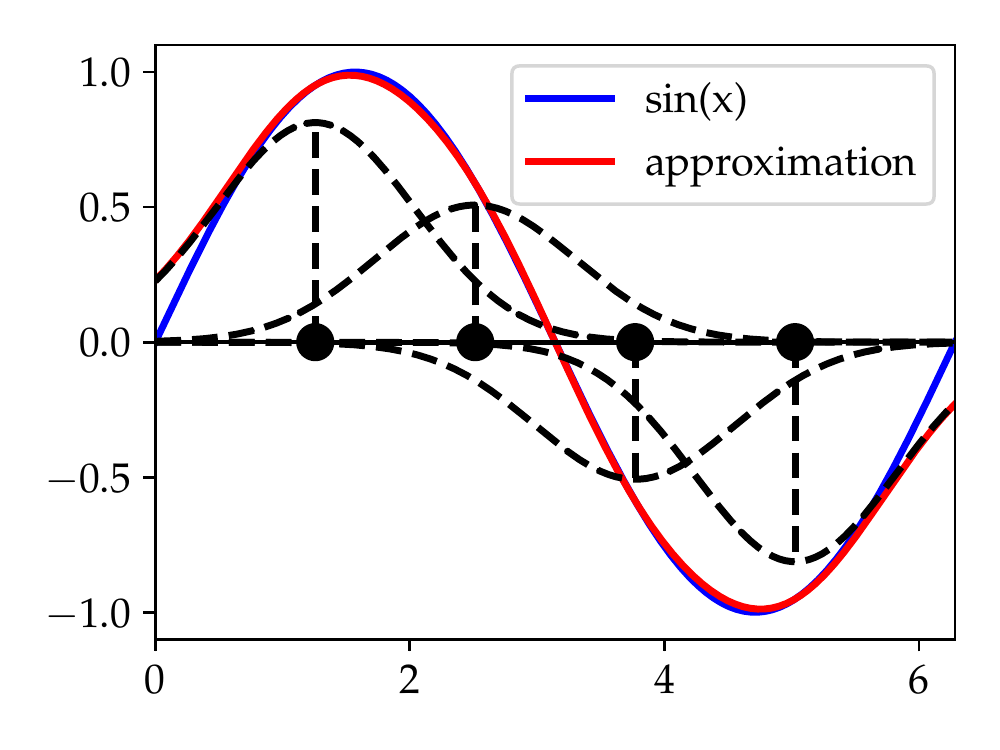}
\caption{Radial Basis Function approximation of sin(x). We plot the four
Gaussian bumps that sum to approximate the function.}
\end{figure}

How to choose the~\(z_k\)? In low dimensions,~\(z_k\) could be placed on
a regular grid. But the number of bases would then need to scale
exponentially with dimension. In higher dimensions, there are several
other possibilities. One is to use the set of training data. This is
well motivated by the theory of \emph{reproducing kernels}. Another
option would be to pick the~\(z_k\) at random, inducing \emph{random
features}. A third idea would be to search for the best~\(z_i\). This
motivates our study of \emph{neural networks}. As we will now see, all
three of these methods are powerful enough to approximate any desired
function, and they are intimately related to each other.

\hypertarget{kernels}{%
\subsection{Kernels}\label{kernels}}

\index{kernel}

One way around high dimensionality is to constrain the space of
prediction function to lie in a low dimensional subspace. Which subspace
would be useful? In the case of linear functions, there is a natural
choice: the span of the training data. By the fundamental theorem of
linear algebra, any vector in~\(\mathbb{R}^d\) can be written as sum of
a vector in the span of the training data and a vector orthogonal to all
of the training data. That is, \[
    w = \sum_{i=1}^n \alpha_i x_i + v
\] where~\(v\) is orthogonal to the~\(x_i\). But if~\(v\) is orthogonal
to every training data point, then in terms of prediction, \[
    w^T x_i = \sum_{j=1}^n \alpha_j x_j^T x_i \,.
\] That is, the~\(v\) has no bearing on in-sample prediction whatsoever.
Also note that prediction is only a function of the dot products between
the training data points. In this section, we consider a family of
prediction functions built with such observations in mind: we will look
at functions that expressly let us compute dot products between liftings
of points, noting that our predictions will be linear combinations of
such lifted dot products.

Let~\(\Phi(x)\) denote any lifting function. Then \[
    k(x,z) := \Phi(x)^T \Phi(z)
\] is called the \emph{kernel function} associated with the feature map
\(\Phi\). Such kernel functions have the property that for any
\(x_1,\ldots, x_n\), the matrix~\(K\) with
entries\index{kernel!function} \[
    K_{ij} = k(x_i,x_j)
\] is positive semidefinite. This turns out to be the key property to
define kernel functions. We say a symmetric function
\(k:\mathbb{R}^d \times \mathbb{R}^d \rightarrow \mathbb{R}\) is a
kernel function if it has this positive semidefiniteness property.

It is clear that positive combinations of kernel functions are kernel
functions, since this is also true for positive semidefinite matrices.
It is also true that if~\(k_1\) and~\(k_2\) are kernel functions, then
so is \(k_1 k_2\). This follows because the elementwise product of two
positive semidefinite matrices is positive semidefinite.

Using these rules, we see that the function \[
    k(x,z) = (a +b x^Tz)^p
\] where~\(a,b\geq 0\),~\(p\) a positive integer is a kernel function.
Such kernels are called a polynomial kernels. For every polynomial
kernel, there exists an associated lifting function~\(\Phi\) with each
coordinate of~\(\Phi\) being a monomial such that \[
    k(x,z) =\Phi(x)^T \Phi(z) \,.
\] As a simple example, consider the 1-dimensional case of a kernel \[
    k(u,v) = (1+uv)^2\,.
\] Then~\(k(u,v) =\Phi(u)^T \Phi(v)\) for \[
\Phi(u) = \begin{bmatrix} 1\\ \sqrt{2} u \\ u^2
\end{bmatrix}\,.
\]

We can generalize polynomial kernels to \emph{Taylor Series} kernels.
Suppose that the one dimensional function~\(h\) has a convergent Taylor
series for all~\(t \in [-R,R]\): \[
    h(t) = \sum_{j=1}^\infty a_j t^j
\] where~\(a_j \geq 0\) for all~\(j\). Then the function \[
    k(x,z) = h(\langle x, z \rangle)
\] is a positive definite kernel. This follows because each term
\(\langle x,z\rangle^j\) is a kernel, and we are taking a nonnegative
combination of these polynomial kernels. The feature space of this
kernel is the span of the monomials of degrees where the~\(a_j\) are
nonzero.

Two example kernels of this form are the \emph{exponential kernel} \[
    k(x,z) = \exp(\gamma \langle x, z\rangle)
\] and the \emph{arcsine kernel} \[
    k(x,z) = \sin^{-1}(\langle x,z \rangle)\,,
\] which is a kernel for~\(x\),~\(z\) on the unit sphere.

Another important kernel is the Gaussian kernel:\index{kernel!Gaussian}
\[
    k(x,z) = \exp\big(-\tfrac{\gamma}{2} \Vert x-z\Vert^2\big)\,.
\] The Gaussian kernel can be thought of as first lifting data using the
exponential kernel then projecting onto the unit sphere in the lifted
space.

We note that there are many kernels with the same feature space. Any
Taylor Series kernel with positive coefficients will have the same set
of features. The feature space associated Gaussian kernel is equivalent
to the span of radial basis functions. What distinguishes the kernels
beyond the features they represent? The key is to look at the norm.
Suppose we want to find a fit of the form \[
    f(x_j) = w^T \Phi(x_j)\qquad \text{for}~j=1,2,\ldots,n\,.
\] In the case when~\(\Phi\) maps into a space with more dimensions than
the number of data points we have acquired, there will be an infinite
number of~\(w\) vectors that perfectly interpolate the data. As we saw
in our introduction to supervised learning, a convenient means to pick
an interpolator is to choose the one with smallest norm. Let's see how
the norm interacts with the form of the kernel. Suppose our kernel is a
Taylor series kernel \[
    h(t) = \sum_{j=1}^\infty a_j \langle x, z \rangle^j\,.
\] Then the smaller~\(a_j\), the larger the corresponding~\(w_j\) should
have to be. Thus, the~\(a_j\) in the kernel expansion govern how readily
we allow each feature in a least-norm fit. If we consider the
exponential kernel with parameter~\(\gamma\), then
\(a_j = \frac1{j!}\gamma^j\). Hence, for large values of~\(\gamma\),
only low degree terms will be selected. As we decrease~\(\gamma\), we
allow for higher degree terms to enter the approximation. Higher degree
terms tend to be more sensitive to perturbations in the data than lower
degree ones, so~\(\gamma\) should be set as large as possible while
still providing desirable predictive performance.

The main appeal of working with kernel representations is they translate
into simple algorithms with bounded complexity. Since we restrict our
attention to functions in the span of the data, our functions take the
form \[
    f(x) = \left(\sum\nolimits_i\alpha_i \Phi(x_i)^T \right)\Phi(x) = \sum\nolimits_i \alpha_i k(x_i,x)\,.
\] We can thus pose all of our optimization problems in terms of the
coefficients~\(\alpha_i\). This means that any particular problem will
have at most~\(n\) parameters to search for. Even the norm of~\(f\) can
be computed without ever explicitly computing the feature embedding.
Recall that when~\(f(x) = w^T \Phi(x)\) with
\(w = \sum\nolimits_i\alpha_i \Phi(x_i)\), we have \[
 \Vert w \Vert^2=\left \Vert\sum\nolimits_i \alpha_i \Phi(x_i)\right\Vert^2 = \alpha^T K \alpha\,,
\] where~\(K\) is the matrix with~\(ij\)th entry~\(k(x_i,x_j)\). As we
will see in the optimization chapter, such representations turn out to
be optimal in most machine learning problems. Functions learned by ERM
methods on kernel spaces are weighted sums of the similarity (dot
product) between the training data and the new data. When~\(k\) is a
Gaussian kernel, this relationship is even more evident: The optimal
solution is simply a radial basis function whose anchor points are given
by the training data points.

\hypertarget{neural-networks}{%
\subsection{Neural networks}\label{neural-networks}}

\index{neural networks}

Though the name originates from the study of neuroscience, modern neural
nets arguably have little to do with the brain. Neural nets are
mathematically a composition of differentiable functions, typically
alternating between componentwise nonlinearities and linear maps. The
simplest example of a neural network would be \[
    f(x) = w^T \sigma(Ax+b)
\] Where~\(w\) and~\(b\) are vectors,~\(A\) is a matrix, and~\(\sigma\)
is a componentwise nonlinearity, applying the same nonlinear function to
each component of its input.

The typically used nonlinearities are not Gaussian bumps. Indeed, until
recently most neural nets used \emph{sigmoid nonlinearities} where
\(\sigma(t)\) is some function that is~\(0\) at negative infinity,~\(1\)
at positive infinity, and strictly increasing. Popular choices of such
functions include~\(\sigma(t) = \tanh(t)\) or
\(\sigma(t) = \frac{1}{\exp(-t)+1}\). More recently, another
nonlinearity became overwhelmingly popular, the \emph{rectified linear
unit} or ReLU: \[
    \sigma(t) = \max\big\{t,\,0\big\}
\] This simple nonlinearity is easy to implement and differentiate in
hardware.

Though these nonlinearities are all different, they all generate similar
function spaces that can approximate each other. In fact, just as was
the case with kernel feature maps, neural networks are powerful enough
to approximate any continuous function if enough bases are used. A
simple argument by Cybenko clarifies why only a bit of nonlinearity is
needed for universal
approximation.\citep{cybenko1989approximation}\index{universal approximation}

Suppose that we can approximate the unit step function
\(u(t) = \mathbb{1}\{t>0\}\) as a linear combination of shifts of
\(\sigma(t)\). A sigmoidal function like~\(\tanh\) is already such an
approximation and~\(\tanh(\alpha t)\) converges to the unit step as
\(\alpha\) approaches~\(\infty\). For ReLU functions we have for
any~\(c>0\) that \[
    \frac{1}{2c} \max\big\{t+c,\,0\big\} - \frac{1}{2c} \max\big\{t-c,\,0\} =
    \begin{cases}
        0 & t<-c\\
        1 & t>c\\
        \frac{t+c}{2c} & \text{otherwise}
    \end{cases}\,.
\] It turns out that approximation of such step functions is all that is
needed for universal approximation.

\begin{figure}
\centering
\includegraphics[width=0.5\textwidth,height=\textheight]{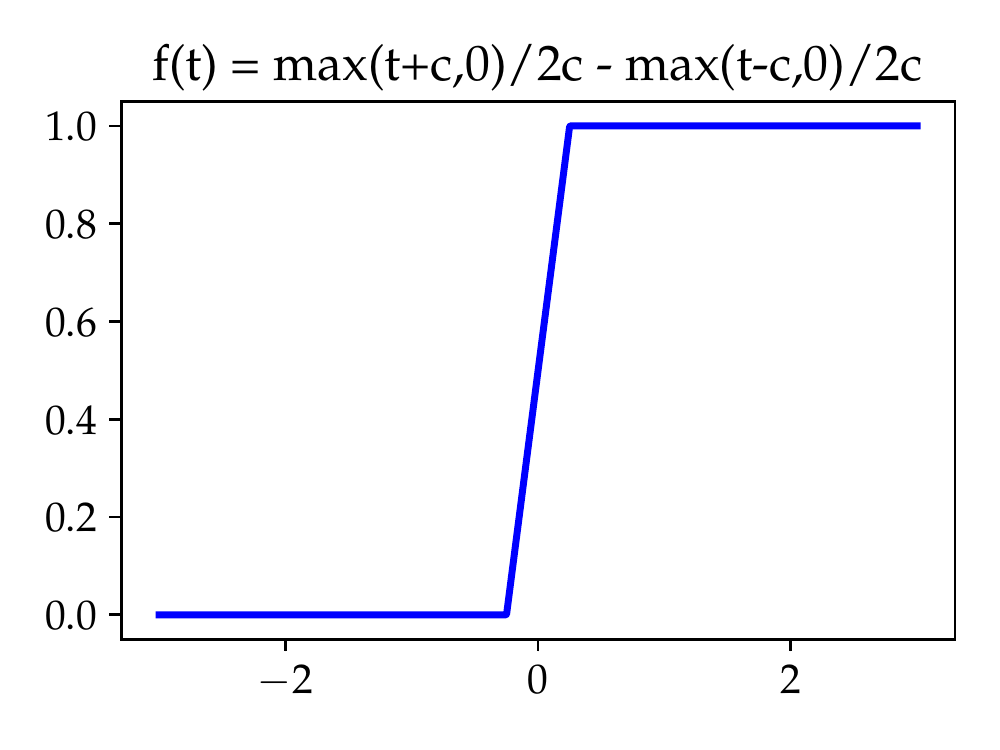}
\caption{Creating a step function from ReLUs. Here, c=1/4.}
\end{figure}

To see why, suppose we have a nonzero function~\(g\) that is not
well-approximated by a sum of the form \[
    \sum_{i=1}^N w_i \sigma(a_i^Tx +b_i)\,.
\] This means that we must have a nonzero function~\(f\) that lies
outside the span of sigmoids. We can take~\(f\) to be the projection
of~\(g\) onto the orthogonal complement of the span of the sigmoidal
functions. This function~\(f\) will satisfy \[
    \int \sigma(a^Tx +b) f(x) \dif x =0
\] for all vectors~\(a\) and scalars~\(b\). In turn, since our
nonlinearity can approximate step functions, this implies that for
any~\(a\) and any \(t_0\) and~\(t_1\), \[
    \int \mathbb{1}\{t_0 \leq a^Tx \leq t_1\} f(x) \dif x = 0\,.
\] We can approximate any continuous function as a sum of the indicator
function of intervals, which means \[
    \int h(a^T x) f(x) \dif x = 0
\] for any continuous function~\(h\) and any vector~\(a\). Using
\(h(t)=\exp(it)\) proves that the Fourier transform of~\(f\) is equal to
zero, and hence that~\(f\) itself equals zero. This is a contradiction.

The core of Cybenko's argument is a reduction to approximation in one
dimension. From this perspective, it suffices to find a nonlinearity
that can well-approximate ``bump functions'' which are nearly equal to
zero outside a specified interval and are equal to~\(1\) at the center
of the interval. This opens up a variety of potential nonlinearities for
use as universal approximators.

While elegant and simple, Cybenko's argument does not tell us \emph{how
many} terms we need in our approximation. More refined work on this
topic was pursued in the 1990s. Barron\citep{Barron93} used a similar
argument about step functions combined with a powerful randomized
analysis due to Maurey\citep{Pisier81}. Similar results were derived for
sinusoids\citep{Jones92} by Jones and ReLU
networks\citep{breiman1993hinging} by Breiman. All of these results
showed that two-layer networks sufficed for universal approximation, and
quantified how the number of basis functions required scaled with
respect to the complexity of the approximated function.

\hypertarget{random-features}{%
\subsection{Random features}\label{random-features}}

\index{random features}

Though the idea seems a bit absurd, a powerful means of choosing basis
function is by random selection. Suppose we have a parametric family of
basis functions~\(b(x;\vartheta)\). A random feature map chooses
\(\vartheta_1,\ldots,\vartheta_D\) from some distribution
on~\(\vartheta\), and use the feature map \[
    \Phi_{\text{rf}}(x) = \begin{bmatrix}
        b(x;\vartheta_1) \\b(x;\vartheta_2)\\ \vdots \\b(x;\vartheta_D)
    \end{bmatrix}\,.
\] The corresponding prediction functions are of the form \[
    f(x) = \sum_{k=1}^{D} w_k b(x;\vartheta_k)
\] which looks very much like a neural network. The main difference is a
matter of emphasis: here we are stipulating that the parameters
\(\vartheta_k\) are random variables, whereas in neural networks,
\(\vartheta_k\) would be considered parameters to be determined by the
particular function we aim to approximate.

Why might such a random set of functions work well to approximate
complex functional behavior? First, from the perspective of
optimization, it might not be too surprising that a random set of
nonlinear basis functions will be linearly independent. Hence, if we
choose enough of them, we should be able to fit any set of desired
labels.

Second, random feature maps are closely connected with kernel spaces.
The connections were initially drawn out in work by Rahimi and
Recht.\citep{RahimiRecht07, RahimiRechtNIPS08} Any random feature map
generates an empirical kernel,
\(\Phi_{\text{rf}}(x)^T \Phi_{\text{rf}}(z)\). The expected value of
this kernel can be associated with some Reproducing Kernel Hilbert
Space. \[
\begin{aligned}
\E[\tfrac{1}{D} \Phi_{\text{rf}}(x)^T \Phi_{\text{rf}}(z) ]
&= \E\left[\frac{1}{D} \sum_{k=1}^D b(x;\vartheta_k) b(z;\vartheta_k) \right]\\
&= \E[b(x;\vartheta_1) b(z;\vartheta_1) ]\\
& = \int p(\vartheta) b(x;\vartheta) b(z;\vartheta) \dif\vartheta
\end{aligned}
\] In expectation, the random feature map yields a kernel given by the
final integral expressions. There are many interesting kernels that can
be written as such an integral. In particular, the Gaussian kernel would
arise if \[
\begin{aligned}
p(\vartheta) &= \mathcal{N}(0,\gamma I)\\
b(x;\vartheta) &= [\cos(\vartheta^T x), \,\, \sin(\vartheta^T x)]
\end{aligned}
\] To see this, recall that the Fourier transform of a Gaussian is a
Gaussian, and write: \[
\begin{aligned}
    & k(x,z) \\
    &= \exp(-\tfrac{\gamma}{2} \Vert x-z \Vert^2)\\
    &= \frac{1}{(2 \pi \gamma)^{d/2}}\int e^{-\frac{ \Vert v\Vert^2}{2\gamma}} \exp(i v^T(x-z)) \dif v\\
    &= \frac{1}{(2 \pi \gamma)^{d/2}}\int e^{-\frac{ \Vert v\Vert^2}{2\gamma}} \left\{\cos(v^T x) \cos(v^Tz) + \sin(v^T x)\sin(v^Tz)\right\} \dif v\,.
\end{aligned}
\] This calculation gives new insights into the feature space associated
with a Gaussian kernel. It shows that the Gaussian kernel is a
continuous mixture of inner products of sines and cosines. The sinusoids
are weighted by a Gaussian function on their frequency: high frequency
sinusoids have vanishing weight in this expansion. The parameter
\(\gamma\) controls how quickly the higher frequencies are damped.
Hence, the feature space here can be thought of as low frequency
sinusoids. If we sample a frequency from a Gaussian distribution, it
will be low frequency (i.e., have small norm) with high probability.
Hence, a random collection of low frequency sinusoids approximately
spans the same space as that spanned by a Gaussian kernel.

If instead of using sinusoids, we chose our random features to be
\(\mathrm{ReLU}(v^Tx)\), our kernel would become \[
\begin{aligned}
    k(x,z)
    &= \frac{1}{(2 \pi)^{d/2}}\int \exp \left(-\frac{ \Vert v\Vert^2}{2}\right)\operatorname{ReLU}(v^T x)\operatorname{ReLU}(v^T z) \dif v\\
    &= \Vert x\Vert \Vert z\Vert \left\{sin(\vartheta) + (\pi-\vartheta) \cos(\vartheta)\right\}\,,
\end{aligned}
\] where \[
        \vartheta = \cos^{-1} \left(\frac{\langle x, z \rangle}{\Vert x \Vert \Vert z \Vert}\right)\,.
\] This computation was first made by Cho and Saul.\citep{Cho09} Both
the Gaussian kernel and this ``ReLU kernel'' are universal Taylor
kernels, and, when plotted, we see even are comparable on unit norm
data.

\begin{figure}
\centering
\includegraphics[width=0.75\textwidth,height=\textheight]{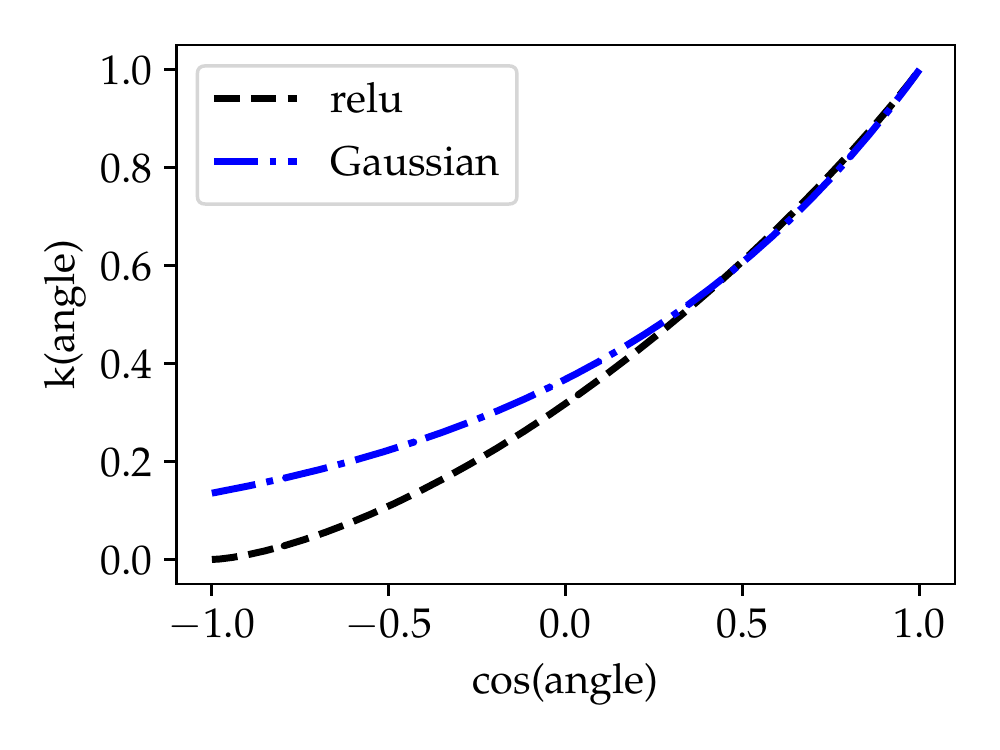}
\caption{Comparison of the Gaussian and Arccosine Kernels. Plotting the
kernel value as a function of the angle between two unit norm vectors.}
\end{figure}

Prediction functions built with random features are just randomly wired
neural networks. This connection is useful for multiple reasons. First,
as we will see in the next chapter, optimization of the weights~\(w_k\)
is far easier than joint optimization of the weights and the parameters
\(\vartheta\). Second, the connection to kernel methods makes the
generalization properties of random features straightforward to analyze.
Third, much of the recent theory of neural nets is based on connections
between random feature maps and the randomization at initialization of
neural networks. The main drawback of random features is that, in
practice, they often require large dimensions before they provide
predictions on par with neural networks. These tradeoffs are worth
considering when designing and implementing nonlinear prediction
functions.

Returning to the radial basis expansion \[
    f(x) = \sum_{j=1}^k w_j  e^{-\gamma \Vert x-z_j \Vert^2}\,,
\] we see that this expression could be a neural network, a kernel
machine, or a random feature network. The main distinction between these
three methods is how the~\(z_j\) are selected. Neural nets search
require some sort of optimization procedure to find the~\(z_j\). Kernel
machines place the~\(z_j\) on the training data. Random features
select~\(z_j\) at random. The best choice for any particular prediction
problem will always be dictated by the constraints of practice.

\hypertarget{chapter-notes-3}{%
\section{Chapter notes}\label{chapter-notes-3}}

To our knowledge, there is no full and holistic account of measurement
and quantization, especially when quantization is motivated by data
science applications. From the statistical signal processing viewpoint,
we have a full and complete theory of quantization in terms of
understanding what signals can be reconstructed from digital
measurements. The Nyquist-Shannon theory allows us to understand what
parts of signals may be lost and what artifacts are introduced. Such
theory is now canon in undergraduate courses on signal processing. See,
e.g., Oppenheim and Willsky.\citep{willsky1997signals} For task-driven
sampling, the field remains quite open. The theory of compressive
sensing led to many recent and exciting developments in this space,
showing that task-driven sampling where we combined domain knowledge,
computation, and device design could reduce the data acquisition burdens
of many pattern recognition tasks.\citep{candes2008introduction} The
theory of experiment design and survey design has taken some cues from
task-driven sampling.

Reproducing kernels have been used in pattern recognition and machine
learning for nearly as long as neural networks. Kernels and Hilbert
spaces were first used in time series prediction in the late 1940s, with
fundamental work by Karhunen--Loève showing that the covariance function
of a time series was a Mercer
Kernel.\citep{karhunen1947lineare, loeve1946functions} Shortly
thereafter, Reproducing Kernel Hilbert Spaces were formalized by
Aronszajn in 1950.\citep{Aronszajn50} Parzen was likely the first to
show that time series prediction problem could be reduced to solving a
least-squares problem in an RKHS and hence could be computed by solving
a linear system.\citep{parzen1961approach} Wahba's survey of RKHS
techniques in statistics covers many other developments
post-Parzen.\citep{Wahba90} For further reading on the theory and
application of kernels in machine learning consult the texts by
Schölkopf and Smola\citep{SchoelkopfKernelBook} and Shawe-Taylor and
Cristianini.\citep{shawe2004kernel}

Also since its inception, researchers have been fascinated by the
approximation power of neural networks. Rosenblatt discussed properties
of universal approximation in his monograph on
neurodynamics.\citep{rosenblatt1962principles} It was in the 80s when it
became clear that though neural networks were able to approximate any
continuous function, they needed to be made more complex and intricate
in order to achieve high quality approximation. Cybenko provided a
simple proof that neural nets were dense in the space of continuous
functions, though did not estimate how large such networks might need to
be.\citep{cybenko1989approximation} An elegant, randomized argument by
Maurey\citep{Pisier81} led to a variety of approximation results which
quantified how many basis terms were needed. Notably, Jones showed that
a simple greedy method could approximate any continuous function with a
sum of sinusoids.\citep{Jones92} Barron shows that similar greedy
methods could be used \citep{Barron93} to build neural nets that
approximated any function. Breiman analyzed ReLU networks using the same
framework.\citep{breiman1993hinging} The general theory of approximation
by bases is rich, and a Pinkus' book details some of the necessary and
sufficient conditions to achieve high quality approximations with as few
bases as possible.\citep{PinkusBook}

That randomly wired neural networks could solve pattern recognition
problems also has a long history. Minsky's first electronic neural
network, SNARC, was randomly wired. The story of SNARC (Stochastic
Neural Analog Reinforcement Calculator) is rather apocryphal. There are
no known photographs of the assembled device, although a 2019 article by
Akst has a picture of one of the ``neurons''.\citep{akst2019machine} The
commonly referenced publication, a Harvard technical report, appears to
not be published. However, the ``randomly wired'' story lives on, and it
is one that Minsky told countless times through his life. Many years
later, Rahimi and Recht built upon the approximation theory of Maurey,
Barron, and Jones to show that random combinations of basis functions
could approximate continuous functions well, and that such random
combinations could be thought of as approximately solving prediction
problems in a
RKHS.\citep{RahimiRechtNIPS08, RahimiRecht07, RahimiRechtAllerton08}
This work was later used as a means to understand neural networks, and,
in particular, the influence of their random initializations. Daniely et
al.~computed the kernel spaces associated with that randomly initialized
neural networks\citep{daniely2016toward}, and Jacot et al.~pioneered a
line of work using kernels to understand the dynamics of neural net
training.\citep{jacot2018neural}

There has been noted cultural tension between the neural-net and kernel
``camps.'' For instance, the tone of the introduction of work by Decoste
and Schölkopf telegraphs a disdain by neural net proponents of the
Support Vector Machine.\citep{decoste2002training}

\begin{quote}
Initially, SVMs had been considered a theoretically elegant spin-off of
the general but, allegedly, largely useless VC-theory of statistical
learning. In 1996, using the first methods for incorporating prior
knowledge, SVMs became competitive with the state of the art in the
handwritten digit classification benchmarks that were popularized in the
machine learning community by AT\&T and Bell Labs. At that point,
practitioners who are not interested in theory, but in results, could no
longer ignore SVMs.
\end{quote}

With the rise of deep learning, however, there are a variety of machine
learning benchmarks where SVMs or other kernel methods fail to match the
performance of neural networks. Many have dismissed kernel methods as a
framework whose time has past. However, kernels play an evermore active
role in helping to better understand neural networks and insights from
deep learning have helped to improve the power of kernel methods on
pattern recognition tasks.\citep{Shankar20a} Neural nets and kernels are
complementary, and active research in machine learning aims to bring
these two views more closely together.

\chapter{Optimization}

In Chapter 2, we devised a closed form expression for the optimal
decision rule assuming we have a probability model for the data. Then we
turned to empirical risk minimization (ERM) where we instead rely on
numerical methods to discover good decision rules when we don't have
such a probability model. In this chapter, we take a closer look at how
to solve empirical risk minimization problems effectively. We focus on
the core optimization methods commonly used to solve empirical risk
minimization problems and on the mathematical tools used to analyze
their running times.

Our main subject will be \emph{gradient descent} algorithms and how to
shape loss functions so that gradient descent succeeds. Gradient descent
is an iterative procedure that iterates among possible models, at each
step replacing the old model with one with lower empirical risk. We show
that the class of optimization problems where gradient descent is
guaranteed to find an optimal solution is the set of \emph{convex
functions}. When we turn to risk minimization, this means that gradient
descent will find the model that minimizes the empirical risk whenever
the loss function is convex and the decision function is a linear
combination of features.\index{gradient descent}

We then turn to studying \emph{stochastic} gradient descent, the
workhorse of machine learning. Stochastic gradient descent is
effectively a generalization of the perceptron learning rule. Its
generality enables us to apply it to a variety of function classes and
loss functions and guarantee convergence even if the data may not be
separable. We spend a good deal of time looking at the dynamics of the
stochastic gradient method to try to motivate why it is so successful
and popular in machine learning.

Starting from the convex case, we work towards more general nonconvex
problems. In particular, we highlight two salient features of gradient
descent and stochastic gradient descent that are particular to empirical
risk minimization and help to motivate the resilience of these methods.

First, we show that even for problems that are not convex, gradient
descent for empirical risk minimization has an \emph{implicit convexity}
property that encourages convergence. Though we explicitly optimize over
function representations which are computationally intractable to
optimize in the worst-case, it turns out that we can still reason about
the convergence of the predictions themselves.\index{convexity!implicit}

Second, we show that gradient descent implicitly manages the complexity
of the prediction function, encouraging solutions of low complexity in
cases where infinitely many solutions exist. We close the chapter with a
discussion of other methods for empirical risk minimization that more
explicitly account for model complexity and stable convergence.

\hypertarget{optimization-basics}{%
\section{Optimization basics}\label{optimization-basics}}

Stepping away from empirical risk minimization for a moment, consider
the general minimization problem \[
\begin{array}{ll}
    \text{minimize}_w  & \Phi(w)
\end{array}
\] where~\(\Phi\colon\R^d\to\R\) is a real-valued function over the
domain \(\R^d\).

When and how can we minimize such a function? Before we answer this
question, we need to formally define what we're shooting
for.\index{minimizer}\index{minimizer!global}\index{minimizer!local}

\begin{Definition}

A point~\(w_\star\) is a \emph{minimizer} of~\(\Phi\) if
\(\Phi(w_\star)\leq \Phi(w)\) for all~\(w\). It is a \emph{local
minimizer} of \(\Phi\) if for
some~\(\epsilon> 0\),~\(\Phi(w_\star)\leq \Phi(w)\) for all \(w\) such
that~\(\Vert w-w_\star \Vert\leq \epsilon\).

Sometimes we will refer to minimizers as \emph{global minimizers} to
contrast against local minimizers.

\end{Definition}

The figure below presents example functions and their minima. In the
first illustration, there is a unique minimizer. In the second, there
are an infinite number of minimizers, but all local minimizers are
global minimizers. In the third example, there are many local minimizers
that are not global minimizers.

\begin{figure}
\centering
\includegraphics[width=1\textwidth,height=\textheight]{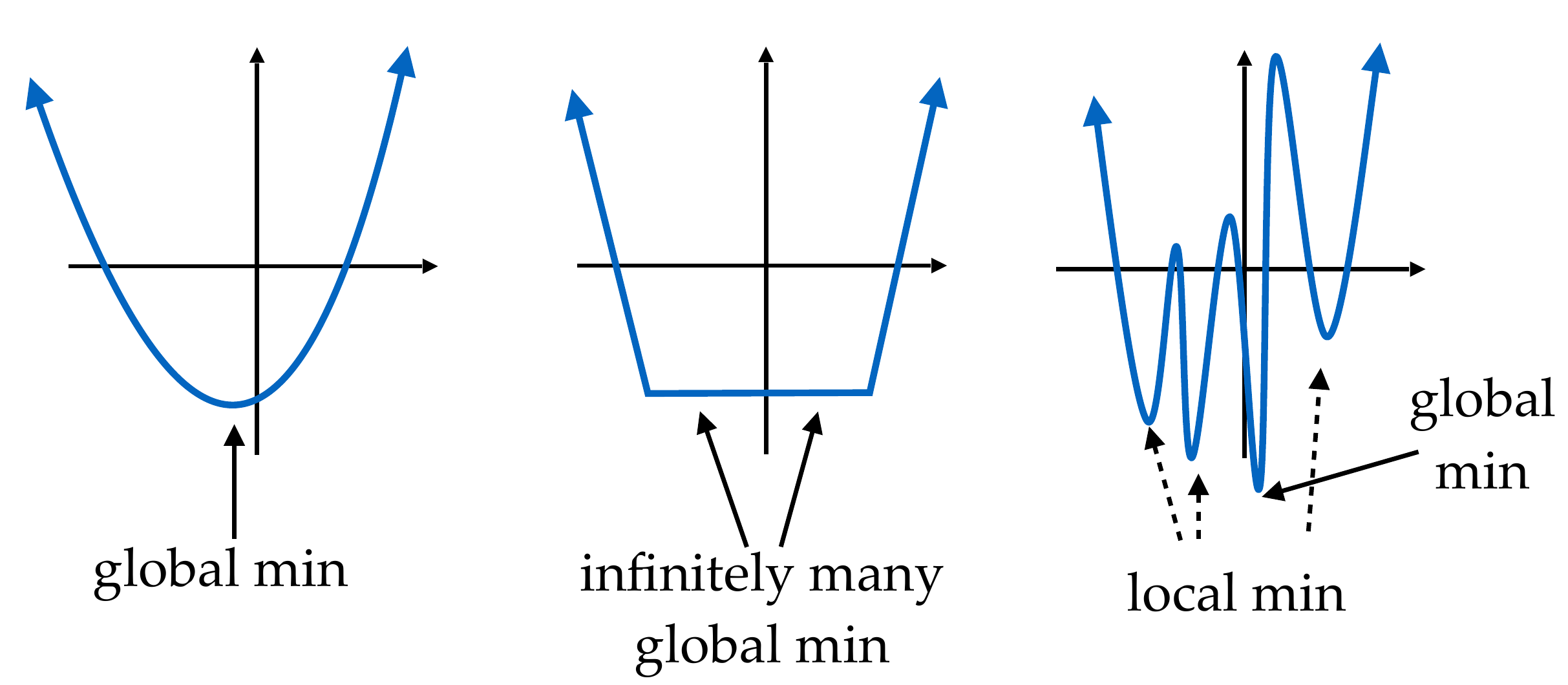}
\caption{Examples of minima of functions.}
\end{figure}

Note that in our example, the two functions without suboptimal local
minimizers share the property that for any two points~\(w_1\)
and~\(w_2\), the line segment connecting~\((w_1,\Phi(w_1))\)
to~\((w_2,\Phi(w_2))\) lies completely above the graph of the function.
Such functions are called \emph{convex functions}.\index{convexity}

\begin{Definition}

A function~\(\Phi\) is \emph{convex} if for all~\(w_1\),~\(w_2\)
in~\(\mathbb{R}^d\) and~\(\alpha \in [0,1]\), \[
        \Phi(\alpha w_1 + (1-\alpha) w_2 )\leq \alpha\Phi(w_1)+(1-\alpha)\Phi(w_2)\,.
\]

\end{Definition}

We will see shortly that convex functions are the class of functions
where gradient descent is guaranteed to find an optimal solution.

\begin{figure}
\centering
\includegraphics[width=0.66\textwidth,height=\textheight]{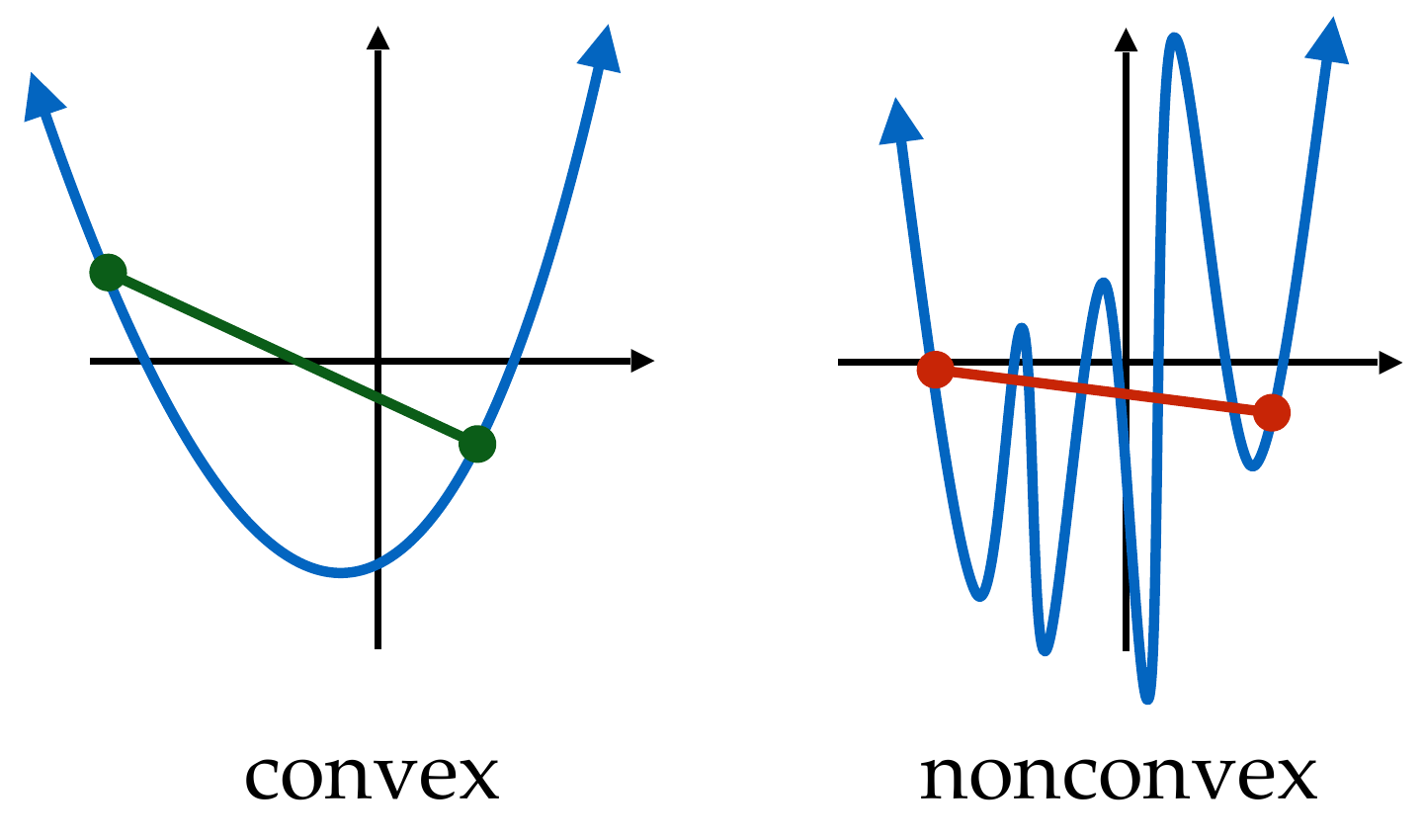}
\caption{Convex vs nonconvex functions.}
\end{figure}

\hypertarget{gradient-descent}{%
\section{Gradient descent}\label{gradient-descent}}

Suppose we want to minimize a differentiable function
\(\Phi\colon\mathbb{R}^d \rightarrow \mathbb{R}\). Most of the
algorithms we will consider start at some point~\(w_0\) and then aim to
find a new point~\(w_1\) with a lower function value. The simplest way
to do so is to find a direction~\(v\) such that~\(\Phi\) is decreasing
when moving along the direction~\(v\). This notion can be formalized by
the following definition:\index{descent direction}

\begin{Definition}

A vector~\(v\) is a \emph{descent direction} for~\(\Phi\) at~\(w_0\) if
\(\Phi(w_0 + tv) < \Phi(w_0)\) for some~\(t>0\).

\end{Definition}

For continuously differentiable functions, it's easy to tell if~\(v\) is
a descent direction: if~\(v^T \nabla \Phi(w_0) <0\) then~\(v\) is a
descent direction.

To see this note that by Taylor's theorem, \[
\Phi(w_0 + \alpha v) = \Phi(w_0) + \alpha \nabla \Phi(w_0 + \tilde{\alpha} v)^T v
\] for some~\(\tilde{\alpha} \in [0,\alpha]\). By continuity,
if~\(\alpha\) is small, we'll
have~\(\nabla \Phi(w_0 + \tilde{\alpha} v)^Tv <0\).
Therefore~\(\Phi(w_0 + \alpha v) < \Phi(w_0)\) and~\(v\) is a descent
direction.

This characterization of descent directions allows us to provide
conditions as to when~\(w\) minimizes \(\Phi\).\index{minimizer!local}

\begin{Proposition}

The point~\(w_\star\) is a local minimizer only if
\(\nabla \Phi(w_\star) = 0\,.\)

\end{Proposition}

Why is this true? Well, the point~\(-\nabla \Phi(w_\star)\) is always a
descent direction if it's not zero. If~\(w_\star\) is a local minimum,
there can be no descent directions. Therefore, the gradient must vanish.

Gradient descent uses the fact that the negative gradient is always a
descent direction to construct an algorithm: repeatedly compute the
gradient and take a step in the opposite direction to minimize~\(\Phi\).

\begin{Algorithm}

\textbf{Gradient Descent}

\begin{itemize}
\tightlist
\item
  Start from an initial point \(w_0 \in \R^d.\)
\item
  At each step \(t=0,1,2,\dots\):

  \begin{itemize}
  \tightlist
  \item
    Choose a step size \(\alpha_t>0\)
  \item
    Set \(w_{t+1} = w_t - \alpha_t \nabla \Phi(w_t)\)
  \end{itemize}
\end{itemize}

\end{Algorithm}

Gradient descent terminates whenever the gradient is so small that the
iterates~\(w_t\) no longer substantially change. Note now that there can
be points where the gradient vanishes but where the function is not
minimized. For example, maxima have this property. In general, points
where the gradient vanishes are called \emph{stationary points}. It is
critically important to remember that not all stationary points are
minimizers.

\begin{figure}
\centering
\includegraphics[width=1\textwidth,height=\textheight]{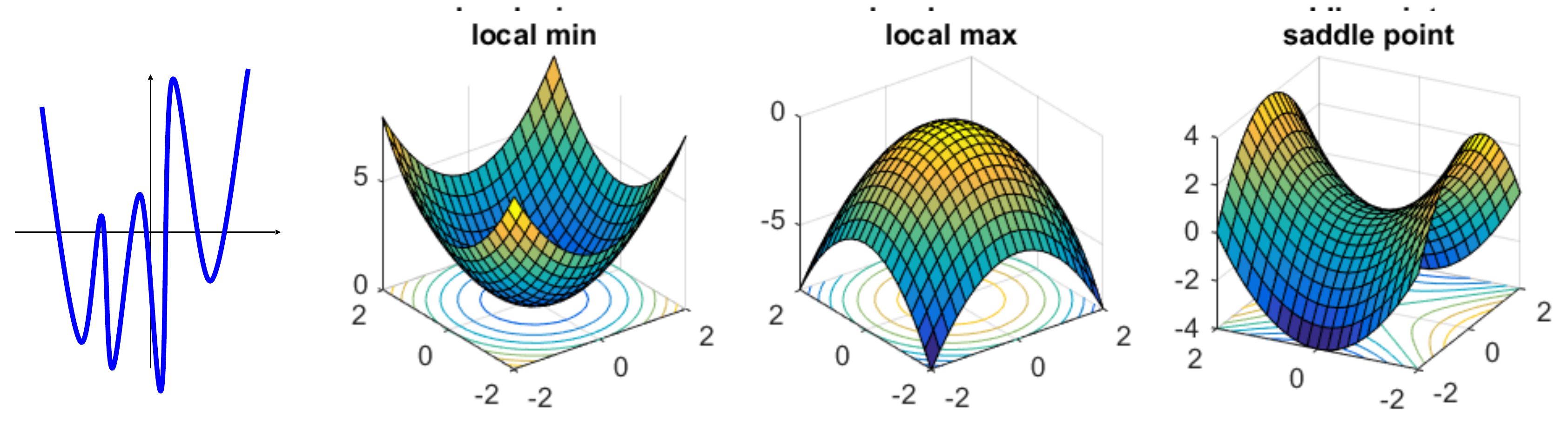}
\caption{Examples of stationary points.}
\end{figure}

For convex~\(\Phi\), the situation is dramatically simpler. This is part
of the reason why convexity is so appealing.

\begin{Proposition}

Let~\(\Phi:\mathbb{R}^d\rightarrow \mathbb{R}\) be a differentiable
convex function. Then~\(w_\star\) is a global minimizer of~\(\Phi\) if
and only if \(\nabla \Phi(w_\star)=0\).

\end{Proposition}

\begin{Proof}

To prove this, we need our definition of convexity: for any
\(\alpha \in [0,1]\) and~\(w\in\mathbb{R}^d\), \[
    \Phi(w_\star + \alpha(w-w_\star)) =  \Phi((1-\alpha)w_\star + \alpha w) \leq (1-\alpha) \Phi(w_\star) + \alpha \Phi(w)
\] Here, the inequality is just our definition of convexity. Now, if we
rearrange terms, we have \[
    \Phi(w) \geq \Phi(w_\star) + \frac{\Phi(w_\star + \alpha(w-w_\star)) - \Phi(w_\star)}{\alpha}
\] Now apply Taylor's theorem: there is now some
\(\tilde{\alpha}\in[0,1]\) such that
\(\Phi(w_\star + \alpha(w-w_\star)) - \Phi(w_\star)=\alpha \nabla \Phi(w_\star+ \tilde{\alpha}(w-w_\star))^T(w-w_\star)\).
Taking the limit as~\(\alpha\) goes to zero yields \[
    \Phi(w) \geq \Phi(w_\star) +\nabla \Phi(w_\star)^T(w-w_\star)\,.
\]

But if~\(\nabla \Phi(w_\star)=0\), that means,
\(\Phi(w) \geq \Phi(w_\star)\) for all~\(w\), and hence~\(w_\star\) is a
global minimizer.

\end{Proof}

Tangent hyperplanes always fall below the graphs of convex functions.

\begin{Proposition}

Let~\(\Phi:\mathbb{R}^d\rightarrow \mathbb{R}\) be a differentiable
convex function. Then for any~\(u\) and~\(v\), we have \[
        \Phi(u) \geq \Phi(v) + \nabla \Phi(v)^T(u-v)\,.
\]

\end{Proposition}

\begin{figure}
\centering
\includegraphics[width=0.6\textwidth,height=\textheight]{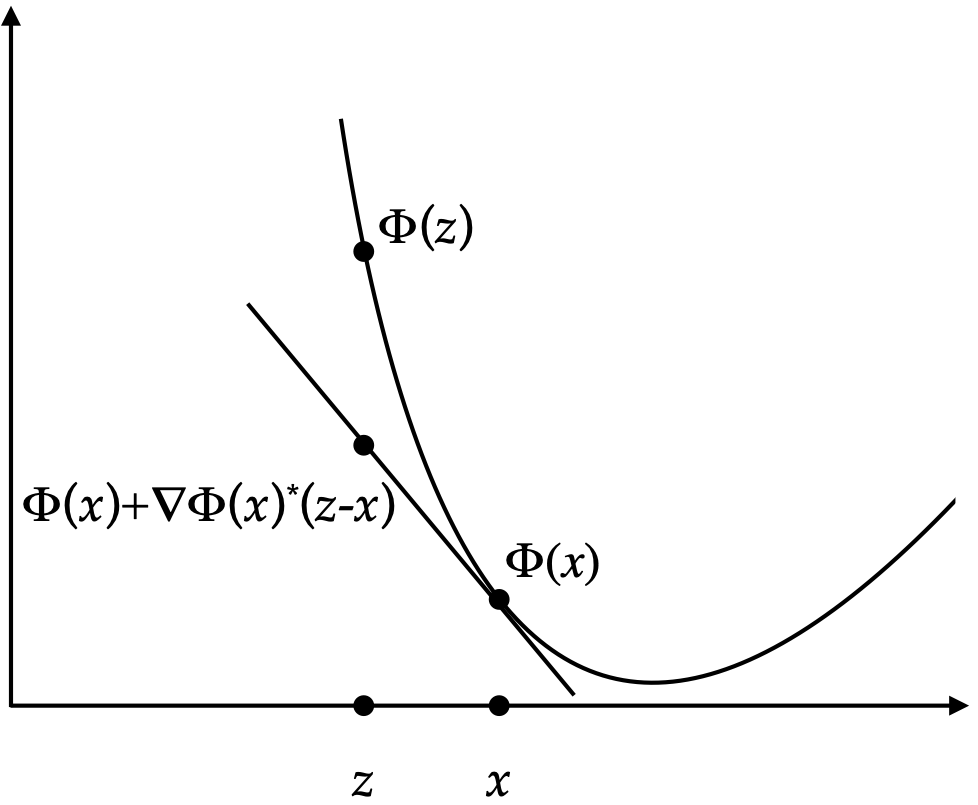}
\caption{Tangent planes to graphs of functions are defined by the
gradient.}
\end{figure}

\hypertarget{a-convex-function-cookbook}{%
\subsection{A convex function
cookbook}\label{a-convex-function-cookbook}}

Testing if a function is convex can be tricky in more than two
dimensions. But here are 5 rules that generate convex functions from
simpler functions. In machine learning, almost all convex cost functions
are built using these rules.

\begin{enumerate}
\def\labelenumi{\arabic{enumi}.}
\item
  All norms are convex (this follows from the triangle inequality).
\item
  If~\(\Phi\) is convex and~\(\alpha \geq 0\), then~\(\alpha \Phi\) is
  convex.
\item
  If~\(\Phi\) and~\(\Psi\) are convex, then~\(\Phi+\Psi\) is convex.
\item
  If~\(\Phi\) and~\(\Psi\) are convex, then
  \(h(w) = \max \{\Phi(w),\Psi(w)\}\) is convex.
\item
  If~\(\Phi\) is convex and~\(A\) is a matrix and~\(b\) is a vector,
  then the function~\(h(w)=\Phi(Aw+b)\) is convex.
\end{enumerate}

All of these properties can be verified using only the definition of
convex functions. For example, consider the 4th property. This is
probably the trickiest of the list. Take two points~\(w_1\) and~\(w_2\)
and \(\alpha \in [0,1]\). Suppose, without loss of generality, that
\(\Phi((1-\alpha)w_1+ \alpha w_2) \geq \Psi((1-\alpha) w_1 + \alpha w_2)\)
\[
\begin{aligned}
    h((1-\alpha) w_1 + \alpha w_2) &= \max \{\Phi((1- \alpha) w_1+ \alpha w_2),\Psi((1-\alpha) w_1 + \alpha w_2)\}\\
    &= \Phi((1-\alpha) w_1+ \alpha w_2)\\
    & \leq  (1-\alpha) \Phi(w_1)+ \alpha \Phi(w_2)\\
    &\leq (1-\alpha) \max\{\Phi(w_1),\Psi(w_1)\} + \alpha \max\{\Phi(w_2),\Psi(w_2)\}\\
    & = (1-\alpha) h(w_1) + \alpha h(w_2)
\end{aligned}
\] Here, the first inequality follows because~\(\Phi\) is convex.
Everything else follows from the definition that~\(h\) is the max of
\(\Phi\) and~\(\Psi\). The reader should verify the other four
assertions as an exercise. Another useful exercise is to verify that the
SVM cost in the next section is convex by just using these five basic
rules and the fact that the one dimensional function~\(f(x)=mx+b\) is
convex for any scalars~\(m\) and~\(b\).

\hypertarget{applications-to-empirical-risk-minimization}{%
\section{Applications to empirical risk
minimization}\label{applications-to-empirical-risk-minimization}}

For decision theory problems, we studied the zero-one loss that counts
errors: \[
    \loss(\hat{y},y) = \mathbb{1}\{ y\hat{y} < 0\}
\] Unfortunately, this loss is not useful for the gradient method. The
gradient is zero almost everywhere. As we discussed in the chapter on
supervised learning, machine learning practice always turns to surrogate
losses that are easier to optimize. Here we review three popular
choices, all of which are convex loss functions.\index{loss!zero-one}
Each choice leads to a different important optimization problem that has
been studied in its own right.

\hypertarget{the-support-vector-machine}{%
\subsection{The support vector
machine}\label{the-support-vector-machine}}

\index{support vector machine}\index{loss!hinge}

Consider the canonical problem of support vector machine classification.
We are provided pairs~\(\left(x_{i},y_{i}\right)\), with
\(x_{i}\in\mathbb{R}^{d}\) and~\(y_{i}\in\left\{ -1,1\right\}\) for
\(i=1,\ldots n\) (Note, the~\(y\) labels are now in~\(\{-1,1\}\) instead
of \(\{0,1\}\).) The goal is to find a vector~\(w\in \mathbb{R}^d\) such
that: \[
\begin{aligned}
w^{T}x_{i} & >  0\quad\text{ for }y_{i}=1\\
w^{T}x_{i} & < 0\quad\text{ for }y_{i}=-1
\end{aligned}
\] Such a~\(w\) defines a half-space where we believe all of the
positive examples lie on one side and the negative examples on the
other.

Rather than classifying these points exactly, we can allow some slack.
We can pay a penalty of~\(1-y_i w^T x_i\) points that are not strongly
classified. This motivates the hinge loss we encountered earlier and
leads to the \emph{support vector machine objective}: \[
    \text{minimize}_w\quad  \sum_{i=1}^n \max\left\{1-y_i w^T x_i,0\right\} \,.
\] Defining the function~\(e(z) = \mathbb{1}\{ z\leq 1 \},\) we can
compute that the gradient of the SVM cost is \[
    -\sum_{i=1}^n e(y_i w^T x_i) y_i x_i \,.
\] Hence, gradient descent for this ERM problem would follow the
iteration \[
    w_{t+1} = w_t + \alpha \sum_{i=1}^n e(y_i w^T x_i) y_i x_i
\] Although similar, note that this isn't quite the perceptron method
yet. The time to compute one gradient step is~\(O(n)\) as we sum over
all \(n\) inner products. We will soon turn to the stochastic gradient
method that has constant iteration complexity and will subsume the
perceptron algorithm.

\hypertarget{logistic-regression}{%
\subsection{Logistic regression}\label{logistic-regression}}

\index{logistic regression}\index{loss!logistic}

Logistic regression is equivalent to using the loss function \[
\loss(\hat y,y) = \log \left(1+\exp(-y\hat y)\right)\,.
\] Note that even though this loss has a probabilistic interpretation,
it can also just be seen as an approximation to the error-counting
zero-one loss.

\hypertarget{least-squares-classification}{%
\subsection{Least squares
classification}\label{least-squares-classification}}

\index{least squares}\index{loss!squared}

Least squares classification uses the loss function \[
\loss(\hat y,y) =  \tfrac{1}{2}(\hat y-y)^2\,.
\] Though this might seem like an odd approximation to the
error-counting loss, it leads to the maximum a posteriori (MAP) decision
rule when minimizing the population risk. Recall the MAP rule selects
the label that has highest probability conditional on the observed
data.\index{MAP}\index{maximum a posteriori}

It is helpful to keep the next picture in mind that summarizes how each
of these different loss functions approximate the zero-one loss. We can
ensure that the squared loss is an upper bound on the zero-one loss by
dropping the factor~\(1/2.\)

\begin{figure}
\centering
\includegraphics{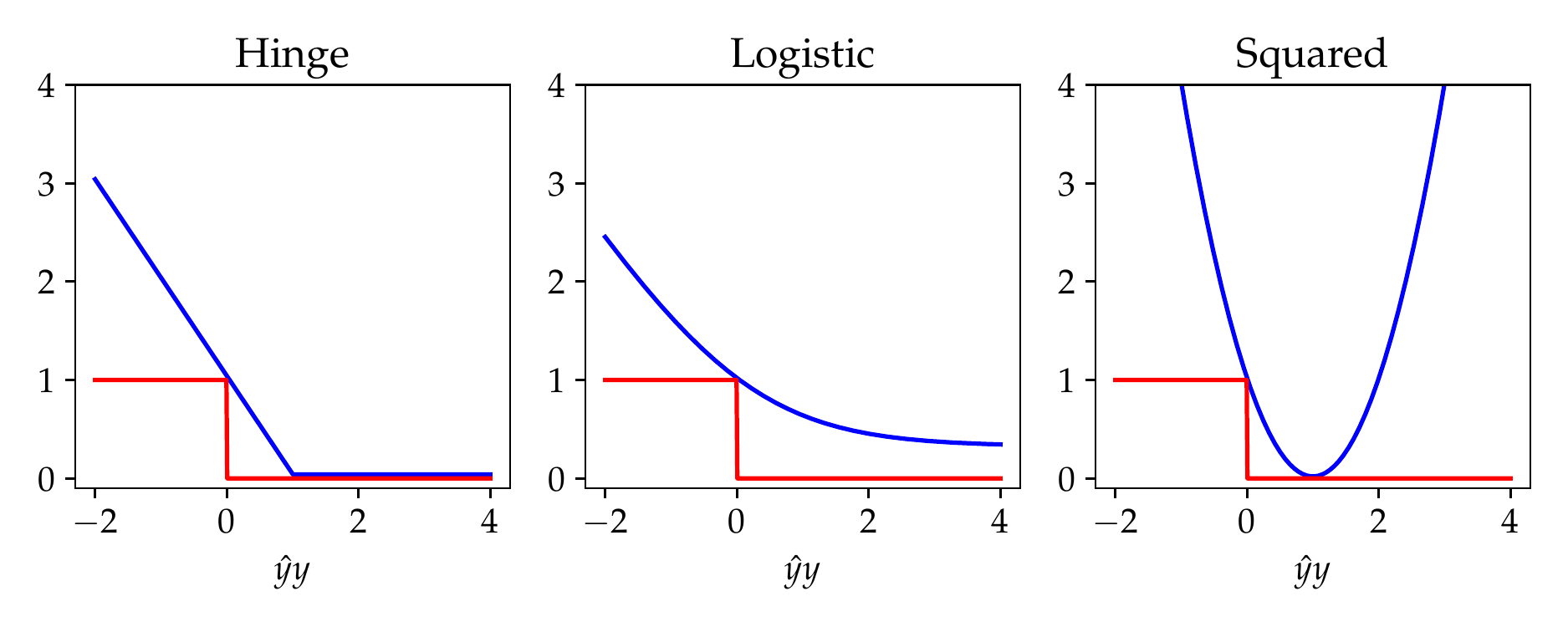}
\caption{Three different convex losses compared with the zero-one loss.}
\end{figure}

\hypertarget{insights-from-quadratic-functions}{%
\section{Insights from quadratic
functions}\label{insights-from-quadratic-functions}}

Quadratic functions are the prototypical example that motivate
algorithms for differentiable optimization problems. Though not all
insights from quadratic optimization transfer to more general functions,
there are several key features of the dynamics of iterative algorithms
on quadratics that are notable. Moreover, quadratics are a good test
case for reasoning about optimization algorithms: if a method doesn't
work well on quadratics, it typically won't work well on more
complicated nonlinear optimization problems. Finally, note that ERM with
linear functions and a squared loss is a quadratic optimization problem,
so such problems are indeed relevant to machine learning practice.

The general quadratic optimization problem takes the form \[
    \Phi(w) = \tfrac{1}{2} w^T Q w - p^T w + r\,,
\] where~\(Q\) is a symmetric matrix,~\(p\) is a vector, and~\(r\) is a
scalar. The scalar~\(r\) only affects the value of the function, and
plays no role in the dynamics of gradient descent. The gradient of this
function is \[
    \nabla \Phi(w) = Qw - p\,.
\] The stationary points of~\(\Phi\) are the~\(w\) where~\(Qw = p\).
If~\(Q\) is full rank, there is a unique stationary point.

The gradient descent algorithm for quadratics follows the iterations \[
    w_{t+1} = w_t - \alpha (Qw_t - p)\,.
\] If we let~\(w_\star\) be \emph{any} stationary point of~\(\Phi\), we
can rewrite this iteration as \[
    w_{t+1}-w_\star = (I-\alpha Q) (w_t - w_\star)\,.
\] Unwinding the recursion yields the ``closed form'' formula for the
gradient descent iterates \[
    w_{t}-w_\star = (I-\alpha Q)^t (w_0 - w_\star)\,.
\]

This expression reveals several possible outcomes. Let
\(\lambda_1 \geq \lambda_2 \geq \ldots \geq \lambda_d\) denote the
eigenvalues of~\(Q\). These eigenvalues are real because~\(Q\) is
symmetric. First suppose that~\(Q\) has a negative
eigenvalue~\(\lambda_d<0\) and~\(v\) is an eigenvector such
that~\(Qv = \lambda_d v\). Then
\((I-\alpha Q)^t v = (1+\alpha |\lambda_d|)^t v\) which tends to
infinity as~\(t\) grows. This is because~\(1+\alpha |\lambda_d|\) is
greater than~\(1\) if~\(\alpha>0\). Hence,
if~\(\langle v, w_0-w_\star \rangle \neq 0\), gradient descent
\emph{diverges}. For a random initial condition~\(w_0\), we'd expect
this dot product will not equal zero, and hence gradient descent will
almost surely not converge from a random initialization.

In the case that all of the eigenvalues of~\(Q\) are positive, then
choosing~\(\alpha\) greater than zero and less than~\(1/\lambda_1\) will
ensure that~\(0\leq 1-\alpha \lambda_k <1\) for all~\(k\). In this case,
the gradient method converges exponentially quickly to the optimum
\(w_\star:\) \[
\begin{aligned}
    \Vert w_{t+1}-w_\star\Vert &= \Vert (I-\alpha Q) (w_t - w_\star)\Vert\\
    &\leq \Vert I-\alpha Q\Vert  \Vert w_t - w_\star \Vert
    \leq \left(1-\frac{\lambda_d}{\lambda_1}\right)\Vert w_t-w_\star\Vert\,.
    \end{aligned}
\] When the eigenvalues of~\(Q\) are all positive, the function~\(\Phi\)
is strongly convex. Strongly convex functions turn out to be the set of
functions where gradient descent with a constant step size converges
exponentially from any starting point.

Note that the ratio of~\(\lambda_1\) to~\(\lambda_d\) governs how
quickly all of the components converge to~\(0\). Defining the
\emph{condition number} of~\(Q\) to be~\(\kappa = \lambda_1/\lambda_d\)
and setting the step size \(\alpha = 1/\lambda_1\), gives the
bound\index{condition number} \[
 \Vert w_{t}-w_\star\Vert \leq \left(1 - \kappa^{-1}\right)^t  \Vert w_{0}-w_\star \Vert\,.
\] This rate reflects what happens in practice: when there are small
singular values, gradient descent tends to bounce around and oscillate
as shown in the figure below. When the condition number of~\(Q\) is
small, gradient descent makes rapid progress towards the optimum.

\includegraphics[width=0.4\textwidth,height=\textheight]{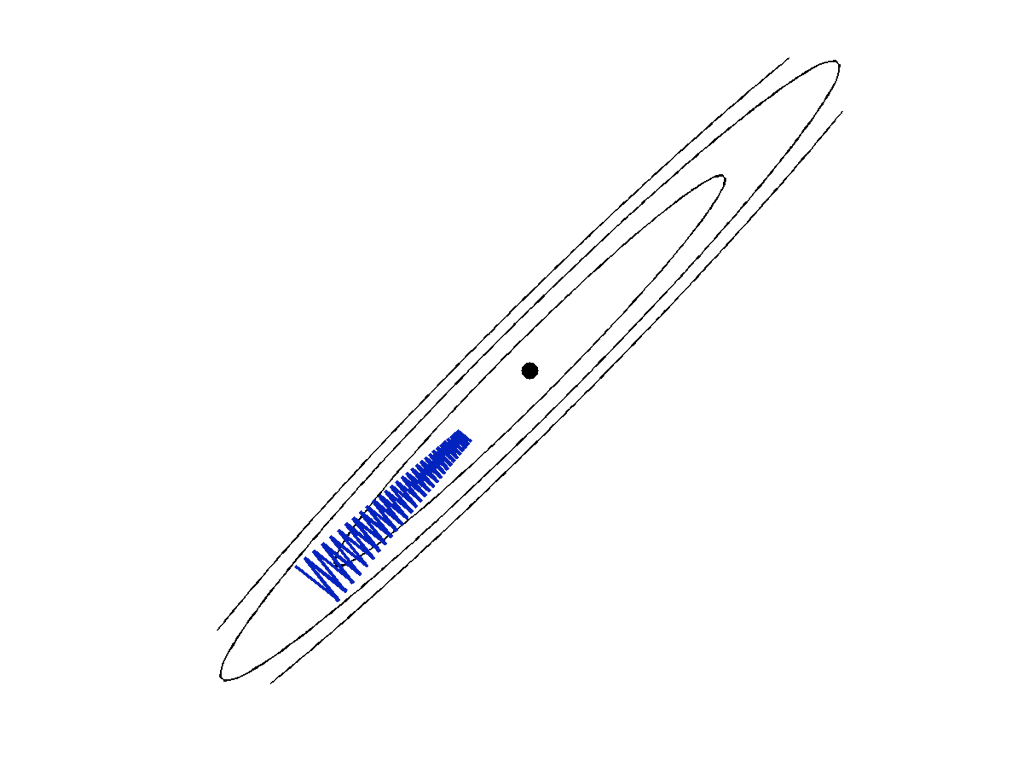}
\includegraphics[width=0.3\textwidth,height=\textheight]{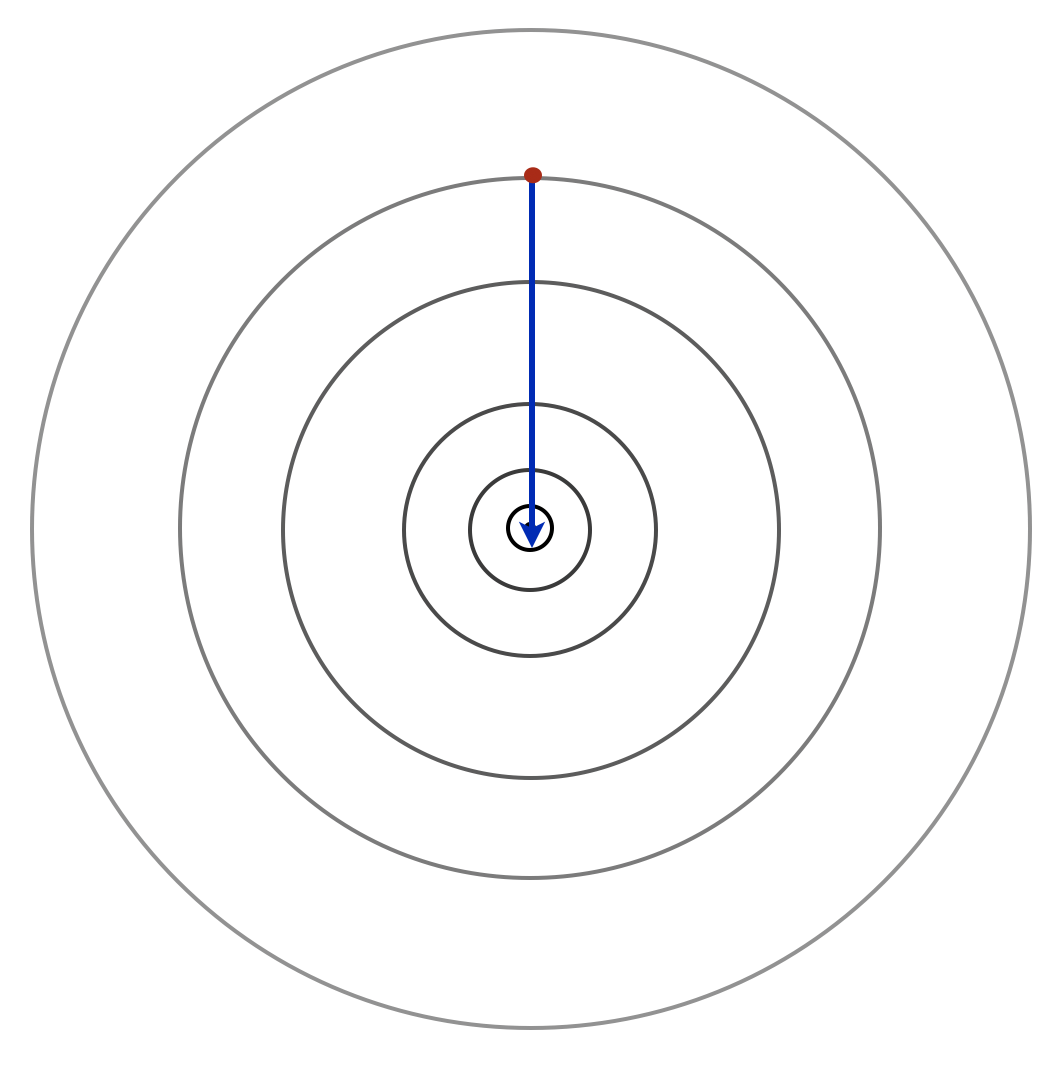}

There is one final case that's worth considering. When all of the
eigenvalues of~\(Q\) are nonnegative but some of them are zero, the
function~\(\Phi\) is convex but not strongly convex. In this case,
exponential convergence to a unique point cannot be guaranteed. In
particular, there will be an infinite number of global minimizers of
\(\Phi\). If~\(w_\star\) is a global minimizer and~\(v\) is any vector
with \(Qv=0\), then~\(w_\star+v\) is also a global minimizer. However,
in the span of the eigenvectors corresponding to positive eigenvalues,
gradient descent still converges exponentially. For general convex
functions, it will be important to consider different parts of the
parameter space to fully understand the dynamics of gradient methods.

\hypertarget{stochastic-gradient-descent}{%
\section{Stochastic gradient
descent}\label{stochastic-gradient-descent}}

The stochastic gradient method is one of the most popular algorithms for
contemporary data analysis and machine learning. It has a long history
and has been ``invented'' several times by many different communities
(under the names ``least mean squares,'' ``backpropagation,'' ``online
learning,'' and the ``randomized Kaczmarz method''). Most researchers
attribute this algorithm to the initial work of Robbins and Monro from
1951 who solved a more general problem with the same
method.\citep{robbins1951stochastic}

\index{gradient descent!stochastic}

Consider again our main goal of minimizing the empirical risk with
respect to a vector of parameters~\(w\), and consider the simple case of
linear classification where~\(w\) is~\(d\)-dimensional and \[
f(x_i;w) = w^Tx_i\,.
\]

The idea behind the stochastic gradient method is that since the
gradient of a sum is the sum of the gradients of the summands, each
summand provides useful information about how to optimize the total sum.
Stochastic gradient descent minimizes empirical risk by following the
gradient of the risk evaluated on a \emph{single, random} example.

\begin{Algorithm}

\textbf{Stochastic Gradient Descent}

\begin{itemize}
\tightlist
\item
  Start from an initial point \(w_0 \in \mathbb{R}^d\).
\item
  At each step \(t=0,1,2,\dots\):

  \begin{itemize}
  \tightlist
  \item
    Choose a step size \(\alpha_t>0\) and random index \(i \in [n]\).
  \item
    Set \(w_{t+1} = w_t - \alpha_t \nabla_{w_t} \loss(f(x_i;w_t),y_i)\)
  \end{itemize}
\end{itemize}

\end{Algorithm}

The intuition behind this method is that by following a descent
direction in expectation, we should be able to get close to the optimal
solution if we wait long enough. However, it's not quite that simple.
Note that even when the gradient of the sum is zero, the gradients of
the individual summands may not be. The fact that~\(w_\star\) is no
longer a fixed point complicates the analysis of the method.

\hypertarget{example-revisiting-the-perceptron}{%
\subsection{Example: revisiting the
perceptron}\label{example-revisiting-the-perceptron}}

Let's apply the stochastic gradient method to the support vector machine
cost loss. We initialize our half-space at some~\(w_{0}\). At iteration
\(t\), we choose a random data point~\((x_{i},y_{i})\) and update \[
w_{t+1}= w_{t}+\eta\begin{cases}
y_{i}x_{i} &\text{if}~y_{i}w_{t}^{T}x_{i}\leq 1\\
0 & \text{otherwise}
\end{cases}
\] As we promised earlier, we see that using stochastic gradient descent
to minimize empirical risk with a hinge loss is completely equivalent to
Rosenblatt's Perceptron algorithm.

\hypertarget{example-computing-a-mean}{%
\subsection{Example: computing a mean}\label{example-computing-a-mean}}

Let's now try to examine the simplest example possible. Consider
applying the stochastic gradient method to the function \[
    \frac{1}{2n} \sum_{i=1}^n  (w-y_i)^2\,,
\] where~\(y_1,\dots,y_n\) are fixed scalars. This setup corresponds to
a rather simple classification problem where the~\(x\) features are all
equal to~\(1\). Note that the gradient of one of the increments is \[
    \nabla \loss (f(x_i;w),y) = w-y_i\,.
\]

To simplify notation, let's imagine that our random samples are coming
to us in order~\(\{1,2,3,4,...\}\) Start with~\(w_1=0\), use the step
size \(\alpha_k = 1/k\). We can then write out the first few equations:
\[
\begin{aligned}
w_{2} & =  w_{1}-w_{1}+y_{1}=y_{1}\\
w_{3} & =  w_{2}-\frac{1}{2}\left(w_{2}-y_{2}\right)=\frac{1}{2}y_{1}+\frac{1}{2}y_{2}\\
w_{4} & =   w_{3}-\frac{1}{3}\left(w_{3}-y_{3}\right) = \frac{1}{3}y_{1}+\frac{1}{3}y_{2}+\frac{1}{3}y_{3}
\end{aligned}
\] Generalizing from here, we can conclude by induction that \[
w_{k+1} =  \left(\frac{k-1}{k}\right)w_{k}+\frac{1}{k}y_{k}=\frac{1}{k}\sum_{i=1}^{k}y_{i}\,.
\] After~\(n\) steps,~\(w_n\) is the mean of the~\(y_i\), and you can
check by taking a gradient that this is indeed the minimizer of the ERM
problem.

The~\(1/k\) step size was the originally proposed step size by Robbins
and Monro. This simple example justifies why: we can think of the
stochastic gradient method as computing a running average. Another
motivation for the~\(1/k\) step size is that the steps tend to zero, but
the path length is infinite.

Moving to a more realistic random setting where the data might arrive in
any order, consider what happens when we run the stochastic gradient
method on the function \[
    R(w) = \tfrac{1}{2} \mathbb{E}[ (w-Y)^2]\,.
\] Here~\(Y\) is some random variable with mean~\(\mu\) and variance
\(\sigma^2\). If we run for~\(k\) steps with i.i.d. samples~\(Y_i\) at
each iteration, the calculation above reveals that \[
    w_k = \frac{1}{k} \sum_{i=1}^k Y_i\,.
\] The associated cost is \[
    R(w_k) = \frac{1}{2} \E\left[ \left(\frac{1}{k} \sum_{i=1}^k Y_i-Y\right)^2\right]
    = \frac{1}{2k}\sigma^2 + \frac{1}{2}\sigma^2\,.
\] Compare this to the minimum achievable risk~\(R_\star\). Expanding
the definition \[
    R(w) = \tfrac{1}{2} \mathbb{E}[w^2 - 2 Y w + Y^2] = \tfrac{1}{2}w^2 - \mu w + \tfrac{1}{2}\sigma^2 + \tfrac{1}{2}\mu^2\,,
\] we find that the minimizer is~\(w_\star=\mu\). Its cost is \[
    R_\star=R(w_\star) = \frac{1}{2} \sigma^2\,,
\] and after~\(n\) iterations, we have the expected \emph{optimality
gap} \[
    \E\left[R(w_n)-R_\star\right] = \frac{1}{2n}\sigma^2\,.
\] This is the best we could have achieved using any estimator for
\(w_\star\) given the collection of random draws. Interestingly, the
incremental ``one-at-a-time'' method finds as good a solution as one
that considers all of the data together. This basic example reveals a
fundamental limitation of the stochastic gradient method: we can't
expect to generically get fast convergence rates without additional
assumptions. Statistical fluctuations themselves prevent the optimality
gap from decreasing exponentially quickly.

This simple example also helps give intuition on the convergence as we
sample stochastic gradients. The figure below plots an example of each
individual term in the summand, shaded with colors of blue to
distinguish each term. The minimizing solution is marked with a red
star. To the far left and far right of the figure, all of the summands
will have gradients pointing in the same direction to the solution.
However, as our iterate gets close to the optimum, we will be pointed in
different directions depending on which gradient we sample. By reducing
the step size, we will be more likely to stay close and eventually
converge to the optimal solution.

\begin{figure}
\centering
\includegraphics{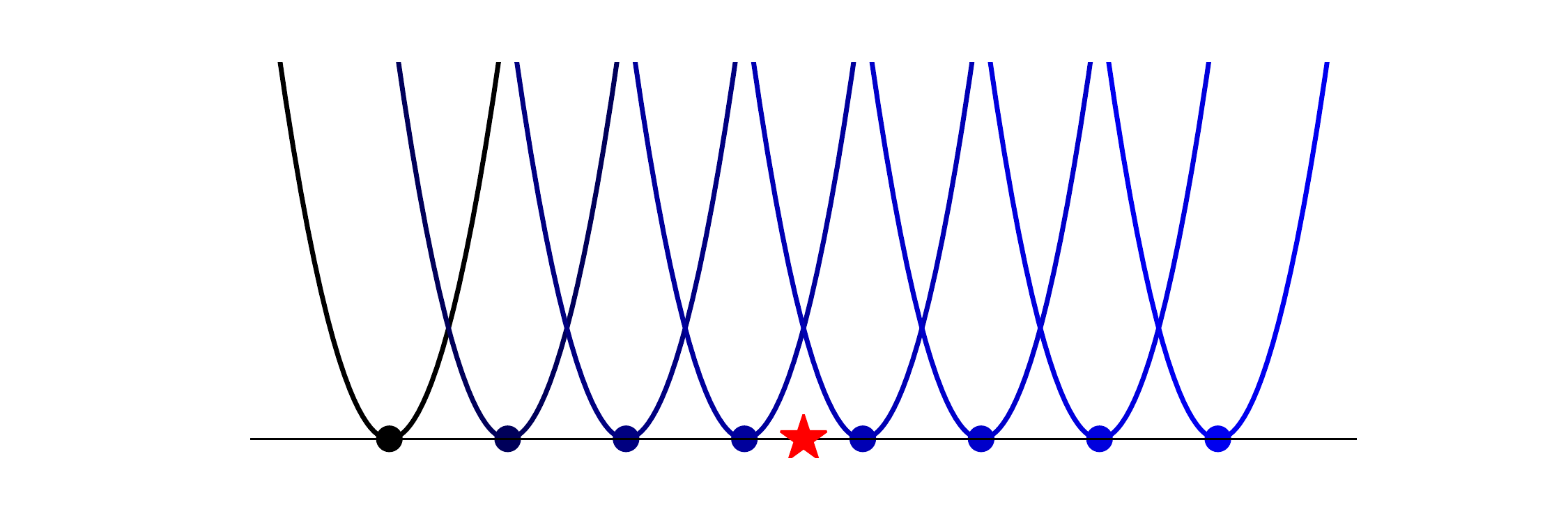}
\caption{Plot of the different increments of
\(\tfrac{1}{2n} \sum_{i=1}^n (w-y_i)^2\). The red star denotes the
optimal solution.}
\end{figure}

\hypertarget{example-stochastic-minimization-of-a-quadratic-function}{%
\subsection{Example: Stochastic minimization of a quadratic
function}\label{example-stochastic-minimization-of-a-quadratic-function}}

Let's consider a more general version of stochastic gradient descent
that follows the gradient plus some unbiased noise. We can model the
algorithm as minimizing a function~\(\Phi(w)\) where we follow a
direction \(\nabla \Phi(w) + \nu\) where~\(\nu\) is a random vector with
mean zero and \(\E[\Vert \nu \Vert^2] = \sigma^2\). Consider the special
case where \(\Phi(w)\) is a quadratic function \[
    \Phi(w) = \tfrac{1}{2} w^T Q w - p^T w +r\,.
\] Then the iterations take the form \[
    w_{t+1} = w_t - \alpha (  Qw_t -p  + \nu_t)\,.
\] Let's consider what happens when~\(Q\) is positive definite with
maximum eigenvalue~\(\lambda_1\) and minimum
eigenvalue~\(\lambda_d >0\). Then, if~\(w_\star\) is a global minimizer
of~\(Q\), we can again rewrite the iterations as \[
    w_{t+1}-w_\star = (I-\alpha Q) (w_t-w_\star) - \alpha\nu_t\,.
\] Since we assume that the noise~\(\nu_t\) is independent of all of the
\(w_k\) with~\(k \leq t\), we have \[
    \E[ \Vert w_{t+1}-w_\star \Vert^2 ]\leq \Vert I-\alpha Q \Vert^2 \E[ \Vert w_t-w_\star\Vert^2 ] + \alpha^2\sigma^2\,.
\] which looks like the formula we derived for quadratic functions, but
now with an additional term from the noise. Assuming
\(\alpha < 1/\lambda_1\), we can unwind this recursion to find \[
    \E[ \Vert w_{t}-w_\star \Vert^2 ]\leq ( 1-\alpha \lambda_d)^{2t} \Vert w_0-w_\star\Vert^2  + \frac{\alpha \sigma^2}{\lambda_d}\,.
\] From this expression, we see that gradient descent converges
exponentially quickly to some ball around the optimal solution. The
smaller we make~\(\alpha\), the closer we converge to the optimal
solution, but the rate of convergence is also slower for smaller
\(\alpha\). This tradeoff motivates many of the step size selection
rules in stochastic gradient descent. In particular, it is common to
start with a large step size and then successively reduce the step size
as the algorithm progresses.

\hypertarget{tricks-of-the-trade}{%
\subsection{Tricks of the trade}\label{tricks-of-the-trade}}

In this section, we describe key engineering techniques that are useful
for tuning the performance of stochastic gradient descent. Every machine
learning practitioner should know these simple tricks.

\textbf{Shuffling.} Even though we described the stochastic gradient
method a sampling each gradient with replacement from the increments, in
practice better results are achieved by simply randomly permuting the
data points and then running SGD in this random order. This is called
``shuffling,'' and even a single shuffle can eliminate the pathological
behavior we described in the example with highly correlated data.
Recently, beginning with work by Gürbüzbalaban et al., researchers have
shown that in theory, shuffling outperforms independent sampling of
increments\citep{gurbuzbalaban2019random}. The arguments for
without-replacement sampling remain more complicated than the
with-replacement derivations, but optimal sampling for SGD remains an
active area of optimization research.

\textbf{Step size selection.} Step size selection in SGD remains a hotly
debated topic. We saw above a decreasing stepsize~\(1/k\) solved our
simple one dimensional ERM problem. However, a rule that works for an
unreasonable number of cases is to simply pick the largest step size
which does not result in divergence. This step will result in a model
that is not necessarily optimal, but significantly better than
initialization. By slowly reducing the step size from this initial large
step size to successively smaller ones, we can zero in on the optimal
solution.

\textbf{Step decay.} The step size is usually reduced after a fixed
number of passes over the training data. A pass over the entire dataset
is called an \emph{epoch}. In an epoch, some number of iterations are
run, and then a choice is made about whether to change the step size. A
common strategy is to run with a constant step size for some fixed
number of iterations \(T\), and then reduce the step size by a constant
factor~\(\gamma\). Thus, if our initial step size is~\(\alpha\), on
the~\(k\)th epoch, the step size is~\(\alpha \gamma^{k-1}\). This method
is often more robust in practice than the diminishing step size rule.
For this step size rule, a reasonable heuristic is to choose~\(\gamma\)
between~\(0.8\) and~\(0.9\). Sometimes people choose rules as aggressive
as~\(\gamma = 0.1\).

Another possible schedule for the step size is called \emph{epoch
doubling}. In epoch doubling, we run for~\(T\) steps with step
size~\(\alpha\), then run~\(2T\) steps with step size~\(\alpha/2\), and
then~\(4T\) steps with step size~\(\alpha/4\) and so on. Note that this
provides a piecewise constant approximation to the
function~\(\alpha/k\).

\textbf{Minibatching.} A common technique used to take advantage of
parallelism is called \emph{minibatching}. A minibatch is an average of
many stochastic gradients. Suppose at each iteration we sample a
\(\mathrm{batch}_k\) with~\(m\) data points. The update rule then
becomes \[
    w_{k+1} = w_{k} -\alpha_k  \frac{1}{m}\sum_{j\in \mathrm{batch}_k} \nabla_w  \loss(f(x_j;w_k),y_j)\,.
\] Minibatching reduces the variance of the stochastic gradient estimate
of the true gradient, and hence tends to be a better descent direction.
Of course, there are tradeoffs in total computation time versus the size
of the minibatch, and these typically need to be handled on a case by
case basis.

\textbf{Momentum.} Finally, we note that one can run stochastic gradient
descent with \emph{momentum}. Momentum mixes the current gradient
direction with the previously taken step. The idea here is that if the
previous weight update was good, we may want to continue moving along
this direction. The algorithm iterates are defined as \[
w_{k+1}=w_{k}-\alpha_{k}g_k\left(w_{k}\right) + \beta (w_{k}+w_{k-1})\,,
\] where~\(g_k\) denotes a stochastic gradient. In practice, these
methods are very successful. Typical choices for~\(\beta\) here are
between~\(0.8\) and~\(0.95\). Momentum can provide significant
accelerations, and should be considered an option in any implementation
of SGM.

\hypertarget{the-sgd-quick-start-guide}{%
\subsection{The SGD quick start guide}\label{the-sgd-quick-start-guide}}

Newcomers to stochastic gradient descent often find all of these design
choices daunting, and it's useful to have simple rules of thumb to get
going. We recommend the following:

\begin{enumerate}
\def\labelenumi{\arabic{enumi}.}
\tightlist
\item
  Pick as large a minibatch size as you can given your computer's RAM.
\item
  Set your momentum parameter to either 0 or 0.9. Your call!
\item
  Find the largest constant stepsize such that SGD doesn't diverge. This
  takes some trial and error, but you only need to be accurate to within
  a factor of 10 here.
\item
  Run SGD with this constant stepsize until the empirical risk plateaus.
\item
  Reduce the stepsize by a constant factor (say, 10)
\item
  Repeat steps 4 and 5 until you converge.
\end{enumerate}

While this approach may not be the most optimal in all cases, it's a
great starting point and is good enough for probably 90\% of
applications we've encountered.

\hypertarget{analysis-of-the-stochastic-gradient-method}{%
\section{Analysis of the stochastic gradient
method}\label{analysis-of-the-stochastic-gradient-method}}

We now turn to a theoretical analysis of the general stochastic gradient
method. Before we proceed, let's set up some conventions. We will assume
that we are trying to minimize a convex function
\(R:\mathbb{R}^d \rightarrow \mathbb{R}\). Let~\(w_\star\) denote any
optimal solution of~\(R\). We will assume we gain access at every
iteration to a \emph{stochastic function} \(g(w;\xi)\) such that \[
    \mathbb{E}_\xi[g(w; \xi)] = \nabla R(w)\,.
\] Here~\(\xi\) is a random variable which determines what our direction
looks like. We additionally assume that these stochastic gradients are
\emph{bounded} so there exists a non-negative constants~\(B\) such that
\[
     \Vert g(w;\xi)\Vert \leq  B\,.
\]

We will study the stochastic gradient iteration \[
w_{t+1}=w_{t}-\alpha_{t}g\left(w_{t};\xi_t\right)\,.
\] Throughout, we will assume that the sequence~\(\{\xi_j\}\) is
selected i.i.d. from some fixed distribution.

We begin by expanding the distance to the optimal solution: \[
\begin{aligned}
 \Vert w_{t+1} - w_\star \Vert^2 &=
 \Vert  w_t - \alpha_t g_t(w_t;\xi_t) - w_\star \Vert^2  \\
 &= \Vert w_t - w_\star\Vert^2 -2 \alpha_t \langle g_t( w_t;\xi_t), w_t- w_\star \rangle + \alpha_t^2 \Vert g_t(w_t; \xi_t) \Vert^2
\end{aligned}
\]

We deal with each term in this expansion separately. First note that if
we apply the law of iterated expectation \[
\begin{aligned}
    \mathbb{E}[ \langle g_t(w_t;\xi_t), w_t - w_\star \rangle ]
    &= \mathbb{E}\left[\mathbb{E}_{\xi_t}[  \langle g_t(w_t;\xi_t), w_t - w_\star \rangle \mid  \xi_0,\ldots,\xi_{t-1} ]\right]\\
    &= \mathbb{E}\left[  \langle\mathbb{E}_{\xi_t}[  g_t(w_t;\xi_t)  \mid  \xi_0,\ldots,\xi_{t-1} ] , w_t - w_\star \rangle\right]\\
    &= \mathbb{E}\left[ \langle \nabla R(w_t) , w_t - w_\star \rangle\right]\,.
\end{aligned}
\] Here, we are simply using the fact that~\(\xi_t\) being independent
of all of the preceding~\(\xi_i\) implies that it is independent
of~\(w_t\). This means that when we iterate the expectation, the
stochastic gradient can be replaced by the gradient. The last term we
can bound using our assumption that the gradients are bounded: \[
\begin{aligned}
\mathbb{E}[ \Vert g(w_t; \xi_t)\Vert^2]  \leq B^2
\end{aligned}
\]

Letting~\(\delta_t := \mathbb{E}[\Vert w_{t}-w_\star\Vert^2]\), this
gives \[
 \delta_{t+1} \leq \delta_t - 2 \alpha_t \mathbb{E}\left[ \langle \nabla R(w_t) , w_t - w_\star \rangle\right] + \alpha_t^2 B^2\,.
\]

Now let~\(\lambda_t = \sum_{j=0}^t \alpha_j\) denote the sum of all the
step sizes up to iteration~\(t\). Also define the average of the
iterates weighted by the step size \[
    \bar{w}_t = \lambda_t^{-1} \sum_{j=0}^t \alpha_j w_j\,.
\] We are going to analyze the deviation of~\(R(\bar{w}_t)\) from
optimality.

Also let~\(\rho_0 = \Vert w_0-w_\star\Vert\).~\(\rho_0\) is the initial
distance to an optimal solution. It is not necessarily a random
variable.

To proceed, we just expand the following expression: \[
\begin{aligned}
\mathbb{E}\left[  R\left (\bar{w}_T \right)  - R(w_\star)  \right]  & \leq
    \mathbb{E}\left[ \lambda_T^{-1} \sum_{t=0}^T \alpha_t( R(w_t)  - R(w_\star) ) \right] \\
    &\leq   \lambda_T^{-1} \sum_{t=0}^T \alpha_t \mathbb{E}[ \langle \nabla R(x_t), w_t-w_\star \rangle ]\\
    &\leq   \lambda_T^{-1} \sum_{t=0}^T \tfrac{1}{2}(\delta_{t}-\delta_{t+1})+ \tfrac{1}{2} \alpha_t^2 B^2\\
    & = \frac{\delta_0 - \delta_{T+1} +  B^2 \sum_{t=0}^T \alpha_t^2}{2 \lambda_T}\\
    & \leq \frac{\rho_0^2  + B^2 \sum_{t=0}^T \alpha_t^2}{2 \sum_{t=0}^T \alpha_t}
\end{aligned}
\] Here, the first inequality follows because~\(R\) is convex (the line
segments lie above the function, i.e.,
\(R(w_\star) \geq R(w_t) + \langle\nabla R(w_t), w_t-w_\star\rangle\)).
The second inequality uses the fact that gradients define tangent planes
to~\(R\) and always lie below the graph of~\(R\), and the third
inequality uses the recursion we derived above for~\(\delta_t\).

The analysis we saw gives the following result.\citep{Nemirovski09}

\begin{Theorem}

Suppose we run the SGM on a convex function~\(R\) with minimum value
\(R_\star\) for~\(T\) steps with step size~\(\alpha\). Define \[
\alpha_{\mathrm{opt}} = \frac{\rho_0}{B\sqrt{T}}\qquad\text{and}\qquad
\theta =  \frac\alpha{\alpha_{\mathrm{opt}}}\,.
\] Then, we have the bound \[
\mathbb{E}[R(\bar{w}_T) - R_\star] \leq \left(\tfrac{1}{2} \theta + \tfrac{1}{2} \theta^{-1}\right) \frac{B \rho_0}{\sqrt{T}}  \,.
\]

\end{Theorem}

This proposition asserts that we pay linearly for errors in selecting
the optimal constant step size. If we guess a constant step size that is
two-times or one-half of the optimal choice, then we need to run for at
most twice as many iterations. The optimal step size is found by
minimizing our upper bound on the suboptimality gap. Other step sizes
could also be selected here, including diminishing step size. But the
constant step size turns out to be optimal for this upper bound.

What are the consequences for risk minimization? First, for
\emph{empirical risk}, assume we are minimizing a convex loss function
and searching for a linear predictor. Assume further that there exists a
model with zero empirical risk. Let~\(C\) be the maximum value of the
gradient of the loss function,~\(D\) be the largest norm of any
example~\(x_i\) and let~\(\rho\) denote the minimum norm~\(w\) such
that~\(R_S[w]=0\). Then we have \[
\mathbb{E}[R_S[\bar{w}_T]] \leq \frac{CD\rho}{\sqrt{T}}  \,.
\] we see that with appropriately chosen step size, the stochastic
gradient method converges at a rate of~\(1/\sqrt{T}\), the same rate of
convergence observed when studying the one-dimensional mean computation
problem. Again, the stochasticity forces us into a slow,~\(1/\sqrt{T}\)
rate of convergence, but high dimensionality does not change this rate.

Second, if we only operate on samples exactly once, and we assume our
data is i.i.d., we can think of the stochastic gradient method as
minimizing the \emph{population risk} instead of the empirical risk.
With the same notation, we'll have \[
\mathbb{E}[R[\bar{w}_T]] -R_\star \leq \frac{CD\rho}{\sqrt{T}}  \,.
\] The analysis of stochastic gradient gives our second
\emph{generalization bound} of the book. What it shows is that by
optimizing over a fixed set of~\(T\) data points, we can get a solution
that will have low cost on new data. We will return to this observation
in the next chapter.

\hypertarget{implicit-convexity}{%
\section{Implicit convexity}\label{implicit-convexity}}

\index{convexity!implicit}

We have thus far focused on convex optimization problems, showing that
gradient descent can find global minima with modest computational means.
What about nonconvex problems? Nonconvex optimization is such a general
class of problems that in general it is hard to make useful guarantees.
However, ERM is a special optimization problem, and its structure
enables nonconvexity to enter in a graceful way.

As it turns out there's a ``hidden convexity'' of ERM problems which
shows that the \emph{predictions} should converge to a global optimum
even if we can't analyze to where exactly the model converges. We will
show this insight has useful benefits when models are overparameterized
or nonconvex.

Suppose we have a loss function~\(\loss\) that is equal to zero when
\(\hat{y} = y\) and is nonnegative otherwise. Suppose we have a
generally parameterized function
class~\(\{ f(x;w) \colon w\in\mathbb{R}^d\}\) and we aim to find
parameters~\(w\) that minimize the empirical risk. The empirical risk \[
R_S[w] =\frac{1}{n} \sum_{i=1}^n \loss(f(x_i;w),y_i)
\] is bounded below by~\(0\). Hence if we can find a solution with
\(f(x_i;w)=y_i\) for all~\(i\), we would have a \emph{global minimum}
not a local minimum. This is a trivial observation, but one that helps
focus our study. If during optimization all predictions~\(f(x_i;w)\)
converge to \(y_i\) for all~\(i\), we will have computed a global
minimizer. For the sake of simplicity, we specialize to the square loss
in this section: \[
 \loss(f(x_i;w),y_i) =  \tfrac{1}{2} (f(x_i;w)-y_i)^2
\] The argument we develop here is inspired by the work of Du \emph{et
al.} who use a similar approach to rigorously analyze the convergence of
two layer neural networks.\citep{Du18gradient} Similar calculations can
be made for other losses with some modifications of the argument.

\hypertarget{convergence-of-overparameterized-linear-models}{%
\subsection{Convergence of overparameterized linear
models}\label{convergence-of-overparameterized-linear-models}}

Let's first consider the case of linear prediction functions \[
    f(x;w) = w^T x\,.
\] Define \[
    y = \begin{bmatrix}
        y_1\\ \vdots \\ y_n
    \end{bmatrix}\qquad \qquad \text{and}\qquad \qquad X = \begin{bmatrix} x_1^T \\ \vdots \\x_n^T \end{bmatrix}\,.
\] We can then write the empirical risk objective as \[
R_S[w]=\frac1{2n}\|Xw-y\|^2\,.
\] The gradient descent update rule has the form \[
    w_{t+1} = w_t - \alpha X^T (X w_t-y)\,.
\] We pull the scaling factor~\(1/n\) into the step size for notational
convenience. Now define the vector of predictions \[
\hat{y}_t = \begin{bmatrix}
        f(x_1;w_t) \\ \vdots \\ f(x_n;w_t)
    \end{bmatrix}\,.
\] For the linear case, the predictions are given by
\(\hat{y}_k = X w_k\). We can use this definition to track the evolution
of the \emph{predictions} instead of the parameters~\(w\). The
predictions evolve according to the rule \[
    \hat{y}_{t+1} = \hat{y}_t - \alpha X X^T (\hat{y}_t-y)\,.
\] This looks a lot like the gradient descent algorithm applied to a
strongly convex quadratic function that we studied earlier. Subtracting
\(y\) from both sides and rearranging shows \[
    \hat{y}_{t+1}-y = (I - \alpha X X^T) (\hat{y}_t-y)\,.
\] This expression proves that as long as~\(XX^T\) is strictly positive
definite and~\(\alpha\) is small enough, then the predictions converge
to the training labels. Keep in mind that~\(X\) is~\(n\times d\) and a
model is overparameterized if~\(d>n\). The~\(n\times n\) matrix~\(XX^T\)
has a chance of being strictly positive definite in this case.

When we use a sufficiently small and constant step size~\(\alpha\), our
predictions converge at an \emph{exponential} rate. This is in contrast
to the behavior we saw for gradient methods on overdetermined problems.
Our general analysis of the weights showed that the convergence rate
might be only inverse polynomial in the iteration counter. In the
overparameterized regime, we can guarantee the predictions converge more
rapidly than the weights themselves.

The rate in this case is governed by properties of the matrix~\(X\). As
we have discussed we need the eigenvalues of~\(XX^T\) to be positive,
and we'd ideally like that the eigenvalues are all of similar magnitude.

First note that a \emph{necessary} condition is that~\(d\), the
dimension, must be larger than~\(n\), the number of data points. That
is, we need to have an overparameterized model in order to ensure
exponentially fast convergence of the predictions. We have already seen
that such overparameterized models make it possible to interpolate any
set of labels and to always force the data to be linearly separable.
Here, we see further that overparameterization encourages optimization
methods to converge in fewer iterations by improving the condition
number of the data matrix.

Overparameterization can also help accelerate convergence. Recall that
the eigenvalues of~\(XX^T\) are the squares of the \emph{singular
values} of \(X\). Let us write out a singular value
decomposition~\(X = USV^T\,,\) where~\(S\) is a diagonal matrix of
singular values \((\sigma_1,\ldots, \sigma_n)\). In order to improve the
condition number of this matrix, it suffices to add a feature that is
concentrated in the span of the singular vectors with small singular
values. How to find such features is not always apparent, but does give
us a starting point as to where to look for new, informative features.

\hypertarget{convergence-of-nonconvex-models}{%
\subsection{Convergence of nonconvex
models}\label{convergence-of-nonconvex-models}}

\index{nonconvex}

Surprisingly, this analysis naturally extends to nonconvex models. With
some abuse of notation, let~\(\hat y = f(x;w)\in\R^n\) denote the~\(n\)
predictions of some nonlinear model parameterized by the weights~\(w\)
on input~\(x\). Our goal is to minimize the squared loss objective \[
\frac12\|f(x;w)-y\|^2\,.
\] Since the model is nonlinear this objective is no longer convex.
Nonetheless we can mimic the analysis we did previously for
overparameterized linear models.

Running gradient descent on the weights gives \[
    w_{t+1}
    = w_t - \alpha \jac f (x;w_t)(\hat{y}_t - y)\,,
\] where~\(\hat y_t=f(x;w_t)\) and~\(\jac f\) is the Jacobian of the
predictions with respect to~\(w\). That is,~\(\jac f (x;w)\) is the
\(d\times n\) matrix of first order derivatives of the
function~\(f(x;w)\) with respect to~\(w\). We can similarly define the
Hessian operator~\(H(w)\) to be the~\(n \times d \times d\) array of
second derivatives of~\(f(x;w)\). We can think of~\(H(w)\) as a
quadratic form that maps pairs of vectors \((u,v) \in \R^{d\times d}\)
to~\(\R^n\). With these higher order derivatives, Taylor's theorem
asserts \[
\begin{aligned}
    \hat{y}_{t+1} &= f(x,w_{t+1})\\
    &= f(x,w_t) +  \jac f (x;w_t)^T(w_{t+1}-w_t) \\
    &\qquad\qquad+  \int_0^1 H(w_t+s (w_{t+1}-w_t))  (w_{t+1}-w_t, w_{t+1}-w_t) \dif s\,.
\end{aligned}
\] Since~\(w_t\) are the iterates of gradient descent, this means that
we can write the prediction as \[
\begin{aligned}
    \hat{y}_{t+1}   &= \hat{y}_t -\alpha \jac f(x;w_t)^T \jac f(x;w_t)(\hat{y_t} - y)
            + \alpha \epsilon_t \,,
\end{aligned}
\] where \[
\epsilon_t = \alpha \int_0^1 H(w_t+s (w_{t+1}-w_t))
    \left( \jac f(x;w_t) (\hat{y}_t - y),\jac f(x;w_t) (\hat{y}_t - y)\right)\dif s\,.
\] Subtracting the labels~\(y\) from both sides and rearranging terms
gives the recursion \[
    \hat{y}_{t+1}-y = (I -  \alpha \jac f(x;w_t)^T \jac f(x;w_t))(\hat{y}_t - y) + \alpha \epsilon_t\,.
\] If~\(\epsilon_t\) vanishes, this shows that the predictions again
converge to the training labels as long as the eigenvalues of
\(\jac f(x;w_t)^T \jac f(x;w_t)\) are strictly positive. When the error
vector~\(\epsilon_t\) is sufficiently small, similar dynamics will
occur. We expect~\(\epsilon_t\) to not be too large because it is
quadratic in the distance of~\(y_t\) to~\(y\) and because it is
multiplied by the step size~\(\alpha\) which can be chosen to be small.

The nonconvexity isn't particularly disruptive here. We just need to
make sure our Jacobians have full rank most of the time and that our
steps aren't too large. Again, if the number of parameters are larger
than the number of data points, then these Jacobians are likely to be
positive definite as long as we've engineered them well. But how exactly
can we guarantee that our Jacobians are well behaved? We can derive some
reasonable ground rules by unpacking how we compute gradients of
compositions of functions. More on this follows in our chapter on deep
learning.

\hypertarget{regularization}{%
\section{Regularization}\label{regularization}}

\index{regularization}

The topic of \emph{regularization} belongs somewhere between
optimization and generalization and it's one way of connecting the two.
Hence, we will encounter it in both chapters. Indeed, one complication
with optimization in the overparameterized regime is that there is an
\emph{infinite collection} of models that achieve zero empirical risk.
How do we break ties between these and which set of weights should we
prefer?

To answer this question we need to take a step back and remember that
the goal of supervised learning is not just to achieve zero training
error. We also care about performance on data \emph{outside} the
training set, and having zero loss on its own doesn't tell us anything
about data outside the training set.

As a toy example, imagine we have two sets of
data~\(X_{\mathrm{train}}\) and~\(X_{\mathrm{test}}\)
where~\(X_{\mathrm{train}}\) has shape \(n \times d\)
and~\(X_{\mathrm{test}}\) is~\(m \times d\). Let \(y_{\mathrm{train}}\)
be the training labels and let~\(q\) be an \(m\)-dimensional vector of
random labels. Then if~\(d>(m+n)\) we can find weights~\(w\) such that
\[
    \begin{bmatrix}
     X_{\mathrm{train}} \\ X_{\mathrm{test}}
    \end{bmatrix} w =
    \begin{bmatrix}
     y_{\mathrm{train}} \\ q
    \end{bmatrix}\,.
\] That is, these weights would produce zero error on the training set,
but error no better than random guessing on the testing set. That's not
desired behavior! Of course this example is pathological, because in
reality we would have no reason to fit random labels against the test
set when we create our model.

The main challenge in supervised learning is to design models that
achieve low training error while performing well on new data. The main
tool used in such problems is called \emph{regularization}.
Regularization is the general term for taking a problem with infinitely
many solutions and biasing its solution towards a smaller subset of
solution space. This is a highly encompassing notion.

Sometimes regularization is \emph{explicit} insofar as we have a desired
property of the solution in mind which we exercise as a constraint on
our optimization problem. Sometimes regularization is \emph{implicit}
insofar as algorithm design choices lead to a unique solution, although
the properties of this solution might not be immediately
apparent.\index{regularization!implicit}\index{regularization!explicit}

Here, we take an unconventional tack of working from implicit to
explicit, starting with stochastic gradient descent.

\hypertarget{implicit-regularization-by-optimization}{%
\subsection{Implicit regularization by
optimization}\label{implicit-regularization-by-optimization}}

\index{regularization!implicit}

Consider again the linear case of gradient descent or stochastic
gradient descent \[
    w_{t+1} = w_t - \alpha e_t x_t\,,
\] where~\(e_t\) is the gradient of the loss at the current prediction.
Note that if we initialize~\(w_0=0\), then~\(w_t\) is always in the span
of the data. This can be seen by simple induction. This already shows
that even though general weights lie in a high dimensional, SGD searches
over a space with dimension at most~\(n\), the number of data points.

Now suppose we have a nonnegative loss with
\(\frac{\partial \loss(z,y)}{\partial z} = 0\) if and only if~\(y=z\).
This condition is satisfied by the square loss, but not hinge and
logistic losses. For such losses, at optimality we have for some
vector~\(v\) that:

\begin{enumerate}
\def\labelenumi{\arabic{enumi}.}
\tightlist
\item
  \(Xw = y\,,\) because we have zero loss.
\item
  \(w = X^T v\,,\) because we are in the span of the data.
\end{enumerate}

Under the mild assumption that our examples are linearly independent, we
can combine these equations to find that \[
    w=X^T(XX^T)^{-1}y\,.
\] That is, when we run stochastic gradient descent we converge to a
very specific solution. Even though we were searching through an
\(n\)-dimensional space, we converge to a unique point in this space.

This special~\(w\) is the \emph{minimum Euclidean norm solution}
of~\(Xw=y\). In other words, out of all the linear prediction functions
that interpolate the training data, SGD selects the solution with the
minimal Euclidean norm. To see why this solution has minimal norm,
suppose that \(\hat{w} = X^T\alpha + v\) with~\(v\) orthogonal to
all~\(x_i\). Then we have \[
X\hat{w} = XX^T\alpha + Xv = XX^T\alpha \,.
\] Which means~\(\alpha\) is completely determined and hence
\(\hat{w} = X^T(XX^T)^{-1}y+ v\). But now \[
    \Vert \hat{w}\Vert^2 = \Vert X^T(XX^T)^{-1}y\Vert^2 + \Vert v \Vert^2\,.
\] Minimizing the right hand side shows that~\(v\) must equal zero.

We now turn to showing that such minimum norm solutions have important
robustness properties that suggest that they will perform well on new
data. In the next chapter, we will prove that these methods are
guaranteed to perform well on new data under reasonable assumptions.

\hypertarget{margin-and-stability}{%
\subsection{Margin and stability}\label{margin-and-stability}}

\index{margin}\index{stability}

Consider a linear predictor that makes no classification errors and
hence perfectly separates the data. Recall that the \emph{decision
boundary} of this predictor is the
hyperplane~\(\mathcal{B}= \{z\,:\,w^Tz = 0\}\) and the \emph{margin} of
the predictor is the distance of the decision boundary from to data: \[
    \mathrm{margin}(w) = \min_i \operatorname{dist}\left(x_i,\mathcal{B}\right)\,.
\] Since we're assuming that~\(w\) correctly classifies all of the
training data, we can write the margin in the convenient form \[
    \operatorname{margin}(w) = \min_i \frac{y_i w^Tx_i}{\Vert w\Vert}\,.
\]

Ideally, we'd like our data to be far away from the boundary and hence
we would like our predictor to have large margin. The reasoning behind
this desideratum is as follows: If we expect new data to be similar to
the training data and the decision boundary is far away from the
training data, then it would be unlikely for a new data point to lie on
the wrong side of the decision boundary. Note that margin tells us how
large a perturbation in the~\(x_i\) can be handled before a data point
is misclassified. It is a robustness measure that tells us how sensitive
a predictor is to changes in the data itself.

Let's now specialize margin to the interpolation regime described in the
previous section. Under the assumption that we interpolate the labels so
that~\(w^Tx_i=y_i\), we have \[
    \operatorname{margin}(w) = \Vert w\Vert^{-1}\,.
\] If we want to simultaneously maximize margin and interpolate the
data, then the optimal solution is to choose the minimum norm solution
of~\(Xw=y\). This is precisely the solution found by SGD and gradient
descent.

Note that we could have directly tried to maximize margin by solving the
constrained optimization problem \[
\begin{array}{ll}
    \text{minimize} & \Vert w \Vert^2\\
    \text{subject to} & y_i w^Tx_i \geq 1\,.
\end{array}
\] This optimization problem is the classic formulation of the support
vector machine. The support vector machine is an example of
\emph{explicit} regularization. Here we declare exactly which solution
we'd like to choose given that our training error is zero. Explicit
regularization of high dimensional models is as old as machine learning.
In contemporary machine learning, however, we often have to squint to
see how our algorithmic decisions are regularizing. The tradeoff is that
we can run faster algorithms with implicit regularizers. But it's likely
that revisiting classic regularization ideas in the context of
contemporary models will lead to many new insights.

\hypertarget{the-representer-theorem-and-kernel-methods}{%
\subsection{The representer theorem and kernel
methods}\label{the-representer-theorem-and-kernel-methods}}

\index{representer theorem}\index{kernel methods}

As we have discussed so far, it is common in linear methods to restrict
the search space to the span of the data. Even when~\(d\) is large (or
even infinite), this reduces the search problem to one in an
\(n\)-dimensional space. It turns out that under broad generality,
solutions in the span of the data are optimal for most optimization
problems in prediction. Here, we make formal an argument we first
introduced in our discussion of features: for most empirical risk
minimization problems, the optimal model will lie in the span of the
training data.

Consider the \emph{penalized} ERM problem \[
    \text{minimize} \frac{1}{n} \sum_{i=1}^n \loss(w^T x_i, y_i) + \lambda \Vert w\Vert^2
\] Here~\(\lambda\) is called a \emph{regularization parameter}. When
\(\lambda=0\), there are an infinite number of~\(w\) that minimize the
ERM problem. But for any~\(\lambda>0\), there is a unique minimizer. The
term regularization refers to adding some prior information to an
optimization problem to make the optimal solution unique. In this case,
the prior information is explicitly encoding that we should prefer~\(w\)
with smaller norms if possible. As we discussed in our chapter on
features, smaller norms tend to correspond to simpler solutions in many
feature spaces. Moreover, we just described that minimum norm solutions
themselves could be of interest in machine learning. A regularization
parameter allows us to explicitly tune the norm of the optimal solution.

For our penalized ERM problem, using the same argument as above, we can
write any~\(w\) as \[
    w = X^T \beta + v
\] for some vectors~\(\beta\) and~\(v\) with~\(v^Tx_i=0\) for all~\(i\).
Plugging this equation into the penalized ERM problem yields \[
    \text{minimize}_{\beta,v} \frac{1}{n} \sum_{i=1}^n \loss(\beta^T X x_i,y_i) + \lambda \Vert X^T \beta \Vert^2 + \lambda \Vert v\Vert^2\,.
\] Now we can minimize with respect to~\(v\), seeing that the only
option is to set~\(v=0\). Hence, we must have that the optimum model
lies in the span of the data: \[
    w = X^T \beta
\] This derivation is commonly called the \emph{representer theorem} in
machine learning. As long as the cost function only depends on function
evaluations~\(f(x_i)=w^Tx_i\) and the cost increases as a function of
\(\Vert w\Vert\), then the empirical risk minimizer will lie in the span
of the data.

Define the kernel matrix of the training data~\(K = XX^T\). We can then
rewrite the penalized ERM problem as \[
    \text{minimize}_{\beta} \frac{1}{n} \sum_{i=1}^n \loss(e_i^T K\beta,y_i) + \lambda \beta^T K \beta\,,
\] where~\(e_i\) is the standard Euclidean basis vector. Hence, we can
solve the machine learning problem only using the values in the matrix
\(K\), searching only for the coefficients~\(\beta\) in the kernel
expansion.

The representer theorem (also known as the kernel trick) tells us that
most machine learning problems reduce to a search in~\(n\) dimensional
space, even if the feature space has much higher dimension. Moreover,
the optimization problems only care about the values of dot products
between data points. This motivates the use of the kernel functions
described in our discussion of representation: kernel functions allow us
to evaluate dot products of vectors in high dimensional spaces often
without ever materializing the vectors, reducing high-dimensional
function spaces to the estimation of weightings of individual data
points in the training sample.

\hypertarget{squared-loss-methods-and-other-optimization-tools}{%
\section{Squared loss methods and other optimization
tools}\label{squared-loss-methods-and-other-optimization-tools}}

\index{loss!squared}

This chapter focused on gradient methods for minimizing empirical risk,
as these are the most common methods in contemporary machine learning.
However, there are a variety of other optimization methods that may be
useful depending on the computational resources available and the
particular application in question.

There are a variety of optimization methods that have proven fruitful in
machine learning, most notably constrained quadratic programming for
solving support vector machines and related problems. In this section we
highlight least squares methods which are attractive as they can be
solved by solving linear systems. For many problems, linear systems
solves are faster than iterative gradient methods, and the computed
solution is exact up to numerical precision, rather than being
approximate.

Consider the optimization problem \[
\begin{array}{ll}
    \text{minimize}_w  & \tfrac{1}{2}\sum_{i=1}^n (y_i-w^T x_i)^2 \,.
    \end{array}
\] The gradient of this loss function with respect to~\(w\) is given by
\[
    -\sum_{i=1}^n (y_i-w^T x_i)x_i \,.
\] If we let~\(y\) denote the vector of~\(y\) labels and~\(X\) denote
the \(n\times d\) matrix \[
    X = \begin{bmatrix} x_1^T\\ x_2^T \\ \vdots \\ x_n^T \end{bmatrix}\,,
\] then setting the gradient of the least squares cost equal to zero
yields the solution \[
    w = (X^T X)^{-1} X^T y\,.
\] For many problems, it is faster to compute this closed form solution
than it is to run the number of iterations of gradient descent required
to find a~\(w\) with small empirical risk.

Regularized least squares also has a convenient closed form solution.
The penalized ERM problem where we use a square loss is called the
\emph{ridge regression problem}:\index{ridge regression} \[
\begin{array}{ll}
    \text{minimize}_w  & \tfrac{1}{2}\sum_{i=1}^n (y_i-w^T x_i)^2 +\lambda \Vert w\Vert^2
    \end{array}
\] Ridge regression can be solved in the same manner as above, yielding
the optimal solution \[
    w = (X^T X+ \lambda I)^{-1} X^T y\,.
\] There are a few other important problems solvable by least squares.
First, we have the identity \[
    (X^T X+ \lambda I_d)^{-1} X^T = X (X X^T+ \lambda I_n)^{-1}\,.
\] this means that we can solve ridge regression either by solving a
system in~\(d\) equations and~\(d\) unknowns or~\(n\) equations
and~\(n\) unknowns. In the overparameterized regime, we'd choose the
formulation with~\(n\) parameters. Moreover, as we described above, the
minimum norm interpolating problem \[
\begin{array}{ll}
    \text{minimize} & \Vert w \Vert^2\\
    \text{subject to} & w^Tx_i = y_i\,.
\end{array}
\] is solved by~\(w= X (X X^T)^{-1}y.\) This shows that the limit as
\(\lambda\) goes to zero in ridge regression is this minimum norm
solution.

Finally, we note that for kernelized problems, we can simply replace the
matrix~\(X X^T\) with the appropriate kernel~\(K\). Hence, least squares
formulations are extensible to solve prediction problems in arbitrary
kernel spaces.

\hypertarget{chapter-notes-4}{%
\section{Chapter notes}\label{chapter-notes-4}}

Mathematical optimization is a vast field, and we clearly are only
addressing a very small piece of the puzzle. For an expanded coverage of
the material in this chapter with more mathematical rigor and
implementation ideas, we invite the reader to consult the recent book by
Wright and Recht.\citep{Wright2021}

The chapter focuses mostly on iterative, stochastic gradient methods.
Initially invented by Robbins and Monro for solving systems of equations
in random variables\citep{robbins1951stochastic}, stochastic gradient
methods have played a key role in pattern recognition since the
Perceptron. Indeed, it was very shortly after Rosenblatt's invention
that researchers realized the Perceptron was solving a stochastic
approximation problem. Of course, the standard perceptron step size
schedule does not converge to a global optimum when the data is not
separable, and this lead to a variety of methods to fix the problem.
Many researchers employed the Widrow-Hoff ``Least-Mean-Squares'' rule
which in modern terms is minimizing the empirical risk associated with a
square loss by stochastic gradient descent.\citep{WidrowHoffLMS}
Aizerman and his colleagues determined not only how to apply stochastic
gradient descent to linear functions, but how to operate in kernel
spaces as well.\citep{Aizerman65} Of course, all of these methods were
closely related to each other, but it took some time to put them all on
a unified footing. It wasn't until the 1980s with a full understanding
of complexity theory, that optimal step sizes were discovered for
stochastic gradient methods by Nemirovski and
Yudin.\citep{NemirovskiYudinBook} More surprisingly, it was not until
2007 that the first non-asymptotic analysis of the perceptron algorithm
was published.\citep{Pegasos}

Interestingly, it wasn't again until the early 2000s that stochastic
gradient descent became the default optimization method for machine
learning. There tended to be a repeated cycle of popularity for the
various optimization methods. Global optimization methods like linear
programming were determined effective in the 1960s for pattern
recognition problems\citep{mangasarian1965linear}, supplanting interest
in stochastic descent methods. Stochastic gradient descent was rebranded
as back propagation in the 1980s, but again more global methods
eventually took center stage. Mangasarian, who was involved in both of
these cycles, told us in private correspondence that linear programming
methods were always more effective in terms of their speed of
computation and quality of solution.

Indeed this pattern was also followed in optimization. Nemirovski and
Nesterov did pioneering work in iterative gradient and stochastic
gradient methods . But they soon turned to developing the foundations of
interior point methods for solving global optimization
problems.\citep{nesterov-nemirovskii-ip} In the 2000s, they republished
their work on iterative methods, leading to a revolution in machine
learning.

It's interesting to track this history and forgetting in machine
learning. Though these tools are not new, they are often forgotten and
replaced. It's perhaps time to revisit the non-iterative methods in
light of this.

\chapter{Generalization}

Simply put, generalization relates the performance of a model on seen
examples to its performance on \emph{unseen} examples. In this chapter,
we discuss the interplay between representation, optimization, and
generalization, again focusing on models with more parameters than seen
data points. We examine the intriguing empirical phenomena related to
overparameterization and generalization in today's machine learning
practice. We then review available theory---some old and some
emerging---to better understand and anticipate what drives
generalization performance.

\hypertarget{generalization-gap}{%
\section{Generalization gap}\label{generalization-gap}}

\index{generalization gap}

Recall, the risk of a predictor~\(f\colon \mathcal{X}\to \mathcal{Y}\)
with respect to a loss function
\(\loss\colon \mathcal{Y}\times \mathcal{Y}\to\mathbb{R}\) is defined as
\[
R[f] = \E\left[\loss(f(X), Y)\right]\,.
\] Throughout this chapter, it will often be convenient to stretch the
notation slightly by using~\(\loss(f,(x,y))\) to denote the loss of a
predictor~\(f\) on an example~\((x,y)\,.\) For predictors specified by
model parameters~\(w\), we'll also write~\(\loss(w,(x,y))\,.\)

For the purposes of this chapter, it makes sense to think of the~\(n\)
samples as an ordered tuple \[
S=((x_1,y_1),\dots\dots,(x_n,y_n))\in(\mathcal{X}\times \mathcal{Y})^n\,.
\] The empirical risk~\(R_S[f]\) is, as before, \[
R_{S}[f] = \frac{1}{n}\sum_{i=1}^{n}\loss(f(x_i),y_i)\,.
\] Empirical risk minimization seeks to find a predictor~\(f^*\) in a
specified class~\(\mathcal{F}\) that minimizes the empirical risk: \[
R_S[f^*] = \min_{f\in\mathcal{F}} R_S[f]
\] In machine learning practice, the empirical risk is often called
\emph{training error} or \emph{training loss}, as it corresponds to the
loss achieved by some optimization method on the sample. Depending on
the optimization problem, we may not be able to find an exact empirical
risk minimizer and it may not be unique.

Empirical risk minimization is commonly used as a proxy for minimizing
the unknown population risk. But how good is this proxy? Ideally, we
would like that the predictor~\(f\) we find via empirical risk
minimization satisfies~\(R_S[f]\approx R[f].\) However, this may not be
the case, since the risk~\(R[f]\) captures loss on unseen example, while
the empirical risk~\(R_S[f]\) captures loss on seen examples.

Generally, we expect to do much better on seen examples than unseen
examples. This performance gap between seen and unseen examples is what
we call \emph{generalization gap}.

\begin{Definition}

Define the \emph{generalization gap} of a predictor~\(f\) with respect
to a dataset~\(S\) as \[
\gengap(f) = R[f] - R_S[f]\,.
\]

\end{Definition}

This quantity is sometimes also called \emph{generalization error} or
\emph{excess risk}. Recall the following tautological, yet important
identity: \[
R[f] = R_S[f] + \gengap(f)
\] What it says is that if we manage to make the empirical
risk~\(R_S[f]\) small through optimization, then all that remains to
worry about is generalization gap.

The last chapter provided powerful tools to make optimization succeed.
How we can bound the generalization gap is the topic of this chapter. We
first take a tour of evidence from machine learning practice for
inspiration.

\hypertarget{overparameterization-empirical-phenomena}{%
\section{Overparameterization: empirical
phenomena}\label{overparameterization-empirical-phenomena}}

\index{overparameterization}

We previously experienced the advantages of overparameterized models in
terms of their ability to represent complex functions and our ability to
feasily optimize them. The question remains whether they generalize well
to unseen data. Perhaps we simply kicked the can down the road. Does the
model size that was previously a blessing now come back to haunt us? We
will see that not only do large models often generalize well in
practice, but often more parameters lead to better generalization
performance. Model size does, however, challenge some theoretical
analysis. The empirical evidence will orient our theoretical study
towards dimension-free bounds that avoid worst-case analysis.

\hypertarget{effects-of-model-complexity}{%
\subsection{Effects of model
complexity}\label{effects-of-model-complexity}}

Think of a model family with an associated measure of complexity, such
as number of trainable parameters. Suppose that for each setting of the
complexity measure, we can solve the empirical risk minimization
problem. We can then plot what happens to risk and empirical risk as we
vary model complexity.

A traditional view of generalization posits that as we increase model
complexity initially both empirical risk and risk decrease. However,
past a certain point the risk begins to increase again, while empirical
risk decreases.

\begin{figure}
\centering
\includegraphics[width=1\textwidth,height=\textheight]{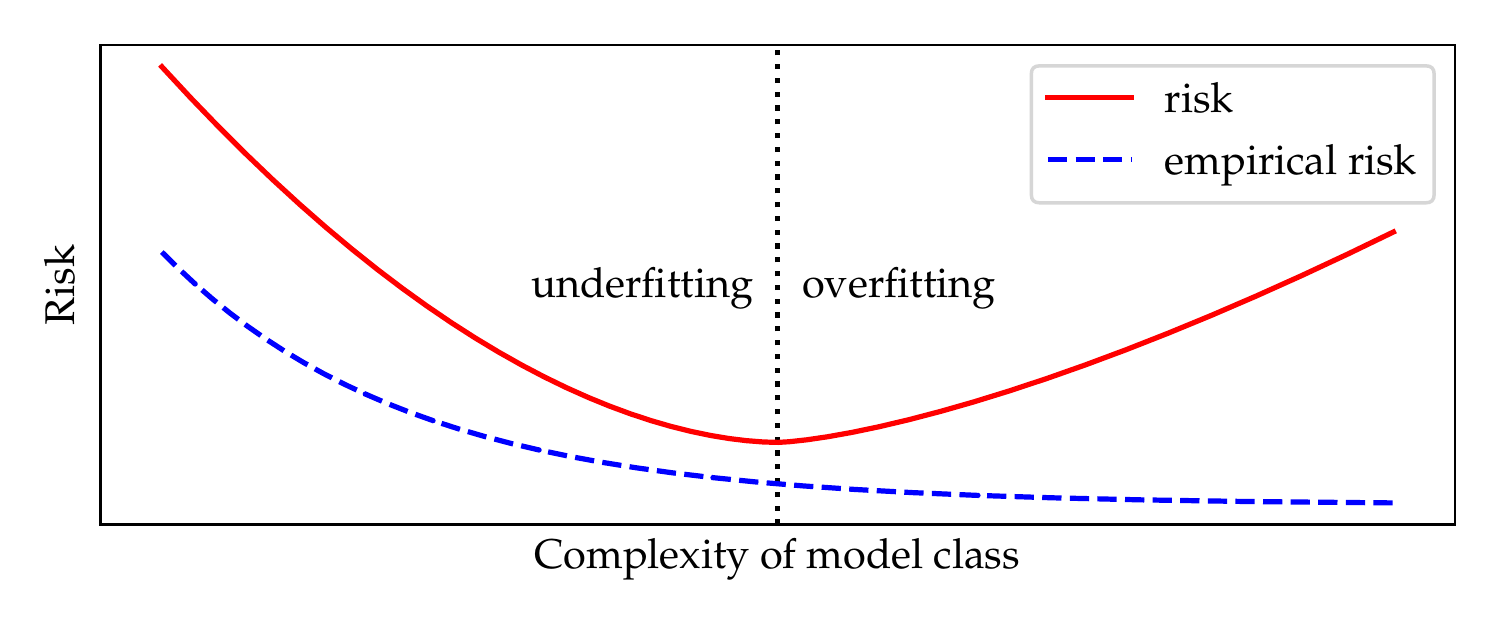}
\caption{Traditional view of generalization}
\end{figure}

The graphic shown in many textbooks is a u-shaped risk curve. The
complexity range below the minimum of the curve is called
\emph{underfitting}. The range above is called
\emph{overfitting}.\index{underfitting}\index{overfitting}

This pictures is often justified using the bias-variance trade-off,
motivated by a least squares regression analysis. However, it does not
seem to bear much resemblance to what is observed in practice.

We have already discussed the example of the Perceptron which achieves
zero training loss and still generalizes well in theory. Numerous
practitioners have observed that other complex models also can
simultaneously achieve close to zero training loss and still generalize
well. Moreover, in many cases risk continues to decreases as model
complexity grows and training data are interpolated exactly down to
(nearly) zero training loss. This empirical relationship between
overparameterization and risk appears to be robust and manifests in
numerous model classes, including overparameterized linear models,
ensemble methods, and neural networks.

\begin{figure}
\centering
\includegraphics[width=1\textwidth,height=\textheight]{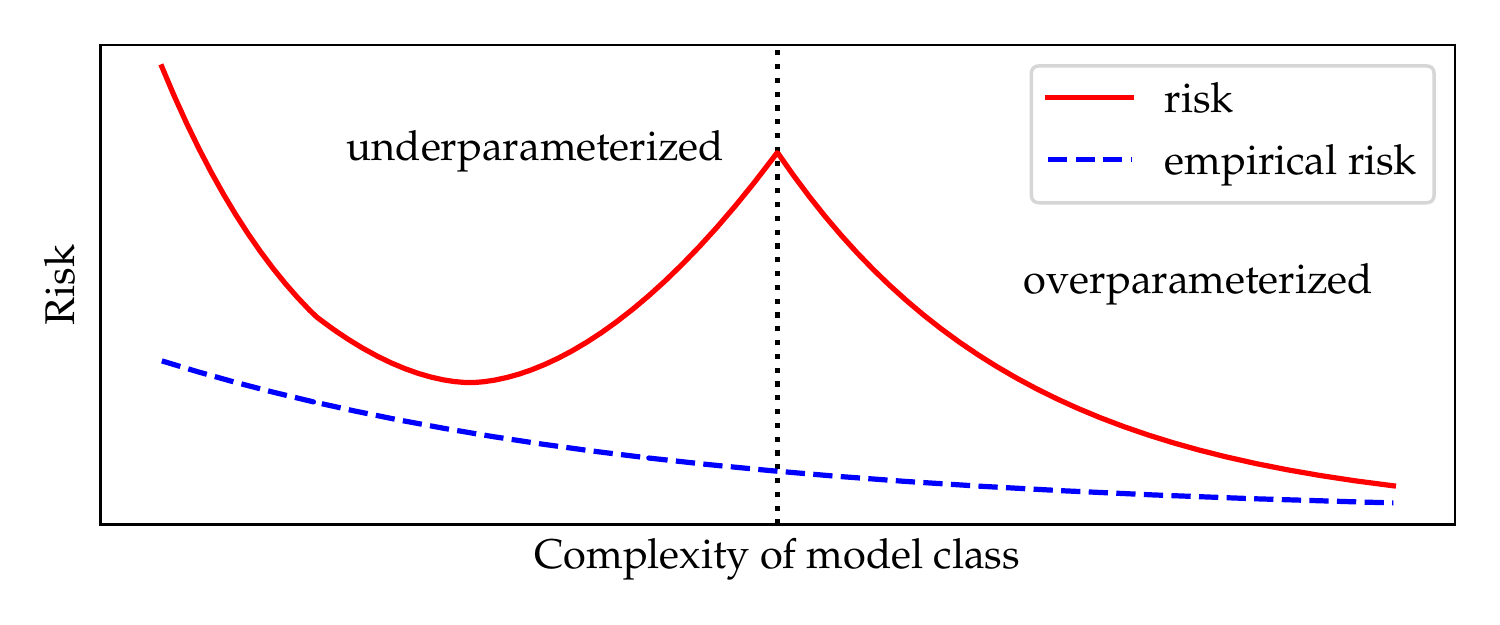}
\caption{Double descent.}
\end{figure}

In the absence of regularization and for certain model families, the
empirical relationship between model complexity and risk is more
accurately captured by the \emph{double descent} curve in the figure
above. There is an interpolation threshold at which a model of the given
complexity can fit the training data exactly. The complexity range below
the threshold is the \emph{underparameterized regime}, while the one
above is the overparameterized regime. Increasing model complexity in
the overparameterized regime continues to decrease risk indefinitely,
albeit at decreasing marginal returns, toward some convergence point.

\begin{figure}
\centering
\includegraphics[width=1\textwidth,height=\textheight]{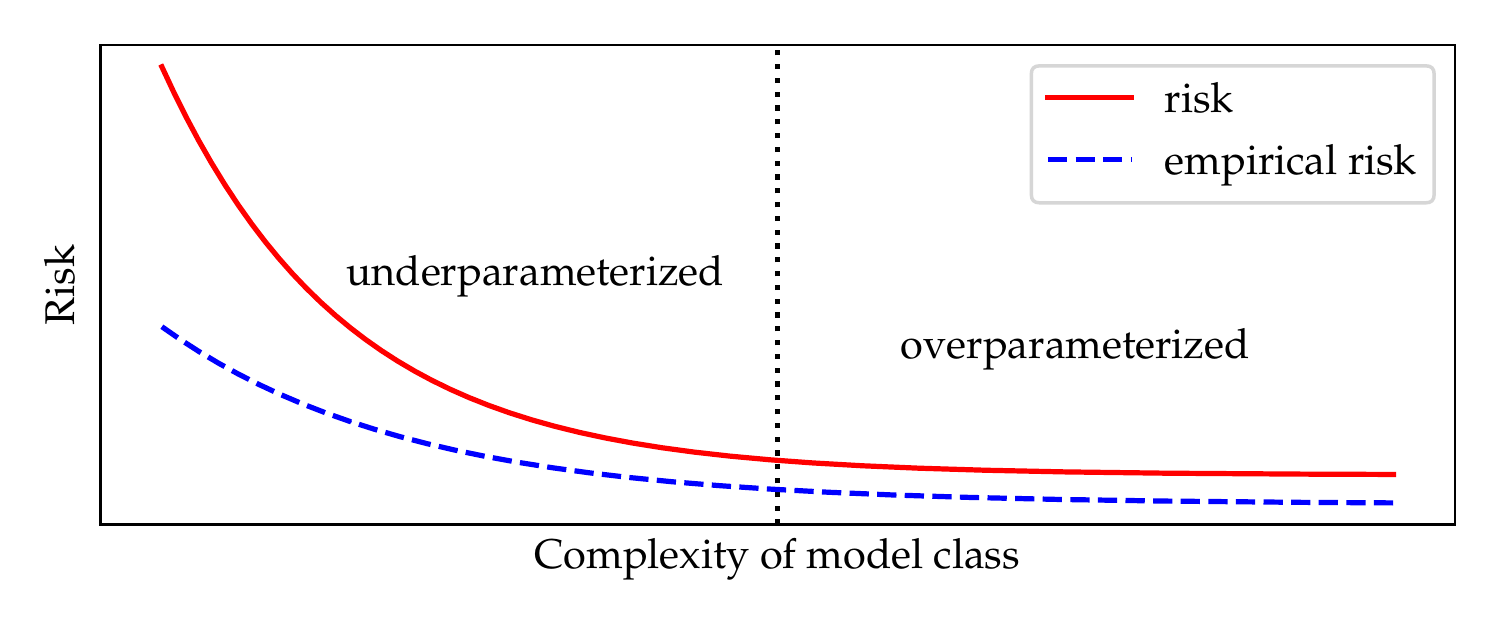}
\caption{Single descent.}
\end{figure}

The double descent curve is not universal. In many cases, in practice we
observe a single descent curve throughout the entire complexity range.
In other cases, we can see multiple bumps as we increase model
complexity.\citep{Liang2020} However, the general point remains. There
is no evidence that highly overparameterized models do not generalize.
Indeed, empirical evidence suggests larger models not only generalize,
but that larger models make better out-of-sample predictors than smaller
ones.\citep{resnet, huang2019gpipe}

\hypertarget{optimization-versus-generalization}{%
\subsection{Optimization versus
generalization}\label{optimization-versus-generalization}}

Training neural networks with stochastic gradient descent, as is
commonly done in practice, attempts to solve a non-convex optimization
problem. Reasoning about non-convex optimization is known to be
difficult. Theoreticians see a worthy goal in trying to prove
mathematically that stochastic gradient methods successfully minimize
the training objective of large artificial neural networks. The previous
chapter discussed some of the progress that has been made toward this
goal.

It is widely believed that what makes optimization easy crucially
depends on the fact that models in practice have many more parameters
than there are training points. While making optimization tractable,
overparameterization puts burden on generalization.

We can force a disconnect between optimization and generalization in a
simple experiment that we will see next. One consequence is that even if
a mathematical proof established the convergence guarantees of
stochastic gradient descent for training some class of large neural
networks, it would not necessarily on its own tell us much about why the
resulting model generalizes well to the test objective.

Indeed, consider the following experiment. Fix training data
\((x_1,y_1),\dots, (x_n, y_n)\) and fix a training algorithm~\(A\) that
achieves zero training loss on these data and achieves good test loss as
well.

Now replace all the labels~\(y_1,\dots,y_n\) by randomly and
independently drawn labels~\(\tilde y_1,\dots, \tilde y_n\,.\) What
happens if we run the same algorithm on the training data with noisy
labels \((x_1,\tilde y_1),\dots, (x_n, \tilde y_n))\)?

One thing is clear. If we choose from~\(k\) discrete classes, we expect
the model trained on the random labels to have no more than~\(1/k\) test
accuracy, that is, the accuracy achieved by random guessing. After all,
there is no statistical relationship between the training labels and the
test labels that the model could learn.

\begin{figure}
\centering
\includegraphics[width=1\textwidth,height=\textheight]{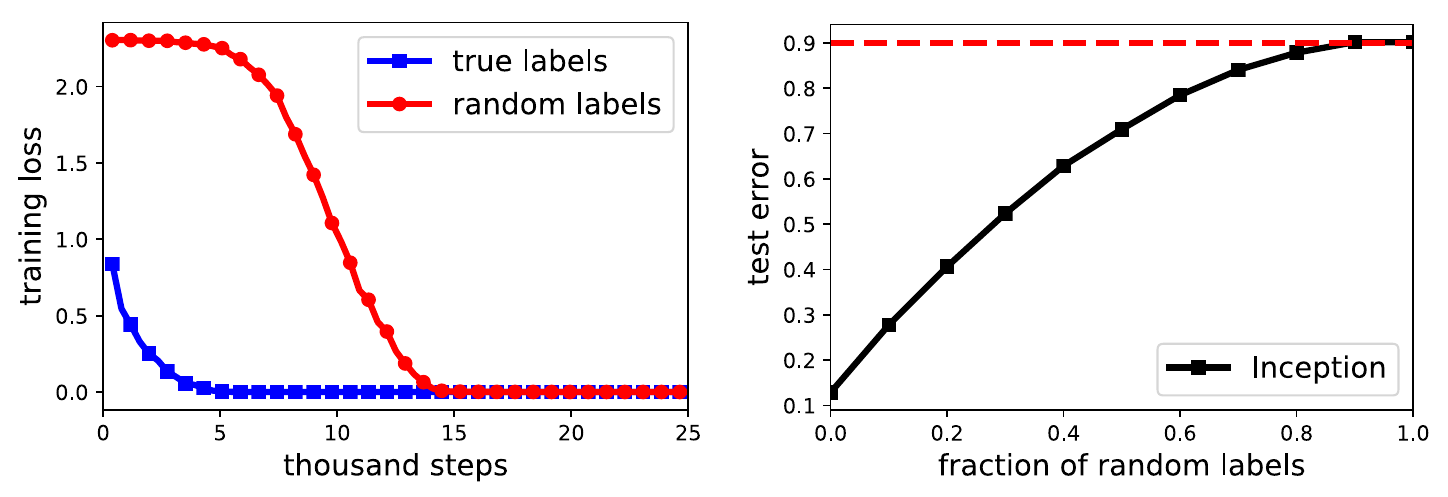}
\caption{Randomization test on CIFAR-10. Left: How randomization affects
training loss. Right: How increasing the fraction of corrupted training
labels affects test error.}
\end{figure}

What is more interesting is what happens to optimization. The left panel
of the figure shows the outcome of this kind of \emph{randomization
test} on the popular CIFAR-10 image classification benchmark for a
standard neural network architecture. What we can see is that the
training algorithm continues to drive the training loss to zero even if
the labels are randomized. The right panel shows that we can vary the
amount of randomization to obtain a smooth degradation of the test
error. At full randomization, the test error degrades to~\(90\%\), as
good as guessing one of the~\(10\) classes. The figure shows what
happens to a specific model architecture, called Inception, but similar
observations hold for most, if not all, overparameterized architectures
that have been proposed.

The randomization experiment shows that optimization continues to work
well even when generalization performance is no better than random
guessing, i.e.,~\(10\%\) accuracy in the case of the CIFAR-10 benchmark
that has~\(10\) classes. The optimization method is moreover insensitive
to properties of the data, since it works even on random labels. A
consequence of this simple experiment is that a proof of convergence for
the optimization method may not reveal any insights into the nature of
generalization.

\hypertarget{the-diminished-role-of-explicit-regularization}{%
\subsection{The diminished role of explicit
regularization}\label{the-diminished-role-of-explicit-regularization}}

\index{regularization}

Regularization plays an important role in the theory of convex empirical
risk minimization. The most common form of regularization used to be
\(\ell_2\)-regularization corresponding to adding a scalar of the
squared Euclidean norm of the parameter vector to the objective
function.

\index{data augmentation}

A more radical form of regularization, called \emph{data augmentation},
is common in the practice of deep learning. Data augmentation transforms
each training point repeatedly throughout the training process by some
operation, such as a \emph{random crop} of the image. Training on such
randomly modified data points is meant to reduce overfitting, since the
model never encounters the exact same data point twice.

Regularization continues to be a component of training large neural
networks in practice. However, the nature of regularization is not
clear. We can see a representative empirical observation in the table
below.

\begin{longtable}[]{@{}ccccc@{}}
\caption{The training and test accuracy (in percentage) with and without
data augmentation and \(\ell_2\)-regularization.}\tabularnewline
\toprule
params & random crop & \(\ell_2\)-regularization & train accuracy & test
accuracy \\
\midrule
\endfirsthead
\toprule
params & random crop & \(\ell_2\)-regularization & train accuracy & test
accuracy \\
\midrule
\endhead
1,649,402 & yes & yes & 100.0 & 89.05 \\
& yes & no & 100.0 & 89.31 \\
& no & yes & 100.0 & 86.03 \\
& no & no & 100.0 & 85.75 \\
\bottomrule
\end{longtable}

The table shows the performance of a common neural model architecture,
called Inception, on the standard CIFAR-10 image classification
benchmark. The model has more than~\(1.5\) million trainable parameters,
even though there are only~\(50,000\) training examples spread
across~\(10\) classes. The training procedure uses two explicit forms of
regularization. One is a form of data augmentation with random crops.
The other is~\(\ell_2\)-regularization. With both forms of
regularization the fully trained model achieves close to~\(90\%\) test
accuracy. But even if we turn both of them off, the model still achieves
close to~\(86\%\) test accuracy (without even readjusting any
hyperparameters such as learning rate of the optimizer). At the same
time, the model fully interpolates the training data in the sense of
making no errors whatsoever on the training data.

These findings suggest that while explicit regularization may help
generalization performance, it is by no means necessary for strong
generalization of heavily overparameterized models.

\hypertarget{theories-of-generalization}{%
\section{Theories of generalization}\label{theories-of-generalization}}

With these empirical facts in hand, we now turn to mathematical theories
that might help explain what we observe in practice and also may guide
future empirical and theoretical work. In the remainder of the chapter,
we tour several different, seemingly disconnected views of
generalization.

We begin with a deep dive into \emph{algorithmic stability}, which
posits that generalization arises when models are insensitive to
perturbations in the data on which they are trained. We then discuss
\emph{VC dimension and Rademacher complexity}, which show how small
generalization gaps can arise when we restrict the complexity of models
we wish to fit to data. We then turn to \emph{margin bounds} which
assert that whenever the data is easily separable, good generalization
will occur. Finally we discuss generalization bounds that arise from
\emph{optimization}, showing how choice of an algorithmic scheme itself
can yield models with desired generalization properties.

In all of these cases, we show that we can recover generalization bounds
of the form we saw in the Perceptron: the bounds will decrease with
number of data points and increase with ``complexity'' of the optimal
prediction function. Indeed, looking back at the proof of the Perceptron
generalization bound, all of the above elements appeared. Our
generalization bound arose because we could remove single data points
from a set and not change the number of mistakes made by the Perceptron.
A large margin assumption was essential to get a small mistake bound.
The mistake bound itself was dependent on the iterations of the
algorithm. And finally, we related the size of the margin to the scale
of the data and optimal separator.

Though starting from different places, we will shows that the four
different views of generalization can all arrive at similar results.
Each of the aforementioned ingredients can alone lead to generalization,
but considerations of all of these aspects help to improve machine
learning methods. Generalization is multifaceted and multiple
perspectives are useful when designing data-driven predictive systems.

Before diving into these four different views, we first take a quick
pause to consider how we hope generalization bounds might look.

\hypertarget{how-should-we-expect-the-gap-to-scale}{%
\subsection{How should we expect the gap to
scale?}\label{how-should-we-expect-the-gap-to-scale}}

Before we turn to analyzing generalization gaps, it's worth first
considering how we should expect them to scale. That is, what is the
relationship between the expected size of~\(\Delta_{\mathrm{gen}}\) and
the number of observations,~\(n\)?

First, note that we showed that for a \emph{fixed} prediction
function~\(f\), the expectation of the empirical risk is equal to the
population risk. That is, the empirical risk of a single function is a
sample average of the population risk of that function. As we discussed
in Chapter 3, i.i.d. sample averages should \emph{generalize} and
approximate the average at the population level. Here, we now turn to
describing \emph{how} they might be expected to scale under different
assumptions.

\hypertarget{quantitative-central-limit-theorems}{%
\subsection{Quantitative central limit
theorems}\label{quantitative-central-limit-theorems}}

The \emph{central limit theorem} formalizes how sample averages estimate
their expectations: If~\(Z\) is a random variable with bounded variance
then~\(\hat{\mu}_Z^{(n)}\) converges in distribution to a Gaussian
random variable with mean zero and variance on the order of~\(1/n\).

The following inequalities are useful \emph{quantitative} forms of the
central limit theorem. They precisely measure how close the sample
average will be to the population average using limited information
about the random quantity.\index{Markov's inequality}

\begin{itemize}
\tightlist
\item
  \textbf{Markov's inequality:} Let \(Z\) be a nonnegative random
  variable. Then, \[
  \Pr[Z \geq t] \leq \frac{\E[Z]}{t}\,.
  \]
\end{itemize}

This can be proven using the inequality
\(I_{[Z\geq t]}(z) \leq \frac{z}{t}\).

\begin{itemize}
\tightlist
\item
  \textbf{Chebyshev's inequality:} Suppose \(Z\) is a random variable
  with mean \(\mu_Z\) and variance \(\sigma_Z^2\). Then, \[
  \Pr[Z\geq t+\mu_Z] \leq \frac{\sigma_Z^2}{t^2}
  \]
\end{itemize}

Chebyshev's inequality\index{Chebyshev's inequality} helps us understand
why sample averages are good estimates of the mean. Suppose
that~\(X_1,\dots, X_n\) are independent samples we were considering
above. Let~\(\hat\mu\) denote the sample
mean~\(\frac1n\sum_{i=1}^n Z_i.\) Chebyshev's inequality implies \[
  \Pr[\hat\mu\geq t+\mu_X] \leq \frac{\sigma_X^2}{n t^2}\,,
\] which tends to zero as~\(n\) grows. A popular form of this inequality
sets~\(t=\mu_X\) which gives \[
  \Pr[\hat{\mu}\geq 2\mu_X] \leq \frac{\sigma_X^2}{n \mu_X^2}\,.
\]

\begin{itemize}
\tightlist
\item
  \textbf{Hoeffding's inequality:} Let \(Z_1, Z_2, \ldots, Z_n\) be
  independent random variables, each taking values in the interval
  \([a_i,b_i]\).Let \(\hat\mu\) denote the sample mean
  \(\frac1n\sum_{i=1}^n Z_i\). Then \[
  \Pr[\hat\mu \geq \mu_Z + t] \leq \exp\left(- \frac{2n^2t^2}{\sum_{i=1}^n(b_i-a_i)^2}\right)\,.
  \]
\end{itemize}

\index{Hoeffding's inequality} An important special case is when
the~\(Z_i\) are identically distributed copies of~\(Z\) and take values
in \([0,1]\). Then we have \[
  \Pr[\hat\mu \geq \mu_Z + t] \leq \exp\left(- 2nt^2\right)\,.
\] This shows that when random variables are bounded, sample averages
concentrate around their mean value exponentially quickly. If we invoke
this bound with~\(t=C/\sqrt{n}\), the point at which it gives
non-trivial results, we have an error of~\(O(1/\sqrt{n})\) with
exponentially high probability. We will see shortly that this
relationship between error and number of samples is ubiquitous in
generalization theory.

These powerful \emph{concentration inequalities} let us precisely
quantify how close the sample average will be to the population average.
For instance, we know a person's height is a positive number and that
there are no people who are taller than nine feet. With these two facts,
Hoeffding's inequality tells us that if we sample the heights of thirty
thousand individuals, our sample average will be within an inch of the
true average height with probability at least 83\%. This assertion is
true no matter how large the population of individuals. The required
sample size is dictated only by the variability of height, not by the
number of total individuals.

You could replace ``height'' in this example with almost any attribute
that you are able to measure well. The quantitative central limits tell
us that for attributes with reasonable variability, a uniform sample
from a general population will give a high quality estimate of the
average value.

``Reasonable variability'' of a random variable is necessary for
quantitative central limit theorems to hold. When random variables have
low variance or are tightly bounded, small experiments quickly reveal
insights about the population. When variances are large or effectively
unbounded, the number of samples required for high precision estimates
might be impractical and our estimators and algorithms and predictions
may need to be rethought.

\hypertarget{bounding-generalization-gaps-for-individual-predictors}{%
\subsection{Bounding generalization gaps for individual
predictors}\label{bounding-generalization-gaps-for-individual-predictors}}

Let us now return to generalization of prediction, considering the
example where the quantity of interest is the prediction error on
individuals in a population. There are effectively two scaling regimes
of interest in generalization theory. In one case when the empirical
risk is large, we expect the generalization gap to decrease inversely
proportional to~\(\sqrt{n}\,.\) When the empirical risk is expected to
be very small, on the other hand, we tend to see the the generalization
gap decrease inversely proportional to~\(n\,.\)

Why we see these two regimes is illustrated by studying the case of a
\emph{single} prediction function~\(f\,,\) chosen \emph{independently}
of the sample \(S\,.\) Our ultimate goal is to reason about the
generalization gap of predictors chosen by an algorithm running on our
data. The analysis we walk through next doesn't apply to data-dependent
predictors directly, but it nonetheless provides helpful intuition about
what bounds we can hope to get.

For a fixed function~\(f\,,\) the zero-one loss on a single randomly
chosen data point is a Bernoulli random variable, equal to~\(1\) with
probability~\(p\) and~\(1-p\) otherwise. The empirical risk~\(R_S[f]\)
is the sample average of this random variable and the risk~\(R[f]\) is
its expectation. To estimate the generalization gap, we can apply
Hoeffding's inequality to find\\
\[
  \Pr[ R[f]-R_S[f] \geq \epsilon] \leq \exp\left( -2n\epsilon^2\right)\,.
\] Hence, we will have with probability~\(1-\delta\) on our sample that
\[
  |\gengap(f)| \leq \sqrt{\frac{\log(1/\delta)}{2n}}\,.
\] That is, the generalization gap goes to zero at a rate of
\(1/\sqrt{n}\,.\)

In the regime where we observe no empirical mistakes, a more refined
analysis can be applied. Suppose that~\(R[f]>\epsilon\,.\) Then the
probability that we observe~\(R_S[f]=0\) cannot exceed \[
\begin{aligned}
  \Pr[ \forall i\colon\operatorname{sign}(f(x_i))=y_i]
  &= \prod_{i=1}^n \Pr[ \operatorname{sign}(f(x_i))=y_i] \\
  & \leq (1-\epsilon)^n \leq e^{-\epsilon n}\,.
  \end{aligned}
\] Hence, with probability~\(1-\delta\), \[
  |\gengap(f)| \leq \frac{\log(1/\delta)}{n}\,,  
\] which is the~\(1/n\) regime. These two rates are precisely what we
observe in the more complex regime of generalization bounds in machine
learning. The main trouble and difficulty in computing bounds on the
generalization gap is that our prediction function~\(f\) depends on the
data, making the above analysis inapplicable.

In this chapter, we will focus mostly on~\(1/\sqrt{n}\) rates. These
rates are more general as they make no assumptions about the expected
empirical risk. With a few notable exceptions, the derivation of
\(1/\sqrt{n}\) rates tends to be easier than the~\(1/n\) counterparts.
However, we note that every one of our approaches to generalization
bounds have analyses for the ``low empirical risk'' or ``large margin''
regimes. We provide references at the end of this chapter to these more
refined analyses.

\hypertarget{algorithmic-stability}{%
\section{Algorithmic stability}\label{algorithmic-stability}}

\index{stability!algorithmic}

We will first see a tight characterization in terms of an algorithmic
robustness property we call \emph{algorithmic stability}. Intuitively,
algorithmic stability measures how sensitive an algorithm is to changes
in a single training example. Whenever a model is insensitive to such
perturbations, the generalization gap will be small. Stability gives us
a powerful and intuitive way of reasoning about generalization.

There are a variety of different notions of perturbation. We could
consider resampling a single data point and look at how much a model
changes. We could also leave one data point out and see how much the
model changes. This was the heart of our Perceptron generalization
argument. More aggressively, we could study what happens when a single
data point is arbitrarily corrupted. All three of these approaches yield
similar generalization bounds, though it is often easier to work with
one than the others. To simplify the exposition, we choose to focus on
only one notion (resampling) here.

To introduce the idea of stability, we first condense our notation to
make the presentation a bit less cumbersome. Recall that we operate on
tuples of~\(n\) labeled examples, \[
S=((x_1,y_1),\dots\dots,(x_n,y_n))\in(\mathcal{X}\times \mathcal{Y})^n\,.
\] We denote a labeled example as~\(z=(x,y).\) We will overload our
notation and denote the loss accrued by a prediction function~\(f\) on a
data point~\(z\) as~\(\loss(f,z).\) That
is,~\(\loss(f,z) = \loss(f(x),y).\) We use the uppercase letters when a
labeled example~\(Z\) is randomly drawn from a population~\((X, Y)\).

With this notation in hand, let's now consider two independent random
samples~\(S = (Z_1, \dots , Z_n)\) and
\(S^{\prime} = (Z^{\prime}_1, \dots ,Z^{\prime}_{n})\), each drawn
independently and identically from a population~\((X, Y).\) We call the
second sample~\(S'\) a \emph{ghost sample} as it is solely an analytical
device. We never actually collect this second sample or run any
algorithm on it.

We introduce~\(n\) \emph{hybrid samples} \(S^{(i)}\),
for~\(i\in\{1, \dots, n\}\) as \[
S^{(i)} = (Z_1,\dots, Z_{i-1}, Z_{i}^{\prime}, Z_{i+1},\dots,Z_{n})\,,
\] where the~\(i\)-th example comes from~\(S',\) while all others come
from \(S.\)

We can now introduce a data-dependent notion of average stability of an
algorithm. For this definition, we think of an algorithm as a
deterministic map~\(A\) that takes a training sample in
\((\mathcal{X}\times\mathcal{Y})^n\) to some prediction function in a
function space~\(\Omega\,.\) That is~\(A(S)\) denotes the function from
\(\mathcal{X}\) to~\(\mathcal{Y}\) that is returned by our algorithm
when run on the sample~\(S\).

\index{stability!average}

\begin{Definition}

The \emph{average stability} of an algorithm
\(A\colon (\mathcal{X} \times \mathcal{Y})^{n} \rightarrow \Omega\) is
\[
\Delta(A) = \E_{S,S^{\prime}}
\left[
\frac{1}{n}\sum_{i=1}^{n}
\left(\loss(A(S),Z_i^{\prime}) -\loss(A(S^{(i)}),Z_{i}^{\prime})\right)
\right]\,.
\]

\end{Definition}

There are two useful ways to parse this definition. The first is to
interpret average stability as the average sensitivity of the algorithm
to a change in a single example. Since we don't know which of its~\(n\)
input samples the algorithm may be sensitive to, we test all of them and
average out the results.

Second, from the perspective of~\(A(S)\), the example~\(Z_i^\prime\) is
\emph{unseen}, since it is not part of~\(S.\) But from the perspective
of \(A(S^{(i)})\) the example~\(Z_i^\prime\) is seen, since it is part
of \(S^{(i)}\) via the substitution that defines the~\(i\)-th hybrid
sample. This shows that the instrument~\(\Delta(A)\) also measures the
average loss difference of the algorithm on seen and unseen examples. We
therefore have reason to suspect that average stability relates to
generalization gap as the next proposition confirms.

\begin{Proposition}

The expected generalization gap equals average stability: \[
\E[\gengap(A(S))] = \Delta(A)
\]

\end{Proposition}

\begin{Proof}

By linearity of expectation, \[
\begin{aligned}
\mathop\mathbb{E}[\gengap(A(S))]
&= \mathop\mathbb{E}\left[R[A(S)] - R_S[A(S)]\right] \\
&= \mathop\mathbb{E}\left[\frac{1}{n}\sum_{i=1}^{n}\loss(A(S),Z_i^{\prime})\right]
- \mathop\mathbb{E}\left[\frac{1}{n}\sum_{i=1}^{n}\loss(A(S),Z_i)\right]\,.
\end{aligned}
\] Here, we used that~\(Z_i^\prime\) is an example drawn from the
distribution that does not appear in the set~\(S,\) while~\(Z_i\) does
appear in~\(S.\) At the same time,~\(Z_i\) and~\(Z_i'\) are identically
distributed and independent of the other examples. Therefore, \[
\mathop\mathbb{E}\loss(A(S),Z_i) = \mathop\mathbb{E}\loss(A(S^{(i)}),Z_i^{\prime})\,.
\] Applying this identity to each term in the empirical risk above, and
comparing with the definition of~\(\Delta(A),\) we conclude \[
\mathop\mathbb{E}[R[A(S)] - R_S[A(S)]] = \Delta(A)\,.\qedhere
\]

\end{Proof}

\hypertarget{uniform-stability}{%
\subsection{Uniform stability}\label{uniform-stability}}

While average stability gave us an exact characterization of
generalization error, it can be hard to work with the expectation over
\(S\) and~\(S'.\) Uniform stability replaces the averages by suprema,
leading to a stronger but useful notion.

\index{stability!uniform}

\begin{Definition}

The \emph{uniform stability} of an algorithm~\(A\) is defined as \[
\Delta_{\mathrm{sup}}(A) = \sup_{\underset{d_H(S, S')=1}{S, S' \in (\mathcal{X}\times\mathcal{Y})^n}}
\sup_{z\in \mathcal{X}\times\mathcal{Y}} |\loss(A(S), z) - \loss(A(S'), z)|,
\] where~\(d_H(S,S')\) is the Hamming distance between tuples~\(S\) and
\(S'\,.\)

\end{Definition}

In this definition, it is important to note that the~\(z\) has nothing
to do with~\(S\) and~\(S'\,.\) Uniform stability is effectively
computing the worst-case difference in the predictions of the learning
algorithm run on two arbitrary datasets that differ in exactly one
point.

Uniform stability upper bounds average stability, and hence uniform
stability upper bounds generalization gap (in expectation). Thus, we
have the corollary \[
\mathop\mathbb{E}[\gengap(A(S))] \leq \Delta_{\mathrm{sup}} (A)
\]

This corollary turns out to be surprisingly useful since many algorithms
are uniformly stable. For example, strong convexity of the loss function
is sufficient for the uniform stability of empirical risk minimization,
as we will see next.

\hypertarget{stability-of-empirical-risk-minimization}{%
\subsection{Stability of empirical risk
minimization}\label{stability-of-empirical-risk-minimization}}

We now show that empirical risk minimization is uniformly stable
provided under strong assumptions on the loss function. One important
assumption we need is that the loss function~\(\loss(w, z)\) is
differentiable and \emph{strongly convex} in the model parameters~\(w\)
for every example~\(z.\) What this means is that for every example~\(z\)
and for all~\(w,w'\in\Omega,\) \[
\loss(w',z)\ge
\loss(w,z)
+ \langle\nabla \loss(w,z), w' - w\rangle
+ \frac{\mu}2\|w-w'\|^2\,.
\] There's only one property of strong convexity we'll need. Namely, if
\(\Phi\colon\mathbb{R}^d\to\mathbb{R}\) is~\(\mu\)-strongly convex
and~\(w^*\) is a stationary point (and hence global minimum) of the
function~\(\Phi\), then we have \[
\Phi(w) - \Phi(w^*) \geq \frac{\mu}{2} \| w - w^* \|^2\,.
\] The second assumption we need is that~\(\loss(w, z)\) is
\emph{\(L\)-Lipschitz} in~\(w\) for every~\(z\), i.e.,
\(\|\nabla \loss(w, z)\|\le L\,.\) Equivalently, this means
\(|\loss(w, z)-\loss(w',z)|\le L\|w-w'\|.\)

\begin{Theorem}

Assume that for every~\(z\),~\(\loss(w, z)\) is~\(\mu\)-strongly convex
in~\(w\) over the domain~\(\Omega\), i.e., Further assume that, that the
loss function~\(\loss(w, z)\) is~\(L\)-Lipschitz in~\(w\) for
every~\(z.\) Then, empirical risk minimization (ERM) satisfies \[
\Delta_{\mathrm{sup}} (\text{ERM}) \leq \frac{4L^2}{\mu n}\,.
\]

\end{Theorem}

\begin{Proof}

Let
\(\hat w_S = \arg \min_{w\in \Omega} \frac{1}{n} \sum_{i=1}^n\loss(w, z_i)\)
denote the empirical risk minimizer on the sample~\(S.\) Fix arbitrary
samples~\(S,S'\) of size~\(n\) that differ in a single index
\(i\in\{1,\dots,n\}\) where~\(S\) contains~\(z_i\) and~\(S'\)
contains~\(z_i'\,.\) Fix an arbitrary example~\(z\,.\) We need to show
that \[
|\loss(\hat w_{S}, z) - \loss(\hat w_{S'}, z)|
\leq \frac{4 L^2}{\mu n}\,.
\] Since the loss function is~\(L\)-Lipschitz, it suffices to show that
\[
\| \hat w_{S} - \hat w_{S'} \| \leq \frac{4L}{\mu n}\,.
\]

On the one hand, since~\(\hat w_S\) minimizes the empirical risk by
definition, it follows from the strong convexity of the empirical risk
that \[
\frac{\mu}{2} \| \hat w_{S} - \hat w_{S'} \|^2
\le R_S[\hat w_{S'}] - R_S[\hat w_{S}]\,.
\]

On the other hand, we can bound the right hand side as \[
\begin{aligned}
& R_S[\hat w_{S'}] - R_S[\hat w_{S}]\\
&=  \frac{1}{n} ( \loss(\hat w_{S'}, z_i) - \loss(\hat w_{S}, z_i)) + \frac{1}{n} \sum_{i \neq j} ( \loss(\hat w_{S'}, z_j) - \loss(\hat w_{S}, z_j))\\
&=\frac{1}{n} (\loss(\hat w_{S'}, z_i) - \loss(\hat w_{S}, z_i))
+ \frac{1}{n} (\loss(\hat w_{S}, z_i') - \loss(\hat w_{S'}, z_i')) \\
&
\qquad + \left(R_{S'}[\hat w_{S'}] - R_{S'}[\hat w_{S}]\right)\\
&\leq \frac{1}{n} | \loss(\hat w_{S'}, z_i) - \loss(\hat w_{S}, z_i)| +
\frac{1}{n} |  \loss(\hat w_{S}, z_i') - \loss(\hat w_{S'}, z_i') |\\
&\leq  \frac{2 L}{n} \| \hat w_{S'} - \hat w_{S}\|\,.
\end{aligned}
\]

Here, we used the assumption that~\(\loss\) is~\(L\)-Lipschitz and the
fact that \[
R_{S'}[\hat w_{S'}] - R_{S'}[\hat w_{S}] \leq 0\,.
\]

Putting together the strong convexity property and our calculation
above, we find \[
\| \hat w_{S'} - \hat w_{S} \| \leq \frac{4 L}{\mu n}\,.
\]

Hence,~\(\Delta_{\mathrm{sup}}(\text{ERM}) \leq \frac{4 L ^ 2}{\mu n}\,.\)

\end{Proof}

An interesting point about this result is that there is no explicit
reference to the complexity of the model class referenced by~\(\Omega.\)

\hypertarget{stability-of-regularized-empirical-risk-minimization}{%
\subsection{Stability of regularized empirical risk
minimization}\label{stability-of-regularized-empirical-risk-minimization}}

\index{regularization}\index{regularized ERM}

Some empirical risk minimization problems, such as the Perceptron (ERM
with hinge loss) we saw earlier, are convex but not strictly convex. We
can turn convex problems into strongly convex problems by adding an
\emph{\(\ell_2\)-regularization} term to the loss function: \[
r(w, z) = \loss(w, z) + \frac{\mu}{2} \| w \|^2\,.
\] The last term is named~\(\ell_2\)-regularization, \emph{weight
decay}, or \emph{Tikhonov regularization} depending on field and
context.\index{weight decay}\index{Tikhonov regularization}

By construction, if the loss is convex, then the regularized loss
\(r(w, z)\) is~\(\mu\)-strongly convex. Hence, our previous theorem
applies. However, by adding regularization we changed the objective
function. The optimizer of the regularized objective is in general not
the same as the optimizer of the unregularized objective.

Fortunately, A simple argument shows that solving the regularized
objective also solves the unregularized objective. The idea is that
assuming~\(\|w\|\le B\) we can set the regularization parameter
\(\mu = \frac{L}{B\sqrt{n}}\,.\) This ensures that the regularization
term \(\mu\|w\|^2\) is at most~\(O(\frac{LB}{\sqrt{n}})\) and therefore
the minimizer of the regularized risk also minimizes the unregularized
risk up to error~\(O(\frac{LB}{\sqrt{n}})\,.\) Plugging this choice
of~\(\mu\) into the ERM stability theorem, the generalization gap will
also be \(O(\frac{LB}{\sqrt{n}})\,.\)

\hypertarget{the-case-of-regularized-hinge-loss}{%
\subsection{The case of regularized hinge
loss}\label{the-case-of-regularized-hinge-loss}}

Let's relate the generalization theory we just saw to the familiar case
of the perceptron algorithm from Chapter 3. This corresponds to the
special case of minimizing the regularized hinge loss \[
r(w, (x,y)) = \max\{1-y\langle w,x\rangle,0\} + \frac{\mu}{2} \| w \|^2\,.
\] Moreover, we assume that the data are are linearly separable with
margin~\(\gamma\).

Denoting by~\(\hat w_S\) the empirical risk minimizer on a random sample
\(S\) of size~\(n\), we know that \[
\frac\mu2\|\hat w_S\|^2
\le R_S(\hat w_S)
\le R_S(0)
= 1\,.
\] Hence,~\(\|\hat w_S\|\le B\) for~\(B=\sqrt{2/\mu}.\) We can therefore
restrict our domain to the Euclidean ball of radius~\(B\). If the data
are also bounded, say~\(\Vert x \Vert \leq D\), we further get that \[
  \Vert \nabla_w r(w,z) \Vert \le \|x\| + \mu\|w\| = D + \mu B\,.
\] Hence, the regularized hinge loss is~\(L\)-Lipschitz with \[
L=D+\mu B=D+\sqrt{2\mu}\,.
\]

Let~\(w_\gamma\) be a maximum margin hyperplane for the sample~\(S\). We
know that the empirical loss will satisfy \[
R_S[\hat w_S] \leq R_S[w_\gamma] = \frac{\mu}{2}\| w_\gamma \|^2 = \frac{\mu }{2\gamma^2}\,.
\] Hence, by Theorem 4, \[
\E[R[\hat w_S]]
\le \E[R_S[\hat w_S]]
+ \Delta_{\mathrm{sup}}(\mathrm{ERM})
\le \frac{\mu }{2\gamma^2}+
\frac{4(D+\sqrt{2\mu})^2}{\mu n}
\] Setting~\(\mu=\frac{2\gamma D}{\sqrt{n}}\) and noting that
\(\gamma\le D\) gives that \[
\E[R[\hat w_S]] \le O\left(\frac{D}{\gamma \sqrt{n}}\right)\,.
\] Finally, since the regularized hinge loss upper bounds the zero-one
loss, we can conclude that \[
  \Pr[Y \hat w_S^T X < 0] \leq O\left(\frac{D}{\gamma \sqrt{n}}\right)\,,
\] where the probability is taken over both sample~\(S\) and test point
\((X, Y)\). Applying Markov's inequality to the sample, we can conclude
the same bound holds for a typical sample up to constant factors.

This bound is proportional to the square root of the bound we saw for
the perceptron in Chapter 3. As we discussed earlier, this rate is
slower than the perceptron rate as it does not explicitly take into
account the fact that the empirical risk is zero. However, it is worth
noting that the relationship between the variables in
question---diameter, margin, and number of samples---is precisely the
same as for the perceptron. This kind of bound is common and we will
derive it a few more times in this chapter.

Stability analysis combined with explicit regularization and convexity
thus give an appealing conceptual and mathematical approach to reasoning
about generalization. However, empirical risk minimization involving
non-linear models is increasingly successful in practice and generally
leads to non-convex optimization problems.

\hypertarget{model-complexity-and-uniform-convergence}{%
\section{Model complexity and uniform
convergence}\label{model-complexity-and-uniform-convergence}}

\index{uniform convergence}

We briefly review other useful tools to reason about generalization.
Arguably, the most basic is based on counting the number of different
functions that can be described with the given model parameters.

Given a sample~\(S\) of~\(n\) independent draws from the same underlying
distribution, the empirical risk~\(R_S[f]\) for a fixed function~\(f\)
is an average of~\(n\) random variables, each with mean equal to the
risk \(R[f]\,.\) Assuming for simplicity that the range of our loss
function is bounded in the interval~\([0,1]\,,\) Hoeffding's bound gives
us the tail bound \[
\Pr\left[ R_S[f] > R[f] + t\right] \le \exp(-2nt^2)\,.
\]

By applying the union bound to a finite set of functions~\(\mathcal{F}\)
we can guarantee that with probability~\(1-\delta,\) we have for all
functions~\(f\in\mathcal{F}\) that \[
\gengap(f)\le\sqrt{\frac{\ln|\mathcal{F}|+\ln(1/\delta)}{n}}\,.\quad\quad(2)
\]

The cardinality bound~\(|\mathcal{F}|\) is a basic measure of the
complexity of the model family~\(\mathcal{F}\,.\) We can think of the
term \(\ln(\mathcal{F})\) as a measure of complexity of the function
class \(\mathcal{F}\,.\) The gestalt of the generalization bound as
``\(\sqrt{\mathrm{complexity}/n}\)'' routinely appears with varying
measures of complexity.

\hypertarget{vc-dimension}{%
\subsection{VC dimension}\label{vc-dimension}}

\index{VC dimension}

Bounding the generalization gap from above for all functions in a
function class is called \emph{uniform convergence}. A classical tool to
reason about uniform convergence is the Vapnik-Chervonenkis dimension
(VC dimension) of a function class
\(\mathcal{F}\subseteq X\rightarrow Y\), denoted
\(\mathrm{VC}(\mathcal{F})\,.\) It's defined as the size of the largest
set~\(Q\subseteq X\) such that for any Boolean function
\(h\colon Q \to \{-1,1\},\) there is a predictor~\(f\in\mathcal{F}\)
such that~\(f(x)= h(x)\) for all~\(x\in Q\,.\) In other words, if there
is a size-\(d\) sample~\(Q\) such that the functions of~\(\mathcal{F}\)
induce all \(2^d\) possible binary labelings of~\(Q\), then the
VC-dimension of \(\mathcal{F}\) is at least~\(d\,.\)

The VC-dimension measures the ability of the model class to conform to
an arbitrary labeling of a set of points. The so-called VC inequality
implies that with probability~\(1-\delta,\) we have for all functions
\(f\in\mathcal{F}\) \[
\gengap(f)\le \sqrt{\frac{\mathrm{VC}(\mathcal{F})\ln n+\ln(1/\delta)}{n}}\,.\quad\quad(3)
\]

We can see that the complexity term~\(\mathrm{VC}(\mathcal{F})\) refines
our earlier cardinality bound since
\(\mathrm{VC}(\mathcal{F})\le\log|\mathcal{F}|+1.\) However VC-dimension
also applies to infinite model classes. Linear models over~\(\R^d\) have
VC-dimension~\(d\), corresponding to the number of model parameters.
Generally speaking, VC dimension tends to grow with the number of model
parameters for many model families of interest. In such cases, the bound
in Equation 3 becomes useless once the number of model parameters
exceeds the size of the sample.

However, the picture changes significantly if we consider notions of
model complexity different than raw counts of parameters. Consider the
set of all hyperplanes in~\(\R^d\) with norm at most~\(\gamma^{-1}\,.\)
Vapnik showed\citep{vapnik1998SLTBook} that when the data had maximum
norm \(\|x\|\leq D\,,\) then the VC dimension of this set of hyperplanes
was \(\frac{D^2}{\gamma^2}\,.\) As described in a survey of support
vector machines by Burges, the worst case arrangement of~\(n\) data
points is a simplex in~\(n-2\) dimensions.\citep{burges1998tutorial}
Plugging this VC-dimension into our generalization bound yields \[
\gengap(f)\le \sqrt{\frac{D^2\ln n+\gamma^2\ln(1/\delta)}{\gamma^2 n}}\,.
\] We again see our Perceptron style generalization bound! This bound
again holds when the empirical risk is nonzero. And the dimension of the
data,~\(d\) does not appear at all in this bound. The difference between
the parametric model and the margin-like bound is that we considered
properties of the data. In the \emph{worst case} bound which counts
parameters, it appears that high-dimensional prediction is impossible.
It is only by considering data-specific properties that we can find a
reasonable generalization bound.

\hypertarget{rademacher-complexity}{%
\subsection{Rademacher complexity}\label{rademacher-complexity}}

\index{Rademacher complexity}

An alternative to VC-dimension is Rademacher complexity, a flexible tool
that often is more amenable to calculations that incorporate
problem-specific aspects such as restrictions on the distribution family
or properties of the loss function. To get a generalization bound in
terms of Rademacher complexity, we typically apply the definition not
the model class~\(\mathcal{F}\) itself but to the class of functions
\(\mathcal{L}\) of the form~\(h(z) = \loss(f, z)\) for some
\(f\in\mathcal{F}\) and a loss function~\(\loss\,.\) By varying the loss
function, we can derive different generalization bounds.

Fix a function class~\(\mathcal{L}\subseteq Z\to\mathbb{R}\), which will
later correspond to the composition of a predictor with a loss function,
which is why we chose the symbol~\(\mathcal{L}\,.\) Think of the domain
\(Z\) as the space of labeled examples~\(z=(x,y)\,.\) Fix a
distribution~\(P\) over the space~\(Z\,.\)

The \emph{empirical Rademacher complexity} of a function class
\(\mathcal{L}\subseteq Z\to\R\) with respect to a sample
\(\{z_1,\dots,z_n\}\subseteq Z\) drawn i.i.d. from the
distribution~\(P\) is defined as: \[
\widehat{\mathfrak{R}}_n(\mathcal{L})
=\E_{\sigma\in\{-1,1\}^n}
\left[
\frac 1n \sup_{h\in\mathcal{L}}
\left|
\sum_{i=1}^n \sigma_i h(z_i)
\right|
\right]\,.
\]

We obtain the \emph{Rademacher complexity}
\(\mathfrak{R}_n(\mathcal{L})=\E\left[\widehat{\mathfrak{R}}_n(\mathcal{L})\right]\)
by taking the expectation of the empirical Rademacher complexity with
respect to the sample. Rademacher complexity measures the ability of a
function class to interpolate a random sign pattern assigned to a point
set.

One application of Rademacher complexity applies when the loss function
is~\(L\)-Lipschitz in the parameterization of the model class for every
example~\(z\,.\) This bound shows that with probability~\(1-\delta\) for
all functions~\(f\in\mathcal{F},\) we have \[
\gengap(f)\le 2L{\mathfrak{R}}_n(\mathcal{F}) + 3\sqrt{\frac{\log(1/\delta)}{n}}\,.
\] When applied to the hinge loss with the function class being
hyperplanes of norm less than~\(\gamma^{-1}\), this bound again recovers
the perceptron generalization bound \[
\gengap(f)\le 2 \frac{D}{\gamma \sqrt{n}}
 + 3\sqrt{\frac{\log(1/\delta)}{n}}\,.
\]

\hypertarget{margin-bounds-for-ensemble-methods}{%
\subsection{Margin bounds for ensemble
methods}\label{margin-bounds-for-ensemble-methods}}

\index{boosting}\index{ensemble}\index{margin bounds}

Ensemble methods work by combining many weak predictors into one strong
predictor. The combination step usually involves taking a weighted
average or majority vote of the weak predictors. Boosting and random
forests are two ensemble methods that continue to be highly popular and
competitive in various settings. Both methods train a sequence of small
decision trees, each on its own achieving modest accuracy on the
training task. However, so long as different trees make errors that
aren't too correlated, we can obtain a higher accuracy model by taking,
say, a majority vote over the individual predictions of the trees.

Researchers in the 1990s already observed that boosting often continues
to improve test accuracy as more weak predictors are added to the
ensemble. The complexity of the entire ensemble was thus often far too
large to apply standard uniform convergence bounds.

A proffered explanation was that boosting, while growing the complexity
of the ensemble, also improved the \emph{margin} of the ensemble
predictor. Assuming that the final predictor~\(f\colon X\to \{-1,1\}\)
is binary, its \emph{margin} on an example~\((x,y)\) is defined as the
value~\(yf(x)\,.\) The larger the margin the more ``confident'' the
predictor is about its prediction. A margin~\(yf(x)\) just above~\(0\)
shows that the weak predictors in the ensemble were nearly split evenly
in their weighted votes.

An elegant generalization bound relates the risk of any predictor~\(f\)
to the fraction of correctly labeled training examples at a given margin
\(\theta.\) Below let~\(R[f]\) be the risk of~\(f\) w.r.t. zero-one
loss. However, let~\(R^\theta_S(f)\) be the empirical risk with respect
to \emph{margin errors} at level~\(\theta\), i.e., the loss
\(\mathbf{1}(yf(x)\le\theta)\) that penalizes errors where the predictor
is within an additive~\(\theta\) margin of making a mistake.

\begin{Theorem}

With probability~\(1-\delta\), every convex combination~\(f\) of base
predictors in~\(\mathcal{H}\) satisfies the following bound for every
\(\theta>0:\)

\[
R[f]
- R^\theta_S[f]
\le O\left(
\frac1{\sqrt{n}}\left(\frac{
\mathrm{VC}(\mathcal{H})\log n}{\theta^2}
+\log(1/\delta)
\right)^{1/2}
\right)
\]

\end{Theorem}

The theorem can be proved using Rademacher complexity. Crucially, the
bound only depends on the VC dimension of the base class~\(\mathcal{H}\)
but not the complexity of ensemble. Moreover, the bound holds for all
\(\theta>0\) and so we can choose~\(\theta\) after knowing the margin
that manifested during training.

\hypertarget{margin-bounds-for-linear-models}{%
\subsection{Margin bounds for linear
models}\label{margin-bounds-for-linear-models}}

Margins also play a fundamental role for linear prediction. We saw one
margin bound for linear models in our chapter on the Perceptron
algorithm. Similar bounds hold for other variants of linear prediction.
We'll state the result here for a simple least squares problem: \[
w^* = \arg\min_{w\colon\|w\|\le B} \frac1n\sum_{i=1}^n\left(\langle x_i, w\rangle - y\right)^2
\]

In other words, we minimize the empirical risk w.r.t. the squared loss
over norm bounded linear separators, call this
class~\(\mathcal{W}_B\,.\) Further assume that all data points
satisfy~\(\|x_i\|\le 1\) and \(y\in\{-1,1\}.\) Analogous to the margin
bound in Theorem \(5\), it can be shown that with probability
\(1-\delta\) for every linear predictor~\(f\) specified by weights in
\(\mathcal{W}_B\) we have \[
R[f]
- R_S^\theta[f]
\le 4\frac{{\mathfrak{R}}(\mathcal{W}_B)}{\theta}
+ O\left(\frac{\log(1/\delta)}{\sqrt{n}}\right)\,.
\]

Moreover, given the assumptions on the data and model class we made, the
Rademacher complexity satisfies
\({\mathfrak{R}}(\mathcal{W})\le B/\sqrt{n}.\) What we can learn from
this bound is that the relevant quantity for generalization is the ratio
of complexity to margin~\(B/\theta\,.\)

It's important to understand that margin is a scale-sensitive notion; it
only makes sense to talk about it after suitable normalization of the
parameter vector. If the norm didn't appear in the bound we could scale
up the parameter vector to achieve any margin we want. For linear
predictors the Euclidean norm provides a natural and often suitable
normalization.

\hypertarget{generalization-from-algorithms}{%
\section{Generalization from
algorithms}\label{generalization-from-algorithms}}

In the overparameterized regime, there are always an infinite number of
models that minimize empirical risk. However, when we run a particular
algorithm, the algorithm usually returns only one from this continuum.
In this section, we show how directly analyzing algorithmic iteration
can itself yield generalization bounds.

\hypertarget{one-pass-optimization-of-stochastic-gradient-descent}{%
\subsection{One pass optimization of stochastic gradient
descent}\label{one-pass-optimization-of-stochastic-gradient-descent}}

As we briefly discussed in the optimization chapter, we can interpret
the convergence analysis of stochastic gradient descent as directly
providing a generalization bound for a particular variant of SGD. Here
we give the argument in full detail. Suppose that we choose a loss
function that upper bounds the number of mistakes. That is
\(\loss(\hat{y},y) \geq \mathbb{1}\{y\hat{y}<0\}\,.\) The hinge loss
would be such an example. Choose the function~\(R\) to be the risk (not
empirical risk!) with respect to this loss function: \[
    R[w] = \E[ \loss(w^T x, y) ]
\] At each iteration, suppose we gain access to an example pair
\((x_i, y_i)\) sampled i.i.d. from the a data generating distribution.
Then when we run the stochastic gradient method, the iterates are \[
    w_{t+1} = w_t - \alpha_t e(w_t^Tx_t,y_t) x_t\,,\quad
\text{where}\quad
    e(z,y) = \frac{\partial \loss(z,y)}{\partial z}\,.
\] Suppose that for all~\(x\),~\(\Vert x \Vert \leq D\,.\) Also suppose
that \(|e(z,y)|\leq C\,.\) Then the SGD convergence theorem tells us
that after \(n\) steps, starting at~\(w_0=0\) and using an appropriately
chosen constant step size, the average of our iterates~\(\bar{w}_n\)
will satisfy \[
\Pr[\operatorname{sign}(\bar{w}_n^T x ) \neq y ] \leq
\mathbb{E}[R[\bar{w}_n]] \leq R[w_\star]  + \frac{C D \Vert w_\star \Vert }{\sqrt{n}}\,.
\] This inequality tells us that we will find a distribution boundary
that has low \emph{population} risk after seeing~\(n\) samples. And the
population risk itself lets us upper bound the probability of our model
making an error on new data. That is, this inequality is a
generalization bound.

We note here that this importantly does not measure our empirical risk.
By running stochastic gradient descent, we can find a low-risk model
without ever computing the empirical risk.

Let us further assume that the population can be separated with large
margin. As we showed when we discussed the Perceptron, the margin is
equal to the inverse of the norm of the corresponding hyperplane.
Suppose we ran the stochastic gradient method using a hinge loss. In
this case,~\(C=1\), so, letting~\(\gamma\) denote the maximum margin, we
get the simplified bound \[
\Pr[\operatorname{sign}(\bar{w}_n^T x ) \neq y ] \leq
 \frac{D}{\gamma\sqrt{n}}\,.
\] Note that the Perceptron analysis did not have a step size parameter
that depended on the problem instance. But, on the other hand, this
analysis of SGD holds regardless of whether the data is separable or
whether zero empirical risk is achieved after one pass over the data.
The stochastic gradient analysis is more general but generality comes at
the cost of a looser bound on the probability of error on new examples.

\hypertarget{uniform-stability-of-stochastic-gradient-descent}{%
\subsection{Uniform stability of stochastic gradient
descent}\label{uniform-stability-of-stochastic-gradient-descent}}

Above we showed that empirical risk minimization is stable no matter
what optimization method we use to solve the objective. One weakness is
that the analysis applied to the exact solution of the optimization
problem and only applies for strongly convex loss function. In practice,
we might only be able to compute an approximate empirical risk minimizer
and may be interested in losses that are not strongly convex.
Fortunately, we can also show that some optimization methods are stable
even if they don't end up computing a minimizer of a strongly convex
empirical risk. Specifically, this is true for the stochastic gradient
method under suitable assumptions. Below we state one such result which
requires the assumption that the loss function is \emph{smooth}. A
continuously differentiable
function~\(f\colon\mathbb{R}^d\to\mathbb{R}\) is~\(\beta\)-smooth
if~\(\|\nabla f(y)-\nabla f(x)\|\le \beta\|y-x\|.\)

\begin{Theorem}

Assume a continuously differentiable loss function that is
\(\beta\)-smooth and~\(L\)-Lipschitz on every example and convex.
Suppose that we run the stochastic gradient method (SGM) with step sizes
\(\eta_t\le 2/\beta\) for~\(T\) steps. Then, we have \[
\Delta_{\mathrm{sup}} (\text{SGM}) \leq \frac{2L^2}{n}\sum_{t=1}^T\eta_t\,.
\]

\end{Theorem}

The theorem allows for SGD to sample the same data points multiple
times, as is common practice in machine learning. The stability approach
also extends to the non-convex case albeit with a much weaker
quantitative bound.

\hypertarget{what-solutions-does-stochastic-gradient-descent-favor}{%
\subsection{What solutions does stochastic gradient descent
favor?}\label{what-solutions-does-stochastic-gradient-descent-favor}}

We reviewed empirical evidence that explicit regularization is not
necessary for generalization. Researchers therefore believe that a
combination of data generating distribution and optimization algorithm
perform \emph{implicit regularization}. Implicit regularization
describes the tendency of an algorithm to seek out solutions that
generalize well on their own on a given a dataset without the need for
explicit correction. Since the empirical phenomena we reviewed are all
based on gradient methods, it makes sense to study implicit
regularization of gradient descent. While a general theory for
non-convex problems remains elusive, the situation for linear models is
instructive.

Consider again the linear case of gradient descent or stochastic
gradient descent: \[
    w_{t+1} = w_t - \alpha e_t x_t
\] where~\(e_t\) is the gradient of the loss at the current prediction.
As we showed in the optimization chapter, if we run this algorithm to
convergence, we must have the resulting~\(\hat{w}\) lies in the span of
the data, and that it interpolates the data. These two facts imply that
the optimal~\(\hat{w}\) is the minimum Euclidean norm solution of
\(Xw=y\,.\) That is,~\(w\) solves the optimization problem \[
\begin{array}{ll}
    \text{minimize} & \|w\|^2\\
    \text{subject to} & y_i w^Tx_i = 1\,.
\end{array}
\] Moreover, a closed form solution of this problem is given by \[
    \hat{w}=X^T(XX^T)^{-1}y\,.
\] That is, when we run stochastic gradient descent we converge to a
very specific solution. Now what can we say about the generalization
properties of this minimum norm interpolating solution?

The key to analyzing the generalization of the minimum norm solution
will be a stability-like argument. We aim to control the error of the
model trained on the first~\(m\) data points on the next data point in
the sequence,~\(x_{m+1}\,.\) To do so, we use a simple identity that
follows from linear algebra.

\begin{Lemma}

Let~\(S\) be an arbitrary set of~\(m\geq 2\) data points.
Let~\(w_{m-1}\) and \(w_{m}\) denote the minimum norm solution trained
on the first~\(m-1\) and \(m\) points respectively. Then \[
    (1-y_{m} \langle w_{m-1}, x_m \rangle)^2 = s_{m}^2 (\Vert w_m \Vert^2-\Vert w_{m-1} \Vert^2)\,,
\] where \[
    s_{m} := \operatorname{dist} \left(\operatorname{span}( x_1,\ldots, x_{m-1}),x_{m}\right)\,.
\]

\end{Lemma}

We hold off on proving this lemma and first prove our generalization
result with the help of this lemma.

\begin{Theorem}

Let~\(S_{n+1}\) denote a set of~\(n+1\) i.i.d. samples. Let~\(S_j\)
denote the first~\(j\) samples and~\(w_j\) denote the solution of
minimum norm that interpolates these~\(j\) points. Let~\(R_j\) denote
the maximum norm of \(\Vert x_i \Vert\) for~\(1\leq i\leq j\,.\)
Let~\((x,y)\) denote another independent sample from~\(\mathcal{D}\,.\)
Then if \(\epsilon_j:=\E[(1 - y f_{S_j}(x))^2]\) is a non-increasing
sequence, we have \[
    \Pr[ y \langle w_{n}, x \rangle < 0] \leq  \frac{\E[R_j^2 \Vert w_{n+1} \Vert^2]}{n}\,.
\]

\end{Theorem}

\begin{Proof}

Lemma together with the bound~\(s_i^2 \leq R_{n+1}^2\) yields the
inequality \[
    \E[(1-y \langle w_i, x \rangle)^2] \leq  (\E[ R_{n+1}^2 \Vert w_{i+1} \Vert^2]-\E[ R_{n+1}^2\Vert w_i\Vert^2])\,.
\] Here, we could drop the subscript on~\(x\) and~\(y\) on the left-hand
side as they are identically distributed to~\((x_{i+1},y_{i+1})\,.\)
Adding these inequalities together gives the bound \[
    \frac{1}{n} \sum_{i=1}^n \E[(1 - y f_{S_i}(x))^2]  \leq \frac{\E[R_{n+1}^2 \Vert w_{n+1} \Vert^2]}{n}\,.
\] Assuming the sequence is decreasing means that the minimum summand of
the previous inequality is~\(\E[(1 - y f_{i}(x))^2]\,.\) This and
Markov's inequality prove the theorem.

\end{Proof}

This proof reveals that the minimum norm solution, the one found by
running stochastic gradient descent to convergence, achieves a nearly
identical generalization bound as the Perceptron, even with the fast
\(1/n\) rate. Here, nothing is assumed about margin, but instead we
assume that the complexity of the interpolating solution does not grow
rapidly as we increase the amount of data we collect. This proof
combines ideas from stability, optimization, and model complexity to
find yet another explanation for why gradient methods find high-quality
solutions to machine learning problems.

\hypertarget{proof-of-lemma-lemminimumnorm}{%
\subsection{Proof of Lemma 3}\label{proof-of-lemma-lemminimumnorm}}

We conclude with the deferred proof of Lemma 3.

\begin{Proof}

Let~\(K= XX^T\) denote the kernel matrix for~\(S\,.\) Partition~\(K\) as
\[
    K = \begin{bmatrix} K_{11} & K_{12} \\ K_{21} & K_{22} \end{bmatrix}
\] where~\(K_{11}\) is~\((m-1) \times (m-1)\) and~\(K_{22}\) is a scalar
equal to~\(\langle x_m,x_m \rangle\,.\) Similarly, partition the vector
of labels~\(y\) so that~\(y^{(m-1)}\) denotes the first~\(m-1\) labels.
Under this partitioning, \[
 \langle w_{m-1} , x_m \rangle = K_{21} K_{11}^{-1} y^{(m-1)}\,.
\]

Now note that \[
s_m^2 = K_{22}-K_{21} K_{11}^{-1} K_{12}\,.
\] Next, using the formula for inverting partitioned matrices, we find
\[
    K^{-1} = \begin{bmatrix}
    (K_{11}-K_{12}K_{21}K_{22}^{-1})^{-1} & s_m^{-2} K_{11}^{-1} K_{12} \\
    s_m^{-2} (K_{11}^{-1} K_{12})^T & s_m^{-2}
    \end{bmatrix}\,.
\] By the matrix inversion lemma we have \[
    (K_{11}-K_{12}K_{21}K_{22}^{-1})^{-1}
    = K_{11}^{-1} +s_m^{-2} \left(K_{21} K_{11}^{-1}\right)^T\left(K_{21} K_{11}^{-1}\right) \,.
\] Hence, \[
\begin{aligned}
    \Vert w_{i}\Vert
    &= y^T K^{-1} y \\
    &= s_m^{-2}(y_m^2 - 2y_m \langle w_{m-1} ,x_m\rangle +\langle w_{m-1} ,x_m\rangle^2 ) + {y^{(m-1)}}^T K_{11}^{-1} y^{(m-1)}\,.
\end{aligned}
\] Rearranging terms proves the lemma.

\end{Proof}

\hypertarget{looking-ahead}{%
\section{Looking ahead}\label{looking-ahead}}

Despite significant effort and many recent advances, the theory of
generalization in overparameterized models still lags behind the
empirical phenomenology. What governs generalization remains a matter of
debate in the research community.

Existing generalization bounds often do not apply directly to practice
by virtue of their assumptions, are quantitatively too weak to apply to
heavily overparameterized models, or fail to explain important empirical
observations. However, it is not just a lack of quantitative sharpness
that limits our understanding of generalization.

Conceptual questions remain open: What is it a successful theory of
generalization should do? What are formal success criteria? Even a
qualitative theory of generalization, that is not quantitatively precise
in concrete settings, may be useful if it leads to the successful
algorithmic interventions. But how do we best evaluate the value of a
theory in this context?

Our focus in this chapter was decidedly narrow. We discussed how to
related risk and empirical risk. This perspective can only capture
questions that relate performance on a sample to performance on the very
same distribution that the sample was drawn from. What is left out are
important questions of \emph{extrapolation} from a training environment
to testing conditions that differ from training. Overparameterized
models that generalize well in the narrow sense can fail dramatically
even with small changes in the environment. We will revisit the question
of generalization for overparameterized models in our chapter on deep
learning.

\hypertarget{chapter-notes-5}{%
\section{Chapter notes}\label{chapter-notes-5}}

The tight characterization of generalization gap in terms of average
stability, as well as stability of regularized empirical risk
minimization (Theorem 4), is due to Shalev-Shwartz et
al.\citep{shalev2010learnability} Uniform stability was introduced by
Bousquet and Elisseeff.\citep{bousquet2002stability} For additional
background on VC dimension and Rademacher complexity, see, for example,
the text by Shalev-Shwartz and Ben-David.\citep{shalev2014understanding}

The double descent figure is from work of Belkin et
al.\citep{belkin2019reconciling} Earlier work pointed out similar
empirical risk-complexity relationships.\citep{neyshabur2014search} The
empirical findings related to the randomization test and the role of
regularization are due to Zhang et al.\citep{zhang2017understanding}

Theorem 5 is due to Schapire et al.\citep{schapire1998boosting} Later
work showed theoretically that boosting maximizes
margin.\citep{zhang2005boosting, telgarsky2013margins} The margin bound
for linear models follows from more general results of Kakade,
Sridharan, and Tewari\citep{kakade2009complexity} that build on earlier
work by Bartlett and Mendelson\citep{bartlett2002rademacher}, as well as
work of Koltchinskii and Panchenko\citep{koltchinskii2002empirical}.
Rademacher complexity bounds for family of neural networks go back to
work of Bartlett\citep{bartlett98} and remain and active research topic.
We will see more on this in our chapter on deep learning.

The uniform stability bound for stochastic gradient descent is due to
Hardt, Recht, and Singer\citep{hardt2016train}. Subsequent work further
explores the generalization performance stochastic gradient descent in
terms of its stability properties. Theorem 7 and Lemma 3 are due to
Liang and Recht\citep{LiangRecht2021}.

There has been an explosion of work on generalization and
overparameterization in recent years. See, also, recent work exploring
how other norms shed light on generalization
performance.\citep{neyshabur2017exploring} Our exposition is by no means
a representative survey of the broad literature on this topic. There are
several ongoing lines of work we did not cover: PAC-Bayes
bounds\citep{dziugaite2017computing}, compression
bounds\citep{arora2018stronger}, and arguments about the properties of
the optimization landscape\citep{zhang2017theory}. This chapter builds
on a chapter by Hardt\citep{hardt2021generalization}, but contains
several structural changes as well as different results.

\chapter{Deep learning}

The past chapters have sketched a path towards predictive modeling:
acquire data, construct a set of features that properly represent data
in a way such that relevant conditions can be discriminated, pose a
convex optimization problem that balances fitting training data to
managing model complexity, optimize this problem with a standard solver,
and then reason about generalization via the holdout method or cross
validation. In many ways this pipeline suffices for most predictive
tasks.

However, this standard practice does have its deficiencies. Feature
engineering has many moving pieces, and choices at one part of the
pipeline may influence downstream decisions in unpredictable ways.
Moreover, different software dependencies may be required to intertwine
the various parts of this chain, making the machine learning engineering
more fragile. It's additionally possible that more concise feature
representations are possible if the pieces can all be tuned together.

Though motivated differently by different people, \emph{deep learning}
can be understood as an attempt to ``delayer'' the abstraction
boundaries in the standard machine learning workflow. It enables
holistic design of representation and optimization. This delayering
comes at the cost of loss of convexity and some theoretical guarantees
on optimization and generalization. But, as we will now describe in this
chapter, this cost is often quite modest and, on many machine learning
problems such as image classification and machine translation, the
predictive gains can be dramatic.

Deep learning has been tremendously successful in solving industrial
machine learning problems at many tech companies. It is also the top
performing approach in most academic prediction tasks in computer
vision, speech, and other domains.

The success of deep learning is not just a purely technical matter. Once
the industry had embraced deep learning, an unprecedented amount of
resources has gone into building and refining high quality software for
practicing deep learning. The open source deep learning ecosystem is
vast and evolving quickly. For almost any task, there is already some
code available to start from. Companies are actively competing over open
source frameworks with convenient high-level syntax and extensive
documentation. An honest answer for why practitioners prefer deep
learning at this point over other methods is because it simply seems to
work better on many problems and there is a lot of quality code
available.

We now retrace our path through representation, optimization, and
generalization, highlighting what is different for deep learning and
what remains the same.

\hypertarget{deep-models-and-feature-representation}{%
\section{Deep models and feature
representation}\label{deep-models-and-feature-representation}}

We discussed in the chapter on representation that template matching,
pooling, and nonlinear lifting can all be achieved by affine
transformations followed by pointwise nonlinearities. These mappings can
be chained together to give a series of new feature vectors: \[
    x_{\ell+1} = \phi(A_\ell x_\ell+b_\ell)\,.
\] Here,~\(\ell\) indexes the \emph{layer} of a model. We can chain
several layers together to yield a final representation~\(x_L\).

As a canonical example, suppose~\(x_1\) is a pixel representation of an
image. Let's say this representation has size
\(d_1 \times d_1 \times c_1\), with~\(d_1\) counting spatial dimensions
and \(c_1\) counting the number of color channels. We could
apply~\(c_2\) template matching convolutions to this image, resulting in
a second layer of size~\(d_1 \times d_1 \times c_2\). Since we expect
convolutions to capture local variation, we can compress the size of
this second layer, averaging every~\(2\times 2\) cell to produce~\(x_2\)
of size \(d_2 \times d_2 \times c_2\), with~\(d_2<d_1\) and~\(c_2>c_1\).
Repeating this procedure several times will yield a
representation~\(x_{L-1}\) which has few spatial dimensions (\(d_{L-1}\)
is small) but many channel dimensions (\(c_{L-1}\) is large). We can
then map this penultimate layer through some universal approximator like
a neural network.

A variety of machine learning pipelines can be thought of in this way.
The first layer might correspond to edge detectors like in
SIFT\citep{lowe2004distinctive} or HOG\citep{dalal2005histograms}. The
second layer may look for parts relevant to detection like in a
deformable parts model\citep{felzenszwalb2009object}. The insight in
deep learning is that we can declare the parameters of each
layer~\(A_\ell\) and~\(b_\ell\) to be \emph{optimization variables}.
This way, we do not have to worry about the particular edge or color
detectors to use for our problem, but can instead let the collected data
dictate the best settings of these features.

This abstraction of ``features'' as ``structure linear maps with tunable
parameters'' allows for a set of basic building blocks that can be used
across a variety of domains.

\begin{enumerate}
\def\labelenumi{\arabic{enumi}.}
\item
  \textbf{Fully connected layers.} Fully connected layers are simply
  unstructured neural networks that we discussed in the representation
  chapter. For a fixed nonlinear function~\(\sigma\), a fully connected
  layer maps a vector~\(x\) to a vector~\(z\) with coordinates \[
   z_{i} = \sigma\left(\sum_j A_{ij}x_j + b_i\right)\,.
  \] While it is popular to chain fully connected layers together to get
  deep neural networks, there is little established advantage over just
  using a single layer. Daniely et al.~have backed up this empirical
  observation, showing theoretically that no new approximation power is
  gained by concatenating fully connected
  layers\citep{daniely2016toward}. Moreover, as we will discuss below,
  concatenating many layers together often slows down optimization. As
  with most things in deep learning, there's nothing saying you can't
  chain fully connected layers, but we argue that most of the gains in
  deep learning come from the structured transforms, including the ones
  we highlight here.
\item
  \textbf{Convolutions}. Convolutions are the most important building
  block in all of deep learning. We have already discussed the use of
  convolutions as template matchers that promote spatial invariances.
  Suppose the input is~\(d_0 \times d_0 \times c_0\), with the first two
  components indexing space and the last indexing channels. The
  parameter~\(A\) has size~\(q_0 \times q_0 \times c_0 \times c_1\),
  where \(q_0\) is usually small (greater than~\(2\) and less
  than~\(10\).~\(b\) typically has size~\(c_1\). The number of
  parameters used to define a convolutional layer is dramatically
  smaller than what would be used in a fully connected layer. The
  structured linear map of a convolution can be written as \[
   z_{a,b,c} = \sigma\left( \sum_{i,j,k} A_{i,j,k,c} x_{a-i,b-j,k} + b_{c} \right)\,.
  \]
\item
  \textbf{Recurrent structures.} Recurrent structures let us capture
  repeatable stationary patters in time or space. Suppose we expect
  stationarity in time. In this case, we expect each layer to represent
  a state of the system, and the next time step should be a static
  function of the previous: \[
   x_{t+1} = f(x_t)
  \] When we write~\(f\) as a neural network, this is called a
  \emph{recurrent neural network}. Recurrent neural networks \emph{share
  weights} insofar as the~\(f\) does not change from one time step to
  the next.
\item
  \textbf{Attention mechanisms.} Attention mechanisms have proven
  powerful tools in natural language processing for collections of
  vectors with dependencies that are not necessarily sequential. Suppose
  our layer is a list of~\(m\) vectors of dimension~\(d\). That
  is,~\(x\) has shape \(d\times m\). An attention layer will have two
  matrices~\(U\) and~\(V\), one to operate on the feature dimensions and
  one to operate on the sequential dimensions. The transformation takes
  form \[
   z_{a,b} = \sigma\left( \sum_{i,j} U_{a,i} x_{ij} V_{b,j} \right)\,.
  \] Just as was the case with convolutions, this structured map can
  have fewer dimensions than a fully connected layer, and can also
  respect a separation of the feature dimensions from the sequential
  dimensions in the data matrix~\(x\).
\end{enumerate}

\hypertarget{optimization-of-deep-nets}{%
\section{Optimization of deep nets}\label{optimization-of-deep-nets}}

Once we have settled on a feature representation, typically called
\emph{model architecture} in this context, we now need to solve
empirical risk minimization. Let's group all of the parameters of the
layers into a single large array of weights,~\(w\). We denote the map
from~\(x\) to prediction with weights~\(w\) as~\(f(x;w)\). At an
abstract level, empirical risk minimization amounts to minimizing \[
    R_S[w] = \frac{1}{n}\sum_{i=1}^n \loss(f(x_i;w),y_i)\,.
\] This is a nonconvex optimization problem, but we can still run
gradient methods to try to find minimizers. Our main concerns from an
optimization perspective are whether we run into local optima and how
can we compute gradient steps efficiently.

We will address gradient computation through a discussion of automatic
differentiation. With regards to global optimization, there are
unfortunately computational complexity results proving efficient
optimization of arbitrary neural networks is intractable. Even neural
nets with a single neuron can have exponentially many local
minimizers\citep{auer1996exponentially}, and finding the minimum of a
simple two-layer neural network is NP-hard
\citep{vu1998infeasibility, goel21hardness}. We cannot expect a
perfectly clean mathematical theory guiding our design and
implementation of neural net optimization algorithms.

However, these theoretical results are about the \emph{worst case}. In
practice, optimization of neural nets is often easy. If the loss is
bounded below by zero, any model with zero loss is a global minimizer.
As we discussed in the generalization chapter, one can quickly find a
global optimum of a state-of-the-art neural net by cloning a GitHub repo
and turning off the various regularization schemes. In this section, we
aim to provide some insights about the disconnect between computational
complexity in the worst case and the results achieved in practice. We
provide some partial insights as to why neural net optimization is
doable by studying the convergence of the predictions and how this
convergence can be aided by overparameterization.

\hypertarget{convergence-of-predictions-in-nonlinear-models}{%
\subsection{Convergence of predictions in nonlinear
models}\label{convergence-of-predictions-in-nonlinear-models}}

Consider the special case where we aim to minimize the square-loss. Let
\(\hat{y}_t\) denote the vector of predictions
\(\big(f(x_i;w_t)\big)_{i=1}^n\in\mathbb{R}^n\). Gradient descent
follows the iterations \[
    w_{t+1} = w_t - \alpha \jac_w f(x;w_t) (\hat{y}_t - y)
\] where~\(\jac_w f(x;w_t)\) denotes the~\(d\times n\) Jacobian matrix
of the predictions~\(\hat{y}_t\). Reasoning about convergence of the
weights is difficult, but we showed in the optimization chapter that we
could reason about convergence of the predictions: \[
    \hat{y}_{t+1}-y = (I -  \alpha \jac_w f(x; w_t)^T \jac_w f(x;w_t))(\hat{y}_t - y) + \alpha \epsilon_t\,.
\] Here, \[
    \epsilon_t = \frac{\alpha}{2} \Lambda \Vert \hat{y}_t - y \Vert^2
\] and~\(\Lambda\) bounds the curvature of the~\(f\).
If~\(\epsilon_t=0\) is sufficiently small the predictions will converge
to the training labels.

Hence, under some assumptions, nonconvexity does not stand in the way of
convergence to an empirical risk minimizer. Moreover, control of the
Jacobians~\(\jac_w f\) can accelerate convergence. We can derive some
reasonable ground rules on how to keep~\(\jac_w f\) well conditioned by
unpacking how we compute gradients of compositions of functions.

\hypertarget{automatic-differentiation}{%
\subsection{Automatic differentiation}\label{automatic-differentiation}}

For linear models it's not hard to calculate the gradient with respect
to the model parameters analytically and to implement the analytical
expression efficiently. The situation changes once we stack many
operations on top of each other. Even though in principle we can
calculate gradients using the chain rule, this gets tedious and
error-prone very quickly. Moreover, the straightforward way of
implementing the chain rule is computationally inefficient. What's worse
is that any change to the model architecture would require us to redo
our calculation and update its implementation.

Fortunately, the field of \emph{automatic differentiation} has the tools
to avoid all of these problems. At a high-level, automatic
differentiation provides efficient algorithms to compute gradients of
any function that we can write as a composition of differentiable
building blocks. Though automatic differentiation has existed since the
1960s, the success of deep learning has led to numerous free,
well-engineered, and efficient automatic differentiation packages.
Moreover, the dynamic programming algorithm behind these methods is
quite instructive as it gives us some insights into how to best engineer
model architectures with desirable gradients.

Automatic differentiation serves two useful purposes in deep learning.
First, it lowers the barrier to entry, allowing practitioners to think
more about modeling than about the particulars of calculus and numerical
optimization. Second, a standard automatic differentiation algorithm
helps us write down parseable formulas for the gradients of our
objective function so we can reason about what structures encourage
faster optimization.

To define this dynamic programming algorithm, called
\emph{backpropagation}\index{backpropagation}, it's helpful to move up a
level of abstraction before making the matter more concrete again. After
all, the idea is completely general and not specific to deep models.
Specifically, we consider an abstract computation that proceeds in~\(L\)
steps starting from an input~\(z^{(0)}\) and produces an output
\(z^{(L)}\) with~\(L-1\) intermediate steps~\(z^{(1)},\dots,z^{(L-1)}\):
\[
\begin{aligned}
  \text{input}\quad z^{(0)}\\
    z^{(1)} &= f_1(z^{(0)})\\
     &\vdots\\
    z^{(L-1)} &= f_{L-1}(z^{(L-2)})\\
\text{output}\quad z^{(L)} &=f_L(z^{(L-1)})
\end{aligned}
\] We assume that each \emph{layer} \(z^{(i)}\) is a real-valued vector
and that each function~\(f_i\) maps a real-valued vector of some
dimension to another dimension. Recall that for a
function~\(f\colon\R^n\to\R^m\), the Jacobian matrix~\(\jac(w)\) is
the~\(n\times m\) matrix of first-order partial derivatives evaluated at
the point~\(w\). When~\(m=1\) the Jacobian matrix coincides with the
transpose of the gradient.

In the context of backpropagation, it makes sense to be a bit more
explicit in our notation. Specifically, we will denote the Jacobian of a
function~\(f\) with respect to a variable~\(x\) evaluated at point~\(w\)
by \(\jac_x f(w)\,.\)

The backpropagation algorithm is a computationally efficient way of
computing the partial derivatives of the output~\(z^{(L)}\) with respect
to the input~\(z^{(0)}\) evaluated at a given parameter vector~\(w\),
that is, the Jacobian~\(\jac_{z^{(0)}} z^{(L)}(w).\) Along the way, it
computes the Jacobians for any of the intermediate layers as well.

\begin{Algorithm}

\textbf{Backpropagation}

\begin{itemize}
\tightlist
\item
  Input: parameters \(w\)
\item
  Forward pass:

  \begin{itemize}
  \tightlist
  \item
    Set \(v_0 = w\)
  \item
    For \(i =1,\ldots,L\):

    \begin{itemize}
    \tightlist
    \item
      Store \(v_{i-1}\) and compute \(v_i = f_i(v_{i-1})\)
    \end{itemize}
  \end{itemize}
\item
  Backward pass:

  \begin{itemize}
  \tightlist
  \item
    Set \(\Lambda_L = \jac_{z_{L}} z_L(v_L)= \Id\).
  \item
    For \(i=L,\ldots,1\):

    \begin{itemize}
    \tightlist
    \item
      Set \(\Lambda_{i-1} =\Lambda_i\jac_{z^{(i-1)}} z^{(i)}(v_{i-1})\).
    \end{itemize}
  \end{itemize}
\item
  Output \(\Lambda_0\).
\end{itemize}

\end{Algorithm}

First note that backpropagation runs in time~\(O(LC)\) where~\(C\) is
the computational cost of performing an operation at one step. On the
forward pass, this cost correspond to function evaluation. On the
backward pass, it requires computing the partial derivatives of
\(z^{(i)}\) with respect to~\(z^{(i-1)}\). The computational cost
depends on what the operation is. Typically, the~\(i\)-th step of the
backward pass has the same computational cost as the
corresponding~\(i\)-th step of the forward pass up to constant factors.
What is important is that computing these partial derivatives is an
entirely \emph{local} operation. It only requires the partial
derivatives of a function with respect to its input evaluated at the
value~\(v_{i-1}\) that we computed in the forward pass. This observation
is key to all fast implementations of backpropagation. Each operation in
our computation only has to implement function evaluation, its
first-order derivatives, and store an array of values computed in the
forward pass. There is nothing else we need to know about the
computation that comes before or after.

The main correctness property of the algorithm is that the final output
\(\Lambda_0\) equals the partial derivative of the output layer with
respect to the input layer evaluated at the input~\(w\).

\begin{Proposition}

\textbf{Correctness of backpropagation.} \[
\Lambda_0 = \jac_{z^{(0)}} z^{(L)}(w)
\]

\end{Proposition}

The claim directly follows by induction from the following lemma, which
states that if we have the correct partial derivatives at step~\(i\) of
the backward pass, then we also get them at step~\(i-1.\)

\begin{Lemma}

\textbf{Backpropagation invariant.} \[
\Lambda_i = \jac_{z^{(i)}} z^{(L)}(v_i)
\quad\Longrightarrow\quad
\Lambda_{i-1} = \jac_{z^{(i-1)}} z^{(L)}(v_{i-1})
\]

\end{Lemma}

\begin{Proof}

Assume that the premise holds. Then, we have \[
\begin{aligned}
\Lambda_{i-1}
&=  \Lambda_i\jac_{z^{(i-1)}} z^{(i)}(v_{i-1}) \\
& = \jac_{z^{(i)}} z^{(L)}(v_i)
    \jac_{z^{(i-1)}} z^{(i)}(v_{i-1}) \\
& = \jac_{z^{(i-1)}} z^{(L)}(v_{i-1}) \\
\end{aligned}
\] The last identity is the multivariate chain rule.

\end{Proof}

To aid the intuition, it can be helpful to write the multivariate chain
rule informally in the familiar Leibniz notation: \[
\frac{\partial z^{(L)}}{\partial z^{(i)}}\frac{\partial z^{(i)}}{\partial z^{(i-1)}}
  =\frac{\partial z^{(L)}}{\partial z^{(i-1)}}
\]

\hypertarget{a-worked-out-example}{%
\subsection{A worked out example}\label{a-worked-out-example}}

The backpropagation algorithm works in great generality. It produces
partial derivatives for any variable appearing in any layer~\(z^{(i)}\).
So, if we want partial derivatives with respect to some parameters, we
only need to make sure they appear on one of the layers. This
abstraction is so general that we can easily express all sorts of deep
architectures and the associated objective functions with it.

But let's make that more concrete and get a feeling for the mechanics of
backpropagation in the two cases that are most common: a non-linear
transformation applied coordinate-wise, and a linear transformation.

Suppose at some point in our computation we apply the \emph{rectified
linear unit} \(\mathrm{ReLU}=\max\{u, 0\}\) coordinate-wise to the
vector~\(u\). ReLU units remain one of the most common non-linearities
in deep neural networks. To implement the backward pass, all we need to
be able to do is to compute the partial derivatives
of~\(\mathrm{ReLU}(u)\) with respect to~\(u\). It's clear that
when~\(u_i > 0\), the derivative is~\(1\). When \(u_i < 0,\) the
derivative is~\(0\). The derivative is not defined at \(u_i=0,\) but we
choose to ignore this issue by setting it to be~\(0\). The resulting
Jacobian matrix is a diagonal matrix~\(D(u)\) which has a one in all
coordinates corresponding to positive coordinates of the input vector,
and is zero elsewhere.

\begin{itemize}
\tightlist
\item
  Forward pass:

  \begin{itemize}
  \tightlist
  \item
    Input: \(u\)
  \item
    Store \(u\) and compute the value \(v=\mathrm{ReLU}(u)\).
  \end{itemize}
\item
  Backward pass:

  \begin{itemize}
  \tightlist
  \item
    Input: Jacobian matrix \(\Lambda\)
  \item
    Output \(\Lambda D(u)\)
  \end{itemize}
\end{itemize}

If we were to swap out the rectified linear unit for some other
coordinate-wise nonlinearity, we'd still get a diagonal matrix. What
changes are the coefficients along the diagonal.

Now, let's consider an affine transformation of the input~\(v=Au + b\).
The Jacobian of an affine transformation is simply the matrix~\(A\)
itself. Hence, the backward pass is simple:

\begin{itemize}
\tightlist
\item
  Backward pass:

  \begin{itemize}
  \tightlist
  \item
    Input: Jacobian matrix \(\Lambda\)
  \item
    Output: \(\Lambda A\)
  \end{itemize}
\end{itemize}

We can now easily chain these together. Suppose we have a typical ReLU
network that strings together~\(L\) linear transformations with ReLU
operations in between: \[
f(x)=A_L\mathrm{ReLU}(A_{L-1}\mathrm{ReLU}(\cdots \mathrm{ReLU}(A_1x)))
\] The Jacobian of this chained operation with respect to the
input~\(x\) is given by a chain of linear operations: \[
A_LD_{L-1}A_{L-1}\cdots D_1A_1
\] Each matrix~\(D_i\) is a diagonal matrix that zeroes out some
coordinates and leaves others untouched. Now, in a deep model
architecture the weights of the linear transformation are typically
trainable. That means we really want to be able to compute the gradients
with respect to, say, the coefficients of the matrix~\(A_i\).

Fortunately, the same reasoning applies. All we need to do is to make
\(A_i\) part of the input to the matrix-vector multiplication node and
backpropagation will automatically produce derivatives for it. To
illustrate this point, suppose we want to compute the partial
derivatives with respect to, say, the~\(j\)-th column of~\(A_i\). Let
\(u=\mathrm{ReLU}(A_{i-1}\cdots (A_1x))\) be the vector that we multiply
\(A_i\) with. We know from the argument above that the Jacobian of the
output with respect to the vector~\(A_iu\) is given by the linear map
\(B=A_LD_{L-1}A_{L-1}\cdots D_i\). To get the Jacobian with respect to
the \(j\)-th column of~\(A_i\) we therefore only need to find the
partial derivative of~\(A_iu\) with respect to the~\(j\)-th column
of~\(A_i\). We can verify by taking derivatives that this
equals~\(u_j\Id\). Hence, the partial derivatives of~\(f(x)\) with
respect to the~\(j\)-th column of~\(A_i\) are given by~\(Bu_j\).

Let's add in the final ingredient, a loss function. Suppose~\(A_L\) maps
into one dimension and consider the squared loss~\(\frac12(f(x)-y)^2.\)
The only thing this loss function does is to scale our Jacobian by a
factor~\(f(x)-y.\) In other words, the partial derivatives of the loss
with respect to the~\(j\)-th columns of the weight matrix~\(A_i\) is
given by~\((f(x)-y)Bu_j\).

As usual with derivatives, we can interpret this quantity as the
\emph{sensitivity} of our loss to changes in the model parameters. This
interpretation will be helpful next.

\hypertarget{vanishing-gradients}{%
\section{Vanishing gradients}\label{vanishing-gradients}}

The previous discussion revealed that the gradients of deep models are
produced by chains of linear operations. Generally speaking, chains of
linear operations tend to either blow up the input exponentially with
depth, or shrink it, depending on the singular values of the matrices.
It is helpful to keep in mind the simple case of powering the same
symmetric real matrix~\(A\) a number of times. For almost all
vectors~\(u\), the norm~\(\|A^Lu\|\) grows
as~\(\Theta\left(\lambda_1(A)^L\right)\) asymptotically with~\(L\).
Hence it vanishes exponentially quickly if \(\lambda_1(A)<1\) and it
grows exponentially quickly if~\(\lambda_1(A)>1.\)

When it comes to gradients, these two cases correspond to the
\emph{vanishing gradients problem} and the \emph{exploding gradients
problem}, respectively. Both result in a failure case for gradient-based
optimization methods. Huge gradients are numerically unstable. Tiny
gradients preclude progress and stall any gradient-based optimization
method. We can always avoid one of the problems by scaling the weights,
but avoiding both can be delicate.

Vanishing and exploding gradients are not the fault of an optimization
method. They are a property of the model architecture. As a result, to
avoid them we need to engineer our model architectures carefully. Many
architecture innovations in deep learning aim to address this problem.
We will discuss two of them. The first are residual connections, and the
second are layer normalizations.

\hypertarget{residual-connections}{%
\subsection{Residual connections}\label{residual-connections}}

The basic idea behind \emph{residual networks} is to make each layer
close to the identity map. Traditional building blocks of neural
networks typically looked like two affine transformations~\(A, B\), with
a nonlinearity in the middle: \[
f(x)= B(\mathrm{ReLU}(A(x)))
\] Residual networks modify these building blocks by adding the input
\(x\) back to the output: \[
h(x)= x + B(\mathrm{ReLU}(A(x)))
\] In cases where the output of~\(C\) differs in its dimension
from~\(x\), practitioners use different padding or truncation schemes to
match dimensions. Thinking about the computation graph, we create a
connection from the input to the output of the building block that
\emph{skips} the transformation. Such connections are therefore called
\emph{skip connections}.

This seemingly innocuous change was hugely successful. Residual networks
took the computer vision community by storm, after they achieved leading
performance in numerous benchmarks upon their release in 2015. These
networks seemed to avoid the vanishing gradients better than prior
architectures and allowed for model depths not seen before.

Let's begin to get some intuition for why residual layers are reasonable
by thinking about what they do to the gradient computation. Let
\(J = \jac_x f(x)\) be the Jacobian of the function~\(f\) with respect
to its input. We can verify that the Jacobian of the residual block
looks like \[
J' = \jac_x h(x) = \jac_x \Id(x) + \jac_x f(x) = \Id + J\,.
\] In other words, the Jacobian of a residual block is the Jacobian of a
regular block plus the identity. This means that if we scale down the
weights of the regular block, the Jacobian~\(J'\) approaches the
identity in the sense that all its singular values are
between~\(1-\epsilon\) and \(1+\epsilon\). We can think of such a
transformation as locally well-conditioned. It neither blows up nor
shrinks down the gradient much. Since the full Jacobian of a deep
residual network will be a product of such matrices, our reasoning
suggests that suitably scaled residual networks have well-conditioned
Jacobians, and hence, as we discussed above, predictions should converge
rapidly.

Our analyses of non-convex ERM thus far have described scenarios where
we could prove the \emph{predictions} converge to the training labels.
Residual networks are interesting as we can construct cases where
\emph{the weights} converge to a unique optimal solution. This perhaps
gives even further motivation for their use.

Let's consider the simple case of where the activation function is the
identity. The resulting residual blocks are linear transformations of
the form~\(\Id + A.\) We can chain them as
\(A=(\Id + A_L)\cdots (\Id + A_1).\) Such networks are no longer
universal approximators, but they are non-convex parameterizations of
linear functions. We can turn this parameterization into an optimization
problem by solving a least squares regression problem in this residual
parameterization: \[
\begin{array}{ll}
    \text{minimize}_{A_1,\dots,A_L}  & \E[\frac12\|A(X)-Y\|^2]
\end{array}
\] Assume~\(X\) is a random vector with covariance matrix~\(\Id\) and
\(Y=B(X)+G\) where~\(B\) is a linear transformation and~\(G\) is
centered random Gaussian noise. A standard trick shows that up to an
additive constant the objective function is equal to \[
f(A_1,\dots,A_L)=\frac12\|A-B\|_F^2\,.
\] What can we say about the gradients of this function? We can verify
that the Jacobian of~\(f\) with respect to~\(A_i\) equals \[
\jac_{A_i} f(A_1,\dots, A_L)
= P^T E Q^T
\] where~\(P=(\Id+A_L)\cdots(\Id+A_{i+1})\) and
\(Q=(\Id + A_{i-1})\cdots(\Id + A_1).\)

Note that when~\(P\) and~\(Q\) are non-singular, then the gradient can
only vanish at~\(E=0\). But that means we're at the global minimum of
the objective function~\(f\). We can ensure that~\(P\) and~\(Q\) are
non-singular by making the largest singular value of each~\(A_i\) to be
less than \(1/L\).

The property we find here is that the gradient vanishes only at the
optimum. This is implied by convexity, but it does not imply convexity.
Indeed, the objective function above is not convex. However, this weaker
property is enough to ensure that gradient-based methods do not get
stuck except at the optimum. In particular, the objective has no saddle
points. This desirable property does not hold for the standard
parameterization~\(A=A_L\cdots A_1\) and so it speaks to the benefit of
the residual parameterization.

\hypertarget{normalization}{%
\subsection{Normalization}\label{normalization}}

Consider a feature vector~\(x\). We can partition this vector into
subsets so that \[
x = \begin{bmatrix} x_1 & \cdots & x_P \end{bmatrix}\,,
\] and \emph{normalize} each subset to yield a vector with partitions \[
    x'_i = 1/s_i (x_i - \mu_i)
\] where~\(\mu_i\) is the mean of the components of~\(x_i\) and~\(s_i\)
is the standard deviation.

Such normalization schemes have proven powerful for accelerating the
convergence of stochastic gradient descent. In particular, it is clear
why such an operation can improve the conditioning of the Jacobian.
Consider the simple linear case where~\(\hat{y} = Xw\). Then
\(\jac (w) = X\). If each row of~\(X\) has a large mean, i.e.,
\(X \approx X_0 + c11^T\), then,~\(X\) may be ill conditioned, as the
rank one term will dominate first singular value, and the remaining
singular values may be small. Removing the mean improves the condition
number. Rescaling by the variance may improve the condition number, but
also has the benefit of avoiding numerical scaling issues, forcing each
layer in a neural network to be on the same scale.

Such whole vector operations are expensive. Normalization in deep
learning chooses parts of the vector to normalize that can be computed
quickly. Batch Normalization normalizes along the data-dimension in
batches of data used to as stochastic gradient
minibatches\citep{ioffe2015batch}. Group Normalization generalizes this
notion to arbitrary partitioning of the data, encompassing a variety of
normalization proposals\citep{wu2018group}. The best normalization
scheme for a particular task is problem dependent, but there is a great
deal of flexibility in how one partitions features for normalization.

\hypertarget{generalization-in-deep-learning}{%
\section{Generalization in deep
learning}\label{generalization-in-deep-learning}}

While our understanding of optimization of deep neural networks has made
significant progress, our understanding of generalization is
considerably less settled. In the previous chapter, we highlighted four
paths towards generalization: stability, capacity, margin, optimization.
It's plausible deep neural networks have elements of all four of these
core components. The evidence is not as cut and dry as it is for linear
models, but some mathematical progress has been made to understand how
deep learning leverages classic foundations of generalization. In this
section we review the partial evidence gathered so far.

In the next chapter we will extend this discussion by taking a closer
look at the role that data and community practices play in the study of
generalization for deep learning.

\hypertarget{algorithmic-stability-of-deep-neural-networks}{%
\subsection{Algorithmic stability of deep neural
networks}\label{algorithmic-stability-of-deep-neural-networks}}

We discussed how stochastic gradient descent trained on convex models
was algorithmically stable. The results we saw in Chapter 6 extend to
some to the non-convex case. However, results that go through uniform
stability ultimately cannot explain the generalization performance of
training large deep models. The reason is that uniform stability does
not depend on the data generating distribution. In fact, uniform
stability is invariant under changes to the labeling of the data. But we
saw in Chapter 6 that we can make the generalization gap whatever we
want by randomizing the labels in the training set.

Nevertheless, it's possible that a weaker notion of stability that is
sensitive to the data would work. In fact, it does appear to be the case
in experiment that stochastic gradient descent is relatively stable in
concrete instances. To measure stability empirically we can look at the
Euclidean distance between the parameters of two identical models
trained on the datasets which differ only by a single example. If these
are close for many independent runs, then the algorithm appears to be
stable. Note that parameter distance is a \emph{stronger} notion than
stability of the loss.

The plot below displays the parameter distance for the venerable AlexNet
model trained on the ImageNet benchmark. We observe that the parameter
distance grows sub-linearly even though our theoretical analysis is
unable to prove this to be true.

\begin{figure}
\centering
\includegraphics[width=0.5\textwidth,height=\textheight]{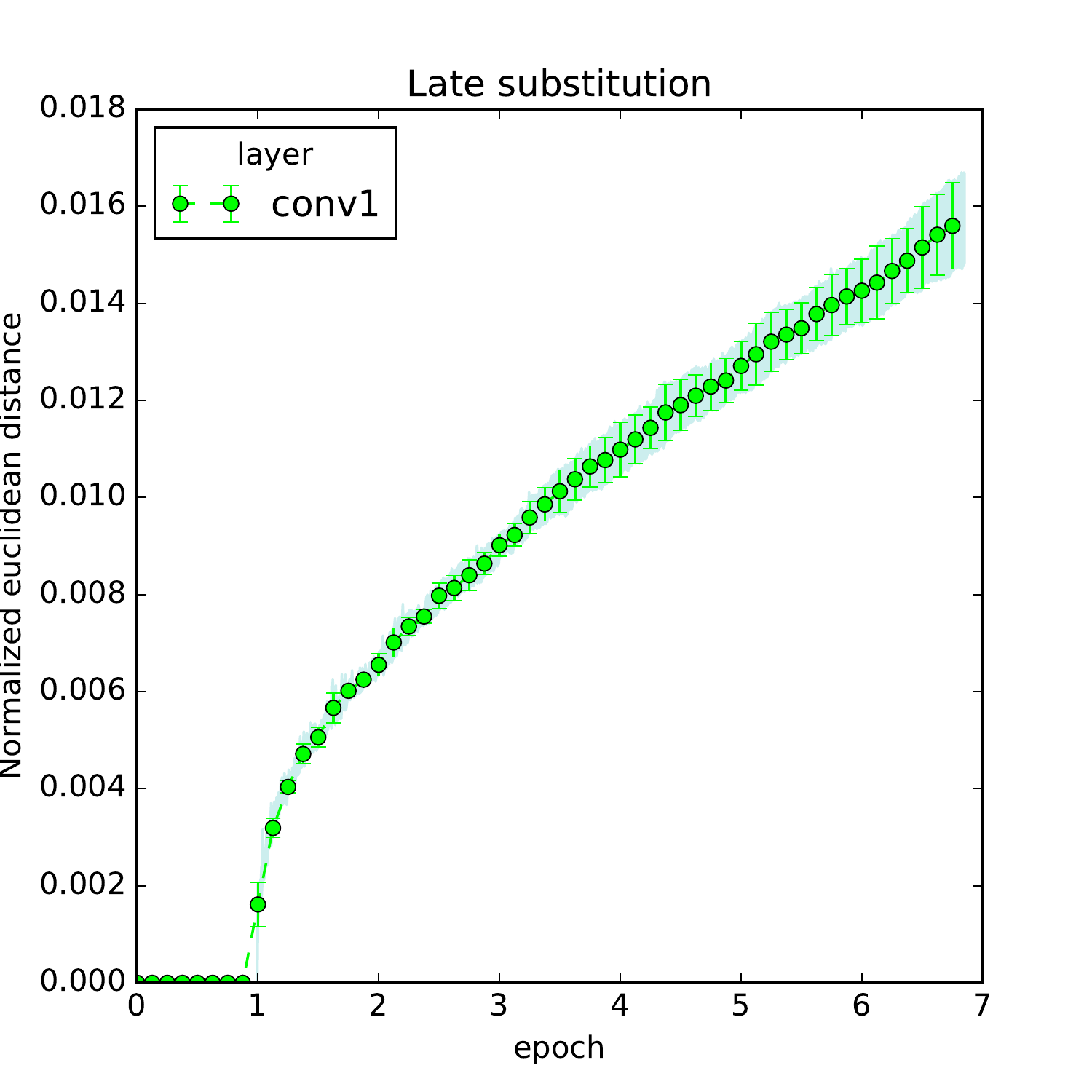}
\caption{Parameter divergence on AlexNet trained on ImageNet. The two
models differ only in a single example.}
\end{figure}

\hypertarget{capacity-of-deep-neural-networks}{%
\subsection{Capacity of deep neural
networks}\label{capacity-of-deep-neural-networks}}

Many researchers have attempted to compute notions like VC-dimension or
Rademacher complexity of neural networks. The earliest work by Baum and
Haussler bounded the VC-dimension of small neural networks and showed
that as long these networks could be optimized well in the
\emph{underparameterized} regime, then the neural networks would
generalize\citep{Baum88}. Later, seminal work by Bartlett showed that
the size of the weights was more important than the number of weights in
terms of understanding generalization capabilities of neural
networks\citep{bartlett1998sample}. These bounds were later sharpened
using Rademacher complexity arguments \citep{bartlett2002rademacher}.

\hypertarget{margin-bounds-for-deep-neural-networks}{%
\subsection{Margin bounds for deep neural
networks}\label{margin-bounds-for-deep-neural-networks}}

The margin theory for linear models conceptually extends to neural
networks. The definition of margin is unchanged. It simply quantifies
how close the network is to making an incorrect prediction. What changes
is that for multi-layer neural networks the choice of a suitable norm is
substantially more delicate.

To see why, a little bit of notation is necessary. We consider
multi-layer neural networks specified by a composition of~\(L\) layers.
Each layer is a linear transformation of the input, followed by a
coordinate-wise non-linear map:

\[
\mathrm{Input}\, x
\rightarrow Ax
\rightarrow \sigma (Ax)
\]

The linear transformation has trainable parameters, while the non-linear
map does not. For notational simplicity, we assume we have the same
non-linearity~\(\sigma\) at each layer scaled so that the map is
\(1\)-Lipschitz. For example, the popular coordinate-wise ReLU
\(\max\{x,0\}\) operation satisfies this assumption.

Given~\(L\) weight matrices~\(\mathcal{A} = (A_1,\ldots,A_L)\) let
\(f_\mathcal{A}\colon\R^d\to\R^k\) denote the function computed by the
corresponding network: \[
  f_\mathcal{A}(x) := A_L\sigma(A_{L-1} \cdots \sigma(A_1 x)\cdots))\,.
\] The network output~\(F_\mathcal{A}(x)\in\R^{k}\) is converted to a
class label in~\(\{1,\ldots,k\}\) by taking the~\(\arg\max\) over
components, with an arbitrary rule for breaking ties. We
assume~\(d\ge k\) only for notational convenience.

Our goal now is to define a complexity measure of the neural network
that will allow us to prove a margin bound. Recall that margins are
meaningless without a suitable normalization of the network. Convince
yourself that the Euclidean norm can no longer work for multi-layer ReLU
networks. After all we can scale the linear transformation on one later
by a constant~\(c\) and a subsequent layer by a constant~\(1/c\). Since
the ReLU non-linearity is piecewise linear, this transformation changes
the Euclidean norm of the weights arbitrarily without changing the
function that the network computes.

There's much ongoing work about what good norms are for deep neural
networks. We will introduce one possible choice here.

Let~\(\|\cdot\|_{\mathrm{op}}\) denote the spectral norm. Also, let
\(\|A\|_{2,1}=\left\| (\|{}A_{:,1}\|_2, \ldots, \|{}A_{:,m} \|_2)\right\|_1\)
the matrix norm where we apply the~\(\ell_2\)-norm to each column of the
matrix and then take the~\(\ell_1\)-norm of the resulting vector.

The \emph{spectral complexity} \(R_\mathcal{A}\) of a
network~\(F_\mathcal{A}\) with weights~\(\mathcal{A}\) is the defined as
\begin{equation}   R_{\mathcal{A}}   :=   \left(\prod_{i=1}^L\|A_i\|_{\mathrm{op}}\right)   \left(\sum_{i=1}^L \left(\frac{ \|A_i^{\top} - M_i^{\top}\|_{2,1}}{\|A_i\|_{\mathrm{op}}}\right)^{2/3}\right)^{3/2}\,.   \label{eq:spec_comp} \end{equation}
Here, the matrices~\(M_1,\dots, M_L\) are free parameters that we can
choose to minimize the bound. Random matrices tend to be good choices.

The following theorem provides a generalization bound for neural
networks with fixed nonlinearities and weight matrices~\(\mathcal{A}\)
of bounded spectral complexity~\(R_{\mathcal{A}}\).

\begin{Theorem}

Assume data~\((x_1,y_1),\dots,(x_n,y_n)\) are drawn i.i.d. from any
probability distribution over~\(\R^d\times\{1,\ldots,k\}.\) With
probability at least~\(1-\delta\), for every margin~\(\theta > 0\) and
every network~\(f_{\mathcal{A}} \colon \R^d \to \R^k,\) \[
    R[f_{\mathcal{A}}]- R_S^{\theta}[f_{\mathcal{A}}]
    \leq \widetilde{\mathcal{O}}
    \left( \frac {R_{\mathcal{A}}\sqrt{ \sum_i \|x_i\|_2^2 }\ln(d)}{\theta n} + \sqrt{\frac {\ln(1/\delta)}{n}} \right),
  \] where
\(R_S^{\theta}[f] \leq n^{-1} \sum_i \mathbf{1}\left[ f(x_i)_{y_i} \leq \theta + \max_{j\neq y_i} f(x_i)_j \right]\).

\end{Theorem}

The proof of the theorem involves Rademacher complexity and so-called
data-dependent covering arguments. Although it can be shown empirically
that the above complexity measure~\(R_{\mathcal{A}}\) is somewhat
correlated with generalization performance in some cases, there is no
reason to believe that it is the ``right'' complexity measure. The bound
has other undesirable properties, such as an exponential dependence on
the depth of the network as well as an implicit dependence on the size
of the network.

\hypertarget{implicit-regularization-by-stochastic-gradient-descent-in-deep-learning}{%
\subsection{Implicit regularization by stochastic gradient descent in
deep
learning}\label{implicit-regularization-by-stochastic-gradient-descent-in-deep-learning}}

There have been recent attempts to understand the dynamics of stochastic
gradient descent on deep neural networks. Using an argument similar to
the one we used to reason about convergence of the \emph{predictions} of
deep neural networks, Jacot et al.~used a differential equations
argument to understand to which \emph{function} stochastic gradient
converged \citep{jacot2018neural}.

Here we can sketch the rough details of their argument. Recall our
expression for the dynamics of the predictions in gradient descent: \[
    \hat{y}_{t+1}-y = (I -  \alpha \jac(w_t)^T \jac(w_t))(\hat{y}_t - y) + \alpha \epsilon_t\,.
\] If~\(\alpha\) is very small, we can further approximate this as \[
    \hat{y}_{t+1}-y = (I -  \alpha \jac(w_0)^T \jac(w_0))(\hat{y}_t - y) + \alpha \epsilon'_t\,.
\] Now~\(\epsilon'_t\) captures both the curvature of the function
representation and the deviation of the weights from their initial
condition. Note that in this case,~\(\jac(w_0)\) is a constant for all
time. The matrix~\(\jac(w_0)^T \jac(w_0)\) is~\(n \times n\), positive
semidefinite, and only depends on the data and the initial setting of
the weights. When the weights are random, this is a kernel induced by
random features. The expected value of this random feature embedding is
called a \emph{Neural Tangent Kernel}. \[
    k(x_1,x_2) = \E_{w_0} \left[\langle \nabla_w f(x_1; w_0), \nabla_w f(x_2;w_0)
    \rangle \right] \,.
\]

Using a limit where~\(\alpha\) tends to zero, Jacot et. al argue that a
deep neural net will find a minimum norm solution in the RKHS of the
Neural Tangent Kernel. This argument was made non-asymptotic by Heckel
and Soltanolkotabi\citep{heckel2020compressive}. In the generalization
chapter, we showed that the minimum norm solution of in RKHS generalized
with a rate of~\(O(1/n)\). Hence, this argument suggests a similar rate
of generalization may occur in deep learning, provided the norm of the
minimum norm solution does not grow to rapidly with the number of data
points and that the true function optimized by stochastic gradient
descent isn't too dissimilar from the Neural Tangent Kernel limit. We
note that this argument is qualitative, and there remains work to be
done to make these arguments fully rigorous.

Perhaps the most glaring issue with NTK arguments is that they do not
reflect practice. Models trained with Neural Tangent Kernels do not
match the predictive performance of the corresponding neural network.
Moreover, simpler kernels inspired by these networks can out perform
Neural Tangent Kernels\citep{Shankar20a}. There is a significant gap
between the theory and practice here, but the research is new and
remains active and this gap may be narrowed in the coming years.

\hypertarget{chapter-notes-6}{%
\section{Chapter notes}\label{chapter-notes-6}}

Deep learning is at this point a vast field with tens of thousands of
papers. We don't attempt to survey or systematize this vast literature.
It's worth emphasizing though the importance of learning about the area
by experimenting with available code on publicly available datasets. The
next chapter will cover datasets and benchmarks in detail.

Apart from the development of benchmark datasets, one of the most
important advances in deep learning over the last decade has been the
development of high quality, open source software. This software makes
it easier than ever to prototype deep neural network models. One such
open source project is PyTorch (\url{pytorch.org}), which we recommend
for the researcher interested in experimenting with deep neural
networks. The best way to begin to understand some of the nuances of
architecture selection is to find pre-existing code and understand how
it is composed. We recommend the tutorial by David Page which
demonstrates how the many different pieces fit together in a deep
learning pipeline.\citep{page2020how}

For a more in depth understanding of automatic differentiation, we
recommend Griewank and Walther's \emph{Evaluating
Derivatives}.\citep{griewank2008evaluating} This text works through a
variety of unexpected scenarios---such as implicitly defined
functions---where gradients can be computed algorithmically. Jax is open
source automatic differentiation package that incorporates many of these
techniques, and is useful for any application that could be assisted by
automatic differentiation.\citep{jax2018github}

The AlexNet architecture was introduced by Krizhevsky, Sutskever, and
Hinton in 2012.\citep{KrizhevskySuHi12} Achieving the best known
performance on the ImageNet benchmark at the time, it was pivotal in
kicking of the most recent wave of research on deep learning.

Residual networks were introduced by He, Zhang, Ren, and Sun in
2016.\citep{resnet} The observation about linear residual networks is
due to Hardt and Ma.\citep{hardt2017identity}

Theorem 8 is due to Bartlett, Foster, and
Telgarsky.\citep{bartlett2017spectrally} The dimension dependence of the
theorem can be removed.\citep{golowich2018size}

\chapter{Datasets}

It's become commonplace to point out that machine learning models are
only as good as the data they're trained on. The old slogan ``garbage
in, garbage out'' no doubt applies to machine learning practice, as does
the related catchphrase ``bias in, bias out''. Yet, these aphorisms
still understate---and somewhat misrepresent---the significance of data
for machine learning.

It's not only the output of a learning algorithm that may suffer with
poor input data. A dataset serves many other vital functions in the
machine learning ecosystem. The dataset itself is an integral part of
the problem formulation. It implicitly sorts out and operationalizes
what the problem is that practitioners end up solving. Datasets have
also shaped the course of entire scientific communities in their
capacity to measure and benchmark progress, support competitions, and
interface between researchers in academia and practitioners in industry.

If so much hinges on data in machine learning, it might come as a
surprise that there is no simple answer to the question of what makes
data good for what purpose. The collection of data for machine learning
applications has not followed any established theoretical framework,
certainly not one that was recognized a priori.

In this chapter, we take a closer look at popular datasets in the field
of machine learning and the benchmarks that they support. We trace out
the history of benchmarks and work out the implicit scientific
methodology behind machine learning benchmarks. We limit the scope of
this chapter in some important ways. Our focus will be largely on
publicly available datasets that support training and testing purposes
in machine learning research and applications. Primarily, we critically
examine the train-and-test paradigm machine learning practitioners take
for granted today.

\hypertarget{the-scientific-basis-of-machine-learning-benchmarks}{%
\section{The scientific basis of machine learning
benchmarks}\label{the-scientific-basis-of-machine-learning-benchmarks}}

Methodologically, much of modern machine learning practice rests on a
variant of \emph{trial and error}, which we call the \emph{train-test
paradigm}. Practitioners repeatedly build models using any number of
heuristics and test their performance to see what works. Anything goes
as far as training is concerned, subject only to computational
constraints, so long as the performance looks good in testing. Trial and
error is sound so long as the testing protocol is robust enough to
absorb the pressure placed on it. We will examine to which extent this
is the case in machine learning.

From a theoretical perspective, the best way to test the performance of
a predictor~\(f\) is to collect a sufficiently large fresh dataset~\(S\)
and to compute the empirical risk~\(R_S[f]\). We already learned that
the empirical risk in this case is an unbiased estimate of the risk of
the predictor. For a bounded loss function and a test set of size~\(n\),
an appeal to Hoeffding's inequality proves the generalization gap to be
no worse than~\(O(1/\sqrt{n})\). We can go a step further and observe
that if we take union bound over~\(k\) fixed predictors, our fresh
sample will simultaneously provide good estimates for all~\(k\)
predictors up to a maximum error of~\(O(\sqrt{\log(k)/n})\). In fact, we
can apply any of the mathematical tools we saw in the Generalization
chapter so long as the sample~\(S\) really is a fresh sample with
respect to the set of models we want to evaluate.

Data collection, however, is a difficult and costly task. In most
applications, practitioners cannot sample fresh data for each model they
would like to try out. A different practice has therefore become the de
facto standard. Practitioners split their dataset into typically two
parts, a \emph{training set} used for training a model, and a \emph{test
set} used for evaluating its performance. Sometimes practitioners divide
their data into multiple splits, e.g., training, validation, and test
sets. However, for our discussion here that won't be necessary. Often
the split is determined when the dataset is created. Datasets used for
benchmarks in particular have one fixed split persistent throughout
time. A number of variations on this theme go under the name
\emph{holdout method}.\index{holdout method}

Machine learning competitions have adopted the same format. The company
Kaggle, for example, has organized hundreds of competitions since it was
founded. In a competition, a holdout set is kept secret and is used to
rank participants on a public leaderboard as the competition unfolds. In
the end, the final winner is whoever scores highest on a separate secret
test set not used to that point.\index{Kaggle}\index{competition}

In all applications of the holdout method the hope is that the test set
will serve as a fresh sample that provides good risk estimates for all
the models. The central problem is that practitioners don't just use the
test data once only to retire it immediately thereafter. The test data
are used incrementally for building one model at a time while
incorporating feedback received previously from the test data. This
leads to the fear that eventually models begin to \emph{overfit} to the
test data.\index{overfit}

Duda and Hart summarize the problem aptly in their 1973 textbook:

\begin{quote}
In the early work on pattern recognition, when experiments were often
done with very small numbers of samples, the same data were often used
for designing and testing the classifier. This mistake is frequently
referred to as ``testing on the training data.'' A related but less
obvious problem arises when a classifier undergoes a long series of
refinements guided by the results of repeated testing on the same data.
This form of ``training on the testing data'' often escapes attention
until new test samples are obtained.\citep{duda1973pattern}
\end{quote}

Nearly half a century later, Hastie, Tibshirani, and Friedman still
caution in the 2017 edition of their influential textbook:

\begin{quote}
Ideally, the test set should be kept in a ``vault,'' and be brought out
only at the end of the data analysis. Suppose instead that we use the
test-set repeatedly, choosing the model with smallest test-set error.
Then the test set error of the final chosen model will underestimate the
true test error, sometimes substantially.\citep{hastie2017elements}
\end{quote}

Indeed, reuse of test data---on the face of it---invalidates the
statistical guarantees of the holdout method. The predictors created
with knowledge about prior test-set evaluations are no longer
independent of the test data. In other words, the sample isn't fresh
anymore. While the suggestion to keep the test data in a ``vault'' is
safe, it couldn't be further from the reality of modern practice.
Popular test datasets often see tens of thousands of evaluations.

We could try to salvage the situation by relying on uniform convergence.
If all models we try out have sufficiently small complexity in some
formal sense, such as VC-dimension, we could use the tools from the
Generalization chapter to negotiate some sort of a bound. However, the
whole point of the train-test paradigm is not to constrain the
complexity of the models a priori, but rather to let the practitioner
experiment freely. Moreover, if we had an actionable theoretical
generalization guarantee to begin with, there would hardly be any need
for the holdout method whose purpose is to provide an empirical estimate
where theoretical guarantees are lacking.

Before we discuss the ``training on the testing data'' problem any
further, it's helpful to get a better sense of concrete machine learning
benchmarks, their histories, and their impact within the community.

\hypertarget{a-tour-of-datasets-in-different-domains}{%
\section{A tour of datasets in different
domains}\label{a-tour-of-datasets-in-different-domains}}

The creation of datasets in machine learning does not follow a clear
theoretical framework. Datasets aren't collected to test a specific
scientific hypothesis. In fact, we will see that there are many
different roles data plays in machine learning. As a result, it makes
sense to start by looking at a few influential datasets from different
domains to get a better feeling for what they are, what motivated their
creation, how they organized communities, and what impact they had.

\hypertarget{timit}{%
\subsection{TIMIT}\label{timit}}

Automatic speech recognition is a machine learning problem of
significant commercial interest. Its roots date back to the early 20th
century.\citep{li2019vocal}

Interestingly, speech recognition also features one of the oldest
benchmarks datasets, the TIMIT (Texas Instruments/Massachusetts
Institute for Technology) data. The creation of the dataset was funded
through a 1986 DARPA program on speech recognition. In the mid-eighties,
artificial intelligence was in the middle of a ``funding winter'' where
many governmental and industrial agencies were hesitant to sponsor AI
research because it often promised more than it could deliver. DARPA
program manager Charles Wayne proposed a way around this problem was
establishing more rigorous evaluation methods. Wayne enlisted the
National Institute of Standards and Technology to create and curate
shared datasets for speech, and he graded success in his program based
on performance on recognition tasks on these datasets.

Many now credit Wayne's program with kick starting a revolution of
progress in speech
recognition.\citep{liberman10obituary, Church18, Liberman20} According
to Church,

\begin{quote}
It enabled funding to start because the project was
glamour-and-deceit-proof, and to continue because funders could measure
progress over time. Wayne's idea makes it easy to produce plots which
help sell the research program to potential sponsors. A less obvious
benefit of Wayne's idea is that it enabled hill climbing. Researchers
who had initially objected to being tested twice a year began to
evaluate themselves every hour.\citep{Church18}
\end{quote}

A first prototype of the TIMIT dataset was released in December of 1988
on a CD-ROM. An improved release followed in October 1990. TIMIT already
featured the training/test split typical for modern machine learning
benchmarks. There's a fair bit we know about the creation of the data
due to its thorough documentation.\citep{garofolo1993darpa}

TIMIT features a total of about 5 hours of speech, composed of 6300
utterances, specifically, 10 sentences spoken by each of 630 speakers.
The sentences were drawn from a corpus of 2342 sentences such as the
following.\index{TIMIT}

\begin{verbatim}
She had your dark suit in greasy wash water all year. (sa1)
Don't ask me to carry an oily rag like that. (sa2)
This was easy for us. (sx3)
Jane may earn more money by working hard. (sx4)
She is thinner than I am. (sx5)
Bright sunshine shimmers on the ocean. (sx6)
Nothing is as offensive as innocence. (sx7)
\end{verbatim}

The TIMIT documentation distinguishes between 8 major dialect regions in
the United States:

\begin{quote}
New England, Northern, North Midland, South Midland, Southern, New York
City, Western, Army Brat (moved around)
\end{quote}

Of the speakers, 70\% are male and 30\% are female. All native speakers
of American English, the subjects were primarily employees of Texas
Instruments at the time. Many of them were new to the Dallas area where
they worked.

Racial information was supplied with the distribution of the data and
coded as ``White'', ``Black'', ``American Indian'',
``Spanish-American'', ``Oriental'', and ``Unknown''. Of the 630
speakers, 578 were identified as White, 26 as Black, 2 as American
Indian, 2 as Spanish-American, 3 as Oriental, and 17 as unknown.

\begin{longtable}[]{@{}llll@{}}
\caption{Demographic information about the TIMIT
speakers}\tabularnewline
\toprule
& Male & Female & Total (\%) \\
\midrule
\endfirsthead
\toprule
& Male & Female & Total (\%) \\
\midrule
\endhead
White & 402 & 176 & 578 (91.7\%) \\
Black & 15 & 11 & 26 (4.1\%) \\
American Indian & 2 & 0 & 2 (0.3\%) \\
Spanish-American & 2 & 0 & 2 (0.3\%) \\
Oriental & 3 & 0 & 3 (0.5\%) \\
Unknown & 12 & 5 & 17 (2.6\%) \\
\bottomrule
\end{longtable}

The documentation notes:

\begin{quote}
In addition to these 630 speakers, a small number of speakers with
foreign accents or other extreme speech and/or hearing abnormalities
were recorded as ``auxiliary'' subjects, but they are not included on
the CD-ROM.
\end{quote}

It comes to no surprise that early speech recognition models had
significant demographic and racial biases in their performance.

Today, several major companies, including Amazon, Apple, Google, and
Microsoft, all use speech recognition models in a variety of products
from cell phone apps to voice assistants. Today, speech recognition
lacks a major open benchmark that would support the training models
competitive with the industrial counterparts. Industrial speech
recognition pipelines are often complex systems that use proprietary
data sources that not a lot is known about. Nevertheless, even today's
speech recognition systems continue to have racial
biases.\citep{koenecke2020racial}

\hypertarget{uci-machine-learning-repository}{%
\subsection{UCI Machine Learning
Repository}\label{uci-machine-learning-repository}}

The UCI Machine Learning Repository currently hosts more than 500
datasets, mostly for different classification and regression tasks. Most
datasets are relatively small, many of them structured tabular datasets
with few attributes.\index{UCI}

The UCI Machine Learning Repository contributed to the adoption of the
train-test paradigm in machine learning in the late 1980s. Langley
recalls:

\begin{quote}
The experimental movement was aided by another development. David Aha,
then a PhD student at UCI, began to collect datasets for use in
empirical studies of machine learning. This grew into the UCI Machine
Learning Repository (\url{http://archive.ics.uci.edu/ml/}), which he
made available to the community by FTP in 1987. This was rapidly adopted
by many researchers because it was easy to use and because it let them
compare their results to previous findings on the same
tasks.\citep{langley2011changing}
\end{quote}

Aha's PhD work involved evaluating nearest-neighbor methods, and he
wanted to be able to compare the utility of his algorithms to decision
tree induction algorithms, popularized by Ross Quinlan. Aha describes
his motivation for building the UCI repository as follows.

\begin{quote}
I was determined to create and share it, both because I wanted to use
the datasets for my own research and because I thought it was ridiculous
that the community hadn't fielded what should have been a useful
service. I chose to use the simple attribute-value representation that
Ross Quinlan was using so successfully for distribution with his TDIDT
implementations.\citep{aha2020personal}
\end{quote}

The UCI dataset was wildly successful, and partially responsible for the
renewed interest in pattern recognition methods in machine learning.
However, this success came with some detractors. By the mid 1990s, many
were worried that evaluation-by-benchmark encouraged chasing
state-of-the-art results and writing incremental papers. Aha reflects:

\begin{quote}
By ICML-95, the problems ``caused'' by the repository had become
popularly espoused. For example, at that conference Lorenza Saitta had,
in an invited workshop that I co-organized, passionately decried how it
allowed researchers to publish dull papers that proposed small
variations of existing supervised learning algorithms and reported their
small-but-significant incremental performance improvements in comparison
studies.
\end{quote}

Nonetheless, the UCI repository remains one of the most popular source
for benchmark datasets in machine learning, and many of the early
datasets still are used for benchmarking in machine learning research.
The most popular dataset in the UCI repository is Ronald A. Fisher's
Iris Data Set that Fisher collected for his 1936 paper on ``The use of
multiple measurements in taxonomic problems''.

As of writing, the second most popular dataset in the UCI repository is
the \emph{Adult} dataset. Extracted from the 1994 Census database, the
dataset features nearly 50,000 instances describing individual in the
United States, each having 14 attributes. The goal is to classify
whether an individual earns more than 50,000 US dollars or less.

The Adult dataset became popular in the algorithmic fairness community,
largely because it is one of the few publicly available datasets that
features demographic information including \emph{gender} (coded in
binary as male/female), as well as \emph{race} (coded as
Amer-Indian-Eskimo, Asian-Pac-Islander, Black, Other, and White).

Unfortunately, the data has some idiosyncrasies that make it less than
ideal for understanding biases in machine learning
models.\citep{ding2021retiring} Due to the age of the data, and the
income cutoff at \$50,000, almost all instances labeled \emph{Black} are
below the cutoff, as are almost all instances labeled \emph{female}.
Indeed, a standard logistic regression model trained on the data
achieves about 85\% accuracy overall, while the same model achieves 91\%
accuracy on Black instances, and nearly 93\% accuracy on female
instances. Likewise, the ROC curves for the latter two groups enclose
actually more area than the ROC curve for male instances. This is a
rather untypical situation since often machine learning models perform
more poorly on historically disadvantaged groups.

\hypertarget{highleymans-data}{%
\subsection{Highleyman's data}\label{highleymans-data}}

The first machine learning benchmark dates back to the late 1950s. Few
used it and even fewer still remembered it by the time benchmarks became
widely used in machine learning in the late 1980s.

In 1959 at Bell Labs, Bill Highleyman and Louis Kamenstky designed a
scanner to evaluate character recognition
techniques.\citep{highleyman1959generalized} Their goal was ``to
facilitate a systematic study of character-recognition techniques and an
evaluation of methods prior to actual machine development.'' It was not
clear at the time which part of the computations should be done in
special purpose hardware and which parts should be done with more
general computers. Highleyman later patented an optical character
recognition (OCR) scheme that we recognize today as a convolutional
neural network with convolutions optically computed as part of the
scanning.\citep{highleyman1961character}

Highleyman and Kamentsky used their scanner to create a data set of 1800
alphanumeric characters. They gathered the 26 capital letters of the
English alphabet and 10 digits from 50 different writers. Each character
in their corpus was scanned in binary at a resolution of 12 x 12 and
stored on punch cards that were compatible with the IBM 704, the first
mass-produced computer with floating-point arithmetic hardware.

\begin{figure}
\centering
\includegraphics{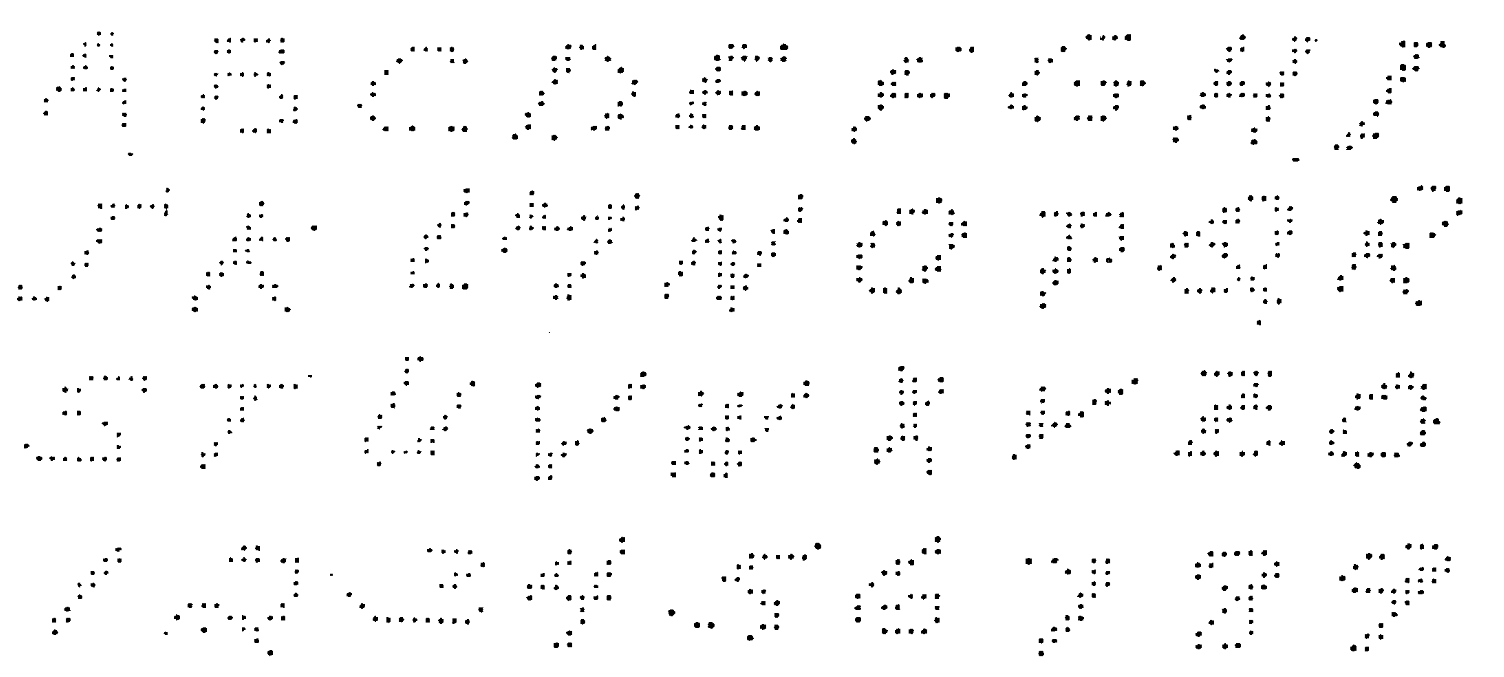}
\caption{A look at Highleyman's data}
\end{figure}

With the data in hand, Highleyman and Kamenstky began studying various
proposed techniques for recognition. In particular, they analyzed a
method of Woody Bledsoe's and published an analysis claiming to be
unable to reproduce the results.\citep{highleyman1960comments} Bledsoe
found the numbers to be considerably lower than he expected, and asked
Highleyman to send him the data. Highleyman obliged, mailing the package
of punch cards across the country to Sandia Labs. Upon receiving the
data, Bledsoe conducted a new experiment. In what may be the first
application of the train-test split, he divided the characters up, using
40 writers for training and 10 for testing. By tuning the
hyperparameters, Bledsoe was able to achieve approximately 60\%
error.\citep{bledsoe1961further} Bledsoe also suggested that the high
error rates were to be expected as Highleyman's data was too small.
Prophetically, he declared that 1000 alphabets might be needed for good
performance.

By this point, Highleyman had also shared his data with Chao Kong Chow
at the Burroughs Corporation (a precursor to Unisys). A pioneer of using
decision theory for pattern recognition,\citep{chow1957optimum} Chow
built a pattern recognition system for characters. Using the same
train-test split as Bledsoe, Chow obtained an error rate of
41.7\%.\citep{chow1962recognition}

Highleyman made at least six additional copies of the data he had sent
to Bledsoe and Chow, and many researchers remained interested. He thus
decided to publicly offer to send a copy to anyone willing to pay for
the duplication and shipping fees.\citep{highleyman1963data} Of course,
the dataset was sent by US Postal Service. Electronic transfer didn't
exist at the time, resulting in sluggish data transfer rates on the
order of a few bits per minute.

Highleyman not only created the first machine learning benchmark. He
authored the the first formal study of train-test
splits\citep{highleyman1962design} and proposed empirical risk
minimization for pattern classification\citep{highleyman1962linear} as
part of his 1961 dissertation. By 1963, however, Highleyman had left his
research position at Bell Labs and abandoned pattern recognition
research.

We don't know how many people requested Highleyman's data, but the total
number of copies may have been less than twenty. Based on citation
surveys, we determined there were at least six additional copies made
after Highleyman's public offer for duplication, sent to researchers at
UW Madison, CMU, Honeywell, SUNY Stony Brook, Imperial College in
London, and Stanford Research Institute (SRI).

The SRI team of John Munson, Richard Duda, and Peter Hart performed some
of the most extensive experiments with Highleyman's
data.\citep{munson1968experiments} A 1-nearest-neighbors baseline
achieved an error rate of 47.5\%. With a more sophisticated approach,
they were able to do significantly better. They used a multi-class,
piecewise linear model, trained using Kesler's multi-class version of
the perceptron algorithm. Their feature vectors were 84 simple pooled
edge detectors in different regions of the image at different
orientations. With these features, they were able to get a test error of
31.7\%, 10 points better than Chow. When restricted only to digits, this
method recorded 12\% error. The authors concluded that they needed more
data, and that the error rates were ``still far too high to be
practical.'' They concluded that ``larger and higher-quality datasets
are needed for work aimed at achieving useful results.'' They suggested
that such datasets ``may contain hundreds, or even thousands, of samples
in each class.''

Munson, Duda, and Hart also performed informal experiments with humans
to gauge the readability of Highleyman's characters. On the full set of
alphanumeric characters, they found an average error rate of 15.7\%,
about 2x better than their pattern recognition machine. But this rate
was still quite high and suggested the data needed to be of higher
quality. They concluded that ``an array size of at least 20X20 is
needed, with an optimum size of perhaps 30X30.''

Decades passed until such a dataset, the MNIST digit recognition task,
was created and made widely available.

\hypertarget{mnist}{%
\subsection{MNIST}\label{mnist}}

The MNIST dataset contains images of handwritten digits. Its most common
version has 60,000 training images and 10,000 test images, each having
28x28 grayscale pixels.\index{MNIST}

\begin{figure}
\centering
\includegraphics{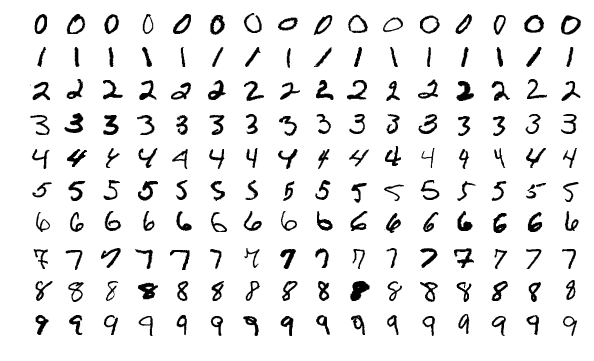}
\caption{A sample of MNIST digits}
\end{figure}

MNIST was created by researchers Burges, Cortes, and LeCun from data by
the National Institute of Standards and Technology (NIST). The dataset
was introduced in a research paper in 1998 to showcase the use of
gradient-based deep learning methods for document recognition
tasks.\citep{lecun1998gradient} However, the authors released the
dataset to provide a convenient benchmark of image data, in contrast to
UCI's predominantly tabular data. The MNIST website states

\begin{quote}
It is a good database for people who want to try learning techniques and
pattern recognition methods on real-world data while spending minimal
efforts on preprocessing and formatting.\citep{mnisturl}
\end{quote}

MNIST became a highly influential benchmark in the machine learning
community. Two decades and over 30,000 citation later, researchers
continue to use the data actively.

The original NIST data had the property that training and test data came
from two different populations. The former featured the handwriting of
two thousand American Census Bureau employees, whereas the latter came
from five hundred American high school students.\citep{grother1995nist}
The creators of MNIST reshuffled these two data sources and split them
into training and test set. Moreover, they scaled and centered the
digits. The exact procedure to derive MNIST from NIST was lost, but
recently reconstructed by matching images from both data
sources.\citep{yadav2019cold}\index{NIST}

The original MNIST test set was of the same size as the training set,
but the smaller test set became standard in research use. The 50,000
digits in the original test set that didn't make it into the smaller
test set were later identified and dubbed \emph{the lost
digits}.\citep{yadav2019cold}

From the beginning, MNIST was intended to be a benchmark used to compare
the strengths of different methods. For several years, LeCun maintained
an informal leaderboard on a personal website that listed the best
accuracy numbers that different learning algorithms achieved on
MNIST.\index{leaderboard}

\begin{longtable}[]{@{}ll@{}}
\caption{A snapshot of the original MNIST leaderboard from February 2,
1999. Source: Internet Archive (Retrieved: December 4,
2020)}\tabularnewline
\toprule
Method & Test error (\%) \\
\midrule
\endfirsthead
\toprule
Method & Test error (\%) \\
\midrule
\endhead
linear classifier (1-layer NN) & 12.0 \\
linear classifier (1-layer NN) {[}deskewing{]} & 8.4 \\
pairwise linear classifier & 7.6 \\
K-nearest-neighbors, Euclidean & 5.0 \\
K-nearest-neighbors, Euclidean, deskewed & 2.4 \\
40 PCA + quadratic classifier & 3.3 \\
1000 RBF + linear classifier & 3.6 \\
K-NN, Tangent Distance, 16x16 & 1.1 \\
SVM deg 4 polynomial & 1.1 \\
Reduced Set SVM deg 5 polynomial & 1.0 \\
Virtual SVM deg 9 poly {[}distortions{]} & 0.8 \\
2-layer NN, 300 hidden units & 4.7 \\
2-layer NN, 300 HU, {[}distortions{]} & 3.6 \\
2-layer NN, 300 HU, {[}deskewing{]} & 1.6 \\
2-layer NN, 1000 hidden units & 4.5 \\
2-layer NN, 1000 HU, {[}distortions{]} & 3.8 \\
3-layer NN, 300+100 hidden units & 3.05 \\
3-layer NN, 300+100 HU {[}distortions{]} & 2.5 \\
3-layer NN, 500+150 hidden units & 2.95 \\
3-layer NN, 500+150 HU {[}distortions{]} & 2.45 \\
LeNet-1 {[}with 16x16 input{]} & 1.7 \\
LeNet-4 & 1.1 \\
LeNet-4 with K-NN instead of last layer & 1.1 \\
LeNet-4 with local learning instead of ll & 1.1 \\
LeNet-5, {[}no distortions{]} & 0.95 \\
LeNet-5, {[}huge distortions{]} & 0.85 \\
LeNet-5, {[}distortions{]} & 0.8 \\
Boosted LeNet-4, {[}distortions{]} & 0.7 \\
\bottomrule
\end{longtable}

In its capacity as a benchmark, it became a showcase for the emerging
kernel methods of the early 2000s that temporarily achieved top
performance on MNIST.\citep{decoste2002training} Today, it is not
difficult to achieve less than 0.5\% classification error with a wide
range of convolutional neural network architectures. The best models
classify all but a few pathological test instances correctly. As a
result, MNIST is widely considered \emph{too easy} for today's research
tasks.

MNIST wasn't the first dataset of handwritten digits in use for machine
learning research. Earlier, the US Postal Service (USPS) had released a
dataset of 9298 images (7291 for training, and 2007 for testing). The
USPS data was actually a fair bit harder to classify than MNIST. A
non-negligible fraction of the USPS digits look unrecognizable to
humans,\citep{bromley1991neural} whereas humans recognize essentially
all digits in MNIST.

\hypertarget{imagenet}{%
\subsection{ImageNet}\label{imagenet}}

ImageNet is a large repository of labeled images that has been highly
influential in computer vision research over the last decade. The image
labels correspond to nouns from the WordNet lexical database of the
English language. WordNet groups nouns into cognitive synonyms, called
\emph{synsets}. The words \emph{car} and \emph{automobile}, for example,
would fall into the same synset. On top of these categories, WordNet
provides a hierarchical structure according to a super-subordinate
relationship between synsets. The synset for \emph{chair}, for example,
is a child of the synset for \emph{furniture} in the wordnet hierarchy.
WordNet existed before ImageNet and in part inspired the creation of
ImageNet.\index{ImageNet}\index{WordNet}

The initial release of ImageNet included about 5000 image categories
each corresponding to a synset in WordNet. These ImageNet categories
averaged about 600 images per category.\citep{deng2009imagenet} ImageNet
grew over time and its Fall 2011 release had reached about 32,000
categories.

The construction of ImageNet required two essential steps:

\begin{enumerate}
\def\labelenumi{\arabic{enumi}.}
\tightlist
\item
  The first was the retrieval of candidate images for each synset.
\item
  The second step in the creation process was the labeling of images.
\end{enumerate}

Scale was an important consideration due to the target size of the image
repository.

This first step utilized available image databases with a search
interface, specifically, Flickr. Candidate images were taken from the
image search results associated with the synset nouns for each category.

For the second labeling step, the creators of ImageNet turned to
Amazon's Mechanical Turk platform (MTurk). MTurk is an online labor
market that allows individuals and corporations to hire on-demand
workers to perform simple tasks. In this case, MTurk workers were
presented with candidate images and had to decide whether or not the
candidate image was indeed an image corresponding to the category that
it was putatively associated with.\index{MTurk}

It is important to distinguish between this ImageNet database and a
popular machine learning benchmark and competition, called ImageNet
Large Scale Visual Recognition Challenge (ILSVRC), that was derived from
it.\citep{russakovsky2015imagenet} The competition was organized yearly
from 2010 until 2017 to ``measure the progress of computer vision for
large scale image indexing for retrieval and annotation.\index{ILSVRC}''
In 2012, ILSVRC reached significant notoriety in both industry and
academia when a deep neural network trained by Krizhevsky, Sutskever,
and Hinton outperformed all other models by a significant
margin.\citep{KrizhevskySuHi12} This result---yet again an evaluation in
a train-test paradigm---helped usher in the latest era of exuberant
interest in machine learning and neural network models under the
rebranding as \emph{deep learning}.\citep{MalikCACM}

When machine learning practitioners say ``ImageNet'' they typically
refer to the data used for the image classification task in the 2012
ILSVRC benchmark. The competition included other tasks, such as object
recognition, but image classification has become the most popular task
for the dataset. Expressions such as ``a model trained on ImageNet''
typically refer to training an image classification model on the
benchmark dataset from 2012.

Another common practice involving the ILSVRC data is
\emph{pre-training}. Often a practitioner has a specific classification
problem in mind whose label set differs from the 1000 classes present in
the data. It's possible nonetheless to use the data to create useful
features that can then be used in the target classification problem.
Where ILSVRC enters real-world applications it's often to support
pre-training.\index{pre-training}

This colloquial use of the word ImageNet can lead to some confusion, not
least because the ILSVRC-2012 dataset differs significantly from the
broader database. It only includes a subset of 1000 categories.
Moreover, these categories are a rather skewed subset of the broader
ImageNet hierarchy. For example, of these 1000 categories only three are
in the \emph{person} branch of the WordNet hierarchy, specifically,
\emph{groom}, \emph{baseball player}, and \emph{scuba diver}. Yet, more
than 100 of the 1000 categories correspond to different dog breeds. The
number is 118, to be exact, not counting wolves, foxes, and wild dogs
that are also present among the 1000 categories.

What motivated the exact choice of these 1000 categories is not entirely
clear. The apparent canine inclination, however, isn't just a quirk
either. At the time, there was an interest in the computer vision
community in making progress on prediction with many classes, some of
which are very similar. This reflects a broader pattern in the machine
learning community. The creation of datasets is often driven by an
intuitive sense of what the technical challenges are for the field. In
the case of ImageNet, \emph{scale}, both in terms of the number of data
points as well as the number of classes, was an important motivation.

The large scale annotation and labeling of datasets, such as we saw in
the case of ImageNet, fall into a category of labor that anthropologist
Gray and computer scientist Suri coined \emph{Ghost Work} in their book
of the same name.\citep{gray2019ghost} They point out:

\begin{quote}
MTurk workers are the AI revolution's unsung heroes.
\end{quote}

Indeed, ImageNet was labeled by about 49,000 MTurk workers from 167
countries over the course of multiple years.

\hypertarget{longevity-of-benchmarks}{%
\section{Longevity of benchmarks}\label{longevity-of-benchmarks}}

The holdout method is central to the scientific and industrial
activities of the machine learning community. Thousands of research
papers have been written that report numbers on popular benchmark data,
such as MNIST, CIFAR-10, or ImageNet. Often extensive tuning and
hyperparameter optimization went into each such research project to
arrive at the final accuracy numbers reported in the paper.

Does this extensive reuse of test sets not amount to what Duda and Hart
call the ``training on the testing data'' problem? If so, how much of
the progress that has been made is real, and how much amounts of
overfitting to the test data?

To answer these questions we will develop some more theory that will
help us interpret the outcome of empirical meta studies into the
longevity of machine learning benchmarks.

\hypertarget{the-problem-of-adaptivity}{%
\subsection{The problem of adaptivity}\label{the-problem-of-adaptivity}}

Model building is an iterative process where the performance of a model
informs subsequent design choices. This iterative process creates a
closed feedback loop between the practitioner and the test set. In
particular, the models the practitioner chooses are not independent of
the test set, but rather
\emph{adaptive}.\index{adaptive}\index{adaptivity}

Adaptivity can be interpreted as a form of overparameterization. In an
adaptively chosen sequence of predictors~\(f_1,\dots,f_k,\) the~\(k\)-th
predictor had the ability to incorporate at least~\(k-1\) bits of
information about the performance of previously chosen predictors. This
suggests that as~\(k\ge n,\) the statistical guarantees of the holdout
method become vacuous. This intuition is formally correct, as we will
see.

To reason about adaptivity, it is helpful to frame the problem as an
interaction between two parties. One party holds the dataset~\(S\).
Think of this party as implementing the holdout method. The other party,
we call \emph{analyst}, can \emph{query} the dataset by requesting the
empirical risk \(R_S[f]\) of a given predictor~\(f\) on the
dataset~\(S\). The parties interact for some number~\(k\) of rounds,
thus creating a sequence of adaptively chosen
predictors~\(f_1,\dots,f_k.\) Keep in mind that this sequence depends on
the dataset! In particular, when~\(S\) is drawn at
random,~\(f_2,\dots, f_k\) become random variables, too, that are in
general not independent of each
other.\index{analyst}\index{adaptive!analyst}

\begin{figure}
\centering
\includegraphics[width=1\textwidth,height=\textheight]{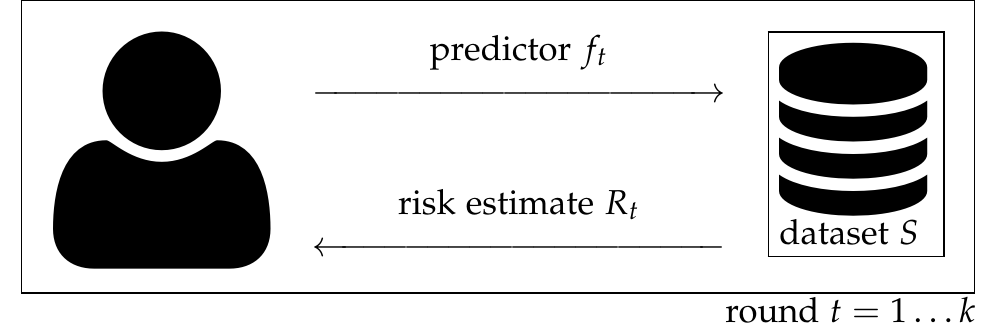}
\caption{The adaptive analyst model.}
\end{figure}

In general, the estimate~\(R_t\) returned by the holdout mechanism at
step \(t\) need not be equal to~\(R_S[f_t]\). We can often do better
than the standard holdout mechanism by limiting the information revealed
by each response. Throughout this chapter, we restrict our attention to
the case of the zero-one loss and binary prediction, although the theory
extends to other settings.

As it turns out, the guarantee of the standard holdout mechanism in the
adaptive case is exponentially worse in~\(k\) compared with the
non-adaptive case. Indeed, there is a fairly natural sequence of~\(k\)
adaptively chosen predictors, resembling the practice of ensembling, on
which the empirical risk is off by at least~\(\Omega(\sqrt{k/n}).\) This
is a lower bound on the gap between risk and empirical risk in the
adaptive setting. Contrast this with the~\(O(\sqrt{\log(k)/n})\) upper
bound that we observed for the standard holdout mechanism in the
non-adaptive case. We present the idea for the zero-one loss in a binary
prediction problem.\index{ensembling}

\begin{Algorithm}

\textbf{Overfitting by ensembling:}

\begin{enumerate}
\def\labelenumi{\arabic{enumi}.}
\tightlist
\item
  Choose \(k\) of random binary predictors \(f_1,\dots, f_k.\)
\item
  Compute the set \(I=\{i\in[k]\colon R_S[f_i] < 1/2\}.\)
\item
  Output the predictor \(f=\mathrm{majority}\{ f_i \colon i\in I \}\)
  that takes a majority vote over all the predictors computed in the
  second step.
\end{enumerate}

\end{Algorithm}

The key idea of the algorithm is to select all the predictors that have
accuracy strictly better than random guessing. This selection step
creates a bias that gives each selected predictor an advantage over
random guessing. The majority vote in the third step amplifies this
initial advantage into a larger advantage that grows with~\(k\). The
next proposition confirms that indeed this strategy finds a predictor
whose empirical risk is bounded away from~\(1/2\) (random guessing) by a
margin of~\(\Omega(\sqrt{k/n}).\) Since the predictor does nothing but
taking a majority vote over random functions, its risk is of course no
better than~\(1/2\).

\begin{Proposition}

For sufficiently large~\(k\le n\), overfitting by ensembling returns a
predictor~\(f\) whose classification error satisfies with probability
\(1/3\), \[
R_S[f]\le \frac12-\Omega(\sqrt{k/n})\,.
\] In particular,~\(\gengap(f)\ge\Omega(\sqrt{k/n})\).

\end{Proposition}

We also have a nearly matching upper bound that essentially follows from
a Hoeffding's concentration inequality just as the cardinality bound in
the previous chapter. However, in order to apply Hoeffding's inequality
we first need to understand a useful idea about how we can analyze the
adaptive setting.

The idea is that we can think of the interaction between a fixed analyst
\(\cA\) and the dataset as a \emph{tree}. The root node is labeled by
\(f_1=\cA(\emptyset)\), i.e., the first function that the analyst
queries without any input. The response~\(R_S[f_1]\) takes on~\(n+1\)
possible values. This is because we consider the zero-one loss, which
can only take the values~\(\{0, 1/n, 2/n,\dots, 1\}\). Each possible
response value \(a_1\) creates a new child node in the tree
corresponding to the function \(f_2=\cA(a_1)\) that the analyst queries
upon receiving answer~\(a_1\) to the first query~\(f_1\). We recursively
continue the process until we built up a tree of depth~\(k\) and
degree~\(n+1\) at each node.

\begin{figure}
\centering
\includegraphics[width=1\textwidth,height=\textheight]{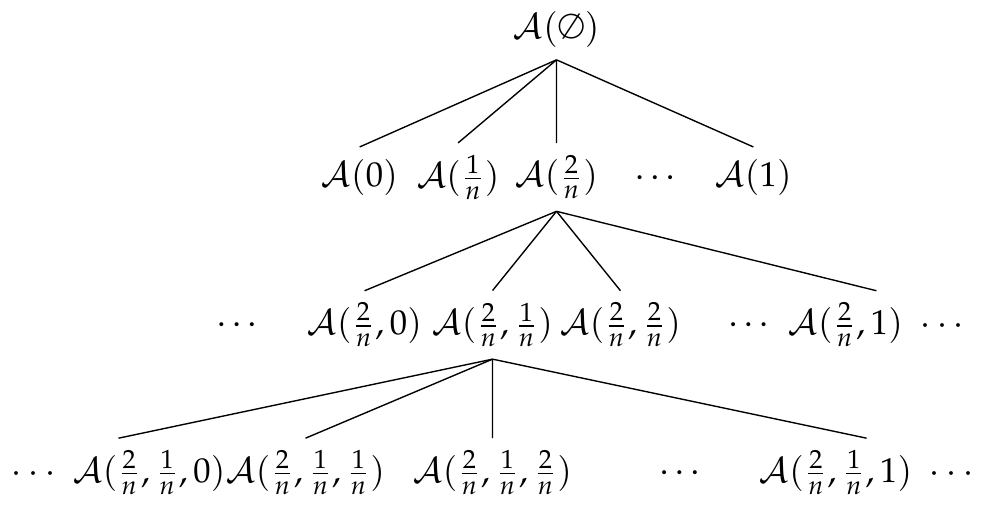}
\caption{Constructing a tree of depth \(k\) and degree \(n+1\) given an
adaptive analyst. Each node corresponds to the predictor the analyst
chooses based on the responses seen so far.}
\end{figure}

Note that this tree only depends on the analyst and how it responds to
possible query answers, but it does not depend on the actual query
answers we get out of the sample~\(S\). The tree is therefore
data-independent. This argument is useful in proving the following
proposition.\index{adaptive!tree}

\begin{Proposition}

For any sequence of~\(k\) adaptively chosen
predictors~\(f_1,\dots, f_k\), the holdout method satisfies with
probability~\(2/3\), \[
\max_{1\le t\le k} \gengap(f_t) \le O\left(\sqrt{k\log(n+1)/n}\right)\,.
\]

\end{Proposition}

\begin{Proof}

The adaptive analyst defines a tree of depth~\(k\) and degree~\(n+1\).
Let \(F\) be the set of functions appearing at any of the nodes in the
tree. Note that~\(|F|\le (n+1)^k.\)

Since this set of functions is data independent, we can apply the
cardinality bound from the previous chapter to argue that the maximum
generalization gap for any function in~\(F\) is bounded by
\(O(\sqrt{\log|F|/n})\) with any constant probability. But the functions
\(f_1,\dots, f_k\) are contained in~\(F\) by construction. Hence, the
claim follows.

\end{Proof}

These propositions show that the principal concern of ``training on the
testing data'' is not unfounded. In the worst case, holdout data can
lose its guarantees rather quickly. If this pessimistic bound manifested
in practice, popular benchmark datasets would quickly become useless.
But does it?

\hypertarget{replication-efforts}{%
\subsection{Replication efforts}\label{replication-efforts}}

In recent replication efforts, researchers carefully recreated new test
sets for the CIFAR-10 and ImageNet classification benchmarks, created
according to the very same procedure as the original test sets. The
researchers then took a large collection of representative models
proposed over the years and evaluated all of them on the new test sets.
In the case of MNIST, researchers used the \emph{lost digits} as a new
test set, since these digits hadn't been used in almost all of the
research on MNIST.

The results of these studies teach us a couple of important lessons that
we will discuss in turn.

\begin{figure}
\centering
\includegraphics[width=1\textwidth,height=\textheight]{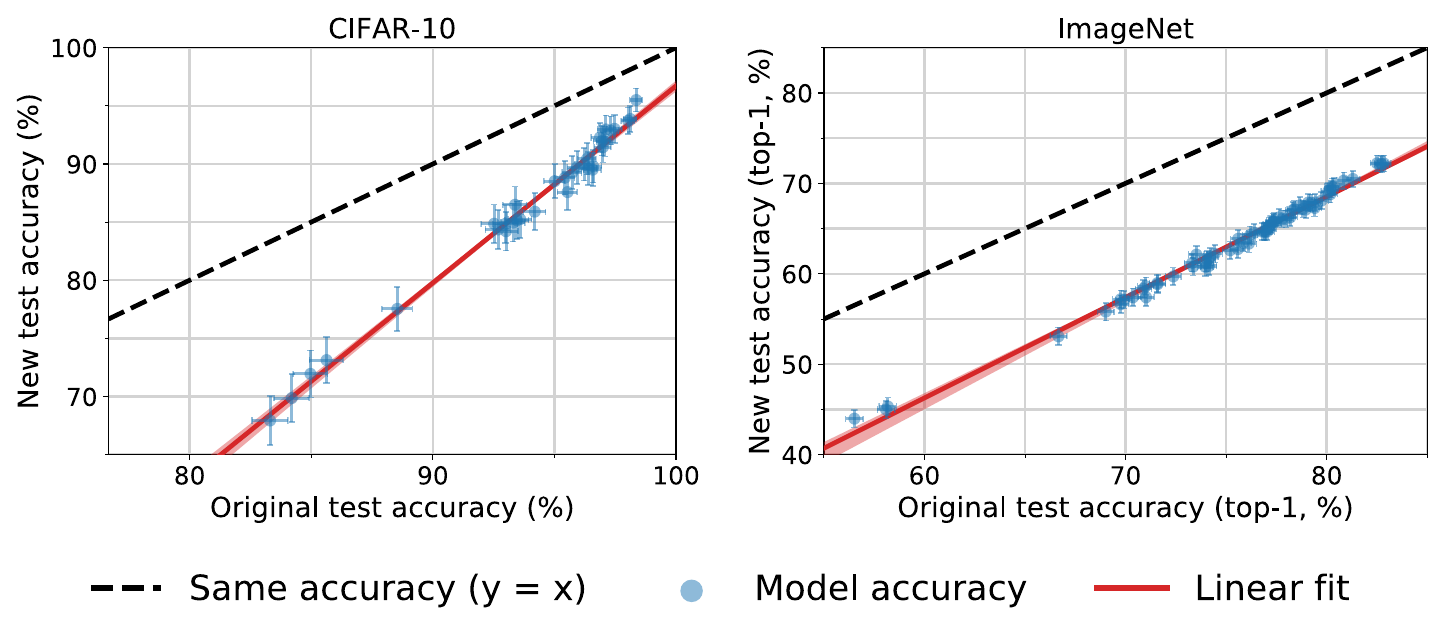}
\caption{Model accuracy on the original test sets vs.~new test sets for
CIFAR-10 and ImageNet. Each data point corresponds to one model in a
test bed of representative models.}
\end{figure}

First, all models suffer a significant drop in performance on the new
test set. The accuracy on the new data is substantially lower than on
the old data. This shows that these models \emph{transfer} surprisingly
poorly from one dataset to a very similar dataset that was constructed
in much the same way as the original data. This observation resonates
with a robust fact about machine learning. Model fitting will do exactly
that. The model will be good on exactly the data it is trained on, but
there is no good reason to believe that it will perform well on other
data. Generalization as we cast it in the preceding chapter is thus
about \emph{interpolation}. It's about doing well on more data from the
same source. It is decidedly \emph{not} about doing well on data from
other sources.

The second observation is relevant to the question of adaptivity; it's a
bit more subtle. The scatter plots admit a clean linear fit with
positive slope. In other words, the better a model is on the old test
set the better it is on the new test set, as well. But notice that newer
models, i.e., those with higher performance on the original test set,
had \emph{more} time to adapt to the test set and to incorporate more
information about it. Nonetheless, the better a model performed on the
old test set the better it performs on the new set. Moreover, on
CIFAR-10 we even see clearly that the absolute performance drops
diminishes with increasing accuracy on the old test set. In particular,
if our goal was to do well on the new test set, seemingly our best
strategy is to continue to inch forward on the old test set. This might
seem counterintuitive.

We will discuss each of these two observations in more detail, starting
with the one about adaptivity.

\hypertarget{benign-adaptivity}{%
\subsection{Benign adaptivity}\label{benign-adaptivity}}

The experiments we just discussed suggest that the effect of adaptivity
was more benign than our previous analysis suggested. This raises the
question what it is that prevents more serious overfitting. There are a
number of pieces to the puzzle that researchers have found. Here we
highlight two.

The main idea behind both mechanisms that damp adaptivity is that the
set of possible nodes in the adaptive tree may be much less than~\(n^k\)
because of empirical conventions. The first mechanism is \emph{model
similarity}. Effectively, model similarity notes that the leaves of the
adaptive tree may be producing similar predictions, and hence the
adaptivity penalty is smaller than our worst case count. The second
mechanism is the \emph{leaderboard principle}. This more subtle effect
states that since publication biases force researchers to chase
state-of-the-art results, they only publish models if they see
significant improvements over prior models.

While we don't believe that these two phenomena explain the entirety of
why overfitting is not observed in practice, they are simple mechanisms
that significantly reduce the effects of adaptivity. As we said, these
are two examples of norms in machine learning practice that diminish the
effects of overfitting.

\hypertarget{model-similarity}{%
\subsection{Model similarity}\label{model-similarity}}

Naively, there are~\(2^n\) total assignments of binary labels to a
dataset of size~\(n\). But how many such labeling assignments do we see
in practice? We do not solve pattern recognition problems using the
ensembling attack described above. Rather, we use a relatively small set
of function approximation architectures, and tune the parameters of
these architectures. While we have seen that these architectures can
yield any of the~\(2^n\) labeling patterns, we expect that a much
smaller set of predictions is returned in practice when we run standard
empirical risk minimization.

\emph{Model similarity} formalizes this notion as follows. Given an
adaptively chosen sequence of predictors~\(f_1,\dots,f_k\), we have a
corresponding sequence of empirical risks~\(R_1,\dots,R_k\).

\begin{Definition}

We say that a sequence of models~\(f_1,\ldots,f_k\)
are~\(\zeta\)-\emph{similar} if for all pairs of models~\(f_i\)
and~\(f_j\) with empirical risks \(R_i \leq R_j\), respectively, we have
\[
  \Pr\left[\{f_j(x) = y\} \cap \{f_i(x) \neq y\} \right] \leq \zeta\,.
\]

\end{Definition}

This definition states that there is low probability of a model with
small empirical risk misclassifying an example where a model with higher
empirical risk was correct. It effectively grades the set of
\emph{examples} as being easier or harder to classify, and suggests that
models with low risk usually get the easy examples correct.

Though simple, this definition is sufficient to reduce the size of the
adaptive tree, thus leading to better theoretical
bound.\citep{Mania-Model-Sim} Empirically, deep learning models appear
to have high similarity beyond what follows from their
accuracies.\citep{Mania-Model-Sim}

The definition of similarity can also help explain the scatter plots we
saw previously: When we consider the empirical risks of high similarity
models on two different test sets, the scatter plot the~\((R_i, R_i')\)
pairs cluster around a line.\citep{mania-sra}

\hypertarget{the-leaderboard-principle}{%
\subsection{The leaderboard principle}\label{the-leaderboard-principle}}

The leaderboard principle postulates that \emph{a researcher only cares
if their model improved over the previous best or
not.}\index{leaderboard!principle}\index{leaderboard!error} This
motivates a notion of \emph{leaderboard error} where the holdout method
is only required to track the risk of the best performing model over
time, rather than the risk of all models ever evaluated.

\begin{Definition}

Given an adaptively chosen sequence of predictors~\(f_1,\dots,f_k,\) we
define the \emph{leaderboard error} of a sequence of estimates
\(R_1,\dots,R_k\) as \[
\mathrm{lberr}(R_1,\dots,R_k)=
\max_{1\le t\le k}\left|\min_{1\le i\le t} R[f_i] - R_t\right|\,.
\]

\end{Definition}

We discuss an algorithm called the Ladder algorithm that achieves small
leaderboard accuracy. The algorithm is simple. For each given predictor,
it compares the empirical risk estimate of the predictor to the
previously smallest empirical risk achieved by any predictor encountered
so far. If the loss is below the previous best by some margin, it
announces the empirical risk of the current predictor and notes it as
the best seen so far. Importantly, if the loss is not smaller by a
margin, the algorithm releases no new information and simply continues
to report the previous best.

Again, we focus on risk with respect to the zero-one loss, although the
ideas apply more generally.\index{Ladder algorithm}

\begin{Algorithm}

\textbf{Input:} Dataset~\(S,\) threshold~\(\eta>0\)

\begin{itemize}
\tightlist
\item
  Assign initial leaderboard error \(R_0\leftarrow 1.\)
\item
  For each round \(t \leftarrow 1,2 \ldots:\)

  \begin{enumerate}
  \def\labelenumi{\arabic{enumi}.}
  \tightlist
  \item
    Receive predictor \(f_t\colon X\to Y\)
  \item
    If \(R_S[f_t] < R_{t-1} - \eta,\) update leaderboard error to
    \(R_t\leftarrow R_S[f_t].\) Else keep previous leaderboard error
    \(R_t\leftarrow R_{t-1}.\)
  \item
    Output leaderboard error \(R_t\)
  \end{enumerate}
\end{itemize}

\end{Algorithm}

The next theorem follows from a variant of the adaptive tree argument we
saw earlier, in which we carefully prune the tree and bound its
size.\index{adaptive!tree}

\begin{Theorem}

For a suitably chosen threshold parameter, for any sequence of
adaptively chosen predictors~\(f_1,\dots,f_k,\) the Ladder algorithm
achieves with probability~\(1-o(1)\): \[
\mathrm{lberr}(R_1,\dots,R_k)
\le O\left(\frac{\log^{1/3}(kn)}{n^{1/3}}\right)\,.
\]

\end{Theorem}

\begin{Proof}

Set~\(\eta = \log^{1/3}(kn)/n^{1/3}.\) With this setting of~\(\eta,\) it
suffices to show that with probability~\(1-o(1)\) we have for all
\(t\in[k]\) the bound
\(|R_S[f_t]-R[f_t]|\le O(\eta) = O(\log^{1/3}(kn)/n^{1/3})\).

Let~\(\cA\) be the adaptive analyst generating the function sequence.
The algorithm~\(\cA\) naturally defines a rooted tree~\(\cT\) of
depth~\(k\) recursively defined as follows:

\begin{enumerate}
\def\labelenumi{\arabic{enumi}.}
\tightlist
\item
  The root is labeled by \(f_1 = \cA(\emptyset).\)
\item
  Each node at depth \(1<i\le k\) corresponds to one realization
  \((h_1,r_1,\dots,h_{i-1},r_{i-1})\) of the tuple of random variables
  \((f_1,R_1,\dots,f_{i-1},R_{i-1})\) and is labeled by
  \(h_i = \cA(h_1,r_1,\dots,h_{i-1},r_{i-1}).\) Its children are defined
  by each possible value of the output \(R_i\) of Ladder Mechanism on
  the sequence \(h_1,r_1,\dots,r_{i-1},h_i.\)
\end{enumerate}

Let~\(B = (1/\eta+1)\log(4k(n+1)).\) We claim that the size of the tree
satisfies~\(|\cT|\le 2^B.\) To prove the claim, we will uniquely encode
each node in the tree using~\(B\) bits of information. The claim then
follows directly.

The compression argument is as follows. We use
\(\lceil\log k\rceil\le \log(2k)\) bits to specify the depth of the node
in the tree. We then specify the index of each~\(i\in[k]\) for which the
Ladder algorithm performs an ``update'' so that~\(R_i \le R_{i-1}-\eta\)
together with the value~\(R_i.\) Note that since~\(R_i \in[0,1]\) there
can be at most~\(\lceil 1/\eta\rceil\le (1/\eta)+1\) many such steps.
This is because the loss is lower bounded by~\(0\) and decreases
by~\(\eta\) each time there is an update.

\begin{figure}
\centering
\includegraphics[width=0.5\textwidth,height=\textheight]{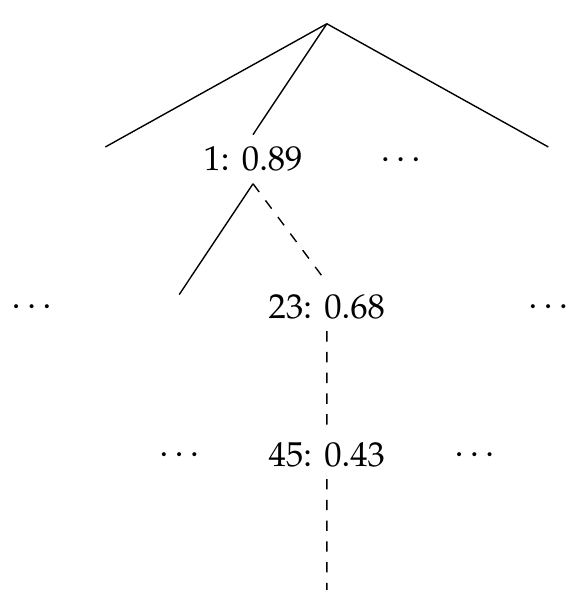}
\caption{Low bit encoding of the adaptive tree. Dashed lines correspond
to rounds with no update.}
\end{figure}

Moreover, there are at most~\(n+1\) many possible values for~\(R_i\),
since we're talking about the zero-one loss on a dataset of size~\(n\).
Hence, specifying all such indices requires at most
\((1/\eta + 1)(\log(n+1)+\log(2k))\) bits. These bits of information
uniquely identify each node in the graph, since for every index~\(i\)
not explicitly listed we know that~\(R_i=R_{i-1}.\) The total number of
bits we used is: \[
(1/\eta+1)(\log(n+1)+\log(2k))+\log(2k)
\le (1/\eta +1)\log(4k(n+1)) = B\mper
\] This establishes the claim we made. The theorem now follows by
applying a union bound over all nodes in~\(\cT\) and using Hoeffding's
inequality for each fixed node. Let~\(\mathcal{F}\) be the set of all
functions appearing in~\(\cT.\) By a union bound, we have \[
\begin{aligned}
\Pr\left\{\exists f\in \mathcal{F}\colon \left|R[f]-R_S[f]\right|>\epsilon\right\}
& \le 2|\mathcal{F}|\exp(-2\epsilon^2n)\\
& \le 2\exp(-2\epsilon^2n+B)\mper
\end{aligned}
\] Verify that by putting~\(\epsilon=5\eta\), this expression can be
upper bounded by~\(2\exp(-n^{1/3})=o(1)\), and thus the claim follows.

\end{Proof}

\hypertarget{harms-associated-with-data}{%
\section{Harms associated with data}\label{harms-associated-with-data}}

The collection and use of data often raises serious ethical concerns. We
will walk through some that are particularly relevant to machine
learning.\index{harm}

\hypertarget{representational-harm-and-biases}{%
\subsection{Representational harm and
biases}\label{representational-harm-and-biases}}

As we saw earlier, we have no reason to expect a machine learning model
to perform well on any population that differs significantly from the
training data. As a result, underrepresentation or misrepresentation of
populations in the training data has direct consequences on the
performance of any model trained on the
data.\index{harm!representational}\index{bias}

In a striking demonstration of this problem, Buolamwini and
Gebru\citep{buolamwini2018gender} point out that two facial analysis
benchmarks, IJB-A and Adience, overwhelmingly featured lighter-skinned
subjects. Introducing a new facial analysis dataset, which is balanced
by gender and skin type, Buolamwini and Gebru demonstrated that
commercial face recognition software misclassified darker-skinned
females at the highest rate, while misclassifying lighter-skinned males
at the lowest rate.

Images are not the only domain where this problem surfaces. Models
trained on text corpora reflect the biases and stereotypical
representations present in the training data. A well known example is
the case of word embeddings. Word embeddings map words in the English
language to a vector representation. This representation can then be
used as a feature representation for various other downstream tasks. A
popular word embedding method is Google's \texttt{word2vec} tool that
was trained on a corpus of Google news articles. Researchers
demonstrated that the resulting word embeddings encoded stereotypical
gender representations of the form \emph{man is to computer programmer
as woman is to homemaker}.\citep{bolukbasi2016man} Findings like these
motivated much work on \emph{debiasing} techniques that aim to remove
such biases from the learned representation. However, there is doubt
whether such methods can successfully remove bias after the
fact.\citep{gonen2019lipstick}\index{word embedding}

\hypertarget{privacy-violations}{%
\subsection{Privacy violations}\label{privacy-violations}}

\index{privacy}

The Netflix Prize was one of the most famous machine learning
competitions. Starting on October 2, 2006, the competition ran for
nearly three years ending with a grand prize of \$1M, announced on
September 18, 2009. Over the years, the competition saw 44,014
submissions from 5169 teams.\index{Netflix Prize}

The Netflix training data contained roughly 100 million movie ratings
from nearly 500 thousand Netflix subscribers on a set of~\(17770\)
movies. Each data point corresponds to a tuple
\texttt{\textless{}user,\ movie,\ date\ of\ rating,\ rating\textgreater{}}.
At about 650 megabytes in size, the dataset was just small enough to fit
on a CD-ROM, but large enough to be pose a challenge at the time.

The Netflix data can be thought of as a matrix with~\(n=480189\) rows
and \(m=17770\) columns. Each row corresponds to a Netflix subscriber
and each column to a movie. The only entries present in the matrix are
those for which a given subscriber rated a given movie with rating in
\(\{1,2,3,4,5\}\). All other entries---that is, the vast majority---are
\emph{missing}. The objective of the participants was to predict the
missing entries of the matrix, a problem known as \emph{matrix
completion}, or \emph{collaborative filtering} somewhat more broadly. In
fact, the Netflix challenge did so much to popularize this problem that
it is sometimes called the \emph{Netflix problem}. The idea is that if
we could predict missing entries, we'd be able to recommend unseen
movies to users accordingly.

The holdout data that Netflix kept secret consisted of about three
million ratings. Half of them were used to compute a running leaderboard
throughout the competition. The other half determined the final
winner.\index{leaderboard}

The Netflix competition was hugely influential. Not only did it attract
significant participation, it also fueled much academic interest in
collaborative filtering for years to come. Moreover, it popularized the
competition format as an appealing way for companies to engage with the
machine learning community. A startup called Kaggle, founded in April
2010, organized hundreds of machine learning competitions for various
companies and organizations before its acquisition by Google in 2017.

But the Netflix competition became infamous for another reason. Although
Netflix had replaced usernames by random numbers, researchers Narayanan
and Shmatikov were able to re-identify many of the Netflix subscribers
whose movie ratings were in the dataset.\citep{narayanan2008robust} In a
nutshell, their idea was to link movie ratings in the Netflix dataset
with publicly available movie ratings on IMDB, an online movie database.
Some Netflix subscribers had also publicly rated an overlapping set of
movies on IMDB under their real name. By matching movie ratings between
the two sources of information, Narayanan and Shmatikov succeeded in
associating anonymous users in the Netflix data with real names from
IMDB. In the privacy literature, this is called a \emph{linkage attack}
and it's one of the many ways that seemingly anonymized data can be
de-anonymized.\citep{dwork2017exposed}\index{deanonymization}

What followed were multiple class action lawsuits against Netflix, as
well as an inquiry by the Federal Trade Commission over privacy
concerns. As a consequence, Netflix canceled plans for a second
competition, which it had announced on August 6, 2009.

To this day, privacy concerns are a highly legitimate obstacle to public
data release and dataset creation. Deanonymization techniques are mature
and efficient. There provably is no algorithm that could take a dataset
and provide a rigorous privacy guarantee to all participants, while
being useful for all analyses and machine learning purposes. Dwork and
Roth call this the Fundamental Law of Information Recovery:
\emph{``overly accurate answers to too many questions will destroy
privacy in a spectacular way.''}\citep{dwork2014algorithmic}

\hypertarget{copyright}{%
\subsection{Copyright}\label{copyright}}

Privacy concerns are not the only obstruction to creating public
datasets and using data for machine learning purposes. Almost all data
sources are also subject to copyright. Copyright is a type of
intellectual property, protected essentially worldwide through
international treaties. It gives the creator of a piece of work the
exclusive right to create copies of it. Copyright expires only decades
after the creator dies. Text, images, video, digital or not, are all
subject to copyright. Copyright infringement is a serious crime in many
countries.\index{copyright}

The question of how copyright affects machine learning practice is far
from settled. Courts have yet to set precedents on whether copies of
content that feed into machine learning training pipelines may be
considered copyright infringement.

Legal scholar Levendowski argues that copyright law biases creators of
machine learning systems toward ``biased, low-friction data''. These are
data sources that carry a low risk of creating a liability under
copyright law, but carry various biases in the data that affect how the
resulting models perform.\citep{levendowski2018copyright}

One source of low-friction data is what is known as ``public domain''.
Under current US law, works enter public domain 75 years after the death
of the copyright holder. This means that most public domain works were
published prior to 1925. If the creator of a machine learning system
relied primarily on public domain works for training, it would likely
bias the data toward older content.

Another example of a low-friction dataset is the \emph{Enron email
corpus} that contains 1.6 million emails sent among Enron employees over
the course of multiple years leading up to the collapse of the company
in 2001. The corpus was released by the Federal Energy Regulatory
Commission (FERC) in 2003, following its investigation into the serious
accounting fraud case that became known as ``Enron scandal''. The Enron
dataset is one of the few available large data sources of emails sent
between real people. Even though the data were released by regulators to
the public, that doesn't mean that they are ``public domain''. However,
it is highly unlikely that a former Enron employee might sue for
copyright infringement. The dataset has numerous biases. The emails are
two decades old, and sent by predominantly male senior managers in a
particular business sector.

An example of a dataset that is \emph{not} low-friction is the corpus of
news articles that became the basis for Google's famous word embedding
tool called word2vec that we mentioned earlier. Due to copyright
concerns with the news articles contained in the corpus, the dataset was
never released, only the trained model.

\hypertarget{problem-framing-and-comparisons-with-humans}{%
\subsection{Problem framing and comparisons with
humans}\label{problem-framing-and-comparisons-with-humans}}

A long-standing ambition of artificial intelligence research is to match
or exceed human cognitive abilities by an algorithm. This desire often
leads to comparisons between humans and machines on various tasks.
Judgments about human accuracy often also enter the debate around when
to use statistical models in high stakes decision making settings.

The comparison between human decision makers and statistical models is
by no means new. For decades, researchers have compared the accuracy of
human judgments with that of statistical
models.\citep{dawes1989clinical}

Even within machine learning, the debate dates way back. A 1991 paper by
Bromley and Sackinger explicitly compared the performance of artificial
neural networks to a measure of human accuracy on the USPS digits
dataset that predates the famous MNIST data.\citep{bromley1991neural} A
first experiment put the human accuracy at 2.5\%, a second experiment
found the number 1.51\%, while a third reported the number
2.37\%.\citep{chaaban2007human}

Comparison with so-called \emph{human baselines} has since become widely
accepted in the machine learning community. The Electronic Frontier
Foundation (EFF), for example, hosts a major repository of AI progress
measures that compares the performance of machine learning models to
\emph{reported human accuracies} on numerous
benchmarks.\citep{eckersley2017eff}

For the ILSVRC 2012 data, the reported human error rate is 5.1\%. To be
precise, this number is referring to the fraction of times that the
correct image label was not contained in the top 5 predicted labels.
This often quoted number corresponds to the performance of a single
human annotator who was ``trained on 500 images and annotated 1500 test
images''. A second annotator who was ``trained on 100 images and then
annotated 258 test images'' achieved an error rate of
12\%.\citep{russakovsky2015imagenet}

Based on this number of 5.1\%, researchers announced in 2015 that their
model was ``the first to surpass human-level
performance''.\citep{he2015delving} Not surprisingly, this claim
received significant attention throughout the media.

However, a later more careful investigation into ``human accuracy'' on
ImageNet revealed a very different picture.\citep{shankar2020evaluating}
The researchers found that only models from 2020 are actually on par
with the strongest human labeler. Moreover, when restricting the data to
590 object classes out of 1000 classes in total, the best human labeler
performed much better at less than 1\% error than even the best
predictive models. Recall, that the ILSVRC 2012 data featured 118
different dog breeds alone, some of which are extremely hard to
distinguish for anyone who is not a trained dog expert. In fact, the
researchers had to consult with experts from the American Kennel Club
(AKC) to disambiguate challenging cases of different dog breeds. Simply
removing dog classes alone increases the performance of the best human
labeler to less than 1.3\% error.

There is another troubling fact. Small variations in the data collection
protocol turn out to have a significant effect on the performance of
machine predictors: ``the accuracy scores of even the best image
classifiers are still highly sensitive to minutiae of the data cleaning
process.''\citep{recht2019imagenet}

These results cast doubt not only on how me measure ``human accuracy'',
but also on the validity of the presumed theoretical construct of
``human accuracy'' itself. It is helpful to take a step back and reflect
on measurement more broadly. Recall from Chapter 4 that the field of
measurement theory distinguishes between a measurement procedure and the
target \emph{construct} that we wish to measure. For any measurement to
be valid, the target construct has to be \emph{valid} in the first
place.

However, the machine learning community has adopted a rather casual
approach to measuring human accuracy. Many researchers assume that the
construct of \emph{human accuracy} exists unambiguously and it is
whatever number comes out of some ad-hoc testing protocol for some set
of human beings. These ad-hoc protocols often result in anecdotal
comparisons of questionable scientific value.

There is a broader issue with the idea of \emph{human accuracy}. The
notion presupposes that we have already accepted the prediction task to
be the definitive task that we ought to solve, thus forgoing alternative
solutions. But in many cases the problem formulation in itself is the
subject of normative debate.

Consider the case of predicting \emph{failure to appear in court}. This
prediction problem is at the center of an ongoing criminal justice
reform in the United states. Many proponents seek to replace, or at
least augment, human judges by statistical models that predict whether
or not a defendant would fail to appear in court, if released ahead of a
future trial in court. Defendants of high risk are jailed, often for
months without a verdict, until their court appointment. An alternative
to prediction is to understand the \emph{causes} of failure to appear in
court, and to specifically address these. We will turn to causality in
subsequent chapters, where we will see that it often provides an
important alternative to prediction.

As it turns out, defendants often fail to appear in court for lack of
transportation, lack of childcare, inflexible work hours, or simply too
many court appointments. Addressing these fundamental problems, in fact,
is part of a settlement in Harris County, Texas.

To conclude, invalid judgments about human performance relative to
machines are not just a scientific error, they also have the potential
to create narratives that support poor policy choices in high stakes
policy questions around the use of predictive models in consequential
decisions.

\hypertarget{toward-better-data-practices}{%
\section{Toward better data
practices}\label{toward-better-data-practices}}

The practices of data collection and dataset creation in the machine
learning community leave much room for improvement. We close this
chapter highlighting a few practices that can be immediately adopted.

\hypertarget{data-annotation}{%
\subsection{Data annotation}\label{data-annotation}}

Many existing datasets in machine learning are poorly documented, and
details about their creation are often missing. This leads to a range of
issues from lack of reproducibility and concerns of scientific validity
to misuse and ethical concerns. Fortunately, there is some emerging
literature on how to better execute and document the creation of
datasets for machine learning.

\emph{Datasheets for datasets} is an initiative to promote a more
detailed and systematic annotation for
datasets.\citep{gebru2018datasheets} A datasheet requires the creator of
a dataset to answer questions relating to several areas of interest:
Motivation, composition, collection process,
preprocessing/cleaning/labeling, uses, distribution,
maintenance.\index{datasheets}

One goal is that process of creating a datasheet will help anticipate
ethical issues with the dataset. But datasheets also aim to make data
practices more reproducible, and help practitioners select more adequate
data sources.

Going a step beyond datasheets, researchers Jo and
Gebru\citep{jo2020lessons} draw lessons from archival and library
sciences for the construction and documentation of machine learning
datasets. These lessons draw attention to issues of consent,
inclusivity, power, transparency, ethics and privacy.

\hypertarget{lessons-from-measurement}{%
\subsection{Lessons from measurement}\label{lessons-from-measurement}}

Measurement theory is an established science with ancient roots. In
short, measurement is about assigning numbers to objects in the real
world in a way that reflects relationships between these objects.
Measurement draws an important distinction between a \emph{construct}
that we wish to measure and the measurement procedure that we used to
create a numerical representation of the construct.\index{measurement}

For example, we can think of a well-designed math exam as measuring the
mathematical abilities of a student. A student with greater mathematical
ability than another is expected to score higher on the exam. Viewed
this way, an exam is a \emph{measurement procedure} that assigns numbers
to students. The \emph{mathematical ability} of a student is the
construct we hope to measure. We desire that the ordering of these
numbers reflects the sorting of students by their mathematical
abilities. A measurement procedure operationalizes a
construct.\index{measurement!construct}\index{measurement!procedure}

Recall that in a prediction problem we have covariates~\(X\) from which
we're trying to predict a variable~\(Y\). This variable~\(Y\) is what we
call the \emph{target variable} in our prediction problem. The
definition and choice of a target variable is one way that measurement
theory becomes relevant to machine learning practice.

Consider a machine learning practitioner who attempts to classify the
sentiment of a paragraph of text as ``toxic'' or not. In the language of
measurement, ``toxicity'' is a construct. Whatever labeling procedure
the practitioner comes up with can be thought of as a measurement
procedure that implicitly or explicitly operationalizes this construct.
Before resorting to ad-hoc labeling or survey procedures, machine
learning practitioners should survey available research.

A poor target variable cannot be ironed out with additional data. In
fact, the more data we feed into our model, the better it gets at
capturing the flawed target variable. Improved data quality or diversity
are no cure either.

All formal fairness criteria that involve the target variable,
separation and sufficiency are two prominent examples are either
meaningless or downright misleading when the target variable itself is
the locus of discrimination. Recall from Chapter 2 that separation
requires the protected attribute to be independent of the prediction
conditional on the target variable. Sufficiency requires the target
variable to be independent of the protected attribute given the
prediction.\index{target variable}

To get a better grasp of what makes a target variable more or less
problematic, consider a few examples.

\begin{enumerate}
\def\labelenumi{\arabic{enumi}.}
\tightlist
\item
  Predicting the value of the Standard and Poor 500 Index (S\&P 500) at
  the close of the New York Stock Exchange tomorrow.
\item
  Predicting whether an individual is going to default on a loan.
\item
  Predicting whether an individual is going to commit a crime.
\end{enumerate}

The first example is rather innocuous. It references a fairly robust
target variable, even though it relies on a number of social facts.

The second example is a common application of statistical modeling that
underlies much of modern credit scoring in the United States. At first
sight a default event seems like a clean cut target variable. But the
reality is different. In a public dataset released by the Federal
Reserve\citep{fed2007report} default events are coded by a so-called
\emph{performance} variable that measures a \emph{serious delinquency in
at least one credit line of a certain time period}. More specifically,
the report states that the

\begin{quote}
measure is based on the performance of new or existing accounts and
measures whether individuals have been late 90 days or more on one or
more of their accounts or had a public record item or a new collection
agency account during the performance period.\citep{fed2007report}
\end{quote}

Our third example runs into the most concerning measurement problem. How
do we determine if an individual committed a crime? What we can
determine with certainty is whether or not an individual was arrested
and found guilty of a crime. But this depends crucially on who is likely
to be policed in the first place and who is able to maneuver the
criminal justice system successfully following an arrest.

Sorting out what a good target variable is, in full generality, can
involve the whole apparatus of measurement theory. The scope of
measurement theory, however, goes beyond defining reliable and valid
target variables for prediction. Measurement comes in whenever we create
features for a machine learning problem and should therefore be an
essential part of the data creation process.

Judging the quality of a measurement procedure is a difficult task.
Measurement theory has two important conceptual frameworks for arguing
about what makes measurement good. One is \emph{reliability}. The other
is
\emph{validity}.\index{measurement!reliability}\index{measurement!validity}

Reliability describes the differences observed in multiple measurements
of the same object under identical conditions. Thinking of the
measurement variable as a random variable, reliability is about the
variance between independent identically distributed measurements. As
such, reliability can be analogized with the statistical notion of
variance.

Validity is concerned with how well the measurement procedure in
principle captures the concept that we try to measure. If reliability is
analogous to variance, it is tempting to see validity as analogous to
\emph{bias}. But the situation is a bit more complicated. There is no
simple formal criterion that we could use to establish validity. In
practice, validity is based to a large extent on human expertise and
subjective judgments.

One approach to formalize validity is to ask how well a score predicts
some external criterion. This is called \emph{external validity}. For
example, we could judge a measure of creditworthiness by how well it
predicts default in a lending scenario. While external validity leads to
concrete technical criteria, it essentially identifies good measurement
with predictive accuracy. However, that's certainly not all there is to
validity.

Construct validity is a framework for discussing validity that includes
numerous different types of evidence. Messick highlights six aspects of
construct validity:

\begin{itemize}
\tightlist
\item
  Content: How well does the content of the measurement instrument, such
  as the items on a questionnaire, measure the construct of interest?
\item
  Substantive: Is the construct supported by a sound theoretical
  foundation?
\item
  Structural: Does the score express relationships in the construct
  domain?
\item
  Generalizability: Does the score generalize across different
  populations, settings, and tasks?
\item
  External: Does the score successfully predict external criteria?
\item
  Consequential: What are the potential risks of using the score with
  regards to bias, fairness, and distributive justice?
\end{itemize}

Of these different criteria, external validity is the one most familiar
to the machine learning practitioner. But machine learning practice
would do well to embrace the other, more qualitative, criteria as well.
Ultimately, measurement forces us to grapple with the often surprisingly
uncomfortable question: What are we even trying to do when we predict
something?

\hypertarget{limits-of-data-and-prediction}{%
\section{Limits of data and
prediction}\label{limits-of-data-and-prediction}}

Machine learning fails in many scenarios and it's important to
understand the failure cases as much as the success stories.

The Fragile Families Challenge was a machine learning competition based
on the Fragile Families and Child Wellbeing study
(FFCWS).\citep{reichman2001fragile} Starting from a random sample of
hospital births between 1998 and 2000, the FFCWS followed thousand of
American families over the course of 15 years, collecting detailed
information, about the families' children, their parents, educational
outcomes, and the larger social environment. Once a family agreed to
participate in the study, data were collected when the child was born,
and then at ages 1, 3, 5, 9, and 15.\index{Fragile Families Challenge}

The Fragile Families Challenge took concluded in 2017. The underlying
dataset for the competition contains 4242 rows, one for each family, and
12943 columns, one for each variable plus an ID number of each family.
Of the 12942 variables, 2358 are constant (i.e., had the same value for
all rows), mostly due to redactions for privacy and ethics concerns. Of
the approximately 55 million (4242 x 12942) entries in the dataset,
about 73\% do not have a value. Missing values have many possible
reasons, including non-response of surveyed families, drop out of study
participants, as well as logical relationships between features that
imply certain fields are missing depending on how others are set. There
are six outcome variables, measured at age 15: \emph{1) child grade
point average (GPA), 2) child grit, 3) household eviction, 4) household
material hardship, 5) caregiver layoff, and 6) caregiver participation
in job training.}

The goal of the competition was to predict the value of the outcome
variables at age 15 given the data from age 1 through 9. As is common
for competitions, the challenge featured a three-way data split:
training, leaderboard, and test sets. The training set is publicly
available to all participants, the leaderboard data support a
leaderboard throughout the competition, and the test set is used to
determine a final winner.

Despite significant participation from hundreds of researchers
submitting thousands of models over the course of five months, the
outcome of the prediction challenge was disappointing. Even the winning
model performed hardly better than a simple baseline and predicted
little more than the average outcome value.

What caused the poor performance of machine learning on the fragile
families data? There are a number of technical possibilities, the sample
size, the study design, the missing values. But there is also a more
fundamental reason that remains plausible. Perhaps the dynamics of life
trajectories are inherently unpredictable over the six year time delay
between measurement of the covariates and measurement of the outcome.
Machine learning works best in a static and stable world where the past
looks like the future. Prediction alone can be a poor choice when we're
anticipating dynamic changes, or when we are trying to reason about the
effect that hypothetical actions would have in the real world. In
subsequent chapters, we will develop powerful conceptual tools to engage
more deeply with this observation.

\hypertarget{chapter-notes-7}{%
\section{Chapter notes}\label{chapter-notes-7}}

This chapter overlaps significantly with a chapter on datasets and
measurement in the context of fairness and machine learning in the book
by Barocas, Hardt, and Narayanan\citep{barocas-hardt-narayanan}.

The study of adaptivity in data reuse was subject of work by Dwork,
Hardt, Pitassi, Reingold and
Roth\citep{dwork2015preserving, dwork2015reusable}, showing how tools
from differential privacy lead to statistical guarantees under
adaptivity. Much subsequent work in the area of adaptive data analysis
extended these works. A concern closely related to adaptivity goes under
the name of \emph{inference after selection} in the statistics
community, where it was recognized by Freedman in a 1983
paper.\citep{freedman1983note}

The notion of leaderboard error and the Ladder algorithm come from a
work by Blum and Hardt.\citep{blum2015ladder} The replication study for
CIFAR-10 and ImageNet is due to Recht, Roelofs, Schmidt, and
Shankar.\citep{recht2019imagenet}

The collection and use of large ad-hoc datasets (once referred to as
``big data'') has been scrutinized in several important works,
especially from critical scholars, historians, and social scientists
outside the computer science community. See, for example, boyd and
Crawford\citep{boyd2012critical},
Tufekci\citep{tufekci2014big, tufekci2014engineering}, and
Onuoha\citep{onuoha2016point}. An excellent survey by Paullada, Raji,
Bender, Denton, and Hanna provides a wealth of additional background and
references.\citep{paullada2020data} Olteanu, Castillo, Diaz, and Kiciman
discuss biases, methodological pitfalls, and ethical questions in the
context of social data analysis.\citep{olteanu2019social} In particular,
the article provides comprehensive taxonomies of biases and issues that
can arise in the sourcing, collection, processing, and analysis of
social data. Recently, Couldry and Mejias use the term \emph{data
colonialism} to emphasize the processes by which data are appropriated
and marginalized communities are exploited through data
collection.\citep{couldry2019data}

For an introduction to measurement theory, not specific to the social
sciences, see the books by
Hand.\citep{hand2010measurement, hand2016measurement} The comprehensive
textbook by Bandalos\citep{bandalos2018measurement} focuses on
applications to the social science, including a chapter on fairness.

\chapter{Causality}

Our starting point is the difference between an observation and an
action. What we see in passive observation is how individuals follow
their routine behavior, habits, and natural inclination. Passive
observation reflects the state of the world projected to a set of
features we chose to highlight. Data that we collect from passive
observation show a snapshot of our world as it is.

There are many questions we can answer from passive observation alone:
Do 16 year-old drivers have a higher incidence rate of traffic accidents
than 18 year-old drivers? Formally, the answer corresponds to a
difference of conditional probabilities. We can calculate the
conditional probability of a traffic accident given that the driver's
age is 16 years and subtract from it the conditional probability of a
traffic accident given the age is 18 years. Both conditional
probabilities can be estimated from a large enough sample drawn from the
distribution, assuming that there are both 16 year old and 18 year old
drivers. The answer to the question we asked is solidly in the realm of
observational statistics.

But important questions often are not observational in nature. Would
traffic fatalities decrease if we raised the legal driving age by two
years? Although the question seems similar on the surface, we quickly
realize that it asks for a fundamentally different insight. Rather than
asking for the frequency of an event in our manifested world, this
question asks for the effect of a hypothetical action.

As a result, the answer is not so simple. Even if older drivers have a
lower incidence rate of traffic accidents, this might simply be a
consequence of additional driving experience. There is no obvious reason
why an 18 year old with two months on the road would be any less likely
to be involved in an accident than, say, a 16 year-old with the same
experience. We can try to address this problem by holding the number of
months of driving experience fixed, while comparing individuals of
different ages. But we quickly run into subtleties. What if 18 year-olds
with two months of driving experience correspond to individuals who are
exceptionally cautious and hence---by their natural inclination---not
only drive less, but also more cautiously? What if such individuals
predominantly live in regions where traffic conditions differ
significantly from those in areas where people feel a greater need to
drive at a younger age?

We can think of numerous other strategies to answer the original
question of whether raising the legal driving age reduces traffic
accidents. We could compare countries with different legal driving ages,
say, the United States and Germany. But again, these countries differ in
many other possibly relevant ways, such as, the legal drinking age.

At the outset, causal reasoning is a conceptual and technical framework
for addressing questions about the effect of hypothetical actions or
\emph{interventions}. Once we understand what the effect of an action
is, we can turn the question around and ask what action plausibly
\emph{caused} an event. This gives us a formal language to talk about
cause and effect.

\hypertarget{the-limitations-of-observation}{%
\section{The limitations of
observation}\label{the-limitations-of-observation}}

Before we develop any new formalism, it is important to understand why
we need it in the first place.

To see why we turn to the venerable example of graduate admissions at
the University of California, Berkeley\index{Berkeley} in
1973.\citep{bickel1975sex} Historical data show that 12763 applicants
were considered for admission to one of 101 departments and
inter-departmental majors. Of the 4321 women who applied roughly 35
percent were admitted, while 44 percent of the 8442 men who applied were
admitted. Standard statistical significance tests suggest that the
observed difference would be highly unlikely to be the outcome of sample
fluctuation if there were no difference in underlying acceptance rates.

A similar pattern exists if we look at the aggregate admission decisions
of the six largest departments. The acceptance rate across all six
departments for men is about 44\%, while it is only roughly 30\% for
women, again, a significant difference. Recognizing that departments
have autonomy over who to admit, we can look at the gender bias of each
department.

\begin{longtable}[]{@{}lllll@{}}
\caption{UC Berkeley admissions data from 1973.}\tabularnewline
\toprule
& Men & & Women & \\
\midrule
\endfirsthead
\toprule
& Men & & Women & \\
\midrule
\endhead
Department & Applied & Admitted (\%) & Applied & Admitted (\%) \\
A & 825 & 62 & 108 & \textbf{82} \\
B & 520 & 60 & 25 & \textbf{68} \\
C & 325 & \textbf{37} & 593 & 34 \\
D & 417 & 33 & 375 & \textbf{35} \\
E & 191 & \textbf{28} & 393 & 24 \\
F & 373 & 6 & 341 & \textbf{7} \\
\bottomrule
\end{longtable}

What we can see from the table is that four of the six largest
departments show a higher acceptance ratio among women, while two show a
higher acceptance rate for men. However, these two departments cannot
account for the large difference in acceptance rates that we observed in
aggregate. So, it appears that the higher acceptance rate for men that
we observed in aggregate seems to have reversed at the department level.

Such reversals are sometimes called \emph{Simpson's paradox}, even
though mathematically they are no surprise. It's a fact of conditional
probability that there can be events~\(Y\) (here, acceptance),~\(A\)
(here, female gender taken to be a binary variable) and a random
variable~\(Z\) (here, department choice) such that:

\begin{enumerate}
\def\labelenumi{\arabic{enumi}.}
\tightlist
\item
  \(\mathbb{P}[ Y \mid A ] < \mathbb{P}[ Y \mid \neg A ]\)
\item
  \(\mathbb{P}[ Y \mid A, Z = z ] > \mathbb{P}[ Y \mid \neg A, Z = z]\)
  for all values \(z\) that the random variable \(Z\) assumes.
\end{enumerate}

Simpson's paradox nonetheless causes discomfort to some, because
intuition suggests that a trend which holds for all subpopulations
should also hold at the population level.

The reason why Simpson's paradox is relevant to our discussion is that
it's a consequence of how we tend to misinterpret what information
conditional probabilities encode. Recall that a statement of conditional
probability corresponds to passive observation. What we see here is a
snapshot of the normal behavior of women and men applying to graduate
school at UC Berkeley in 1973.

What is evident from the data is that gender influences department
choice. Women and men appear to have different preferences for different
fields of study. Moreover, different departments have different
admission criteria. Some have lower acceptance rates, some higher.
Therefore, one explanation for the data we see is that women
\emph{chose} to apply to more competitive departments, hence getting
rejected at a higher rate than men.

Indeed, this is the conclusion the original study drew:

\begin{quote}
\emph{The bias in the aggregated data stems not from any pattern of
discrimination on the part of admissions committees, which seems quite
fair on the whole, but apparently from prior screening at earlier levels
of the educational system. Women are shunted by their socialization and
education toward fields of graduate study that are generally more
crowded, less productive of completed degrees, and less well funded, and
that frequently offer poorer professional employment
prospects.\citep{bickel1975sex}}
\end{quote}

In other words, the article concluded that the source of gender bias in
admissions was a \emph{pipeline problem}: Without any wrongdoing by the
departments, women were ``shunted by their socialization'' that happened
at an earlier stage in their lives.

It is difficult to debate this conclusion on the basis of the available
data alone. The question of discrimination, however, is far from
resolved. We can ask why women applied to more competitive departments
in the first place. There are several possible reasons. Perhaps less
competitive departments, such as engineering schools, were unwelcoming
of women at the time. This may have been a general pattern at the time
or specific to the university. Perhaps some departments had a track
record of poor treatment of women that was known to the applicants.
Perhaps the department advertised the program in a manner that
discouraged women from applying.

The data we have also shows no measurement of \emph{qualification} of an
applicant. It's possible that due to self-selection women applying to
engineering schools in 1973 were over-qualified relative to their peers.
In this case, an equal acceptance rate between men and women might
actually be a sign of discrimination.

There is no way of knowing what was the case from the data we have. We
see that at best the original analysis leads to a number of follow-up
questions.

At this point, we have two choices. One is to design a new study and
collect more data in a manner that might lead to a more conclusive
outcome. The other is to argue over which scenario is more likely based
on our beliefs and plausible assumptions about the world.

Causal inference is helpful in either case. On the one hand, it can be
used as a guide in the design of new studies. It can help us choose
which variables to include, which to exclude, and which to hold
constant. On the other hand, causal models can serve as a mechanism to
incorporate scientific domain knowledge and exchange plausible
assumptions for plausible conclusions.

\hypertarget{causal-models}{%
\section{Causal models}\label{causal-models}}

We choose \emph{structural causal models}\index{causal!model} as the
basis of our formal discussion as they have the advantage of giving a
sound foundation for various causal notions we will encounter. The
easiest way to conceptualize a structural causal model is as a program
for generating a distribution from independent noise variables through a
sequence of formal instructions. Imagine instead of samples from a
distribution, somebody gave you a step-by-step computer program to
generate samples on your own starting from a random seed. The process is
not unlike how you would write code. You start from a simple random seed
and build up increasingly more complex constructs. That is basically
what a structural causal model is, except that each assignment uses the
language of mathematics rather than any concrete programming syntax.

\hypertarget{a-first-example}{%
\subsection{A first example}\label{a-first-example}}

Let's start with a toy example not intended to capture the real world.
Imagine a hypothetical population in which an individual exercises
regularly with probability~\(1/2\). With probability~\(1/3\), the
individual has a latent disposition to develop overweight that manifests
in the absence of regular exercise. Similarly, in the absence of
exercise, heart disease occurs with probability~\(1/3\). Denote by~\(X\)
the indicator variable of regular exercise, by~\(W\) that of excessive
weight, and by \(H\) the indicator of heart disease. Below is a
structural causal model to generate samples from this hypothetical
population. Recall a Bernoulli random variable~\(\mathrm{B}(p)\) with
bias~\(p\) is a biased coin toss that assumes value~\(1\) with
probability~\(p\) and value~\(0\) with probability~\(1-p.\)

\begin{enumerate}
\def\labelenumi{\arabic{enumi}.}
\tightlist
\item
  Sample independent Bernoulli random variables
  \(U_1\sim \mathrm{B}(1/2),\) \(U_2\sim \mathrm{B}(1/3),\)
  \(U_3\sim\mathrm{B}(1/3).\)
\item
  \(X := U_1\)
\item
  \(W := \,\) if \(X=1\) then \(0\) else \(U_2\)
\item
  \(H := \,\) if \(X=1\) then \(0\) else \(U_3\)
\end{enumerate}

Contrast this generative description of the population with a usual
random sample drawn from the population that might look like this:

\begin{longtable}[]{@{}ccc@{}}
\toprule
\(X\) & \(W\) & \(H\) \\
\midrule
\endhead
0 & 1 & 1 \\
1 & 0 & 0 \\
1 & 1 & 1 \\
1 & 1 & 0 \\
0 & 1 & 0 \\
\ldots{} & \ldots{} & \ldots{} \\
\bottomrule
\end{longtable}

From the program description, we can immediately see that in our
hypothetical population \emph{exercise} averts both \emph{overweight}
and \emph{heart disease}, but in the absence of exercise the two are
independent. At the outset, our program generates a joint distribution
over the random variables~\((X, W, H).\) We can calculate probabilities
under this distribution. For example, the probability of heart disease
under the distribution specified by our model
is~\(1/2 \cdot 1/3 = 1/6.\) We can also calculate the conditional
probability of heart diseases given overweight. From the event~\(W=1\)
we can infer that the individual does not exercise so that the
probability of heart disease given overweight increases to~\(1/3\)
compared with the baseline of~\(1/6\).

Does this mean that overweight causes heart disease in our model? The
answer is \emph{no} as is intuitive given the program to generate the
distribution. But let's see how we would go about arguing this point
formally. Having a program to generate a distribution is substantially
more powerful than just having sampling access. One reason is that we
can manipulate the program in whichever way we want, assuming we still
end up with a valid program. We could, for example, set~\(W := 1,\)
resulting in a new distribution. The resulting program looks like this:

\begin{enumerate}
\def\labelenumi{\arabic{enumi}.}
\setcounter{enumi}{1}
\tightlist
\item
  \(X := U_1\)
\item
  \(W := 1\)
\item
  \(H := \,\) if \(X=1\) then \(0\) else \(U_3\)
\end{enumerate}

This new program specifies a new distribution. We can again calculate
the probability of heart disease under this new distribution. We still
get~\(1/6.\) This simple calculation reveals a significant insight. The
substitution~\(W:=1\) does not correspond to a conditioning on~\(W=1.\)
One is an action, albeit inconsequential in this case. The other is an
observation from which we can draw inferences. If we observe that an
individual is overweight, we can infer that they have a higher risk of
heart disease (in our toy example). However, this does not mean that
lowering body weight would avoid heart disease. It wouldn't in our
example. The active substitution~\(W:=1\) in contrast creates a new
hypothetical population in which all individuals are overweight with all
that it entails in our model.

Let us belabor this point a bit more by considering another hypothetical
population, specified by the equations:

\begin{enumerate}
\def\labelenumi{\arabic{enumi}.}
\setcounter{enumi}{1}
\tightlist
\item
  \(W := U_2\)
\item
  \(X := \,\) if \(W=0\) then \(0\) else \(U_1\)
\item
  \(H := \,\) if \(X=1\) then \(0\) else \(U_3\)
\end{enumerate}

In this population exercise habits are driven by body weight. Overweight
individuals choose to exercise with some probability, but that's the
only reason anyone would exercise. Heart disease develops in the absence
of exercise. The substitution~\(W:=1\) in this model leads to an
increased probability of exercise, hence lowering the probability of
heart disease. In this case, the conditioning on~\(W=1\) has the same
affect. Both lead to a probability of~\(1/6.\)

What we see is that fixing a variable by substitution may or may not
correspond to a conditional probability. This is a formal rendering of
our earlier point that observation isn't action. A substitution
corresponds to an action we perform. By substituting a value we break
the natural course of action our model captures. This is the reason why
the substitution operation is sometimes called the \emph{do-operator},
written as~\(\mathrm{do}(W:=1)\).\index{do-operator}

Structural causal models give us a formal calculus to reason about the
effect of hypothetical actions. We will see how this creates a formal
basis for all the different causal notions that we will encounter in
this chapter.

\hypertarget{structural-causal-models-more-formally}{%
\subsection{Structural causal models, more
formally}\label{structural-causal-models-more-formally}}

Formally, a structural causal model\index{structural causal model} is a
sequence of assignments for generating a joint distribution starting
from independent noise variables. By executing the sequence of
assignments we incrementally build a set of jointly distributed random
variables. A structural causal model therefore not only provides a joint
distribution, but also a description of how the joint distribution can
be generated from elementary noise variables. The formal definition is a
bit cumbersome compared with the intuitive notion.

\begin{Definition}

A \emph{structural causal model} \(M\) is given by a set of variables
\(X_1,..., X_d\) and corresponding assignments of the form \[
X_i := f_i(P_i, U_i),\quad\quad i=1,..., d\,.
\]

Here,~\(P_i\subseteq\{X_1,...,X_d\}\) is a subset of the variables that
we call the \emph{parents} of~\(X_i\). The random
variables~\(U_1,..., U_d\) are called \emph{noise variables}, which we
require to be jointly independent.

The directed graph corresponding to the model has one node for each
variable~\(X_i,\) which has incoming edges from all the parents~\(P_i.\)
We will call such a graph the \emph{causal graph} corresponding to the
structural causal model.

\end{Definition}

The noise variables that appear in the definition model \emph{exogenous
factors} that influence the system. Consider, for example, how the
weather influences the delay on a traffic route you choose. Due to the
difficulty of modeling the influence of weather more precisely, we could
take the weather induced to delay to be an exogenous factor that enters
the model as a noise variable. The choice of exogenous variables and
their distribution can have important consequences for what conclusions
we draw from a model.

The parent nodes~\(P_i\) of node~\(i\) in a structural causal model are
often called the \emph{direct causes}\index{cause!direct} of~\(X_i.\)
Similarly, we call~\(X_i\) the \emph{direct effect}\index{direct effect}
of its direct causes~\(P_i.\) Recall our hypothetical population in
which weight gain was determined by lack of exercise via the
assignment~\(W:=\min\{U_1,1-X\}.\) Here we would say that exercise (or
lack thereof) is a direct cause of weight gain.

Structural causal model are a collection of formal \emph{assumptions}
about how certain variables interact. Each assignment specifies a
\emph{response function}. We can think of nodes as receiving messages
from their parents and acting according to these messages as well as the
influence of an exogenous noise variable.

To which extent a structural causal model conforms to reality is a
separate and difficult question that we will return to in more detail
later. For now, think of a structural causal model as formalizing and
exposing a set of assumptions about a data generating process. As such
different models can expose different hypothetical scenarios and serve
as a basis for discussion. When we make statements about cause and
effect in reference to a model, we don't mean to suggest that these
relationship necessarily hold in the real world. Whether they do depends
on the scope, purpose, and validity of our model, which may be difficult
to substantiate.

It's not hard to show that a structural causal model defines a unique
joint distribution over the variables~\((X_1,..., X_d)\) such that
\(X_i=f_i(P_i,U_i).\) It's convenient to introduce a notion for
probabilities under this distribution. When~\(M\) denotes a structural
causal model, we will write the probability of an event~\(E\) under the
entailed joint distribution as~\(\mathbb{P}_M\{E\}.\) To gain
familiarity with the notation, let~\(M\) denote the structural causal
model for the hypothetical population in which both weight gain and
heart disease are directly caused by an absence of exercise. We
calculated earlier that the probability of heart disease in this model
is \(\mathbb{P}_M\{H\}=1/6.\)

In what follows we will derive from this single definition of a
structural causal model all the different notions and terminology that
we'll need in this chapter.

Throughout, we restrict our attention to acyclic assignments. Many
real-world systems are naturally described as stateful dynamical system
with feedback loops. For example, often cycles can be broken up by
introducing time dependent variables, such as, investments at time~\(0\)
grow the economy at time~\(1\) which in turn grows investments at time
\(2\), continuing so forth until some chosen time horizon~\(t\). We will
return to a deeper dive into dynamical systems and feedback in later
chapters.

\hypertarget{causal-graphs}{%
\section{Causal graphs}\label{causal-graphs}}

We saw how structural causal models naturally give rise to \emph{causal
graphs}\index{causal!graph} that represent the assignment structure of
the model graphically. We can go the other way as well by simply looking
at directed graphs as placeholders for an unspecified structural causal
model which has the assignment structure given by the graph. Causal
graphs are often called \emph{causal diagrams}. We'll use these terms
interchangeably.

Below we see causal graphs for the two hypothetical populations from our
heart disease example.

\begin{figure}
\centering
\includegraphics[width=0.7\textwidth,height=\textheight]{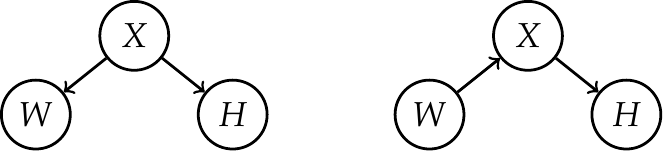}
\caption{Causal diagrams for the heart disease examples.}
\end{figure}

The scenarios differ in the direction of the link between exercise and
weight gain.

Causal graphs are convenient when the exact assignments in a structural
causal models are of secondary importance, but what matters are the
paths present and absent in the graph. Graphs also let us import the
established language of graph theory to discuss causal notions. We can
say, for example, that an \emph{indirect cause}\index{cause!indirect} of
a node is any ancestor of the node in a given causal graph. In
particular, causal graphs allow us to distinguish cause and effect based
on whether a node is an ancestor or descendant of another node.

Let's take a first glimpse at a few important graph structures.

\hypertarget{forks}{%
\subsection{Forks}\label{forks}}

A \emph{fork} is a node~\(Z\) in a graph that has outgoing edges to two
other variables~\(X\) and~\(Y\). Put differently, the node~\(Z\) is a
common cause of~\(X\) and
\(Y\).\index{fork}\index{cause!common}\index{confounder}

\begin{figure}
\centering
\includegraphics[width=0.3\textwidth,height=\textheight]{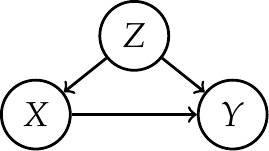}
\caption{Example of a fork.}
\end{figure}

We already saw an example of a fork in our weight and exercise example:
\(W\leftarrow X \rightarrow H\). Here, exercise~\(X\) influences both
weight and heart disease. We also learned from the example that~\(Z\)
has a \emph{confounding}\index{confounding} effect: Ignoring
exercise~\(X\), we saw that~\(W\) and~\(H\) appear to be positively
correlated. However, the correlation is a mere result of confounding.
Once we hold exercise levels constant (via the do-operation), weight has
no effect on heart disease in our example.

Confounding leads to a disagreement between the calculus of conditional
probabilities (observation) and do-interventions (actions).

Real-world examples of confounding are a common threat to the validity
of conclusions drawn from data. For example, in a well known medical
study a suspected beneficial effect of \emph{hormone replacement
therapy} in reducing cardiovascular disease disappeared after
identifying \emph{socioeconomic status} as a confounding
variable.\citep{humphrey2002postmenopausal}

\hypertarget{mediators}{%
\subsection{Mediators}\label{mediators}}

The case of a fork is quite different from the situation where~\(Z\)
lies on a directed path from~\(X\) to~\(Y\):

\begin{figure}
\centering
\includegraphics[width=0.3\textwidth,height=\textheight]{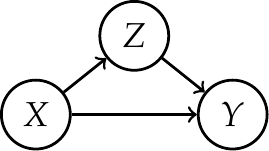}
\caption{Example of a mediator.}
\end{figure}

In this case, the path~\(X\to Z\to Y\) contributes to the total effect
of \(X\) on~\(Y\). It's a causal path and thus one of the ways in
which~\(X\) causally influences~\(Y\). That's why~\(Z\) is not a
confounder. We call~\(Z\) a \emph{mediator} instead.\index{mediator}

We saw a plausible example of a mediator in our UC Berkeley admissions
example. In one plausible causal graph, department choice mediates the
influences of gender on the admissions decision.

\hypertarget{colliders}{%
\subsection{Colliders}\label{colliders}}

Finally, let's consider another common situation: the case of a
\emph{collider}.\index{collider}\index{explaining away}\index{Berkson's law}

\begin{figure}
\centering
\includegraphics[width=0.3\textwidth,height=\textheight]{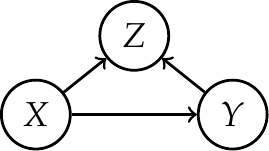}
\caption{Example of a collider.}
\end{figure}

Colliders aren't confounders. In fact, in the above graph,~\(X\)
and~\(Y\) are unconfounded, meaning that we can replace do-statements by
conditional probabilities. However, something interesting happens when
we condition on a collider. The conditioning step can create correlation
between~\(X\) and~\(Y\), a phenomenon called \emph{explaining away}. A
good example of the explaining away effect, or \emph{collider bias}, is
known as Berkson's paradox.\citep{berkson2014limitations} Two
independent diseases can become negatively correlated when analyzing
hospitalized patients. The reason is that when either disease (\(X\)
or~\(Y\)) is sufficient for admission to the hospital (indicated by
variable~\(Z\)), observing that a patient has one disease makes the
other statistically less likely. Berkson's paradox is a cautionary tale
for statistical analysis when we're studying a cohort that has been
subjected to a selection rule.

\hypertarget{interventions-and-causal-effects}{%
\section{Interventions and causal
effects}\label{interventions-and-causal-effects}}

Structural causal models give us a way to formalize the effect of
hypothetical actions or interventions on the population within the
assumptions of our model. As we saw earlier all we needed was the
ability to do substitutions.\index{intervention}\index{causal!effect}

\hypertarget{substitutions-and-the-do-operator}{%
\subsection{Substitutions and the
do-operator}\label{substitutions-and-the-do-operator}}

Given a structural causal model~\(M\) we can take any assignment of the
form \[
X := f(P, U)
\] and replace it by another assignment. The most common substitution is
to assign~\(X\) a constant value~\(x\): \[
X := x
\] We will denote the resulting model by~\(M'=M[X:=x]\) to indicate the
surgery we performed on the original model~\(M\). Under this assignment
we hold~\(X\) constant by removing the influence of its parent nodes and
thereby any other variables in the model.

Graphically, the operation corresponds to eliminating all incoming edges
to the node~\(X\). The children of~\(X\) in the graph now receive a
fixed message~\(x\) from~\(X\) when they query the node's value.

\begin{figure}
\centering
\includegraphics[width=0.6\textwidth,height=\textheight]{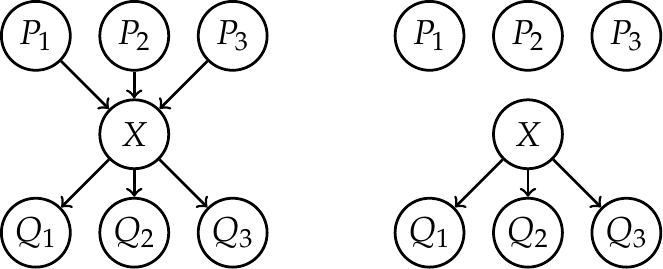}
\caption{Graph before and after substitution.}
\end{figure}

The assignment operator is also called the
\emph{do-operator}\index{do-operator} to emphasize that it corresponds
to performing an action or intervention. We already have notation to
compute probabilities after applying the do-operator,
namely,~\(\mathbb{P}_{M[X:=x]}(E).\)

Another notation is popular and common: \[
\mathbb{P}\{E\mid \mathrm{do}(X:=x)\} = \mathbb{P}_{M[X:=x]}(E)
\]

This notation analogizes the do-operation with the usual notation for
conditional probabilities, and is often convenient when doing
calculations involving the do-operator. Keep in mind, however, that the
do-operator (action) is fundamentally different from the conditioning
operator (observation).

\hypertarget{causal-effects}{%
\subsection{Causal effects}\label{causal-effects}}

The \emph{causal effect} of an action~\(X:=x\) on a variable~\(Y\)
refers to the distribution of the variable~\(Y\) in the
model~\(M[X:=x].\) When we speak of the causal effect of a
variable~\(X\) on another variable~\(Y\) we refer to all the ways in
which setting~\(X\) to any possible value~\(x\) affects the distribution
of~\(Y\).\index{causal!effect}

Often times~\(X\) denotes the presence or absence of an intervention or
\emph{treatment}. In such case,~\(X\) is a binary variable and are
interested in a quantity such as \[
\mathbb{E}_{M[X:=1]}[Y] - \mathbb{E}_{M[X:=0]}[Y]\,.
\] This quantity is called the \emph{average treatment
effect}\index{treatment effect!average}. It tells us how much treatment
(action~\(X:=1\)) increases the expectation of~\(Y\) relative to no
treatment (action~\(X:=0\)).

Causal effects are population quantities. They refer to effects averaged
over the whole population. Often the effect of treatment varies greatly
from one individual or group of individuals to another. Such treatment
effects are called \emph{heterogeneous}.

\hypertarget{confounding}{%
\section{Confounding}\label{confounding}}

Important questions in causality relate to when we can rewrite a
do-operation in terms of conditional probabilities. When this is
possible, we can estimate the effect of the do-operation from
conventional conditional probabilities that we can estimate from data.

The simplest question of this kind asks when a causal effect
\(\mathbb{P}\{Y=y\mid \mathrm{do}(X:=x)\}\) coincides with the condition
probability~\(\mathbb{P}\{Y=y\mid X=x\}.\) In general, this is not true.
After all, the difference between observation (conditional probability)
and action (interventional calculus) is what motivated the development
of causality.

The disagreement between interventional statements and conditional
statements is so important that it has a well-known name:
\emph{confounding}\index{confounding}. We say that~\(X\) and~\(Y\) are
confounded when the causal effect of action~\(X:=x\) on~\(Y\) does not
coincide with the corresponding conditional probability.

When~\(X\) and~\(Y\) are confounded, we can ask if there is some
combination of conditional probability statements that give us the
desired effect of a do-intervention. This is generally possible given a
causal graph by conditioning on the parent nodes~\(\mathit{PA}\) of the
node~\(X\): \[
\mathbb{P}\{Y=y\mid \mathrm{do}(X:=x)\}
= \sum_z \mathbb{P}\{Y=y\mid X=x, \mathit{PA} = z\}
\mathbb{P}\{\mathit{PA} = z\}
\] This formula is called the \emph{adjustment
formula}\index{adjustment formula}. It gives us one way of estimating
the effect of a do-intervention in terms of conditional probabilities.

The adjustment formula is one example of what is often called
\emph{controlling for} a set of variables: We estimate the effect
of~\(X\) on \(Y\) separately in every slice of the population defined by
a condition \(Z=z\) for every possible value of~\(z\). We then average
these estimated sub-population effects weighted by the probability
of~\(Z=z\) in the population. To give an example, when we control for
age, we mean that we estimate an effect separately in each possible age
group and then average out the results so that each age group is
weighted by the fraction of the population that falls into the age
group.

Controlling for more variables in a study isn't always the right choice.
It depends on the graph structure. Let's consider what happens when we
control for the variable~\(Z\) in the three causal graphs we discussed
above.

\begin{itemize}
\tightlist
\item
  Controlling for a confounding variable \(Z\) in a fork
  \(X \leftarrow Z \rightarrow Y\) will deconfound the effect of \(X\)
  on \(Y\).
\item
  Controlling for a mediator \(Z\) will eliminate some of the causal
  influence of \(X\) on \(Y\).
\item
  Controlling for a collider will create correlation between \(X\) and
  \(Y\). That is the opposite of what controlling for \(Z\) accomplishes
  in the case of a fork. The same is true if we control for a descendant
  of a collider.
\end{itemize}

\hypertarget{the-backdoor-criterion}{%
\subsection{The backdoor criterion}\label{the-backdoor-criterion}}

At this point, we might worry that things get increasingly complicated.
As we introduce more nodes in our graph, we might fear a combinatorial
explosion of possible scenarios to discuss. Fortunately, there are
simple sufficient criteria for choosing a set of deconfounding variables
that is safe to control for.

A well known graph-theoretic notion is the
\emph{backdoor}\index{backdoor criterion}
criterion.\citep{pearl2009causality} Two variables are confounded if
there is a so-called \emph{backdoor} path between them. A \emph{backdoor
path} from~\(X\) to \(Y\) is any path starting at~\(X\) with a backward
edge ``\(\leftarrow\)'' into \(X\) such as:
\[ X \leftarrow A \rightarrow B \leftarrow C \rightarrow Y \]
Intuitively, backdoor paths allow information flow from~\(X\) to~\(Y\)
in a way that is not causal. To deconfound a pair of variables we need
to select a \emph{backdoor set} of variables that ``blocks'' all
backdoor paths between the two nodes. A backdoor path involving a chain
\(A\rightarrow B\rightarrow C\) can be blocked by controlling for~\(B\).
Information by default cannot flow through a collider
\(A\rightarrow B\leftarrow C\). So we only have to be careful not to
open information flow through a collider by conditioning on the
collider, or descendant of a collider.

\hypertarget{unobserved-confounding}{%
\subsection{Unobserved confounding}\label{unobserved-confounding}}

The adjustment formula might suggest that we can always eliminate
confounding bias by conditioning on the parent nodes. However, this is
only true in the absence of \emph{unobserved
confounding}\index{confounding!unobserved}. In practice often there are
variables that are hard to measure, or were simply left unrecorded. We
can still include such unobserved nodes in a graph, typically denoting
their influence with dashed lines, instead of solid lines.

\begin{figure}
\centering
\includegraphics[width=0.75\textwidth,height=\textheight]{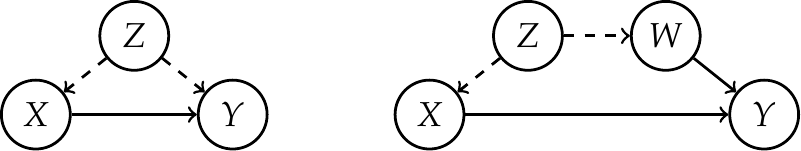}
\caption{Two cases of unobserved confounding.}
\end{figure}

The above figure shows two cases of unobserved confounding. In the first
example, the causal effect of~\(X\) on~\(Y\) is unidentifiable. In the
second case, we can block the confounding backdoor path
\(X\leftarrow Z\rightarrow W\rightarrow Y\) by controlling for~\(W\)
even though~\(Z\) is not observed. The backdoor criterion lets us work
around unobserved confounders in some cases where the adjustment formula
alone wouldn't suffice.

Unobserved confounding nonetheless remains a major obstacle in practice.
The issue is not just lack of measurement, but often lack of
anticipation or awareness of a confounding variable. We can try to
combat unobserved confounding by increasing the number of variables
under consideration. But as we introduce more variables into our study,
we also increase the burden of coming up with a valid causal model for
all variables under consideration. In practice, it is not uncommon to
control for as many variables as possible in a hope to disable
confounding bias. However, as we saw, controlling for mediators or
colliders can be harmful.

\hypertarget{randomization-and-the-backdoor-criterion}{%
\subsection{Randomization and the backdoor
criterion}\label{randomization-and-the-backdoor-criterion}}

The backdoor criterion gives a non-experimental way of eliminating
confounding bias given a causal model and a sufficient amount of
observational data from the joint distribution of the variables. An
alternative experimental method of eliminating confounding bias
randomization.

The idea is simple. If a treatment variable~\(T\) is an unbiased coin
toss, nothing but mere chance influenced it assignment. In particular,
there cannot be a confounding variable exercising influence on both the
treatment variable and a desired outcome variable.

A different way to think about is that randomization breaks natural
inclination. Rather than letting treatment take on its natural value, we
assign it randomly. Thinking in terms of causal models, what this means
is that we eliminate all incoming edges into the treatment variable. In
particular, this closes all backdoor paths and hence avoids confounding
bias. Because randomization is such an important part of causal
inference, we now turn to it in greater detail.

\hypertarget{experimentation-randomization-potential-outcomes}{%
\section{Experimentation, randomization, potential
outcomes}\label{experimentation-randomization-potential-outcomes}}

Let's think about experimentation from first principles. Suppose we have
a population of individuals and we have devised some treatment that can
be applied to each individual. We would like to know the effect of this
treatment on some measurable quantity.

As a simple example, and one which has had great utility, consider the
development of a vaccine for a disease. How can we tell if a vaccine
prevents disease? If we give everyone the vaccine, we'd never be able to
disentangle whether the treatment caused the associated change in
disease we observe or not. The common and widely accepted solution in
medicine is to restrict our attention to a subset of the population, and
leverage randomized assignment to eliminate isolate the effect of the
treatment.

The simplest mathematical formulation underlying randomized experiment
design is as follows. We assume a group of~\(n\) individuals
\(i=1,\dots, n.\) Suppose for an individual, if we apply a treatment,
the quantity of interest is equal to a value~\(Y_1(i)\). If we don't
apply the treatment, the quantity of interest is equal to~\(Y_0(i)\). We
define an outcome~\(Y(i)\) which is equal to~\(Y_1(i)\) if the treatment
is applied and equal to~\(Y_0(i)\) if the treatment is not applied. In
our vaccine example,~\(Y_1(i)\) indicates whether a person contracts the
disease in a specified time period following a vaccination
and~\(Y_0(i)\) indicates whether a person contracts the disease in the
same time period absent vaccination. Now, obviously, one person can only
take one of these paths. Nonetheless, we can imagine two \emph{potential
outcomes}\index{potential outcomes}: one potential outcome \(Y_1(i)\) if
the treatment is applied and another potential outcome \(Y_0(i)\) if the
treatment is not applied. Throughout this section, we assume that the
potential outcomes are fixed deterministic values.

We can write the relationship between observed outcome and potential
outcomes as a mathematical equation by introducing the boolean treatment
indicator~\(T(i)\) which is equal to~\(1\) if subject~\(i\) receives the
treatment and~\(0\) otherwise. In this case, the outcome for individual
\(i\) equals \[
  Y(i) = T(i) Y_1(i) + (1-T(i))Y_0(i)\,.
\] That is, if the treatment is applied, we observe~\(Y_1(i)\) and if
the treatment is not applied we observe~\(Y_0(i)\). While this potential
outcomes formulation is tautological, it lets us apply the same
techniques we use for estimating the mean to the problem of estimating
treatment effects.

The individual treatment effect\index{treatment effect!individual} is a
relation between the quantities~\(Y_1(i)\) and~\(Y_0(i)\), commonly just
the difference \(Y_1(i)-Y_0(i)\). If the difference is positive, we see
that applying the treatment increases the outcome variable for this
individual. But, as we've discussed, our main issue is that we can never
simultaneously observe~\(Y_1(i)\) and~\(Y_0(i)\). Once we choose whether
to apply the treatment or not, we can only measure the corresponding
treated or untreated condition.

While it may be daunting to predict the treatment effect at the level of
each individual, statistical algorithms can be applied to estimate
average treatment effects across the population. There are many ways to
define a measure of the effect of a treatment on a population. For
example, we earlier define the notion of an average treatment effect.
Let~\(\bar{Y}_1\) and~\(\bar{Y}_0\) denote the means of~\(Y_1(i)\) and
\(Y_0(i)\), respectively, averaged over~\(i=1,\dots,n\). We can write \[
\text{Average Treatment Effect} = \bar{Y}_1-\bar{Y}_0
\] In the vaccine example, this would be the difference in the
probability of contracting the illness if one was vaccinated vs if one
was not vaccinated.

The \emph{odds} that an individual catches the disease is the number of
people who catch the disease divided by the number who do not. The
\emph{odds ratio} for a treatment is the odds when every person receives
the vaccine divided by the odds when no one receives the vaccine. We can
write this out as a formula in terms of our quantities~\(\bar Y_1\) and
\(\bar Y_0\). When the potential outcomes take on values 0 or 1, the
average~\(\bar{Y}_1\) is the number of individuals for
which~\(Y_1(i)=1\) divided by the total number of individuals. Hence, we
can write the odds ratio as \[
\text{Odds Ratio} = \frac{\bar{Y}_1}{1-\bar{Y}_1} \cdot \frac{1-\bar{Y}_0}{\bar{Y}_0}\,.
\] This measures the decrease (or increase!) of the odds of a bad event
happening when the treatment is applied. When the odds ratio is less
than 1, the odds of a bad event are lower if the treatment is applied.
When the odds ratio is greater than 1, the odds of a bad event are
higher if the treatment is applied.

Similarly, the \emph{risk} that an individual catches the disease is the
ratio of the number of people who catch the disease to the total
population size. Risk and odds are similar quantities, but some
disciplines prefer one to the other by convention. The \emph{risk ratio}
is the the fraction of bad events when a treatment is applied divided by
the fraction of bad events when not applied: \[
\text{Risk Ratio} = \frac{\bar{Y}_1}{\bar{Y}_0}
\] The risk ratio measures the increase or decrease of relative risk of
a bad event when the treatment is applied. In the recent context of
vaccines, this ratio is popularly reported differently. The
\emph{effectiveness} of a treatment is one minus the risk ratio. This is
precisely the number used when people say a vaccine is 95\% effective.
It is equivalent to saying that the proportion of those treated who fell
ill was 20 times less than the proportion of those not treated who fell
ill.~Importantly, it does not mean that one has a 5\% chance of
contracting the disease.

\hypertarget{estimating-treatment-effects-using-randomization}{%
\subsection{Estimating treatment effects using
randomization}\label{estimating-treatment-effects-using-randomization}}

Let's now analyze how to estimate these effects using a randomized
procedure. In a randomized controlled
trial\index{randomized controlled trial} a group of~\(n\) subjects is
randomly partitioned into a \emph{control group} and a \emph{treatment
group}. We assume participants do not know which group they were
assigned to and neither do the staff administering the trial. The
treatment group receives an actual treatment, such as a drug that is
being tested for efficacy, while the control group receives a placebo
identical in appearance. An outcome variable is measured for all
subjects.\index{control group}\index{treatment group}

Formally, this means each~\(T(i)\) is an unbiased coin toss. Because we
randomly assign treatments we have \[
\E[Y(i) \mid T(i) = 1] = Y_1(i)
\qquad\text{and}\qquad
\E[Y(i)\mid T(i)=0] = Y_0(i)\,.
\] Therefore, for treatment value~\(t\in\{0, 1\}\), \[
\E\left[\frac1n\sum_{i=1}^n Y(i) \mid T(i)=t\right] = \bar Y_t\,.
\] In other words, to get an unbiased estimate of~\(\bar Y_t\) we just
have to average out all outcomes for subjects with treatment assignment
\(t\). This in turn gives us various causal effects we discussed
previously. We can also apply more statistics to this estimate to get
confidence bounds and large deviation bounds. Various things we know for
estimating the mean of a population carry over. For example, we need the
outcome variables to have bounded range in order for our estimates to
have low variance. Similarly, if we are trying to detect a tiny causal
effect, we must choose~\(n\) sufficiently large.

Typically in a randomized control trial, the~\(n\) subjects are supposed
to a uniformly random sample from a larger target population of~\(N\)
individuals. The group average~\(\bar Y_t\) is therefore itself only an
estimate of the population mean. Here, too, conventional statistics
applies in reasoning about how close~\(\bar Y_t\) is to the population
average.

Uniform sampling from a population is an idealization that is hard to
achieve in experimental practice. It is not only hard to independently
sample individuals in a large population, but we also need to be able to
set up identical scenarios to test interventions. For medical
treatments, what if there is variance between the treatment effect at
9AM in the Mayo Clinic on a Tuesday and at 11PM in the Alta Bates
Medical Center on a Saturday? If there are temporal or spatial or other
variabilities, the effective size of the population and the
corresponding variance grow. Accounting for such variability is a
daunting challenge of modern medical and social research that can be at
the root of many failures of replication.

The formulation here also assumes that the potential outcomes~\(Y_t(i)\)
do not vary over time. The framework could be generalized to account for
temporal variation, but such a generalization will not illuminate the
basic issues of statistical methods and modeling. We return to the
practice of causal inference and its challenges in the next chapter. But
before we do so, we will relate what we just learned to the structural
causal models that we saw earlier.

\hypertarget{counterfactuals}{%
\section{Counterfactuals}\label{counterfactuals}}

Fully specified structural causal models allow us to ask causal
questions that are more delicate than the mere effect of an action.
Specifically, we can ask \emph{counterfactual}\index{counterfactual}
questions such as: Would I have avoided the traffic jam had I taken a
different route this morning?

Formally, counterfactuals are random variables that generalize the
potential outcome variables we saw previously. Counterfactuals derive
from a structural causal model, which gives as another useful way to
think about potential outcomes. The procedure for extracting
counterfactuals from a structural causal model is algorithmic, but it
can look a bit subtle at first. It helps to start with a simple example.

\hypertarget{a-simple-counterfactual}{%
\subsection{A simple counterfactual}\label{a-simple-counterfactual}}

Assume every morning we need to decide between two routes~\(T=0\) and
\(T=1\). On bad traffic days, indicated by~\(U=1\), both routes are bad.
On good days, indicated by~\(U=0\), the traffic on either route is good
unless there was an accident on the route.

Let's say that~\(U\sim B(1/2)\) follows the distribution of an unbiased
coin toss. Accidents occur independently on either route with
probability~\(1/2.\) So, choose two Bernoulli random variables
\(U_0, U_1\sim B(1/2)\) that tell us if there is an accident on
route~\(0\) and route~\(1\), respectively. We reject all external route
guidance and instead decide on which route to take uniformly at random.
That is, \(T:=U_T\sim B(1/2)\) is also an unbiased coin toss.

Introduce a variable~\(Y\in\{0,1\}\) that tells us whether the traffic
on the chosen route is good (\(Y=0\)) or bad (\(Y=1\)). Reflecting our
discussion above, we can express~\(Y\) as \[
Y := T\cdot \max\{U, U_1\} + (1-T)\max\{U, U_0\} \,.
\] In words, when~\(T=0\) the first term disappears and so traffic is
determined by the larger of the two values~\(U\) and~\(U_0\). Similarly,
when~\(T=1\) traffic is determined by the larger of~\(U\) and~\(U_1\).

\begin{figure}
\centering
\includegraphics[width=0.25\textwidth,height=\textheight]{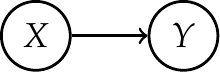}
\caption{Causal diagram for our traffic scenario.}
\end{figure}

Now, suppose one morning we have~\(T=1\) and we observe bad
traffic~\(Y=1.\) Would we have been better off taking the alternative
route this morning?

A natural attempt to answer this question is to compute the likelihood
of~\(Y=0\) after the do-operation~\(T:=0\), that is,
\(\mathbb{P}_{M[T:=0]}(Y=0).\) A quick calculation reveals that this
probability is~\(\frac12 \cdot \frac12 = 1/4.\) Indeed, given the
substitution~\(T:=0\) in our model, for the traffic to be good we need
that~\(\max\{U, U_0\}=0.\) This can only happen when both~\(U=0\)
(probability~\(1/2\)) and~\(U_0=0\) (probability~\(1/2\)).

But this isn't the correct answer to our question. The reason is that we
took route~\(T=1\) and observed that~\(Y=1.\) From this observation, we
can deduce that certain background conditions did not manifest for they
are inconsistent with the observed outcome. Formally, this means that
certain settings of the noise variables~\((U, U_0, U_1)\) are no longer
feasible given the observed event~\(\{Y=1, T=1\}\). Specifically,
if~\(U\) and~\(U_1\) had both been zero, we would have seen no bad
traffic on route \(T=1\), but this is contrary to our observation. In
fact, the available evidence~\(\{Y=1, T=1\}\) leaves only the following
settings for~\(U\) and \(U_1\):

\begin{longtable}[]{@{}cc@{}}
\caption{Possible noise settings after observing evidence. We leave out
\(U_0\) from the table, since its distribution is unaffected by our
observation.}\tabularnewline
\toprule
\(U\) & \(U_1\) \\
\midrule
\endfirsthead
\toprule
\(U\) & \(U_1\) \\
\midrule
\endhead
0 & 1 \\
1 & 1 \\
1 & 0 \\
\bottomrule
\end{longtable}

Each of these three cases is equally likely, which in particular means
that the event~\(U=1\) now has probability~\(2/3.\) In the absence of
any additional evidence, recall,~\(U=1\) had probability~\(1/2.\) What
this means is that the observed evidence~\(\{Y=1, T=1\}\) has biased the
distribution of the noise variable~\(U\) toward~\(1\). Let's use the
letter \(U'\) to refer to this biased version of~\(U\). Formally,~\(U'\)
is distributed according to the distribution of~\(U\) conditional on the
event~\(\{Y=1, T=1\}.\)

Working with this biased noise variable, we can again entertain the
effect of the action~\(T:=0\) on the outcome~\(Y\). For~\(Y=0\) we need
that \(\max\{U', U_0\}=0.\) This means that~\(U'=0\), an event that now
has probability~\(1/3\), and~\(U_0=0\) (probability~\(1/2\) as before).
Hence, we get the probability~\(1/6=1/2\cdot 1/3\) for the event
that~\(Y=0\) under our do-operation~\(T:=0\), and after updating the
noise variables to account for the observation~\(\{Y=1, T=1\}\).

To summarize, incorporating available evidence into our calculation
decreased the probability of no traffic (\(Y=0\)) when choosing
route~\(0\) from~\(1/4\) to~\(1/6\). The intuitive reason is that the
evidence made it more likely that it was generally a bad traffic day,
and even the alternative route would've been clogged. More formally, the
event that we observed biases the distribution of exogenous noise
variables.

We think of the result we just calculated as the \emph{counterfactual}
of choosing the alternative route given the route we chose had bad
traffic.

\hypertarget{the-general-recipe}{%
\subsection{The general recipe}\label{the-general-recipe}}

We can generalize our discussion of computing counterfactuals from the
previous example to a general procedure. There were three essential
steps. First, we incorporated available observational evidence by
biasing the exogenous noise variables through a conditioning operation.
Second, we performed a do-operation in the structural causal model after
we substituted the biased noise variables. Third, we computed the
distribution of a target variable.

These three steps are typically called \emph{abduction}, \emph{action},
and \emph{prediction}, as can be described as follows.

\begin{Definition}

Given a structural causal model~\(M\), an observed event~\(E\), an
action \(T:= t\) and target variable~\(Y,\) we define the
\emph{counterfactual} \(Y_{T:=t}(E)\) by the following three step
procedure:

\begin{enumerate}
\def\labelenumi{\arabic{enumi}.}
\tightlist
\item
  \textbf{Abduction:} Adjust noise variables to be consistent with the
  observed event. Formally, condition the joint distribution of
  \(U=(U_1,...,U_d)\) on the event \(E\). This results in a biased
  distribution \(U'\).
\item
  \textbf{Action:} Perform do-intervention \(T:=t\) in the structural
  causal model \(M\) resulting in the model \(M'=M[T:=t].\)
\item
  \textbf{Prediction:} Compute target counterfactual \(Y_{T:=t}(E)\) by
  using \(U'\) as the random seed in \(M'.\)
\end{enumerate}

\end{Definition}

It's important to realize that this procedure \emph{defines} what a
counterfactual is in a structural causal model. The notation
\(Y_{T:=t}(E)\) denotes the outcome of the procedure and is part of the
definition. We haven't encountered this notation before.

Put in words, we interpret the formal counterfactual~\(Y_{T:=t}(E)\) as
the value~\(Y\) would've taken had the variable~\(T\) been set to
value~\(t\) in the circumstances described by the event~\(E\).

In general, the counterfactual~\(Y_{T:=t}(E)\) is a random variable that
varies with~\(U'\). But counterfactuals can also be deterministic. When
the event~\(E\) narrows down the distribution of~\(U\) to a single point
mass, called \emph{unit}, the variable~\(U'\) is constant and hence the
counterfactual~\(Y_{T:=t}(E)\) reduces to a single number. In this case,
it's common to use the shorthand
notation~\(Y_{t}(u)=Y_{T:=t}(\{U=u\}),\) where we make the
variable~\(t\) implicit, and let~\(u\) refer to a single unit. The
counterfactual random variable~\(Y_t\) refers to~\(Y_t(u)\) for a random
draw of the noise variables~\(u\).

The motivation for the name \emph{unit} derives from the common
situation where the structural causal model describes a population of
entities that form the atomic units of our study. It's common for a unit
to be an individual (or the description of a single individual).
However, depending on application, the choice of units can vary. In our
traffic example, the noise variables dictate which route we take and
what the road conditions are.

Answers to counterfactual questions strongly depend on the specifics of
the structural causal model, including the precise model of how the
exogenous noise variables come into play. It's possible to construct two
models that have identical graph structures, and behave identically
under interventions, yet give different answers to counterfactual
queries.\citep{peters2017elements}

\hypertarget{potential-outcomes}{%
\subsection{Potential outcomes}\label{potential-outcomes}}

Let's return to the \emph{potential outcomes}\index{potential outcomes}
framework that we introduced when discussing randomized experiments.
Rather than deriving potential outcomes from a structural causal model,
we assume their existence as ordinary random variables, albeit some
unobserved. Specifically, we assume that for every unit~\(u\) there
exist random variables~\(Y_t(u)\) for every possible value of the
assignment~\(t\). This potential outcome turns out to equal the
corresponding counterfactual derived from the structrual equation model:
\[
\text{potential outcome}\, Y_t(u) = Y_{T:=t}(\{u\})\, \text{structural counterfactual}
\] In particular, there is no harm in using our potential outcome
notation~\(Y_t(u)\) as a shorthand for the corresponding counterfactual
notation.

In the potential outcomes model, it's customary to think of a binary
\emph{treatment variable} \(T\) that assumes only two values,~\(0\) for
\emph{untreated}, and~\(1\) for \emph{treated}. This gives us two
potential outcome variables~\(Y_0(u)\) and~\(Y_1(u)\) for each
unit~\(u\). There is some potential for notational confusion here.
Readers already familiar with the potential outcomes model may be used
to the notation ``\(Y_i(0), Y_i(1)\)'' for the two potential outcomes
corresponding to unit \(i\). In our notation the unit (or, more
generally, set of units) appears in the parentheses and the subscript
denotes the substituted value for the variable we intervene on.

The key point about the potential outcomes model is that we only observe
the potential outcome~\(Y_1(u)\) for units that were treated. For
untreated units we observe~\(Y_0(u)\). In other words, we can never
simultaneously observe both, although they're both assumed to exist in a
formal sense. Formally, the outcome~\(Y(u)\) for unit~\(u\) that we
observe depends on the binary treatment~\(T(u)\) and is given by the
expression: \[
Y(u)=Y_0(u)\cdot(1-T(u))+Y_1(u) \cdot T(u)
\]

We can revisit our traffic example in this framework. The next table
summarizes what information is observable in the potential outcomes
model. We think of the route we choose as the treatment variable, and
the observed traffic as reflecting one of the two potential outcomes.

\begin{longtable}[]{@{}cccr@{}}
\caption{Traffic example in the potential outcomes model}\tabularnewline
\toprule
Route \(T\) & Outcome \(Y_0\) & Outcome \(Y_1\) & Probability \\
\midrule
\endfirsthead
\toprule
Route \(T\) & Outcome \(Y_0\) & Outcome \(Y_1\) & Probability \\
\midrule
\endhead
0 & 0 & ? & 1/8 \\
0 & 1 & ? & 3/8 \\
1 & ? & 0 & 1/8 \\
1 & ? & 1 & 3/8 \\
\bottomrule
\end{longtable}

Often this information comes in the form of samples. For example, we
might observe the traffic on different days. With sufficiently many
samples, we can estimate the above frequencies with arbitrary accuracy.

\begin{longtable}[]{@{}cccc@{}}
\caption{Traffic data in the potential outcomes model}\tabularnewline
\toprule
Day & Route \(T\) & Outcome \(Y_0\) & Outcome \(Y_1\) \\
\midrule
\endfirsthead
\toprule
Day & Route \(T\) & Outcome \(Y_0\) & Outcome \(Y_1\) \\
\midrule
\endhead
1 & 0 & 1 & ? \\
2 & 0 & 0 & ? \\
3 & 1 & ? & 1 \\
4 & 0 & 1 & ? \\
5 & 1 & ? & 0 \\
\ldots{} & \ldots{} & \ldots{} & \ldots{} \\
\bottomrule
\end{longtable}

In our original traffic example, there were~\(16\) units corresponding
to the background conditions given by the four binary variables
\(U, U_0, U_1, U_T\). When the units in the potential outcome model
agree with those of a structural causal model, then causal effects
computed in the potential outcomes model agree with those computed in
the structural equation model. The two formal frameworks are perfectly
consistent with each other.

As is intuitive from the table above, causal inference in the potential
outcomes framework can be thought of as filling in the missing entries
(``?'') in the table above. This is sometimes called \emph{missing data
imputation} and there are numerous statistical methods for this task. If
we could \emph{reveal} what's behind the question marks, many quantities
would be readily computable. For instance, estimating the average
treatment effect would be as easy as counting rows.

When we were able to directly randomize the treatment variable, we
showed that treatment effects could be imputed from samples. When we are
working with observational data, there is a set of established
conditions under which causal inference becomes possible:

\begin{enumerate}
\def\labelenumi{\arabic{enumi}.}
\tightlist
\item
  \textbf{Stable Unit Treatment Value Assumption} (SUTVA): The treatment
  that one unit receives does not change the effect of treatment for any
  other unit.\index{SUTVA}
\item
  \textbf{Consistency}: Formally, \(Y(u)=Y_0(u)(1-T(u))+Y_1(u)T(u).\)
  That is, \(Y(u)=Y_0(u)\) if \(T(u)=0\) and \(Y(u)=Y_1(u)\) if
  \(T(u)=1\). In words, the outcome \(Y(u)\) agrees with the potential
  outcome corresponding to the treatment indicator.
\item
  \textbf{Ignorability}: The potential outcomes are independent of
  treatment given some deconfounding variables \(Z\), i.e.,
  \(T\bot (Y_0, Y_1)\mid Z\). In words, the potential outcomes are
  conditionally independent of treatment given some set of deconfounding
  variables.\index{ignorability}
\end{enumerate}

The first two assumptions automatically hold for counterfactual
variables derived from structural causal models according to the
procedure described above. This assumes that the units in the potential
outcomes framework correspond to the atomic values of the background
variables in the structural causal model.

The third assumption is a major one. The assumption on its own cannot be
verified or falsified, since we never have access to samples with both
potential outcomes manifested. However, we can verify if the assumption
is consistent with a given structural causal model, for example, by
checking if the set~\(Z\) blocks all backdoor paths from treatment~\(T\)
to outcome~\(Y\).

There's no tension between structural causal models and potential
outcomes and there's no harm in having familiarity with both. It
nonetheless makes sense to say a few words about the differences of the
two approaches.

We can derive potential outcomes from a structural causal model as we
did above, but we cannot derive a structural causal model from potential
outcomes alone. A structural causal model in general encodes more
assumptions about the relationships of the variables. This has several
consequences. On the one hand, a structural causal model gives us a
broader set of formal concepts (causal graphs, mediating paths,
counterfactuals for every variable, and so on). On the other hand,
coming up with a plausibly valid structural causal model is often a
daunting task that might require knowledge that is simply not available.
Difficulty to come up with a plausible causal model often exposes
unsettled substantive questions that require resolution first.

The potential outcomes model, in contrast, is generally easier to apply.
There's a broad set of statistical estimators of causal effects that can
be readily applied to observational data. But the ease of application
can also lead to abuse. The assumptions underpinning the validity of
such estimators are experimentally unverifiable. Our next chapter dives
deeper into the practice of causal inference and some of its
limitations.

\hypertarget{chapter-notes-8}{%
\section{Chapter notes}\label{chapter-notes-8}}

This chapter was developed and first published by Barocas, Hardt, and
Narayanan in the textbook \emph{Fairness and Machine Learning:
Limitations and Opportunities.}\citep{barocas-hardt-narayanan} With
permission from the authors, we include a large part of the original
text here with only slight modifications. We removed a significant
amount of material on discrimination and fairness and added an extended
discussion on randomized experiments.

There are several excellent introductory textbooks on the topic of
causality. For an introduction to causality with an emphasis on causal
graphs and structural equation models turn to Pearl's
primer\citep{pearl2016causal}, or the more comprehensive
textbook\citep{pearl2009causality}. Our exposition of Simpson's paradox
and the UC Berkeley data was influenced by Pearl's discussion, updated
for a new popular audience book\citep{pearl2018book}. The example has
been heavily discussed in various other writings, such as Pearl's recent
discussion \citep{pearl2018book}. We retrieved the Berkeley data from
\url{http://www.randomservices.org/random/data/Berkeley.html}. There is
some discrepancy with the data available on the Wikipedia page for
\href{https://en.wikipedia.org/wiki/Simpson\%27s_paradox}{Simpson's
paradox} that we retrieved on Dec 27, 2018.

For further discussion regarding the popular interpretation of Simpson's
original article\citep{simpson1951interpretation}, see the article by
Hernán, Clayton, and Keiding\citep{hernan2011simpson}, as well as
Pearl's text\citep{pearl2009causality}.

The technically-minded reader will enjoy complementing Pearl's book with
the recent open access text by Peters, Janzing, and
Schölkopf\citep{peters2017elements} that is
\href{https://mitpress.mit.edu/books/elements-causal-inference}{available
online}. The text emphasizes two variable causal models and applications
to machine learning. See Spirtes, Glymour and
Scheines\citep{spirtes2000causation} for a general introduction based on
causal graphs with an emphasis on \emph{graph discovery}, i.e.,
inferring causal graphs from observational data. An article by Schölkopf
provides additional context about the development of causality in
machine learning.\citep{scholkopf2019causality}

The classic formulation of randomized experiment design due to Jerzy
Neyman is now subsumed by and commonly referred to as the framework of
potential outcomes.\citep{neyman1923applications, rubin2005causal}
Imbens and Rubin\citep{imbens2015causal} give a comprehensive
introduction to the technical repertoire of causal inference in the
potential outcomes model. Angrist and Pischke\citep{angrist2008mostly}
focus on causal inference and potential outcomes in econometrics. Hernán
and Robins\citep{hernan2020causal} give another detailed introduction to
causal inference that draws on the authors' experience in epidemiology.
Morgan and Winship\citep{morgan2014counterfactuals} focus on
applications in the social sciences.

\chapter{Causal inference in practice}

The previous chapter introduced the conceptual foundations of causality,
but there's a lot more to learn about how these concepts play out in
practice. In fact, there's a flourishing practice of causal inference in
numerous scientific disciplines. Increasingly, ideas from machine
learning show up in the design of causal estimators. Conversely, ideas
from causal inference can help machine learning practitioners run better
experiments.

In this chapter we focus on estimating the average treatment effect,
often abbreviated as ATE, of a binary treatment~\(T\) on an outcome
variable~\(Y\): \[
\E[Y \mid \doop(T:=1) ] - \E[Y \mid \doop(T:=0)]\,.
\] Causal effects are population quantities that involve two
hypothetical actions, one holding the treatment variable constant at the
treatment value~\(1\), the other holding the treatment constant at its
baseline value~\(0\).

The central question in causal inference is how we can estimate causal
quantities, such as the average treatment effect, from data.

Confounding between the outcome and treatment variable is the main
impediment to causal inference from observational data. Recall that
random variables~\(Y\) and~\(T\) are confounded, if the conditional
probability distribution of~\(Y\) given~\(T\) does not equal its
interventional counterpart: \[
\Pr\{Y=y\mid \doop(T:=t)\}
\ne
\Pr\{Y=y\mid T=t\}
\] If these expressions were equal, we could estimate the average
treatment effect in a direct way by estimating the difference
\(\E[Y\mid T=1]-\E[Y\mid T=0]\) from samples. Confounding makes the
estimation of treatment effects more challenging, and sometimes
impossible. Note that the main challenge here is to arrive at an
expression for the desired causal effect that is free of any causal
constructs, such as the do-operator. Once we have a plain probability
expression at hand, tools from statistics allow us to relate the
population quantity with a finite sample estimate.

\hypertarget{design-and-inference}{%
\section{Design and inference}\label{design-and-inference}}

There are two important components to causal inference, one is
\emph{design}, the other is \emph{inference}.

In short, design is about sorting out various substantive questions
about the data generating process. Inference is about the statistical
apparatus that we unleash on the data in order to estimate a desired
causal effect.

Design requires us to decide on a population, a set of variables to
include, and a precise question to ask. In this process we need to
engage substantively with relevant scientific domain knowledge in order
to understand what assumptions we can make about the data.

Design can only be successful if the assumptions we are able to make
permit the estimation of the causal effect we're interested in. In
particular, this is where we need to think carefully about potential
sources of confounding and how to cope with them.

There is no way statistical estimators can recover from poor design. If
the design does not permit causal inference, there is simply no way that
a clever statistical trick could remedy the shortcoming. It's therefore
apt to think of causal insights as consequences of the substantive
assumptions that we can make about the data, rather than as products of
sophisticated statistical ideas.

Hence, we emphasize design issues throughout this chapter and
intentionally do not dwell on technical statements about rates of
estimation. Such mathematical statements can be valuable, but design
must take precedence.

\hypertarget{experimental-and-observational-designs}{%
\subsection{Experimental and observational
designs}\label{experimental-and-observational-designs}}

Causal inference distinguishes between \emph{experimental} and
\emph{observational} designs. Experimental designs generally are active
in the sense of administering some treatment to some set of experimental
units. Observational designs do not actively assign treatment, but
rather aim to make it possible to identify causal effects from collected
data without implementing any interventions.

The most common and well-established experimental design is a randomized
controlled trial (RCT). The main idea is to assign treatment randomly. A
randomly assigned treatment, by definition, is not influenced by any
other variable. Hence, randomization eliminates any confounding bias
between treatment and outcome.

In a typical implementation of a randomized controlled trial, subjects
are randomly partitioned into a \emph{treatment group} and a
\emph{control group}. The treatment group receives the treatment, the
control group receives no treatment. It is important that subjects do
not know which group they were assigned to. Otherwise knowledge of their
assignment may influence the outcome. To ensure this, subjects in the
control group receive what is called a \emph{placebo}, a device or
procedure that looks indistinguishable from treatment to the study
subject, but lacks the treatment ingredient whose causal powers are in
question. Adequate placebos may not exist depending on what the
treatment is, for example, in the case of a surgery.

Randomized controlled trials have a long history with many success
stories. They've become an important source of scientific knowledge.

Sometimes randomized controlled trials are difficult, expensive, or
impossible to administer. Treatment might be physically or legally
impossible, too costly, or too dangerous. Nor are they free of issues
and pitfalls.\citep{deaton2018understanding} In this chapter, we will
see observational alternatives to randomized controlled trials. However,
these are certainly not without their own set of difficulties and
shortcomings.

The machine learning practitioner is likely to encounter randomization
in the form of so-called \emph{A/B tests}. In an A/B test we randomly
assign one of two treatments to a set of individuals. Such experiments
are common in the tech industry to find out which of two changes to a
product leads to a better outcome.

\index{A/B test}

\hypertarget{the-observational-basics-adjustment-and-controls}{%
\section{The observational basics: adjustment and
controls}\label{the-observational-basics-adjustment-and-controls}}

For the remainder of the chapter we focus on observational causal
inference methods. In the previous chapter we saw that there are
multiple ways to cope with confounding between treatment and outcome.
One of them is to adjust (or control) for the parents (i.e., direct
causes) of~\(T\) via the adjustment formula.\index{adjustment formula}

The extra variables that we adjust for are also called \emph{controls},
and we take the phrase \emph{controlling for} to mean the same thing as
\emph{adjusting for}.

We then saw that we could use any set of random variables satisfying the
graphical backdoor criterion. This is helpful in cases where some direct
causes are unobserved so that we cannot use them in the adjustment
formula.

Let's generalize this idea even further and call a set of variables
\emph{admissible} if it satisfies the adjustment
formula.\index{admissible}

\begin{Definition}

We say that a discrete random variable~\(X\) is \emph{admissible} if it
satisfies the adjustment formula: \[
\Pr[Y=y\mid \doop(T:=t)] = \sum_x \Pr[Y=y\mid T=t, X=x] \Pr[X=x]
\] Here we sum over all values~\(x\) in the support of~\(X\).

\end{Definition}

The definition directly suggests a basic estimator for the
do-intervention.

\begin{Algorithm}

\textbf{Basic adjustment estimator.}

\begin{enumerate}
\def\labelenumi{\arabic{enumi}.}
\tightlist
\item
  Collect samples \(n\) samples \((t_i, y_i, x_i)_{i=1}^n.\)
\item
  Estimate each of the conditional probabilities
  \(\Pr[Y=y\mid T=t,X=x]\) from the collected samples.
\item
  Compute the sum \(\sum_x \Pr[Y=y\mid T=t, X=x] \Pr[X=x].\)
\end{enumerate}

\end{Algorithm}

This estimator can only work if all slices~\(\{T=t,X=x\}\) have nonzero
probability, an assumption often called \emph{overlap} or
\emph{positivity} in causal inference.\index{overlap}

But the basic estimator also fails when the adjustment variable~\(X\)
can take on too many possible values. In general, the variable~\(X\)
could correspond to a tuple of features, such as, age, height, weight,
etc. The support of~\(X\) grows exponentially with the number of
features. This poses an obvious computational problem, but more
importantly a statistical problem as well. By a counting argument some
of the events \(\{T=t,X=x\}\) must have probability as small as the
inverse of size of the support~\(X\). To estimate a probability~\(p>0\)
from samples to within small relative error, we need about~\(O(1/p^2)\)
samples.

Much work in causal inference deals with overcoming the statistical
inefficiency of the basic estimator. Conceptually, however, most
sophisticated estimators work from the same principle. We need to assume
that we have an admissible variable~\(X\) and that positivity holds.
Different estimators then use this assumption in different ways.

\hypertarget{potential-outcomes-and-ignorability}{%
\subsection{Potential outcomes and
ignorability}\label{potential-outcomes-and-ignorability}}

The average treatment effect often appears in the causal inference
literature equivalently in its potential outcome notation
\(\E[Y_1 - Y_0]\). This way of going about it is mathematically
equivalent and either way works for us.

When talking about potential outcomes, it's customary to replace the
assumption that~\(X\) is admissible with another essentially equivalent
assumption called \emph{ignorability} or \emph{unconfoundedness}. To
recall from the previous chapter, this assumption requires that the
potential outcomes variables are conditionally independent of treatment
given~\(X\). Formally,~\(T\bot (Y_0, Y_1)\mid X\). It's not hard to show
that ignorability implies that~\(X\) is
admissible.\index{unconfoundedness}\index{ignorability}

\hypertarget{reductions-to-model-fitting}{%
\section{Reductions to model
fitting}\label{reductions-to-model-fitting}}

Adjustment gives a simple and general way to estimate causal effects
given an admissible set of variables. The primary shortcoming that we
discussed is the sample inefficiency of the formula in high-dimensional
settings.

There's a vast literature of causal estimators that aim to address this
central shortcoming in a range of different settings. While the
landscape of causal estimators might seem daunting to newcomers, almost
all causal inference methods share a fundamental idea. This idea is
reduce causal inference to standard supervised machine learning tasks.

Let's see how this central idea plays out in a few important cases.

\hypertarget{propensity-scores}{%
\subsection{Propensity scores}\label{propensity-scores}}

Propensity scores are one popular way to cope with adjustment variables
that have large support. Let~\(T \in \{ 0, 1 \}\) be a binary treatment
variable. The quantity \[
e(x) = \E[T \mid X=x]
\] is known as the \emph{propensity score} and gives the likelihood of
treatment in the subpopulation defined by the condition
\(X=x\).\index{propensity score}

\begin{Theorem}

Suppose that~\(X\) is admissible, and the propensity scores are positive
\(e(x) \neq 0\) for all~\(X\). Then, \[
\begin{aligned}
\E[ Y \mid \doop(T := 1) ] = \E \left[ \frac{YT}{e(X)} \right]
\end{aligned}
\]

\end{Theorem}

\begin{Proof}

Applying the adjustment formula for a fixed~\(y\), we have \[
\begin{aligned}
\Pr[Y = y \mid \doop(T := 1)]
&= \sum_x \Pr[ Y = y \mid T = 1, X=x] \Pr[X=x]\\
&= \sum_x \frac{\Pr[ Y = y \mid T= 1, X=x] \Pr[X=x] \Pr[T = 1 \mid X=x]}{\Pr[T = 1 \mid X=x]}\\
&= \sum_x \frac{\Pr[Y = y, T = 1, X=x]}{\Pr[T = 1 \mid X=x]}\\
&= \sum_{x, t\in\{0, 1\}} \frac{t \Pr[Y = y,T = t,X=x] }{\Pr[T = 1 \mid X=x]}\,.
\end{aligned}
\] Here, we used that~\(e(x) = \Pr[T = 1 \mid X=x] \neq 0.\) Completing
the proof, \[
\begin{aligned}
\E[ Y \mid \doop(T := 1)]
&= \sum_y y \Pr[Y = y \mid \doop(T := 1)] \\
&= \sum_{y, x, t\in\{0, 1\}}   \frac{y t \Pr[ Y = y, T = t,  X=x ]}{\Pr[T = 1 \mid X=x]} 
= \E \left[ \frac{YT}{e(X)} \right]\,.\qedhere
\end{aligned}
\]

\end{Proof}

The same theorem also shows that \[
\E[Y \mid \doop(T := 0) ] = \E \left[ \frac{Y(1 - T)}{1 - e(X)} \right]
\] and thus the average treatment effect of~\(X\) on~\(Y\) is given by
\[
\E[ Y \mid \doop(T := 1)]
- \E[Y \mid \doop(T := 0)]
= \E \left[ Y\left(\frac{T}{e(X)}-\frac{1-T}{1-e(X)}\right) \right]\,.
\] This formula for the average treatment effect is called \emph{inverse
propensity score weighting}. Let's understand what it buys us compared
with the adjustment formula when working with a finite sample.

One way to approximate the expectation given the theorem above is to
collect many samples from which we estimate the propensity
score~\(e(x)\) separately for each possible setting~\(X\). However, this
way of going about it runs into the very same issues as the basic
estimator. Practitioners therefore choose a different route.

In a first step, we fit a model~\(\hat e\) to the propensity scores
hoping that our model~\(\hat e\) approximates the propensity score
function~\(e\) uniformly well. We approach this step as we would any
other machine learning problem. We create a dataset of
observations~\((x_i, e_i)\) where \(e_i\) is an empirical estimate
of~\(e(x_i)\) that we compute from our sample. We then fit a model to
these data points using our favorite statistical technique, be it
logistic regression or something more sophisticated.

In a second step, we then use our model's estimated propensity scores in
our sample estimate instead of the true propensity scores: \[
\frac{1}{n} \sum_{i=1}^n \frac{t_i y_i}{\hat e (x_i)}.
\] The appeal of this idea is that we can use the entire repertoire of
model fitting to get a good function approximation of the propensity
scores. Depending on what the features are we could use logistic
regression, kernel methods, random forests, or even deep models.
Effectively we're reducing the problem of causal inference to that of
model fitting, which we know how to do.

\hypertarget{double-machine-learning}{%
\subsection{Double machine learning}\label{double-machine-learning}}

Our previous reduction to model fitting has a notable shortcoming. The
propensity score estimate~\(\hat e(x_i)\) appears in the denominator of
our estimator. This has two consequences. First, unbiased estimates of
propensity scores do not imply an unbiased estimate of the causal
effect. Second, when propensity scores are small and samples aren't too
plentiful, this can lead to substantial variance.

There's a popular way to cope, called \emph{double machine learning},
that works in a partially linear structural causal
model:\index{double machine learning} \[
Y = \tau T + g(X) + U\,,
\qquad T = f(X) + V
\] In this model, the variable~\(X\) is an observed confounder between
treatment and outcome. We allow the functions~\(g\) and~\(f\) to be
arbitrary, but note that~\(g\) only depends on~\(X\) but not on~\(T\) as
it could in general. The random variables~\(U, V\) are independent
exogenous noise variables with mean~\(0\). In this model, the effect of
treatment on the outcome is linear and the coefficient~\(\tau\) is the
desired average treatment effect.

The trick behind double machine learning is to subtract~\(\E[Y\mid X]\)
from each side of the first equation and to use the fact that
\(\E[Y\mid X]=\tau\E[T\mid X] + g(X)\). We therefore get the equation \[
Y-\E[Y\mid X] = \tau (T-\E[T\mid X]) + U\,.
\] Denoting~\(\tilde Y = Y-\E[Y\mid X]\) and~\(\tilde T=T-\E[T\mid X]\)
we can see that the causal effect~\(\tau\) is the solution to the
regression problem~\(\tilde Y = \tau \tilde T+U\).

The idea now is to solve \emph{two} regression problems to find good
function approximations of the conditional expectations~\(\E[Y\mid X]\)
and \(\E[T\mid X]\), respectively. We can do this using data drawn from
the joint distribution of~\((X, T, Y)\) by solving two subsequent model
fitting problems, hence the name double machine learning.

Suppose then that we find two function approximations
\(q(X, Y) \approx \E[Y\mid X]\) and~\(r(X, T)\approx\E[T\mid X]\). We
can define the random variables~\(\hat Y = Y - q(X, Y)\) and
\(\hat T = T - r(X, T)\). The final step is to solve the regression
problem~\(\hat Y = \hat\tau\hat T + U\) for the parameter~\(\hat\tau\).

Compared with inverse propensity score weighting, we can see that finite
sample errors in estimating the conditional expectations have a more
benign effect on the causal effect estimate~\(\hat\tau\). In particular,
unbiased estimates of the conditional expectations lead to an unbiased
estimate of the causal effect.

\hypertarget{heterogeneous-treatment-effects}{%
\subsection{Heterogeneous treatment
effects}\label{heterogeneous-treatment-effects}}

In many applications, treatment effects can vary by subpopulation. In
such cases we may be interested in the \emph{conditional average
treatment effect} (CATE) in the subpopulation defined by
\(X=x\):\index{treatment effect!conditional average}\index{treatment effect!heterogeneous}
\[
\tau(x)=\E[Y \mid \doop(T:=1), X=x ] - \E[Y \mid \doop(T:=0), X=x]\,.
\] We're in luck, because the same proof we saw earlier shows that we
can estimate these so-called heterogeneous treatment effects with the
propensity score formula: \[
\tau(x) = \E\left[Y\left(\frac{T}{e(X)}-\frac{1-T}{1-e(X)}\right)\mid X=x\right]
\]

We can also extend double machine learning easily to the heterogeneous
case by replacing the coefficient~\(\tau\) in the first structural
equation with a function~\(\tau(X)\) that depends on~\(X\). The argument
remains the same except that in the end we need to solve the problem
\(\hat Y = \hat\tau(X) \hat T + Y\), which amounts to optimizing over a
function~\(\hat\tau\) in some model family rather than a constant
\(\hat\tau\).

Both inverse propensity score weighting and the double machine learning
can, in principle, estimate heterogeneous treatment effects. These
aren't the only reductions to model fitting, however. Another popular
method, called \emph{causal forests}, constructs decision trees whose
leaves correspond covariate settings that deconfound treatment and
outcome.\citep{wager-causal-forests}\index{causal!forest}

\hypertarget{quasi-experiments}{%
\section{Quasi-experiments}\label{quasi-experiments}}

The idea behind quasi-experimental designs is that sometimes processes
in nature or society are structured in a way that enables causal
inference. The three most widely used quasi-experimental designs are
\emph{regression discontinuities}, \emph{instrumental variables}, and
\emph{differences in differences}. We will review the first two briefly
to see where machine learning comes
in.\index{regression discontinuity}\index{instrumental variables}\index{differences in differences}\index{quasi-experiments}

\hypertarget{regression-discontinuity}{%
\subsection{Regression discontinuity}\label{regression-discontinuity}}

Many consequential interventions in society trigger when a certain score
\(R\) exceeds a threshold value~\(t\). The idea behind a regression
discontinuity design is that units that fall just below the threshold
are indistinguishable from units just above threshold. In other words,
whether or not a unit is just above or just below the threshold is a
matter of pure chance. We can then hope to identify a causal effect of
an intervention by comparing units just below and just above the
threshold.

\begin{figure}
\centering
\includegraphics{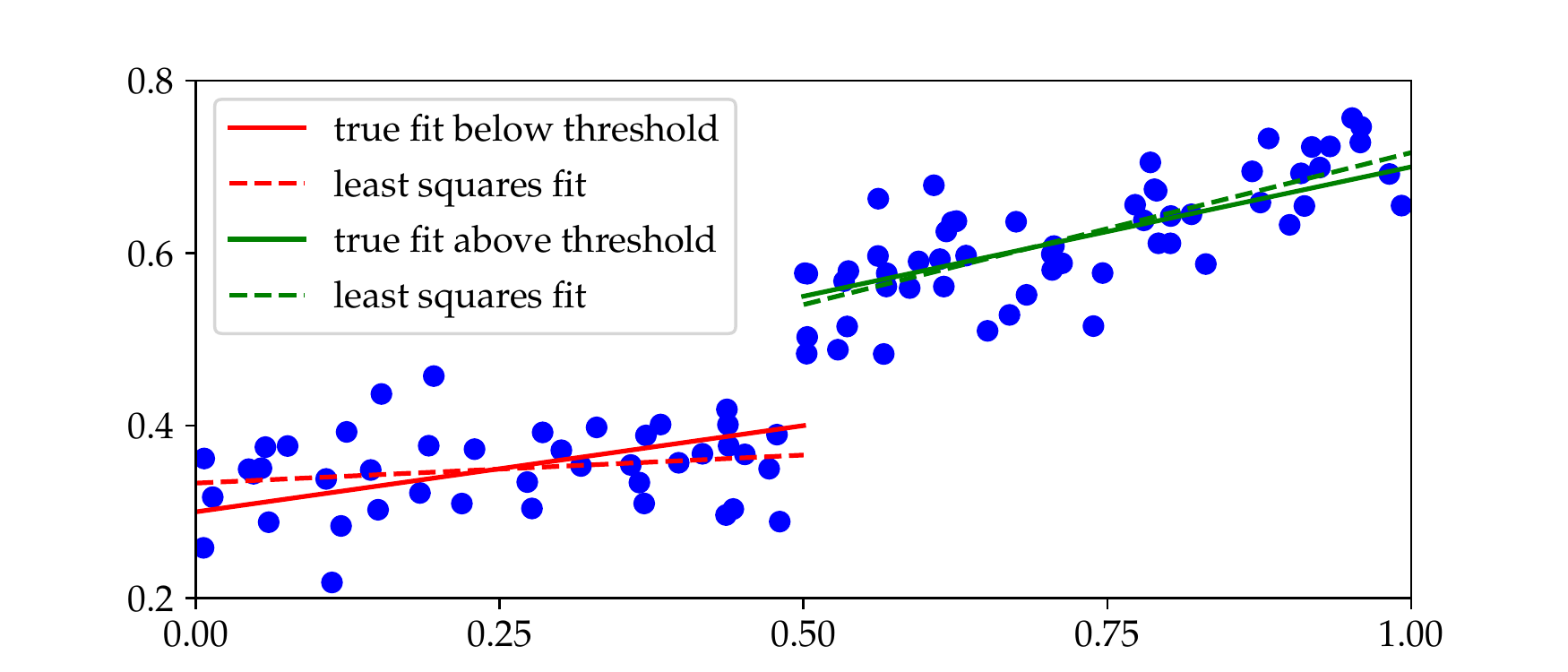}
\caption{Illustration of an idealized regression discontinuity. Real
examples are rarely this clear cut.}
\end{figure}

To illustrate the idea, consider an intervention in a hospital setting
that is assigned to newborn children just below a birth weight of 1500g.
We can ask if the intervention has a causal effect on wellbeing of the
child at a later age as reflected in an outcome variable, such as,
mortality or cumulative hospital cost in their first year. We expect
various factors to influence both birth weight and outcome variable. But
we hope that these confounding factors are essentially held constant
right around the threshold weight of 1500g. Regression discontinuity
designs have indeed been used to answer such questions for a number of
different outcome
variables.\citep{almond2010estimating, bharadwaj2013early}

Once we have identified the setup for a regression discontinuity, the
idea is to perform two regressions. One fits a model to the data below
the threshold. The other fits the model to data above the threshold. We
then take the difference of the values that the two models predict at
the threshold as our estimate of the causal effect. As usual, the idea
works out nicely in an idealized linear setting and can be generalized
in various ways.

There are numerous subtle and not so subtle ways a regression
discontinuity design can fail. One subtle failure mode is when
intervention incentivizes people to strategically make efforts to fall
just below or above the threshold. Manipulation or \emph{gaming} of the
running variable is a well-known issue for instance when it comes to
social program eligibility\citep{camacho2011manipulation}. But there are
other less obvious cases. For example, school class sizes in data from
Chile exhibit irregularities that void regression discontinuity
designs.\citep{urquiola2009class} In turn, researchers have come up with
tests designed to catch such problems.

\hypertarget{instrumental-variables}{%
\subsection{Instrumental variables}\label{instrumental-variables}}

Instrumental variables are a popular quasi-experimental method for
causal inference. The starting point is confounding between a treatment
\(T\) and our outcome of interest~\(Y\). We are in a situation where
we're unable to resolve confounding via the adjustment formula. However,
what we have is the existence of a special variable~\(Z\) called an
\emph{instrument} that will help us estimate the treatment effect.

What makes~\(Z\) a valid instrument is nicely illustrated with the
following causal graph.

\begin{figure}
\centering
\includegraphics[width=0.5\textwidth,height=\textheight]{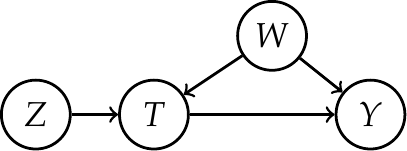}
\caption{Typical graphical model for an instrumental variable setup}
\end{figure}

The graph structure encodes two key assumptions:

\begin{enumerate}
\def\labelenumi{\arabic{enumi}.}
\tightlist
\item
  The instrument \(Z\) and the outcome \(Y\) are unconfounded.
\item
  The instrument \(Z\) has no direct effect on the outcome \(Y\).
\end{enumerate}

Let's walk through how this works out in the one-dimensional linear
structural equation for the outcome: \[
Y= \alpha + \beta T + \gamma W + N
\] Here,~\(N\) is an independent noise term. For convenience, we denote
the \emph{error term} \(U=\gamma W + N\). What we're interested in is
the coefficient~\(\beta\) since we can easily verify that it corresponds
to the average treatment effect: \[
\beta = \E[Y\mid\doop(T:=1)] - \E[Y\mid\doop(T:=0)]
\] To find the coefficient~\(\beta,\) we cannot directly solve the
regression problem~\(Y=\alpha + \beta T + U\), because the error
term~\(U\) is not independent of~\(T\) due to the confounding influence
of~\(W\).

However, there's a way forward after we make a few additional
assumptions:

\begin{enumerate}
\def\labelenumi{\arabic{enumi}.}
\tightlist
\item
  The error term is zero mean: \(\E[U]=0\)
\item
  The instrument is uncorrelated with the error term: \(\Cov(Z, U)=0\)
\item
  Instrument and treatment have nonzero correlation: \(\Cov(Z, T)\ne 0\)
\end{enumerate}

The first two assumptions directly imply \[
 \E[Y-\alpha-\beta T]= 0 \qquad\text{and}\qquad
 \E[Z(Y-\alpha-\beta T)] = 0\,.
\] This leaves us with two linear equations in~\(\alpha\) and~\(\beta\)
so that we can solve for both parameters. Indeed,
\(\alpha=\E[Y]-\beta\E[T]\). Plugging this into the second equation, we
have \[
\E[Z((Y-\E[Y])-\beta(T-\E[T]) )]=0,
\] which implies, via our third assumption~\(\Cov(T, Z)\ne 0,\) \[
\beta = \frac{\Cov(Z, Y)}{\Cov(T, Z)}\,.
\] There's a different intuitive way to derive this solution by solving
a \emph{two step least squares} procedure:

\begin{enumerate}
\def\labelenumi{\arabic{enumi}.}
\tightlist
\item
  Predict the treatment from the instrument via least squares
  regression, resulting in the predictor \(\hat T = cZ\).
\item
  Predict the outcome from the predicted treatment using least squares
  regression, resulting in the predictor \(\hat Y=\beta' \hat T\).
\end{enumerate}

A calculation reveals that indeed~\(\beta'=\beta\), the desired
treatment effect. To see this note that \[
c=\frac{\Cov(Z, T)}{\Var(Z)}\,
\] and hence \[
\beta'
= \frac{\Cov(Y, \hat T)}{\Var(\hat T)}
= \frac{\Cov(Y, Z)}{c\Var(Z)}
= \frac{\Cov(Z, Y)}{\Cov(T, Z)}
= \beta\,.
\] This solution directly generalizes to the multi-dimensional linear
case. The two stage regression approach is in fact the way instrumental
variables is often introduced operationally. We see that again
instrumental variables is a clever way of reducing causal inference to
prediction.

One impediment to instrumental variables is a poor correlation between
the instrument and the treatment. Such instruments are called \emph{weak
instruments}. In this case, the denominator~\(\Cov(T, Z)\) in our
expression for~\(\beta\) is small and the estimation problem is
ill-conditioned. The other impediment is that the causal graph
corresponding to instrumental variables is not necessarily easy to come
by in applications. What's delicate about the graph is that we want the
instrument to have a significant causal effect on the treatment, but at
the same time have no other causal powers that might influence the
outcome in a way that's not mediated by the treatment.

Nonetheless, researchers have found several intriguing applications of
instrumental variables. One famous example goes by the name \emph{judge
instruments}. The idea is that within the United States, at least in
certain jurisdictions and courts, defendants may be assigned randomly to
judges. Different judges then assign different sentences, some perhaps
more lenient, others harsher. The treatment here could be the sentence
length and the outcome may indicate whether or not the defendant went on
to commit another crime upon serving the prison sentence. A perfectly
random assignment of judges implies that the judge assignment and the
outcome are unconfounded. Moreover, the assignment of a judge has a
causal effect on the treatment, but plausibly no direct causal effect on
the outcome. The assignment of judges then serves as an instrumental
variable. The observation that judge assignments may be random has been
the basis of much causal inference about the criminal justice system.
However, the assumption of randomness in judge assignments has also been
challenged.\citep{chilton2015challenging}

\hypertarget{limitations-of-causal-inference-in-practice}{%
\section{Limitations of causal inference in
practice}\label{limitations-of-causal-inference-in-practice}}

It's worth making a distinction between causal modeling broadly speaking
and the practice of causal inference today. The previous chapter covered
the concepts of causal modeling. Structural causal models make it
painfully clear that the model necessarily specifies strong assumptions
about the data generating process. In contrast, the practice of causal
inference we covered in this chapter seems almost \emph{model-free} in
how it reduces to pattern classification via technical assumptions. This
appears to free the practitioner from difficult modeling choices.

The assumptions that make this all work, however, are not verifiable
from data. Some papers seek assurance in statistical robustness checks,
but these too are sample-based estimates. Traditional robustness checks,
such as resampling methods or leave-one-out estimates, may get at issues
of generalization, but cannot speak to the validity of causal
assumptions.

As a result, a certain pragmatic attitude has taken hold. If we cannot
verify the assumption from data anyway, we might as well make it in
order to move forward. But this is a problematic position. Qualitative
and theoretical ways of establishing substantive knowledge remain
relevant where the limitations of data set in. The validity of a causal
claim cannot be established solely based on a sample. Other sources of
substantive knowledge are required.

\hypertarget{validity-of-observational-methods}{%
\subsection{Validity of observational
methods}\label{validity-of-observational-methods}}

The empirical evidence regarding the validity of observational causal
inference studies is mixed and depends on the domain of application.

A well known article compared observational studies in the medical
domain between 1985 and 1998 to the results of randomized controlled
trials.\citep{benson2000comparison} The conclusion was good news for
observational methods:

\begin{quote}
We found little evidence that estimates of treatment effects in
observational studies reported after 1984 are either consistently larger
than or qualitatively different from those obtained in randomized,
controlled trials.
\end{quote}

Another study around the same time came to a similar conclusion:

\begin{quote}
The results of well-designed observational studies (with either a cohort
or a case--control design) do not systematically overestimate the
magnitude of the effects of treatment as compared with those in
randomized, controlled trials on the same
topic.\citep{concato2000randomized}
\end{quote}

One explanation, however, is that medical researchers may create
observational designs with great care on the basis of extensive domain
knowledge and prior investigation.

Freedman's paper \emph{Statistical Models and Shoe Leather} illustrates
this point through the famous example of Jon Snow's discovery from the
1850s that cholera is a waterborne
disease.\citep{freedman1991statistical} Many associate Snow with an
early use of quantitative methods. But the application of those followed
years of substantive investigation and theoretical considerations that
formed the basis of the quantitative analysis.

In other domains, observational methods have been much less successful.
Online advertising, for example, generates hundreds of billions of
dollars in yearly global revenue, but the causal effects of targeted
advertising remain a subject of debate.\citep{hwang2020subprime}
Randomized controlled trials in this domain are rare for technical and
cultural reasons. Advertising platforms are highly optimized toward a
particular way of serving ads that can make true randomization difficult
to implement. As a result, practitioners rely on a range of
observational methods to determine the causal effect of showing an ad.
However, these methods tend to perform poorly as a recent large-scale
study reveals:

\begin{quote}
The observational methods often fail to produce the same effects as the
randomized experiments, even after conditioning on extensive demographic
and behavioral variables. We also characterize the incremental
explanatory power our data would require to enable observational methods
to successfully measure advertising effects. Our findings suggest that
commonly used observational approaches based on the data usually
available in the industry often fail to accurately measure the true
effect of advertising.\citep{gordon2019comparison}
\end{quote}

\hypertarget{interference-interaction-and-spillovers}{%
\subsection{Interference, interaction, and
spillovers}\label{interference-interaction-and-spillovers}}

Confounding is not the only threat to the validity of causal studies. In
a medical setting, it's often relatively easy to ensure that treatment
of one subject does not influence the treatment assignment or outcome of
any other unit. We called this the Stable Unit Treatment Value
Assumption (SUTVA) in the previous chapter and noted that it holds by
default for the units in a structural causal models. Failures of SUTVA,
however, are common and go by many names, such as, interference,
interaction, and spill-over effects.

Take the example of an online social network. Interaction between units
is the default in all online platforms, whose entire purpose is that
people interact. Administering treatment to a subset of the platform's
users typically has some influence on the control group. For example, if
our treatment exposes a group of users to more content of a certain
kind, those users might share the content with others outside the
treatment group. In other words, treatment \emph{spills over} to the
control group. In certain cases, this problem can be mitigated by
assigning treatment to a cluster in the social network that has a
boundary with few outgoing edges thus limiting bias from
interaction.\citep{eckles2016design}

Interference is also common in the economic development context. To
borrow an example from economist John Roemer\citep{roemer19how}, suppose
we want to know if better fishing nets would improve the yield of
fishermen in a town. We design a field experiment in which we give
better fishing nets to a random sample of fishermen. The results show a
significantly improved yield for the treated fishermen. However, if we
scale the intervention to the entire population of fishermen, we might
cause overfishing and hence reduced yield for everyone.

\hypertarget{chapter-notes-9}{%
\section{Chapter notes}\label{chapter-notes-9}}

Aside from the introductory texts from the previous chapter, there are a
few more particularly relevant in the context of this chapter.

The textbook by Angrist and Pischke\citep{angrist2008mostly} covers
causal inference with an emphasis on regression analysis an applications
in econometrics. See Athey and Imbens\citep{athey2017state} for a more
recent survey of the state of causal inference in econometrics.

Marinescu et al.\citep{marinescu2018quasi} give a short introduction to
quasi-experiments and their applications to neuroscience with a focus on
regression discontinuity design, instrumental variables, and differences
in differences.

\chapter{Sequential decision making and dynamic programming}

As the previous chapters motivated we don't just make predictions for
their own sake, but rather use data to inform decision making and
action. This chapter examines sequential decisions and the interplay
between predictions and actions in settings where our repeated actions
are directed towards a concrete end-goal. It will force us to understand
statistical models that evolve over time and the nature of dependencies
in data that is temporally correlated. We will also have to understand
feedback and its impact on statistical decision-making problems.

In machine learning, the subfield of using statistical tools to direct
actions in dynamic environments is commonly called ``reinforcement
learning'' (RL). However, this blanket term tends to lead people towards
specific solution techniques. So we are going to try to maintain a
broader view of the area of \emph{sequential decision making}, including
perspectives from related fields of predictive analytics and optimal
control. These multiple perspectives will allow us to highlight how RL
is different from the machine learning we are most familiar
with.\index{reinforcement learning}\index{decision making!sequential}

This chapter will follow a similar flow to our study of prediction. We
will formalize a temporal mathematical model of sequential decision
making involving notions from \emph{dynamical systems}. We will then
present a common optimization framework for making sequential decisions
when models are known: \emph{dynamic programming}. Dynamic programming
will enable algorithms for finding or approximating optimal decisions
under a variety of scenarios. In the sequel, we will turn to the
learning problem of how to best make sequential decisions when the
mechanisms underlying dynamics and costs are not known in
advance.\index{dynamic programming}

\hypertarget{from-predictions-to-actions}{%
\section{From predictions to
actions}\label{from-predictions-to-actions}}

Let's first begin with a discussion of how sequential decision making
differs from static prediction. In our study of decision theory, we laid
out a framework for making optimal predictions of a binary
covariate~\(Y\) when we had access to data~\(X\), and probabilistic
models of how~\(X\) and \(Y\) were related. Supervised learning was the
resulting problem of making such decisions from data rather than
probabilistic models.

In sequential decision making, we add two new variables. First, we
incorporate \emph{actions} denoted~\(U\) that we aim to take throughout
a procedure. We also introduce rewards~\(R\) that we aim to maximize. In
sequential decision making, the goal is to analyze the data~\(X\) and
then subsequently choose~\(U\) so that~\(R\) is large. We have explicit
agency in choosing~\(U\) and are evaluated based on some quality scoring
of~\(U\) and \(X\). There are an endless number of problems where this
formulation is applied from supply chain optimization to robotic
planning to online engagement maximization. Reinforcement learning is
the resulting problem of taking actions so as to maximize rewards where
our actions are only a function of previously observed data rather than
probabilistic models. Not surprisingly, the optimization problem
associated with sequential decision making is more challenging than the
one that arises in decision theory.

\hypertarget{dynamical-systems}{%
\section{Dynamical systems}\label{dynamical-systems}}

\index{dynamical system}

In addition to the action variable in sequential decision making,
another key feature of sequential decision making problems is the notion
of time and sequence. We assume data is collected in an evolving
process, and our current actions influence our future rewards.

We begin by bringing all of these elements together in the general
definition of a discrete time dynamical system. The definitions both
simple and broad. We will illustrate with several examples shortly.

A \emph{dynamical system model} has a \emph{state} \(X_t\),
\emph{exogenous input} \(U_t\) modeling our \emph{control action}, and
\emph{reward} \(R_t\). The state evolves in discrete time steps
according to the equation \[
    X_{t+1} = f_t(X_t,U_t,W_t)
\] where~\(W_t\) is a random variable and~\(f_t\) is a function. The
reward is assumed to be a function of these variables as well: \[
    R_{t} = g_t(X_t,U_t,W_t)
\] for some function~\(g_t\). To simplify our notation throughout, we
will commonly write~\(R\) explicitly as a function of~\((X,U,W)\), that
is, \(R_t[X_t,U_t,W_t]\).

Formally, we can think of this definition as a structural equation model
that we also used to define causal models. After all, the equations
above give us a way to incrementally build up a data-generating process
from noise variables. Whether or not the dynamical system is intended to
capture any causal relationships in the real world is a matter of
choice. Practitioners might pursue this formalism for reward
maximization without modeling causal relationships. A good example is
the use of sequential decision making tools for revenue maximization in
targeted advertising. Rather than modeling causal relationships between,
say, preferences and clicks, targeted advertising heavily relies on all
sorts of signals, be they causal or not.

\hypertarget{concrete-examples}{%
\subsection{Concrete examples}\label{concrete-examples}}

\textbf{Grocery shopping.} Bob really likes to eat cheerios for
breakfast every morning. Let the state~\(X_t\) denotes the amount of
Cheerios in Bob's kitchen on day~\(t\). The action~\(U_t\) denotes the
amount of Cheerios Bob buys on day~\(t\) and~\(W_t\) denotes the amount
of Cheerios he eats that day. The random variable~\(W_t\) varies with
Bob's hunger. This yields the dynamical system \[
    X_{t+1} = X_t + U_t - W_t\,.
\] While this example is a bit cartoonish, it turns out that such simple
models are commonly used in managing large production supply chains. Any
system where resources are stochastically depleted and must be
replenished can be modeled comparably. If Bob had a rough model for how
much he eats in a given day, he could forecast when his supply would be
depleted. And he could then minimize the number of trips he'd need to
make to the grocer using optimal control.

\textbf{Moving objects.} Consider a physical model of a flying object.
The simplest model of the dynamics of this system are given by Newton's
laws of mechanics. Let~\(Z_t\) denote the position of the vehicle (this
is a three-dimensional vector). The derivative of position is velocity
\[
V_t = \frac{\partial Z_t}{\partial t}\,,
\] and the derivative of velocity is acceleration \[
A_t = \frac{\partial V_t}{\partial t}\,.
\] Now, we can approximate the rules in discrete time with the simple
Taylor approximations \[
\begin{aligned}
    Z_{t+1} &= Z_t + \Delta V_t\\
    V_{t+1} &= V_t + \Delta A_t
\end{aligned}
\] We also know by Newton's second law, that acceleration is equal to
the total applied force divided by the mass of the object:~\(F=mA\). The
flying object will be subject to external forces such as gravity and
wind~\(W_t\) and it will also receive forces from its
propellers~\(U_t\). Then we can add this to the equations to yield a
model \[
\begin{aligned}
    Z_{t+1} &= Z_t + \Delta V_t\\
    V_{t+1} &= V_t + \frac{\Delta}{m} (W_t+U_t)\,.
\end{aligned}
\]

Oftentimes, we only observe the acceleration through an accelerometer.
Then estimating the position and velocity becomes a filtering problem.
Optimal control problems for this model include flying the object along
a given trajectory or flying to a desired location in a minimal amount
of time.

\hypertarget{markov-decision-processes}{%
\subsection{Markov decision processes}\label{markov-decision-processes}}

\index{Markov decision process}

Our definition in terms of structural equations is not the only
dynamical model used in machine learning. Some people prefer to work
directly with probabilistic transition models and conditional
probabilities. In a \emph{Markov Decision Process}, we again have a
state \(X_t\) and input~\(U_t\), and they are linked by a probabilistic
model \[
    \Pr[X_{t+1}\mid X_t,U_t]\,.
\] This is effectively the same as the structural equation model above
except we hide the randomness in this probabilistic notation.

\textbf{Example: machine repair.} The following example illustrates the
elegance of the conditional probability models for dynamical systems.
This example comes from Bertsekas\citep{BertsekasDPBook}. Suppose we
have a machine with ten states of repair. State 10 denotes excellent
condition and 1 denotes the inability to function. Every time one uses
the machine in state~\(j\), it has a probability of falling into
disrepair, given by the probabilities~\(\Pr[X_{t+1} = i\mid X_t=j]\,,\)
where \(\Pr[X_{t+1}=i\mid X_t=j]=0\) if~\(i>j\). The action~\(a\) one
can take at any time is to repair the machine, resetting the system
state to~\(10\). Hence \[
\Pr[X_{t+1}=i\mid X_{t}=j,U_t=0] = \Pr[X_{t+1}=i\mid X_{t}=j]
\] and \[
\Pr[X_{t+1}=i\mid X_{t}=j,U_t=1] = \mathbb{1}\left\{ i=10 \right\} \,.
\] While we could write this dynamical system as a structural equation
model, it is more conveniently expressed by these probability tables.

\hypertarget{optimal-sequential-decision-making}{%
\section{Optimal sequential decision
making}\label{optimal-sequential-decision-making}}

Just as risk minimization was the main optimization problem we studied
in static decision theory, there is an abstract class of optimization
problems that underlie most sequential decision making (SDM) problems.
The main problem is to find a sequence of decision \emph{policies} that
maximize a cumulative reward subject to the uncertain, stochastic system
dynamics. At each time, we assign a reward~\(R_t(X_t,U_t,W_t)\) to the
current state-action pair. The goal is to find a sequence of actions to
make the summed reward as large as possible: \[
\begin{array}{ll}
\text{maximize}_{\{u_t\}} & \E_{W_t}\left[ \sum_{t=0}^T R_t(X_t,u_t,W_t) \right]\\
\text{subject to} & X_{t+1} = f_t(X_t, u_t, W_t)\\
& \text{($x_0$ given)}
\end{array}
\] Here, the expected value is over the sequence of stochastic
disturbance variables~\(W_t\). Note here,~\(W_t\) is a random variable
and \(X_t\) are is hence also a random variable. The sequence of actions
\(\{u_t\}\) is our decision variable. It could be chosen via a random or
deterministic procedure as a matter of design. But it is important to
understand what information is allowed to be used in order to select
\(u_t\).

Since the dynamics are stochastic, the optimal SDM problem typically
allows a policy to observe the state before deciding upon the next
action. This allows a decision strategy to continually mitigate
uncertainty through feedback. This is why we optimize over policies
rather than over a deterministic sequence of actions. That is, our goal
is to find functions of the current state~\(\pi_t\) such that
\(U_t=\pi_t(X_t, X_{t-1}, \ldots)\) is optimal in expected value. By a
\emph{control policy} (or simply ``a policy'') we mean a function that
takes a trajectory from a dynamical system and outputs a new control
action. In order for~\(\pi_t\) to be implementable, it must only have
access only to previous states and actions.

The policies~\(\pi_t\) are the decision variables of the problem: \[
\begin{array}{ll}
\text{maximize}_{\pi_t} & \E_{W_t}\left[ \sum_{t=0}^T R_t(X_t,U_t,W_t) \right]\\
\text{subject to} & X_{t+1} = f(X_t, U_t, W_t) \\
& U_t = \pi_t(X_t, X_{t-1},\ldots)\\
& \text{($x_0$ given)}
\end{array}
\] Now,~\(U_t\) is explicitly a random variable as it is a function of
the state~\(X_t\).

This SDM problem will be the core of what we study in this chapter. And
our study will follow a similar path to the one we took with decision
theory. We will first study how to solve these SDM problems when we know
the model. There is a general purpose solution for these problems known
as \emph{dynamic programming}.

\hypertarget{dynamic-programming}{%
\section{Dynamic programming}\label{dynamic-programming}}

The dynamic programming solution to the SDM problem is based on the
\emph{principle of optimality}: if you've found an optimal control
policy for a time horizon of length~\(T\),~\(\pi_1,\ldots, \pi_T\), and
you want to know the optimal strategy starting at state~\(x\) at
time~\(t\), then you just have to take the optimal policy starting at
time~\(t\), \(\pi_t,\ldots,\pi_T\). The best analogy for this is based
on driving directions: if you have mapped out an optimal route from
Seattle to Los Angeles, and this path goes through San Francisco, then
you must also have the optimal route from San Francisco to Los Angeles
as the tail end of your trip. Dynamic programming is built on this
principle, allowing us to recursively find an optimal policy by starting
at the final time and going backwards in time to solve for the earlier
stages.\index{dynamic programming}

To proceed, define the \emph{Q-function} to be the
mapping:\index{Q-function} \[
\begin{aligned}
\mathcal{Q}_{a\rightarrow b}(x,u) 
&= \max_{\{u_t\}}\E_{W_t}\left[ \sum_{t=a}^b R_t(X_t,u_t,W_t)\right] \\
& \mathrm{s.t.}\quad X_{t+1} = f_t(X_t, u_t, W_t),\quad(X_a,u_a)=(x,u)
\end{aligned}
\] The Q-function determines the best achievable value of the SDM
problem over times~\(a\) to~\(b\) when the action at time~\(a\) is set
to be \(u\) and the initial condition is~\(x\). It then follows that the
optimal value of the SDM problem is
\(\max_u \mathcal{Q}_{0\rightarrow T}(x_0,u)\), and the optimal policy
is \(\pi(x_0) = \arg \max_u \mathcal{Q}_{0\rightarrow T}(x_0,u)\). If we
had access to the Q-function for the horizon~\({0,T}\), then we'd have
everything we'd need to know to take the first step in the SDM problem.
Moreover, the optimal policy is only a function of the current state of
the system. Once we see the current state, we have all the information
we need to predict future states, and hence we can discard the previous
observations.

We can use dynamic programming to compute this Q-function and the
Q-function associated with every subsequent action. That is, clearly we
have that the terminal Q-function is \[
    \mathcal{Q}_{T\rightarrow T}(x,u) = \E_{W_T}\left[ R_T(x,u,W_T) \right]\,,
\] and then compute recursively \[
    \mathcal{Q}_{t\rightarrow T}(x,u) = \E_{W_t}\left[ R_t(x,u,W_t) +  \max_{u'} \mathcal{Q}_{t+1 \rightarrow T}(f_t(x,u,W_t),u')\right]\,.
\] This expression is known as Bellman's equation. We also have that for
all times~\(t\), the optimal policy is
\(u_t = \arg\max_u \mathcal{Q}_{t\rightarrow T} (x_t,u)\) and the policy
depends only on the current state.

To derive this form of the Q-function, we assume inductively that this
form is true for all times beyond~\(t+1\) and then have the chain of
identities \[
\begin{aligned}
    \mathcal{Q}_{t\rightarrow T}(x,u) &= \max_{\pi_{t+1},\ldots,\pi_T} \E_{w}\left[R_t(x,u,W_t) +  \sum_{s=t+1}^T R_s(X_s,\pi_{s}(X_s),W_s)\right]\\
&=  \E_{W_t}\left[R_t(x,u,W_t) +  
\max_{\pi_{t+1},\ldots,\pi_T}\E_{W_{t+1},\ldots,W_T} \left\{
\sum_{s=t+1}^T R_s(X_s,\pi_{s}(X_s),W_s) \right\}\right]\\
 &=\E_{W_t}\left[R_t(x,u,W_t) +     \max_{\pi_{t+1}} Q\left\{f(x,u,W_t),\pi_{t+1}(f(x,u,W_t))\right\}\right]\\
 &= \E_{W_t}\left[R_t(x,u,W_t) +    \max_{u'} Q(f(x,u,W_t),u')\right]\,.
\end{aligned}
\] Here, the most important point is that the maximum can be exchanged
with the expectation with respect to the first~\(W_t\). This is because
the policies are allowed to make decisions based on the history of
observed states, and these states are deterministic functions of the
noise process.

\hypertarget{infinite-time-horizons-and-stationary-policies}{%
\subsection{Infinite time horizons and stationary
policies}\label{infinite-time-horizons-and-stationary-policies}}

The Q-functions we derived for these finite time horizons are \emph{time
varying}. One applies a different policy for each step in time. However,
on long horizons with time invariant dynamics and costs, we can get a
simpler formula. First, for example, consider the limit: \[
\begin{array}{ll}
\text{maximize} & \lim_{N\rightarrow \infty}  \E_{W_t}[ \frac{1}{N} \sum_{t=0}^N R(X_t,U_t,W_t) ]\\
\text{subject to} & X_{t+1} = f(X_t, U_t, W_t),~U_t=\pi_t(X_t)\\
& \text{($x_0$ given).}
\end{array}
\] Such infinite time horizon problems are referred to as \emph{average
cost} dynamic programs. Note that there are no subscripts on the rewards
or transition functions in this model.

Average cost dynamic programming is deceptively difficult. These
formulations are not directly amenable to standard dynamic programming
techniques except in cases with special structure. A considerably
simpler infinite time formulation is known as \emph{discounted} dynamic
programming, and this is the most popular studied formulation.
Discounting is a mathematical convenience that dramatically simplifies
algorithms and analysis. Consider the SDM problem \[
    \begin{array}{ll}
        \text{maximize} &  (1-\gamma) \E_{W_t}[ \sum_{t=0}^\infty \gamma^t R(X_t,U_t,W_t) ]\\
        \text{subject to} & X_{t+1} = f(X_t, U_t, W_t),~U_t=\pi_t(X_t)\\
        & \text{($x_0$ given).}
    \end{array}
\] where~\(\gamma\) is a scalar in~\((0,1)\) called the \emph{discount
factor}.\index{discount factor} For~\(\gamma\) close to~\(1\), the
discounted reward is approximately equal to the average reward. However,
unlike the average cost model, the discounted cost has particularly
clean optimality conditions. If we define~\(\mathcal{Q}_\gamma(x,u)\) to
be the Q-function obtained from solving the discounted problem with
initial condition~\(x\), then we have a discounted version of dynamic
programming, now with the same Q-functions on the left and right hand
sides: \[
    \mathcal{Q}_\gamma(x,u) = \E_{W} \left[ R(x,u,W) + \gamma  \max_{u'} \mathcal{Q}_\gamma(f(x,u,W),u')\right]\,.
\] The optimal policy is now for \emph{all times} to let \[
    u_t= \arg\max_u \mathcal{Q}_\gamma(x_t,u)\,.
\] The policy is time invariant and one can execute it without any
knowledge of the reward or dynamics functions. At every stage, one
simply has to maximize a function to find the optimal action.
Foreshadowing to the next chapter, the formula additionally suggest that
the amount that needs to be ``learned'' in order to ``control'' is not
very large for these infinite time horizon problems.

\hypertarget{computation}{%
\section{Computation}\label{computation}}

Though dynamic programming is a beautiful universal solution to the very
general SDM problem, the generality also suggests computational
barriers. Dynamic programming is only efficiently solvable for special
cases, and we now describe a few important examples.

\hypertarget{tabular-mdps}{%
\subsection{Tabular MDPs}\label{tabular-mdps}}

Tabular MDPs refer to Markov Decision Processes with small number of
states and actions. Say that there are~\(S\) states and~\(A\) actions.
Then the the transition rules are given by tables of conditional
probabilities~\(\Pr[X_{t+1}|X_t,U_t]\), and the size of such tables are
\(S^2 A\). The Q-functions for the tabular case are also tables, each of
size~\(SA\), enumerating the cost-to-go for all possible state action
pairs.\index{tabular} In this case, the maximization \[
 \max_{u'}\mathcal{Q}_{a\rightarrow b}(x,u')
\] corresponds to looking through all of the actions and choosing the
largest entry in the table. Hence, in the case that the rewards are
deterministic functions of~\(x\) and~\(u\), Bellman's equation
simplifies to \[
    \mathcal{Q}_{t\rightarrow T}(x,u) = R_t(x,u) + \sum_{x'} \Pr[X_{t+1}=x'|X_t=x,U_t=u] \max_{u'} \mathcal{Q}_{t+1\rightarrow T}(x',u')\,.
\] This function can be computed by elementary matrix-vector operations:
the Q-functions are~\(S \times A\) arrays of numbers. The ``max''
operation can be performed by operating over each row in such an array.
The summation with respect to~\(x'\) can be implemented by multiplying a
\(SA \times S\) array by an~\(S\)-dimensional vector. We complete the
calculation by summing the resulting expression with the~\(S\times A\)
array of rewards. Hence, the total time to compute
\(\mathcal{Q}_{t\rightarrow T}\) is~\(O(S^2A)\).

\hypertarget{linear-quadratic-regulator}{%
\subsection{Linear quadratic
regulator}\label{linear-quadratic-regulator}}

The other important problem where dynamic programming is efficiently
solvable is the case when the dynamics are \emph{linear} and the rewards
are \emph{quadratic}. In control design, this class of problems is
generally referred to as the problem of the Linear Quadratic Regulator
(LQR):\index{linear quadratic regulator}\index{LQR} \[
\begin{array}{ll}
\text{minimize} \, & \E_{W_t} \left[\frac{1}{2}\sum_{t=0}^T X_t^T \Phi_t X_t + U_t^T \Psi_t U_t\right], \\
\text{subject to} & X_{t+1} = A_t X_t+ B_t U_t + W_t,~U_t=\pi_t(X_t) \\
& \text{($x_0$ given).}
\end{array}
\] Here,~\(\Phi_t\) and~\(\Psi_t\) are most commonly positive
semidefinite matrices.~\(w_t\) is noise with zero mean and bounded
variance, and we assume~\(W_t\) and~\(W_{t'}\) are independent
when~\(t\neq t'\). The state transitions are governed by a linear update
rule with~\(A_t\) and~\(B_t\) appropriately sized matrices. We also
abide by the common convention in control textbooks to pose the problem
as a minimization---not maximization---problem.

As we have seen above, many systems can be modeled by linear dynamics in
the real world. However, we haven't yet discussed cost functions. It's
important to emphasize here that cost functions are \emph{designed} not
given. Recall back to supervised learning: though we wanted to minimize
the number of errors made on out-of-sample data, on in-sample data we
minimized convex surrogate problems. The situation is exactly the same
in this more complex world of dynamical decision making. Cost functions
are designed by the engineer so that the SDM problems are tractable but
also so that the desired outcomes are achieved. Cost function design is
part of the toolkit for online decision making, and quadratic costs can
often yield surprisingly good performance for complex problems.

Quadratic costs are also attractive for computational reasons. They are
convex as long as~\(\Phi_t\) and~\(\Psi_t\) are positive definite.
Quadratic functions are closed under minimization, maximization,
addition. And for zero mean noise~\(W_t\) with covariance~\(\Sigma\), we
know that the noise interacts nicely with the cost function. That is, we
have \[
    \mathbb{E}_W[(x+W)^T M (x+W)] = x^T M x + \operatorname{Tr}(Q\Sigma)
\] for any vector~\(x\) and matrix~\(M\). Hence, when we run dynamic
programming, every Q-function is necessarily quadratic. Moreover, since
the Q-functions are quadratic, the optimal action is a \emph{linear
function} of the state \[
    U_t = -K_t X_t
\] for some matrix~\(K_t\).

Now consider the case where there are static costs~\(\Phi_t=\Phi\) and
\(\Psi_t=\Psi\), and time invariant dynamics such that~\(A_t=A\)
and~\(B_t=B\) for all~\(t\). One can check that the Q-function on a
finite time horizon satisfies a recursion \[
 \mathcal{Q}_{t \rightarrow T}(x,u) = x^T \Phi x + u^T \Psi u + (Ax+Bu)^T M_{t+1} (Ax+Bu) + c_t\,.
\] for some positive definite matrix~\(M_{t+1}\). In the limit as the
time horizon tends to infinity, the optimal control \emph{policy} is
\emph{static, linear state feedback}: \[
    u_t = -K x_t\,.
\] Here the matrix~\(K\) is defined by \[
    K=(\Psi + B^T M B)^{-1} B^T M A
\] and~\(M\) is a solution to the \emph{Discrete Algebraic Riccati
Equation}\index{Riccati Equation} \[
M = \Phi + A^T M A - (A^T M B)(\Psi + B^T M B)^{-1} (B^T M A)\,.
\] Here,~\(M\) is the unique solution of the Riccati equation where all
of the eigenvalues of~\(A-BK\) have magnitude less than~\(1\). Finding
this specific solution is relatively easy using standard linear
algebraic techniques. It is also the limit of the Q-functions computed
above.

\hypertarget{policy-and-value-iteration}{%
\subsection{Policy and value
iteration}\label{policy-and-value-iteration}}

Two of the most well studied methods for solving such discounted
infinite time horizon problems are \emph{value iteration} and
\emph{policy iteration}.\index{policy iteration}\index{value iteration}
Value iteration proceeds by the steps \[
    \mathcal{Q}_{k+1} (x,u) = \E_{W} \left[ R(x,u,W) + \gamma  \max_{u'} \mathcal{Q}_k (f(x,u,W),u')\right]\,.
\] That is, it simply tries to solve the Bellman equation by running a
fixed point operation. This method succeeds when the iteration is a
contraction mapping, and this occurs in many contexts.

On the other hand, Policy Iteration is a two step procedure:
\emph{policy evaluation} followed by \emph{policy improvement}. Given a
policy~\(\pi_k\), the policy evaluation step is given by \[
    \mathcal{Q}_{k+1} (x,u) = \E \left[  R(x,u,W) + \gamma \mathcal{Q}_k (f(x,\pi_k(x),W),\pi_k(x))\right]\,.
\] And then the policy is updated by the rule \[
    \pi_{k+1}(x) = \arg\max_u \mathcal{Q}_{k+1} (x,u)\,.
\] Often times, several steps of policy evaluation are performed before
updating the policy.

For both policy and value iteration, we need to be able to compute
expectations efficiently and must be able to update \emph{all values}
of~\(x\) and~\(u\) in the associated~\(Q\) functions. This is certainly
doable for tabular MDPs. For general low dimensional problems, policy
iteration and value iteration can be approximated by gridding state
space, and then treating the problem as a tabular one. Then, the
resulting~\(Q\) function can be extended to other~\((x,u)\) pairs by
interpolation. There are also special cases where the maxima and minima
yield closed form solutions and hence these iterations reduce to simpler
forms. LQR is a canonical example of such a situation.

\hypertarget{model-predictive-control}{%
\subsection{Model predictive control}\label{model-predictive-control}}

\index{model predictive control}

If the Q-functions in value or policy iteration converge quickly,
long-term planning might not be necessary, and we can effectively solve
infinite horizon problem with short-term planning. This is the key idea
behind one of the most powerful techniques for efficiently and
effectively finding quality policies for SDM problems called \emph{model
predictive control}.

Suppose that we aim to solve the infinite horizon average reward
problem: \[
\begin{array}{ll}
\text{maximize} & \lim_{T\rightarrow \infty} \E_{W_t}[\frac{1}{T}\sum_{t=0}^T R_t(W_t,U_t) ]\\
\text{subject to} & X_{t+1} = f_t(X_t, U_t, W_t)\\
& U_t = \pi_t(X_t)\\
& \text{($x_0$ given).}
\end{array}
\] Model Predictive Control computes an \emph{open loop} policy on a
finite horizon~\(H\) \[
\begin{array}{ll}
\text{maximize}_{u_t} & \E_{W_t}[ \sum_{t=0}^H R_t(X_t,u_t)]\\
\text{subject to} & X_{t+1} = f_t(X_t, u_t, W_t)\\
& \text{($X_0$ = x).}
\end{array}
\] This gives a sequence~\(u_0(x),\ldots, u_H(x)\). The policy is then
set to be~\(\pi(x)=u_0(x)\). After this policy is executed, we observe a
new state,~\(x'\), based on the dynamics. We then recompute the
optimization, now using~\(x_0=x'\) and setting the action to
be~\(\pi(x')= u_0(x')\).

MPC is a rather intuitive decision strategy. The main idea is to plan
out a sequence of actions for a given horizon, taking into account as
much uncertainty as possible. But rather than executing the entire
sequence, we play the first action and then gain information from the
environment about the noise. This direct feedback influences the next
planning stage. For this reason, model predictive control is often a
successful control policy even when implemented with inaccurate or
approximate models. Model Predictive Control also allows us to easily
add a variety of constraints to our plan at little cost, such as bounds
on the magnitude of the actions. We just append these to the
optimization formulation and then lean on the computational solver to
make us a plan with these constraints.

To concretely see how Model Predictive Control can be effective, it's
helpful to work through an example. Let's suppose the dynamics and
rewards are time invariant. Let's suppose further that the reward
function is bounded above, and there is some state-action pair
\((x_\star,u_\star)\) which achieves this maximal
reward~\(R_\mathrm{max}\).

Suppose we solve the finite time horizon problem where we enforce that
\((x,u)\) must be at~\((x_\star,u_\star)\) at the end of the time
horizon: \[
\begin{array}{ll}
\text{maximize} & \E_{W_t}[ \sum_{t=0}^H R(X_t,u_t)]\\
\text{subject to} & X_{t+1} = f(X_t, u_t, W_t)\\
& (X_H,U_H) = (x_\star,u_\star)\\
& \text{($X_0$ = x).}
\end{array}
\] We replan at every time step by solving this optimization problem and
taking the first action.

The following proposition summarizes how this policy performs

\begin{Proposition}

Assume that all rewards are bounded above by~\(R_{\mathrm{max}}\). Then
with the above MPC policy, we have for all~\(T\) that \[
 \mathbb{E}\left[ \frac{1}{T} \sum_{t=0}^T R(x_t,u_t)\right] \geq \frac{Q_{0\rightarrow H}(x_0,u_0)-H R_{\mathrm{max}}}{T}
 + \mathbb{E}_W[R(f(x_\star,u_\star,W),0)]\,.
\]

\end{Proposition}

The proposition asserts that there is a ``burn in'' cost associated with
the initial horizon length. This term goes to zero with~\(T\), but will
have different values for different~\(H\). The policy converges to a
residual average cost due to the stochasticity of the problem and the
fact that we try to force the system to the state~\((x_\star,u_\star)\).

\begin{Proof}

To analyze how the policy performs, we turn to Bellman's equation. For
any time~\(t\), the MPC policy is \[
    u_t = \arg\max_u \mathcal{Q}_{0\rightarrow H}(x_t,u)
\] Now, \[
    \mathcal{Q}_{0\rightarrow H}(x_t,u) = R(x_t,u) + \mathbb{E}[\max_{u'} \mathcal{Q}_{1\rightarrow H}(X_{t+1},u')]\,.
\]

Now consider what to do at the time~\(t+1\). A \emph{suboptimal}
strategy at this time is to try to play the optimal strategy on the
horizon \(1\rightarrow H\), and then do nothing on the last step. That
is, \[
\max_u \mathcal{Q}_{0\rightarrow H}(x_{t+1},u) \geq
                 \max_{u'} \mathcal{Q}_{1\rightarrow H}(x_{t+1},u') + \mathbb{E}[R(f(x_\star,u_\star,W_{t+H}),0)]\,.
\] The last expression follows because the action sequence from
\(1\rightarrow H\) enforces~\((x_{t+H},u_{t+H})=(x_\star,u_\star)\). The
first term on the right hand side was computed in expectation above,
hence we have \[
\mathbb{E}[\max_u \mathcal{Q}_{0\rightarrow H}(X_{t+1},u)] \geq
    \mathbb{E}[\mathcal{Q}_{0\rightarrow H}(x_t,u_t)] - \mathbb{E}[R(x_t,u_t,W)] + \mathbb{E}[R(f(x_\star,u_\star,W),0)] \,.
\] Unwinding this recursion, we find \[
\begin{aligned}
\mathbb{E}[\max_u \mathcal{Q}_{0\rightarrow H}(X_{T+1},u)] &\geq
    \mathcal{Q}_{0\rightarrow H}(x_0,u_0) - \mathbb{E}\left[\sum_{t=0}^T R(x_t,u_t,W_t)\right]\\
    &\qquad\qquad\qquad  + T \mathbb{E}[R(f(x_\star,u_\star,W),0)] \,.
    \end{aligned}
\] Since the rewards are bounded above, we can upper bound the left hand
side by~\(R_{\mathrm{max}} H\). Rearranging terms then proves the
theorem.

\end{Proof}

The main caveat with this argument is that there may not \emph{exist} a
policy that drives~\(x\) to~\(x_\star\) from an arbitrary initial
condition and any realization of the disturbance signal. Much of the
analysis of MPC schemes is devoted to guaranteeing that the problems are
\emph{recursively feasible}, meaning that such constraints can be met
for all time.

This example also shows how it is often helpful to have some sort of
recourse at the end of the planning horizon to mitigate the possibility
of being too greedy and driving the system into a bad state. The
terminal condition of forcing~\(x_H=0\) adds an element of safety to the
planning, and ensures stable execution for all time. More general,
adding some terminal condition to the planning horizon
\(\mathcal{C}(x_{H})\) is part of good Model Predictive Control design
and is often a powerful way to balance performance and robustness.

\hypertarget{partial-observation-and-the-separation-heuristic}{%
\section{Partial observation and the separation
heuristic}\label{partial-observation-and-the-separation-heuristic}}

Let's now move to the situation where instead of observing the state
directly, we observe an output~\(Y_t\): \[
    Y_t = h_t(X_t,U_t,W_t)\,.
\] All of the policies we derived from optimization formulations above
required feeding back a function of the state. When we can only act on
outputs, SDM problems are considerably more difficult.

\begin{enumerate}
\def\labelenumi{\arabic{enumi}.}
\item
  \emph{Static Output Feedback is NP-hard.} Consider the case of just
  building a static policy from the output~\(Y_t\). Let's suppose our
  model is just the simple linear model: \[
  \begin{aligned}
   X_{t+1} &= A X_t + B U_t\\
   Y_t &= C X_t
  \end{aligned}
  \] Here,~\(A\),~\(B\) and~\(C\) are matrices. Suppose we want to find
  a feedback policy~\(U_t = K Y_t\) ( where~\(K\) is a matrix) and all
  we want to guarantee is that for any initial~\(x_0\), the system state
  converges to zero. This problem is called \emph{static state feedback}
  and is surprisingly \emph{NP-Hard}. It turns out that the problem is
  equivalent to finding a matrix~\(K\) such that~\(A+BKC\) has all of
  its eigenvalues inside the unit circle in the complex
  plane.\citep{blondel2000survey} Though in the MDP case, static state
  feedback was not only optimal, but computable for tabular MDPs and
  certain other SDM problems, static output feedback is computationally
  intractable.
\item
  \textbf{POMDPs are PSPACE hard.} Papadimitriou and Tsitsiklis showed
  that optimization of general POMDPs, even on small state spaces, was
  in all likelihood completely
  intractable.\citep{papadimitriou1987complexity} They reduced the
  problem of \emph{quantifier elimination} in logical satisfiability
  problems (QSAT) to POMDPs. QSAT seeks to determine the validity of
  statements like ``there exists \(x\) such that for all \(y\) there
  exists \(z\) such that for all \(w\) this logical formula is true.''
  Optimal action in POMDPs essentially have to keep track of all of the
  possible true states that might have been visited given the partial
  observation and make actions accordingly. Hence, the policies have a
  similar flavor to quantifier elimination as they seek actions that are
  beneficial to all possible occurrences of the unobserved variables.
  Since these policies act over long time horizons, the number of
  counterfactuals that must be maintained grows exponentially large.
\end{enumerate}

Despite these challenges, engineers solve POMDP problems all of the
time. Just because the problems are hard in general, doesn't mean they
are intractable on average. It only means that we cannot expect to have
general purpose optimal algorithms for these problems. Fortunately,
suboptimal solutions are oftentimes quite good for practice, and there
are many useful heuristics for decision making with partial information.
The most common approach to the output feedback problem is the following
two-stage strategy:

\begin{enumerate}
\def\labelenumi{\arabic{enumi}.}
\item
  \textbf{Filtering.} Using all of your past data~\(\{y_s\}\) for
  \(s=0,\ldots,t\), build an estimate,~\(\hat{x}_t\), of your state.
\item
  \textbf{Action based on certainty equivalence.} Solve the desired SDM
  problem as if you had perfect observation of the state~\(X_t\), using
  \(\hat{x}_t\) wherever you would have used an observation~\(X_t=x_t\).
  At run time, plug in the estimator~\(\hat{x}_t\) as if it were a
  perfect measurement of the state.
\end{enumerate}

This strategy uses a \emph{separation principle} between prediction and
action. For certain problems, this two-staged approach is actually
optimal. Notably, if the SDM problem has quadratic rewards/costs, if the
dynamics are linear, and if the noise process is Gaussian, then the
separation between prediction and action is optimal. More commonly, the
separation heuristic is suboptimal, but this abstraction also enables a
simple heuristic that is easy to debug and simple to
design.\index{separation principle}

While we have already covered algorithms for optimal control, we have
not yet discussed state estimation. Estimating the state of a dynamical
system receives the special name \emph{filtering}.\index{filtering}
However, at the end of the day, filtering is a prediction problem.
Define the observed data up to time~\(t\) as \[
\tau_t := (y_t,\ldots, y_1, u_{t-1}, \ldots, u_1)\,.
\] The goal of filtering is to estimate a a function~\(h(\tau_t)\) that
predicts~\(X_t\). We now describe two approaches to filtering.

\hypertarget{optimal-filtering}{%
\subsection{Optimal filtering}\label{optimal-filtering}}

Given a model of the state transition function and the observation
function, we can attempt to compute the maximum \emph{a posteriori}
estimate of the state from the data. That is, we could
compute~\(p(x_t | \tau_t)\) and then estimate the mode of this density.
Here, we show that such an estimator has a relatively simple recursive
formula, though it is not always computationally tractable to compute
this formula.

To proceed, we first need a calculation that takes advantage of the
conditional independence structure of our dynamical system model. Note
that \[
\begin{aligned}
p(y_t, x_t, & x_{t-1}, u_{t-1} | \tau_{t-1} )
= \\
&p(y_t| x_t, x_{t-1}, u_{t-1}, \tau_{t-1} )
p(x_t | x_{t-1},u_{t-1}, \tau_{t-1} )\\
 &\qquad\qquad\qquad\qquad\times p(x_{t-1}|\tau_{t-1} )
p(u_{t-1}|\tau_{t-1} )\\
&= p(y_t| x_t )
p(x_t | x_{t-1},u_{t-1} )
p(x_{t-1}|\tau_{t-1} )
p(u_{t-1}|\tau_{t-1} )\,.
\end{aligned}
\] This decomposition is into terms we now recognize.
\(p(x_t | x_{t-1},u_{t-1} )\) and~\(p(y_t| x_t )\) define the POMDP
model and are known.~\(p(u_t|\tau_t)\) is our policy and it's what we're
trying to design. The only unknown here is~\(p(x_{t-1}|\tau_{t-1} )\),
but this expression gives us a recursive formula
to~\(p(x_{t}|\tau_{t} )\) for all \(t\).

To derive this formula, we apply Bayes rule and then use the above
calculation: \[
\begin{aligned}
p(x_t | \tau_t) &=\frac{\int_{x_{t-1}} p(x_t , y_t, x_{t-1}, u_{t-1} | \tau_{t-1}) }
                                                {\int_{x_t,x_{t-1}} p(x_t , y_t, x_{t-1}, u_{t-1}| \tau_{t-1})}\\
&=\frac{\int_{x_{t-1}} p(y_t| x_t ) p(x_t | x_{t-1},u_{t-1} )
p(x_{t-1}|\tau_{t-1} )
p(u_{t-1}|\tau_{t-1} ) }
{\int_{x_{t},x_{t-1}} p(y_t| x_t ) p(x_t | x_{t-1},u_{t-1} )
p(x_{t-1}|\tau_{t-1} )
p(u_{t-1}|\tau_{t-1} )}\\
&=\frac{\int_{x_{t-1}} p(y_t| x_t ) p(x_t | x_{t-1},u_{t-1} )
p(x_{t-1}|\tau_{t-1} ) }
{\int_{x_t,x_{t-1}} p(y_t| x_t ) p(x_t| x_{t-1},u_{t-1} )
p(x_{t-1}|\tau_{t-1} )}\,.\qquad (4)
\end{aligned}
\] Given a prior for~\(x_0\), this now gives us a formula to compute a
MAP estimate of~\(x_t\) for all~\(t\), incorporating data in a streaming
fashion. For tabular POMDP models with small state spaces, this formula
can be computed simply by summing up the conditional probabilities. In
POMDPs without inputs---also known as \emph{hidden Markov models}---this
formula gives the forward pass of Viterbi's decoding algorithm. For
models where the dynamics are linear and the noise is Gaussian, these
formulas reduce into an elegant closed form solution known as
\emph{Kalman filtering}. In general, this optimal filtering algorithm is
called \emph{belief propagation} and is the basis of a variety of
algorithmic techniques in the field of graphical
models.\index{Kalman filtering}\index{belief propagation}

\hypertarget{kalman-filtering}{%
\subsection{Kalman filtering}\label{kalman-filtering}}

For the case of linear dynamical systems, the above calculation has a
simple closed form solution that looks similar to the solution of LQR.
This estimator is called a \emph{Kalman filter}, and is one of the most
important tools in signal processing and estimation. The Kalman filter
assumes a linear dynamical system driven by Gaussian noise with
observations corrupted by Gaussian noise \[
\begin{aligned}
 X_{t+1} &= A X_t+ B U_t + W_t\\
 Y_{t} & = C X_t + V_t\,.
\end{aligned}
\] Here, assume~\(W_t\) and~\(V_t\) are independent for all time and
Gaussian with means zero and covariances~\(\Sigma_W\) and~\(\Sigma_V\)
respectively. Because of the joint Gaussianity, we can compute a closed
form formula for the density of~\(X_t\) conditioned on the past
observations and actions,~\(p(x_t | \tau_t)\). Indeed,~\(X_t\) is a
Gaussian itself.

On an infinite time horizon, the Kalman filter takes a simple and
elucidating form: \[
\begin{aligned}
    \hat{x}_{t+1} &= A \hat{x}_t + B u_t - L (y_t - \hat{y}_t)\\
    \hat{y}_t &= C \hat{x}_t
\end{aligned}
\] where \[
    L=A P C^T (C P C^T + \Sigma_V)^{-1}
\] and~\(P\) is the positive semidefinite solution to the discrete
algebraic Riccati equation \[
    P = A P A^T + \Sigma_W - (A P C^T) (CPC^T +\Sigma_V)^{-1} (C\Sigma A^T)\,.
\]

The derivation of this form follows from the calculation in Equation
\(4\). But the explanation of the formulas tend to be more insightful
than the derivation. Imagine the case where~\(L=0\). Then our
estimate~\(\hat{x}_t\) simulates the same dynamics as our model with no
noise corruption. The matrix~\(L\) computes a correction for this
simulation based on the observed~\(y_t\). This feedback correction is
chosen in such a way such that~\(P\) is the steady state covariance of
the error~\(X_t-\hat{x}_t\).~\(P\) ends up being the minimum variance
possible with an estimator that is \emph{unbiased} in the sense that
\(\E[\hat{x}_t-X_t]=0\).

Another interesting property of this calculation is that the~\(L\)
matrix is the LQR gain associated with the LQR problem \[
\begin{array}{ll}
\text{minimize} \, & \lim_{T\rightarrow \infty}\E_{W_t} \left[\frac{1}{2}\sum_{t=0}^T X_t^T \Sigma_W X_t + U_t^T \Sigma_V U_t\right], \\
\text{subject to} & X_{t+1} = A^T X_t+ B^T U_t + W_t,~U_t=\pi_t(X_t) \\
& \text{($x_0$ given).}
\end{array}
\] Control theorists often refer to this pairing as the \emph{duality
between estimation and control}.

\hypertarget{feedforward-prediction}{%
\subsection{Feedforward prediction}\label{feedforward-prediction}}

While the optimal filter can be computed in simple cases, we often do
not have simple computational means to compute the optimal state
estimate. That said, the problem of state estimation is necessarily one
of prediction, and the first half of this course gave us a general
strategy for building such estimators from data. Given many simulations
or experimental measurements of our system, we can try to estimate a
function~\(h\) such that~\(X_t \approx h(\tau_t)\). To make this
concrete, we can look at \emph{time lags} of the history \[
\tau_{t-s\rightarrow t} := (y_t,\ldots, y_{t-s}, u_{t-1}, \ldots, u_{t-s})\,.
\] Such time lags are necessarily all of the same length. Then
estimating \[
\begin{array}{ll}
\text{minimize}_h & \sum_t \operatorname{loss}(h(\tau_{t-s\rightarrow t}), x_t)
\end{array}
\] is a supervised learning problem, and standard tools can be applied
to design architectures for and estimate~\(h\).

\hypertarget{chapter-notes-10}{%
\section{Chapter notes}\label{chapter-notes-10}}

This chapter and the following chapter overlap significantly with a
survey of reinforcement learning by Recht\citep{RechtRLSurvey}, which
contains additional connections to continuous control. Those interested
in learning more about continuous control from an optimization viewpoint
should consult the book by Borrelli et al.\citep{BorrelliMPCBook} This
book also provides an excellent introduction to model predictive
control. Another excellent introduction to continuous optimal control
and filtering is Boyd's lecture notes.\citep{BoydOCNotes}

An invaluable introduction to the subject of dynamic programming is by
Bertsekas, who has done pioneering research in this space and has
written some of the most widely read texts.\citep{BertsekasDPBook} For
readers interested in a mathematical introduction to dynamic programming
on discrete processes, we recommend Puterman's text.\citep{PutermanBook}
Puterman also explains the linear programming formulation of dynamic
programming.

\chapter{Reinforcement learning}

Dynamic programming and its approximations studied thus far all require
knowledge of the probabilistic mechanisms underlying how data and
rewards change over time. When these mechanisms are unknown, appropriate
techniques to probe and learn about the underlying dynamics must be
employed in order to find optimal actions. We shall refer to the
solutions to sequential decision making problems when the dynamics are
unknown as \emph{reinforcement learning.}\index{reinforcement learning}
Depending on context, the term may refer to a body of work in artificial
intelligence, the community of researchers and practitioners who apply a
certain set of tools to sequential decision making, and data-driven
dynamic programming. That said, it is a useful name to collect a set of
problems of broad interest to machine learning, control theory, and
robotics communities.

A particularly simple and effective strategy for reinforcement learning
problems is to estimate a predictive model for the dynamical system and
then to use the fit model as if it were the true model in the optimal
control problem. This is an application of the \emph{principle of
certainty equivalence}, an idea tracing back to the dynamic programming
literature of the 1950s.\citep{Simon56, Theil57} Certainty equivalence
is a general solution strategy for the following problem. Suppose you
want to solve some optimization problem with a parameter~\(\vartheta\)
that is unknown. However, suppose we can gather data to
estimate~\(\vartheta\). Then the certainty equivalent solution is to use
a point estimate for~\(\vartheta\) as if it were the true value. That
is, you act as if you were certain of the value of~\(\vartheta\), even
though you have only estimated \(\vartheta\) from data. We will see
throughout this chapter that such certainty equivalent solutions are
powerfully simple and effective baseline for sequential decision making
in the absence of well specified models.\index{certainty equivalence}

Certainty equivalence is a very general principle. We can apply it to
the output of a filtering scheme that predicts state, as we described in
our discussion of partially observed Markov Decision Processes. We can
also apply this principle in the study of MDPs with unknown parameters.
For every problem in this chapter, our core baseline will always be the
certainty equivalent solution. Surprisingly, we will see that certainty
equivalent baselines are typically quite competitive and give a clear
illustration of the best quality one can expect in many reinforcement
learning problems.

\hypertarget{exploration-exploitation-tradeoffs-regret-and-pac-learning}{%
\section{Exploration-exploitation tradeoffs: Regret and PAC
learning}\label{exploration-exploitation-tradeoffs-regret-and-pac-learning}}

In order to compare different approaches to reinforcement learning, we
need to decide on some appropriate rules of comparison. Though there are
a variety of important metrics for engineering practice that must be
considered including ease of implementation and robustness, a
first-order statistical comparison might ask how many samples are needed
to achieve a policy with high reward.

For this question, there are two predominant conventions to compare
methods: \emph{PAC-error} and \emph{regret}. PAC is a shorthand for
\emph{probably approximately correct}. It is a useful notion when we
spend all of our time learning about a system, and then want to know how
suboptimal our solution will be when built from the data gathered thus
far. Regret is more geared towards online execution where we evaluate
the reward accrued at all time steps, even if we are spending that time
probing the system to learn about its dynamics. Our focus in this
chapter will be showing that these two concepts are closely
related.\index{PAC}\index{probably approximately correct}\index{regret}

Let us formalize the two notions. As in the previous chapter, we will be
concerned with sequential decision making problems of the form \[
\begin{array}{ll}
\text{maximize}_{\pi_t} & \E_{W_t}\left[ \sum_{t=0}^T R_t(X_t,U_t,W_t) \right]\\
\text{subject to} & X_{t+1} = f(X_t, U_t, W_t) \\
& U_t = \pi_t(X_t, X_{t-1},\ldots)\\
& \text{($x_0$ given.)}
\end{array}
\] Let~\(\pi_\star\) denote the optimal policy of this problem.

For PAC, let's suppose we allocate~\(N\) samples to probe the system and
use them in some way to build a policy~\(\pi_N\). We can define the
optimization error of this policy to be \[
    \mathcal{E}(\pi_N)=\E\left[\sum_{t=1}^T R_t[X'_t,\pi_\star(X'_t),W_t]\right]-\E\left[\sum_{t=1}^T R_t[X_t,\pi_N(X_t),W_t]\right]\,.
\] Our model has~\((\delta,\epsilon)\)-PAC error if
\(\mathcal{E}(\pi_N)\leq \epsilon\) with probability at
least~\(1-\delta\). The probability here is measured with respect to the
sampling process and dynamics of the system.

Regret is defined similarly, but is subtly different. Suppose we are now
only allowed~\(T\) total actions and we want to understand the
cumulative award achieved after applying these~\(T\) actions. In this
case, we have to balance the number of inputs we use to find a good
policy (exploration) against the number of inputs used to achieve the
best reward (exploitation).

Formally, suppose we use a policy~\(\pi_t\) at each time step to choose
our action. Suppose~\(\pi_\star\) is some other fixed policy.
Let~\(X_t\) denote the states induced by the policy sequence~\(\pi_t\)
and~\(X_t'\) denote the states induced by~\(\pi_\star\). Then the
\emph{regret} of \(\{\pi_t\}\) is defined to be \[
    \mathcal{R}_T(\{\pi_t\}) = \E\left[\sum_{t=1}^T R_t[X'_t,\pi_\star(X'_t),W_t]\right]-\E\left[\sum_{t=1}^T R_t[X_t,\pi_t(X_t),W_t]\right]\,.
\] It is simply the expected difference in the rewards generated under
policy~\(\pi_\star\) as compared to those generated under policy
sequence \(\pi_t\). One major way that regret differs from PAC-error is
the policy can change with each time step.\index{regret}

One note of caution for both of these metrics is that they are comparing
to a policy~\(\pi_\star\). It's possible that the comparison policy
\(\pi_\star\) is not very good. So we can have small regret and still
not have a particularly useful solution to the SDM problem. As a
designer it's imperative to understand~\(\pi_\star\) to formalize the
best possible outcome with perfect information. That said, regret and
PAC-error are valuable ways to quantify how much exploration is
necessary to find nearly optimal policies. Moreover, both notions have
provided successful frameworks for algorithm development: many
algorithms with low regret or PAC-error are indeed powerful in practice.

\hypertarget{multi-armed-bandits}{%
\subsection{Multi-armed bandits}\label{multi-armed-bandits}}

The multi-armed bandit is one of the simplest reinforcement learning
problems, and studying this particular problem provides many insights
into exploration-exploitation
tradeoffs.\index{multi-armed bandits}\index{bandits}

In the multi-armed bandit, we assume \emph{no state whatsoever}. There
are \(K\) total actions, and the reward is a random function of which
action you choose. We can model this by saying there are i.i.d. random
variables~\(W_{t1},\ldots, W_{tk}\), and your reward is the dot product
\[
    R_t = [W_{t1},\ldots, W_{tK}] e_{u_t}
\] where~\(e_i\) is a standard basis vector. Here~\(W_{ti}\) take values
in the range~\([0,1]\). We assume that all of the~\(W_{ti}\) are
independent, and that~\(W_{ti}\) and~\(W_{si}\) are identically
distributed. Let \(\mu_{i} = \E[W_{ti}]\). Then the expected reward at
time~\(t\) is precisely~\(\mu_{u_t}\).

The multi-armed bandit problem is inspired by gambling on slot machines.
Indeed, a ``bandit'' is a colloquial name for a slot machine. Assume
that you have~\(K\) slot machines. Each machine has some probability of
paying out when you play it. You want to find the machine that has the
largest probability of paying out, and then play that machine for the
rest of time. The reader should take an opportunity to ponder the irony
that much of our understanding of statistical decision making comes from
gambling.

First let's understand what the optimal policy is if we know the model.
The total reward is equal to \[
    \E_{W_t}\left[ \sum_{t=0}^T R_t(u_t,W_t) \right] = \sum_{t=1}^T \mu_{u_t}\,.
\] The optimal policy is hence to choose a constant action~\(u_{t}=k\)
where~\(k = \arg\max_i \mu_i\).

When we don't know the model, it makes sense that our goal is to quickly
find the action corresponding to the largest mean. Let's first do a
simple PAC analysis, and then turn to the slightly more complicated
regret analysis. Our simple baseline is one of certainty equivalence. We
will try each action~\(N/K\) times, and compute the empirical return.
The empirical means are: \[
    \hat{\mu}_k = \frac{K}{N} \sum_{i=1}^{N/K} R_i^{(k)}
\] Our policy will be to take the action with the highest observed
empirical return.

To estimate the value of this policy, let's assume that the best action
is~\(u=1\). Then define \[
    \Delta_i  = \mu_1 -\mu_i\,.
\] Then we have \[
    \mathcal{E}(\pi_N) = \sum_{i=1}^K T \Delta_i \Pr[\forall i\colon \hat{\mu}_k \geq \hat{\mu}_i]\,.
\]

We can bound the probability that action~\(k\) is selected as follows.
First, if action~\(k\) has the largest empirical mean, it must have a
larger empirical mean than the true best option, action 1: \[
\Pr[\forall i\colon \hat{\mu}_k \geq \hat{\mu}_i]
\leq \Pr[ \hat{\mu}_k \geq \hat{\mu}_1]\,.
\] We can bound this last term using Hoeffding's inequality. Let
\(m=N/K\). Since each reward corresponds to an independent draw of some
random process, we have~\(\hat{\mu}_k - \hat{\mu}_1\) is the mean
of~\(2m\) independent random variables in the range~\([-1,1]\): \[
    \frac{1}{2} (\hat{\mu}_k - \hat{\mu}_1) =  \frac{1}{2m} \left( \sum_{i=1}^m R_i^{(k)} +\sum_{i=1}^m -R_i^{(1)} \right)\,.
\] Now writing Hoeffding's inequality for this random variable gives the
tail bound \[
\Pr[ \hat{\mu}_k \geq \hat{\mu}_1]
=\Pr[ \tfrac{1}{2}( \hat{\mu}_k- \hat{\mu}_1) \geq 0]
\leq \exp\left(-\frac{m\Delta_k^2}{4}\right)
\] which results in an optimization error \[
    \mathcal{E}(\pi_N) \leq \sum_{i=1}^K T \Delta_i \exp\left(-\frac{N\Delta_i^2}{4K}\right)\,.
\] with probability~\(1\). This expression reveals that the multi-armed
bandit problem is fairly simple. If~\(\Delta_i\) are all small, then any
action will yield about the same reward. But if all of the~\(\Delta_i\)
are large, then finding the optimal action only takes a few samples.
Naively, without knowing anything about the gaps at all, we can use the
fact that~\(x e^{-x^2/2}\leq \tfrac{1}{2}\) for nonnegative~\(x\) to
find \[
    \mathcal{E}(\pi_N) \leq \frac{K^{3/2} T}{\sqrt{N}}\,.
\] This shows that no matter what the gaps are, as long as~\(N\) is
larger than~\(K^3\), we would expect to have a high quality solution.

Let's now turn to analyzing regret of a simple certainty equivalence
baseline. Given a time horizon~\(T\), we can spend the first~\(m\) time
steps searching for the best return. Then we can choose this action for
the remaining~\(T-m\) time steps. This strategy is called
\emph{explore-then-commit}.\index{explore-then-commit}

The analysis of the explore-then-commit strategy for the multi-armed
bandit is a straightforward extension of the PAC analysis. If at round
\(t\), we apply action~\(k\), the expected gap between our policy and
the optimal policy is~\(\Delta_k\). So if we let~\(T_k\) denote the
number of times action~\(k\) is chosen by our policy then we must have
\[
    \mathcal{R}_T = \sum_{k=1}^K \E[T_k]\Delta_k\,.
\] \(T_k\) are necessarily random variables: what the policy learns
about the different means will depend on the observed sequence~\(x_k\)
which are all random variables.

Suppose that for exploration, we mimic our offline procedure, trying
each action~\(m\) times and record the observed rewards for that action
\(r_i^{(k)}\) for~\(i=1,\ldots,m\). At the end of these~\(mk\) actions,
we compute the empirical mean associated with each action as before.
Then we must have that \[
    \E[T_k] = m + (T-mK)\Pr[\forall i\colon \hat{\mu}_k \geq \hat{\mu}_i]\,.
\] The first term just states that each action is performed~\(m\) times.
The second term states that action~\(k\) is chosen for the commit phase
only if its empirical mean is larger than all of the other empirical
means.

Using Hoeffding's inequality again to bound these probabilities, we can
put everything together bound the expected regret as \[
 \mathcal{R}_T \leq \sum_{k=1}^K m\Delta_k + (T-mK) \Delta_k \exp\left(-\frac{m\Delta_k^2}{4}\right) \,.
\]

Let's specialize to the case of \emph{two} actions to see what we can
take away from this decomposition:

\begin{enumerate}
\def\labelenumi{\arabic{enumi}.}
\item
  \textbf{Gap dependent regret.} First, assume we know the gap between
  the means,~\(\Delta_2\), but we don't know which action leads to the
  higher mean. Suppose that \[
   m_0 = \left\lceil \frac{4}{\Delta_2^2} \log \left(\frac{T \Delta_2^2}{4} \right) \right\rceil \geq 1\,.
  \] Then using~\(m=m_0\), we have \[
  \begin{aligned}
   \mathcal{R}_T &\leq m\Delta_2 + T \Delta_2 \exp\left(-\frac{m\Delta_2^2}{4}\right)\\
   &\leq \Delta_2+ \frac{4}{\Delta_2} \left(\log \left(\frac{T \Delta_2^2}{4}\right)+1\right)\,.
  \end{aligned}
  \] If~\(m_0<1\), then~\(\Delta_2 <\frac{2}{\sqrt{T}}\), then choosing
  a random arm yields total expected regret at most \[
   \mathcal{R}_T = \frac{T}{2} \Delta_2 \leq \sqrt{T}\,.
  \] If~\(\Delta_2\) is very small, then we might also just favor the
  bound \[
   \mathcal{R}_T \leq \tfrac{1}{2}\Delta_2 T\,.
  \] Each of these bounds applies in different regimes and tells us
  different properties of this algorithm. The first bound shows that
  with appropriate choice of~\(m\), explore-then-commit incurs regret
  asymptotically bounded by~\(\log(T)\). This is effectively the
  smallest asymptotic growth achievable and is the gold standard for
  regret algorithms. However, this logarithmic regret bound depends on
  the gap~\(\Delta_2\). For small~\(\Delta_2\), the second bound shows
  the regret is never worse than~\(\sqrt{T}\) for any value of the gap.
  \(\sqrt{T}\) is one of the more common values for regret, and though
  it is technically asymptotically worse than logarithmic, algorithms
  with~\(\sqrt{T}\) regret tend to be more stable and robust than their
  logarithmic counterparts. Finally, we note that a very naive algorithm
  will incur regret that grows linearly with the horizon \(T\). Though
  linear regret is not typically an ideal situation, there are many
  applications where it's acceptable. If~\(\Delta_2\) is tiny to the
  point where it is hard to observe the difference between \(\mu_1\)
  and~\(\mu_2\), then linear regret might be completely satisfactory for
  an application.
\item
  \textbf{Gap independent regret.} We can get a gap
  independent,~\(\sqrt{T}\) regret for explore then commit for any value
  of~\(\Delta_2\). This just requires a bit of calculus: \[
  \begin{aligned}
  \frac{4}{\Delta_2} \left(\log \left(\frac{T \Delta^2}{4}\right)+1\right)
  & = 2\sqrt{T} \left( \frac{2}{\Delta_2\sqrt{T}} \left(\log \left(\frac{T \Delta^2}{4}\right)+1\right)   \right)\\
  & = 2\sqrt{T} \sup_{x\geq 0} \frac{2\log(x)+1}{x} \leq 4 e^{-1/2} \sqrt{T} \leq 2.5 \sqrt{T} \,.
  \end{aligned}
  \] Hence, \[
  \mathcal{R}_T \leq \Delta_2 + 2.5 \sqrt{T}
  \] no matter the size of~\(\Delta_2\) the gap is. Often times this
  unconditional bound is smaller than the logarithmic bound we derived
  above.
\item
  \textbf{Gap independent policy.} The stopping rule we described thus
  far requires knowing the value of~\(\Delta_2\). However, if we set
  \(m=T^{2/3}\) then we can achieve sublinear regret no matter what the
  value of~\(\Delta_2\) is. To see this again just requires some
  calculus: \[
  \begin{aligned}
  \mathcal{R}_T &\leq T^{2/3}\Delta_2 + T \Delta_2 \exp\left(-\frac{T^{2/3}\Delta_2^2}{4}\right)\\
  &= T^{2/3} \left(   \Delta_2 + T^{1/3} \Delta_2 \exp\left(-\frac{T^{2/3}\Delta_2^2}{4}\right)   \right)\\
  &\leq T^{2/3} \left(   \Delta_2 + 2 \sup_{x\geq 0} x e^{-x^2} \right) \leq 2 T^{2/3}\,.
  \end{aligned}
  \] \(O(T^{2/3})\) regret is technically ``worse'' than an asymptotic
  regret of~\(O(T^{1/2})\), but often times such algorithms perform well
  in practice. This is because there is a difference between worst case
  and average case behavior, and hence these worst-case bounds on regret
  themselves do not tell the whole story. A practitioner has to weigh
  the circumstances of their application to decide what sorts of
  worst-case scenarios are acceptable.
\end{enumerate}

\hypertarget{interleaving-exploration-and-exploitation}{%
\subsection{Interleaving exploration and
exploitation}\label{interleaving-exploration-and-exploitation}}

Explore-then-commit is remarkably simple, and illustrates most of the
phenomena associated with regret minimization. There are essentially two
main shortcomings in the case of the multi-armed bandit:

\begin{enumerate}
\def\labelenumi{\arabic{enumi}.}
\item
  For a variety of practical concerns, it would be preferable to
  interleave exploration with exploitation.
\item
  If you don't know the gap, you only get a~\(T^{2/3}\) rate.
\end{enumerate}

A way to fix this is called \emph{successive elimination}. As in
explore-then-commit, we try all actions~\(m\) times. Then, we drop the
actions that are clearly performing poorly. We then try the remaining
actions~\(4m\) times, and drop the poorly performing actions. We run
repeated cycles of this pruning procedure, yielding a collection of
better actions on average, aiming at convergence to the best return.
\index{successive elimination}

\begin{Algorithm}

\textbf{Successive Elimination Algorithm:}

\begin{itemize}
\tightlist
\item
  Given number of rounds \(B\) and an increasing sequence of positive
  integers \(\{m_\ell\}\).
\item
  Initialize the active set of options \(\mathcal{A}=\{1,\ldots,K\}\).
\item
  For \(\ell=1,\ldots, B\):

  \begin{enumerate}
  \def\labelenumi{\arabic{enumi}.}
  \tightlist
  \item
    Try every action in \(\mathcal{A}\) for \(m_\ell\) times.
  \item
    Compute the empirical means \(\hat{\mu}_k\) from this iteration
    only.
  \item
    Remove from \(\mathcal{A}\) any action \(j\) with
    \(\mu_j + 2^{-\ell} < \max_{k \in \mathcal{A}} \mu_k\).
  \end{enumerate}
\end{itemize}

\end{Algorithm}

The following theorem bounds the regret of successive elimination, and
was proven by Auer and Ortner.\citep{AuerOrtner10}

\begin{Theorem}

With~\(B=\lfloor \tfrac{1}{2} \log_2 \tfrac{T}{e} \rfloor\) and
\(m_\ell = \lceil 2^{2\ell+1} \log \tfrac{T} {4^{\ell}} \rceil\), the
successive elimination algorithm accrues expected regret \[
\mathcal{R}_T \leq \sum_{i\colon\Delta_i > \lambda} \left( \Delta_i + \frac{32 \log(T\Delta_i^2) + 96}{\Delta_i} \right) + \max_{i\colon\Delta_i \leq \lambda} \Delta_i T
\] for any~\(\lambda>\sqrt{e/T}\).

\end{Theorem}

Another popular strategy is known as \emph{optimism in the face of
uncertainty}.\index{optimism} This strategy is also often called ``bet
on the best.'' At iteration~\(t\), take all of the observations seen so
far and form a set up upper confidence bounds~\(B_i\) such that \[
\Pr[\forall i\colon \mu_i \leq  B_i(t)] \leq 1-\delta
\] This leads to the Upper Confidence Bound (UCB)
algorithm.\index{upper confidence bound}\index{UCB}

\begin{Algorithm}

\textbf{UCB Algorithm}

\begin{itemize}
\tightlist
\item
  For \(t=1,\ldots, T\):

  \begin{enumerate}
  \def\labelenumi{\arabic{enumi}.}
  \tightlist
  \item
    Choose action \(k = \arg \max_i B_i(t-1)\).
  \item
    Play action \(k\) and observe reward \(r_t\).
  \item
    Update the confidence bounds.
  \end{enumerate}
\end{itemize}

\end{Algorithm}

For the simple case of the multi-armed bandit, we can use the bound that
would directly come from Hoeffding's inequality: \[
        B_i(t) = \hat{\mu}_i(t) + \sqrt{\frac{2 \log (1/\delta)}{T_i(t)}}
\] where we remind the reader that~\(T_i(t)\) denotes the number of
times we have tried action~\(i\) up to round~\(t\). Though~\(T_i(t)\) is
a random variable, one can still prove that this choice yields an
algorithm with nearly optimal regret.

More generally, optimistic algorithms work by maintaining an uncertainty
set about the dynamics model underlying the SDM problem. The idea is to
maintain a set~\(S\) where we have confidence our true model lies. The
algorithm then proceeds by choosing the model in~\(S\) which gives the
highest expected reward. The idea here is that either we get the right
model in which case we get a large reward, or we learn quickly that we
have a suboptimal model and we remove it from our set~\(S\).

\hypertarget{contextual-bandits}{%
\subsection{Contextual bandits}\label{contextual-bandits}}

\index{contextual bandits}

Contextual bandits provide a transition from multi-armed bandits to
full-fledged reinforcement learning, introducing \emph{state} or
\emph{context} into the decision problem. Our goal in contextual bandits
is to iteratively update a policy to maximize the total reward: \[
    \text{maximize}_{u_t}~\E_{W_t}\left[\sum_{t=1}^T R(X_t,u_t,W_t)\right]
\] Here, we choose actions~\(u_t\) according to some policy that is a
function of the observations of the random variables~\(X_t\), which are
called \emph{contexts} or \emph{states}. We make no assumptions about
how contexts evolve over time. We assume that the reward function is
unknown and, at every time step, the received reward is given by \[
R(X_t,u_t,W_t)=R(X_t,u_t)+W_t
\] where~\(W_t\) is a random variable with zero mean and independent
from all other variables in the problem.

Contextual bandits are a convenient way to abstractly model engagement
problems on the internet. In this example, contexts correspond to
information about a person. Every interaction a person has with the
website can be scored in term of some sort of reward function that
encodes outcomes such as whether the person clicked on an ad, liked an
article, or purchased an item. Whatever the reward function is, the goal
will be to maximize the total reward accumulated over all time. The
\(X_t\) will be features describing the person's interaction history,
and the action will be related to the content served.

As was the case with the multi-armed bandit, the key idea in solving
contextual bandits is to reduce the problem to a prediction problem. In
fact we can upper bound our regret by our errors in prediction. The
regret accrued by a policy~\(\pi\) is \[
    \mathbb{E}\left\{\sum_{t=1}^T  \max_{u} R(X_t,u) - R(X_t, \pi(X_t)) \right\}\,.
\] This is because if we know the reward function, then the optimal
strategy is to choose the action that maximizes~\(R\). This is
equivalent to the dynamic programming solution when the dynamics are
trivial.

Let's reduce this problem to one of prediction. Suppose that at
time~\(t\) we have built an approximation of the reward function~\(R\)
that we denote \(\hat{R}_t(x,u)\). Let's suppose that our algorithm uses
the policy \[
    \pi(X_t) = \arg \max_u \hat{R}_t(X_t,u)\,.
\] That is, we take our current estimate as if it was the true reward
function, and pick the action that maximizes reward given the context
\(X_t\).

To bound the regret for such an algorithm, note that we have for any
action~\(u\) \[
\begin{aligned}
    0 &\leq \hat{R}_t(X_t,\pi(X_t))-\hat{R}_t(X_t, u)\\
     &\leq R(X_t, \pi(X_t))-R(X_t, u) \\
     &+ \left[\hat{R}_t(X_t,\pi(X_t))
     - R(X_t, \pi(X_t))\right] + \left[\hat{R}_t(X_t,u) - R(X_t, u )\right]\,.   
\end{aligned}
\] Hence, \[
    \sum_{t=1}^T  \max_{u} R(X_t,u) - R(X_t, \pi(X_t))
    \leq 2 \sum_{t=1}^T \max_u | \hat{R}_t(X_t, u)-R(X_t, u)|\,.
\] This final inequality shows that if the prediction error goes to
zero, the associated algorithm accrues sublinear regret.

While there are a variety of algorithms for contextual bandits, we focus
our attention on two simple solutions that leverage the above reduction
to prediction. These algorithms work well in practice and are by far the
most commonly implemented. Indeed, they are so common that most
applications don't even call these implementations of contextual bandit
problems, as they take the bandit nature completely for granted.

Our regret bound naturally suggests the following explore-then-commit
procedure.\index{explore-then-commit}

\begin{Algorithm}

\textbf{Explore-then-commit for contextual bandits}

\begin{itemize}
\tightlist
\item
  For \(t=1,2,\ldots, m\):

  \begin{enumerate}
  \def\labelenumi{\arabic{enumi}.}
  \tightlist
  \item
    Receive new context \(x_t\).
  \item
    Choose a random action \(u_t\).
  \item
    Receive reward \(r_t\).
  \end{enumerate}
\item
  Find a function \(\hat{R}_m\) to minimize prediction error:
  \(\hat{R}_m := \arg\min_f \sum_{s=1}^{m} \loss(f(x_s,u_s),r_s)\,.\)
\item
  Define the policy \(\pi(x) = \arg\max_u \hat{R}_m(x_t,u)\).
\item
  For \(t=m+1,m+2,\ldots\):

  \begin{enumerate}
  \def\labelenumi{\arabic{enumi}.}
  \tightlist
  \item
    Receive new context \(x_t\).
  \item
    Choose the action given by \(\pi(x_t)\).
  \item
    Receive reward \(r_t\).
  \end{enumerate}
\end{itemize}

\end{Algorithm}

Second, an even more popular method is the following greedy
algorithm.\index{greedy algorithm} The greedy algorithm avoids the
initial random exploration stage and instead picks whatever is optimal
for the data seen so far.

\begin{Algorithm}

\textbf{Greedy algorithm for contextual bandits}

\begin{itemize}
\tightlist
\item
  For \(t=1,2,\ldots\)

  \begin{enumerate}
  \def\labelenumi{\arabic{enumi}.}
  \tightlist
  \item
    Find a function \(\hat{R}_t\) to minimize prediction error: \[
       \hat{R}_t := \arg\min_f \sum_{s=1}^{t-1}  \loss(f(x_s,u_s),r_s) \,.
      \]
  \item
    Receive new context \(x_t\).
  \item
    Choose the action given by the policy \[
           \pi_t(x_t) := \arg\max_u \hat{R}_t(x_t,u) \,.
       \]
  \item
    Receive reward \(r_t\).
  \end{enumerate}
\end{itemize}

\end{Algorithm}

In worst-case settings, the greedy algorithm may accrue linear regret.
However, worst-case contexts appear to be rare. In the linear contextual
bandits problem, where rewards are an unknown linear function of the
context, even slight random permutations of a worst-case instance lead
to sublinear regret.\citep{kannan2018smoothed}

The success of the greedy algorithm shows that it is not always
desirable to be exploring random actions to see what happens. This is
especially true for industrial applications where random exploration is
often costly and the value of adding exploration seems
limited.\citep{hummel2016machine, bietti2018contextual} This context is
useful to keep in mind as we move to the more complex problem of
reinforcement learning and approximate dynamic programming.

\hypertarget{when-the-model-is-unknown-approximate-dynamic-programming}{%
\section{When the model is unknown: Approximate dynamic
programming}\label{when-the-model-is-unknown-approximate-dynamic-programming}}

We now bring dynamics back into the picture and attempt to formalize how
to solve general SDM problems when we don't know the dynamics model or
even the reward function. We turn to exploring the three main approaches
in this space: certainty equivalence fits a model from some collected
data and then uses this model as if it were true in the SDM problem.
Approximate Dynamic Programming uses Bellman's principle of optimality
and stochastic approximation to learn Q-functions from data. Direct
Policy Search directly searches for policies by using data from previous
episodes in order to improve the reward. Each of these has their
advantages and disadvantages as we now explore in depth.

\hypertarget{certainty-equivalence-for-sequential-decision-making}{%
\subsection{Certainty equivalence for sequential decision
making}\label{certainty-equivalence-for-sequential-decision-making}}

One of the simplest, and perhaps most obvious strategies to solve an SDM
problem when the dynamics are unknown is to estimate the dynamics from
some data and then to use this estimated model as if it were the true
model in the SDM problem.\index{certainty equivalence}

Estimating a model from data is commonly called ``system
identification'' in the dynamical systems and control literature. System
identification differs from conventional estimation because one needs to
carefully choose the right inputs to excite various degrees of freedom
and because dynamical outputs are correlated over time with the
parameters we hope to estimate, the inputs we feed to the system, and
the stochastic disturbances. Once data is collected, however,
conventional prediction tools can be used to find the system that best
agrees with the data and can be applied to analyze the number of samples
required to yield accurate models.

Let's suppose we want to build a predictor of the state~\(x_{t+1}\) from
the trajectory history of past observed states and actions. A simple,
classic strategy is simply to inject a random probing sequence~\(u_t\)
for control and then measure how the state responds. Up to stochastic
noise, we should have that \[
    x_{t+1} \approx \varphi(x_t,u_t) \,,
\] where~\(\varphi\) is some model aiming to approximate the true
dynamics.~\(\varphi\) might arise from a first-principles physical model
or might be a non-parametric approximation by a neural network. The
state-transition function can then be fit using supervised learning. For
instance, a model can be fit by solving the least-squares problem \[
\begin{array}{ll}
\text{minimize}_{\varphi} & \sum_{t=0}^{N-1} ||x_{t+1} - \varphi(x_t,u_t)||^2\,.
\end{array}
\]

Let~\(\hat{\varphi}\) denote the function fit to the collected data to
model the dynamics. Let~\(\omega_t\) denote a random variable that we
will use as a model for the noise process. With such a point estimate
for the model, we might solve the optimal control problem \[
\begin{array}{ll}
\text{maximize} & \E_{\omega_t}[ \sum_{t=0}^N R(x_t,u_t)  ]\\
\text{subject to} & x_{t+1} = \hat{\varphi}(x_t, u_t)+\omega_t,~u_t = \pi_t(\tau_t)\,.
\end{array}
\] In this case, we are solving the wrong problem to get our control
policies~\(\pi_t\). Not only is the model incorrect, but this
formulation requires some plausible model of the noise process. But we
emphasize that this is standard engineering practice. Though more
sophisticated techniques can be used to account for the errors in
modeling, feedback often can compensate for these modeling errors.

\hypertarget{approximate-dynamic-programming}{%
\subsection{Approximate dynamic
programming}\label{approximate-dynamic-programming}}

Approximate dynamic programming approaches the RL problem by directly
approximating the optimal control cost and then solving this with
techniques from dynamic programming. Approximate Dynamic Programming
methods typically try to infer Q-functions directly from data. The
standard assumption in most practical implementations of Q-learning is
that the Q-functions are static, as would be the case in the infinite
horizon, discounted optimal control problem.

Probably the best known approximate dynamic programming method is
\emph{Q-learning}.\citep{watkins1992q} Q-learning simply attempts to
solve value iteration using \emph{stochastic
approximation}.\index{Q-learning} If we draw a sample trajectory using
the policy given by the optimal policy, then we should have
(approximately and in expectation) \[
    \mathcal{Q}_\gamma(x_t,u_t) \approx R(x_t,u_t) + \gamma \max_{u'} \mathcal{Q}_\gamma(x_{t+1},u')\,.
\] Thus, beginning with some initial guess
\(\mathcal{Q}_\gamma^{(\mathrm{old})}\) for the Q-function, we can
update \[
    \mathcal{Q}_\gamma^{(\mathrm{new})}(x_t,u_t) = (1-\eta) \mathcal{Q}_\gamma^{(\mathrm{old})}(x_t,u_t) + \eta \left(R(x_t,u_t) + \gamma  \max_{u'} \mathcal{Q}_\gamma^{(\mathrm{old})}(x_{t+1},u')\right)
\] where~\(\eta\) is a \emph{step-size} or \emph{learning rate}.

The update here only requires data generated by the policy
\(Q_\gamma^{\mathrm{old}}\) and does not need to know the explicit form
of the dynamics. Moreover, we don't even need to know the reward
function if this is provided online when we generate trajectories.
Hence, Q-learning is often called ``model free.'' We strongly dislike
this terminology and do not wish to dwell on it. Unfortunately,
distinguishing between what is ``model-free'' and what is
``model-based'' tends to just lead to confusion. All reinforcement
learning is inherently based on models, as it implicitly assumes data is
generated by some Markov Decision Process. In order to run Q-learning we
need to know the form of the Q-function itself, and except for the
tabular case, how to represent this function requires \emph{some}
knowledge of the underlying dynamics. Moreover, assuming that value
iteration is the proper solution of the problem is a modeling
assumption: we are assuming a discount factor and time invariant
dynamics. But the reader should be advised that when they read
``model-free,'' this almost always means ``no model of the state
transition function was used when running this algorithm.''

For continuous control problems methods like Q-learning appear to make
an inefficient use of samples. Suppose the internal state of the system
is of dimension~\(d\). When modeling the state-transition function, each
sample provides~\(d\) pieces of information about the dynamics. By
contrast, Q-learning only uses~\(1\) piece of information per time step.
Such inefficiency is often seen in practice. Also troubling is the fact
that we had to introduce the discount factor in order to get a simple
form of the Bellman equation. One can avoid discount factors, but this
requires either considerably more sophisticated analysis. Large discount
factors do in practice lead to brittle methods, and the discount factor
becomes a hyperparameter that must be tuned to stabilize performance.

We close this section by noting that for many problems with high
dimensional states or other structure, we might be interested in not
representing Q-functions as a look up table. Instead, we might
approximate the Q-functions with a parametric family:
\(\mathcal{Q}(x,u;\vartheta)\). Though we'd like to update the parameter
\(\vartheta\) using something like gradient descent, it's not
immediately obvious how to do so. The simplest attempt, following the
guide of stochastic approximation is to run the iterations: \[
\begin{aligned}
\delta_t  &= R(x_{t},u_{t}) + \gamma  \mathcal{Q}(x_{t+1},u_{t+1}; \vartheta_{t})
- \mathcal{Q}(x_t,u_t;\vartheta_t)\\
\vartheta_{t+1} &= \vartheta_t + \eta \delta_t \nabla \mathcal{Q}(x_t,u_t,\vartheta_t)
\end{aligned}
\] This algorithm is called \emph{Q-learning with function
approximation}. A typically more stable version uses momentum to average
out noise in Q-learning. With~\(\delta_t\) as above, we add the
modification \[
\begin{aligned}
e_{t} &= \lambda e_{t-1} + \nabla \mathcal{Q}(x_t,u_t,\vartheta_t)\\
\vartheta_{t+1} &= \vartheta_t + \eta \delta_t e_t
\end{aligned}
\] for~\(\lambda \in [0,1]\). This method is known as
SARSA(\(\lambda\)).\citep{SARSA}

\hypertarget{direct-policy-search}{%
\subsection{Direct policy search}\label{direct-policy-search}}

The most ambitious form of control without models attempts to directly
learn a policy function from episodic experiences without ever building
a model or appealing to the Bellman equation. From the oracle
perspective, these policy driven methods turn the problem of RL into
derivative-free optimization.\index{policy search}

In turn, let's first begin with a review of a general paradigm for
leveraging random sampling to solve optimization problems. Consider the
general unconstrained optimization problem \[
\begin{array}{ll}
    \text{maximize}_{z\in\R^d} & R(z) \,.
    \end{array}
\] Any optimization problem like this is equivalent to an optimization
over probability densities on~\(z\): \[
\begin{array}{ll}
    \text{maximize}_{p(z)} & \mathbb{E}_p[R(z)] \,.
\end{array}
\] If~\(z_\star\) is the optimal solution, then we'll get the same value
if we put a~\(\delta\)-function around~\(z_\star\). Moreover, if~\(p\)
is a density, it is clear that the
\emph{expected value of the reward function} can never be larger than
the maximal reward achievable by a fixed~\(z\). So we can either
optimize over~\(z\) or we can optimize over \emph{densities} over \(z\).

Since optimizing over the space of all probability densities is
intractable, we must restrict the class of densities over which we
optimize. For example, we can consider a family parameterized by a
parameter vector~\(\vartheta\):~\(p(z;\vartheta)\) and attempt to
optimize \[
\begin{array}{ll}
    \text{maximize}_{\vartheta} & \mathbb{E}_{p(z;\vartheta)}[R(z)] \,.
\end{array}
\] If this family of densities contains all of the Delta functions, then
the optimal value will coincide with the non-random optimization
problem. But if the family does not contain the Delta functions, the
resulting optimization problem only provides a lower bound on the
optimal value no matter how good of a probability distribution we find.

That said, this reparameterization provides a powerful and general
algorithmic framework for optimization. In particular, we can compute
the derivative of~\(J(\vartheta):= \mathbb{E}_{p(z;\vartheta)}[R(z)]\)
using the following calculation (called ``the log-likelihood trick''):
\[
\begin{aligned}
    \nabla_{\vartheta} J(\vartheta) &= \int R(z) \nabla_{\vartheta} p(z;\vartheta) dz\\
    &= \int R(z) \left(\frac{\nabla_{\vartheta} p(z;\vartheta)}{p(z;\vartheta)}\right) p(z;\vartheta) dz\\
    &= \int \left( R(z) \nabla_{\vartheta} \log p(z;\vartheta) \right) p(z;\vartheta)dz \\
  &= \mathbb{E}_{p(z;\vartheta)}\left[ R(z) \nabla_{\vartheta} \log p(z;\vartheta) \right]\,.
\end{aligned}
\] This derivation reveals that the gradient of~\(J\) with respect to
\(\vartheta\) is the expected value of the function \[
    G(z,\vartheta) = R(z) \nabla_{\vartheta} \log p(z;\vartheta)
\] Hence, if we sample~\(z\) from the distribution defined by
\(p(z;\vartheta)\), we can compute~\(G(z,\vartheta)\) and will have an
unbiased estimate of the gradient of~\(J\). We can follow this direction
and will be running stochastic gradient descent on~\(J\), defining the
following algorithm:\index{REINFORCE}

\begin{Algorithm}

\textbf{REINFORCE algorithm:}

\begin{itemize}
\tightlist
\item
  \emph{Input Hyperparameters:} step-sizes \(\alpha_j>0\).
\item
  \emph{Initialize:} \(\vartheta_0\) and \(k = 0\).
\item
  Until the heat death of the universe, do:

  \begin{enumerate}
  \def\labelenumi{\arabic{enumi}.}
  \tightlist
  \item
    Sample \(z_k\sim p(z;\vartheta_k)\).
  \item
    Set
    \(\vartheta_{k+1} = \vartheta_k + \alpha_k R(z_k) \nabla_\vartheta \log p(z_k; \vartheta_k)\).
  \item
    \(k\leftarrow k + 1\).
  \end{enumerate}
\end{itemize}

\end{Algorithm}

The main appeal of the REINFORCE Algorithm is that it is not hard to
implement. If you can efficiently sample from~\(p(z;\vartheta)\) and can
easily compute~\(\nabla \log p\), you can run this algorithm on
essentially any problem. But such generality must and does come with a
significant cost. The algorithm operates on stochastic gradients of the
sampling distribution, but the function we cared about
optimizing---\(R\)---is only accessed through function evaluations.
Direct search methods that use the log-likelihood trick are necessarily
derivative free optimization methods, and, in turn, are necessarily less
effective than methods that compute actual gradients, especially when
the function evaluations are noisy. Another significant concern is that
the choice of distribution can lead to very high variance in the
stochastic gradients. Such high variance in turn implies that many
samples need to be drawn to find a stationary point.

That said, the ease of implementation should not be readily discounted.
Direct search methods are easy to implement, and oftentimes reasonable
results can be achieved with considerably less effort than custom
solvers tailored to the structure of the optimization problem. There are
two primary ways that this sort of stochastic search arises in
reinforcement learning: Policy gradient and pure random search.

\hypertarget{policy-gradient}{%
\subsection{Policy gradient}\label{policy-gradient}}

Though we have seen that the optimal solutions of Bellman's equations
are deterministic, probabilistic policies can add an element of
exploration to a control strategy, hopefully enabling an algorithm to
simultaneously achieve reasonable awards and learn more about the
underlying dynamics and reward functions. Such policies are the starting
point for \emph{policy gradient}
methods.\citep{williams1992simple}\index{policy gradient}

Consider a \emph{parametric, randomized policy} such that~\(u_t\) is
sampled from a distribution~\(p(u \vert \tau_t;\vartheta)\) that is only
a function of the currently observed trajectory and a parameter vector
\(\vartheta\). A probabilistic policy induces a probability distribution
over trajectories: \[
    p(\tau;\vartheta) = \prod_{t=0}^{L-1} p(x_{t+1} \vert x_{t},u_{t}) p(u_t \vert \tau_t ;\vartheta)\,.
\] Moreover, we can overload notation and define the reward of a
trajectory to be \[
    R(\tau) = \sum_{t=0}^N R_t(x_t,u_t)
\] Then our optimization problem for reinforcement learning takes the
form of stochastic search. Policy gradient thus proceeds by sampling a
trajectory using the probabilistic policy with
parameters~\(\vartheta_k\), and then updating using REINFORCE.

Using the log-likelihood trick and the factored form of the probability
distribution~\(p(\tau;\vartheta)\), we can see that the gradient
of~\(J\) with respect to~\(\vartheta\) \emph{is not a function of the
underlying dynamics}. However, at this point this should not be
surprising: by shifting to distributions over policies, we push the
burden of optimization onto the sampling procedure.

\hypertarget{pure-random-search}{%
\subsection{Pure random search}\label{pure-random-search}}

\index{random search}

An older and more widely applied method to solve the generic stochastic
search problem is to directly perturb the current decision
variable~\(z\) by random noise and then update the model based on the
received reward at this perturbed value. That is, we apply the REINFORCE
Algorithm with sampling
distribution~\(p(z;\vartheta) = p_0(z-\vartheta)\) for some
distribution~\(p_0\). Simplest examples for~\(p_0\) would be the uniform
distribution on a sphere or a normal distribution. Perhaps less
surprisingly here, REINFORCE can again be run without any knowledge of
the underlying dynamics. The REINFORCE algorithm has a simple
interpretation in terms of gradient approximation. Indeed, REINFORCE is
equivalent to approximate gradient ascent of~\(R\) \[
    \vartheta_{t+1} = \vartheta_{t} + \alpha g_\sigma(\vartheta_k)
\] with the gradient approximation \[
    g_\sigma(\vartheta) = \frac{R(\vartheta + \sigma \epsilon) - R(\vartheta - \sigma \epsilon) }{2\sigma} \epsilon\,.
\] This update says to compute a finite difference approximation to the
gradient along the direction~\(\epsilon\) and move along the gradient.
One can reduce the variance of such a finite-difference estimate by
sampling along multiple random directions and averaging: \[
    g^{(m)}_\sigma(\vartheta) = \frac{1}{m} \sum_{i=1}^m\frac{R(\vartheta + \sigma \epsilon_i) - R(\vartheta - \sigma \epsilon_i) }{2\sigma} \epsilon_i\,.
\] This is akin to approximating the gradient in the random subspace
spanned by the~\(\epsilon_i\)

This particular algorithm and its generalizations goes by many different
names. Probably the earliest proposal for this method is by
Rastrigin.\citep{Rastrigin63} Somewhat surprisingly, Rastrigin initially
developed this method to solve reinforcement learning problems. His main
motivating example was an inverted pendulum. A rigorous analysis using
contemporary techniques was provided by Nesterov and
Spokoiny.\citep{nesterov2017random} Random search was also discovered by
the evolutionary algorithms community and is called
\((\mu,\lambda)\)-Evolution Strategies.\citep{Beyer02, SchwefelThesis}
Random search has also been studied in the context of stochastic
approximation\citep{spall1992multivariate} and
bandits.\citep{flaxman2005online, agarwal2010optimal} Algorithms that
get invented by four different communities probably have something good
going for them.

\hypertarget{deep-reinforcement-learning}{%
\subsection{Deep reinforcement
learning}\label{deep-reinforcement-learning}}

\index{deep reinforcement learning}\index{reinforcement learning!deep}

We have thus far spent no time discussing \emph{deep} reinforcement
learning. That is because there is nothing conceptually different other
than using neural networks for function approximation. That is, if one
wants to take any of the described methods and make them deep, they
simply need to add a neural net. In model-based RL,~\(\varphi\) is
parameterized as a neural net, in ADP, the Q-functions or Value
Functions are assumed to be well-approximated by neural nets, and in
policy search, the policies are set to be neural nets. The algorithmic
concepts themselves don't change. However, convergence analysis
certainly will change, and algorithms like Q-learning might not even
converge. The classic text Neurodynamic Programming by Bertsekas and
Tsitisklis discusses the adaptations needed to admit function
approximation.\citep{bertsekas1996neuro}

\hypertarget{certainty-equivalence-is-often-optimal-for-reinforcement-learning}{%
\section{Certainty equivalence is often optimal for reinforcement
learning}\label{certainty-equivalence-is-often-optimal-for-reinforcement-learning}}

In this section, we give a survey of the power of certainty equivalence
in sequential decision making problems. We focus on the simple cases of
tabular MDPs and LQR as they are illustrative of more general problems
while still being manageable enough to analyze with relatively simple
mathematics. However, these analyses are less than a decade old. Though
the principle of certainty equivalence dates back over 60 years, our
formal understanding of certainty equivalence and its robustness is just
now solidifying.

\hypertarget{certainty-equivalence-for-lqr}{%
\subsection{Certainty equivalence for
LQR}\label{certainty-equivalence-for-lqr}}

\index{linear quadratic regulator}

Consider the linear quadratic regulator problem \[
\begin{array}{ll}
\text{minimize} \, & \lim_{T\rightarrow \infty} \E_{W_t} \left[\frac{1}{2T}\sum_{t=0}^T X_t^T\Phi X_t + U_t^T \Psi U_t\right], \\
\text{subject to} & X_{t+1} = A X_t+ B U_t + W_t,~U_t=\pi_t(X_t) \\
& \text{($x_0$ given).}
\end{array}
\] We have shown that the solution to this problem is static state
feedback~\(U_t = -K_\star X_t\) where \[
    K_\star=(\Psi + B^T M B)^{-1} B^T M A
\] and~\(M\) is the unique stabilizing solution to the Discrete
Algebraic Riccati Equation \[
M = \Phi + A^T M A - (A^T M B)(\Psi + B^T M B)^{-1} (B^T M A)\,.
\]

Suppose that instead of knowing~\((A,B,\Phi,\Psi)\) exactly, we only
have estimates~\((\hat{A},\hat{B},\hat{\Phi},\hat{\Psi})\). Certainty
equivalence would then yield a control policy~\(U_t = -\hat{K} X_t\)
where \(\hat{K}\) can be found
using~\((\hat{A},\hat{B},\hat{\Phi},\hat{\Psi})\) in place
of~\((A,B,\Phi,\Psi)\) in the formulae above. What is the cost of this
model?

The following discussion follows arguments due to Mania et
al.\citep{Mania19} Let~\(J(K)\) denote that cost of using the
policy~\(K\). Note that this cost may be infinite, but it will also be
differentiable in~\(K\). If we unroll the dynamics and compute expected
values, one can see that the cost is the limit of polynomials in~\(K\),
and hence is differentiable.

Suppose that \[
    \epsilon := \max \left\{ \|\hat{A}-A\|,\, \|\hat{B}-B\|,\, \|\hat{\Phi}-\Phi\|,\, \|\hat{\Psi}-\Psi\|\right\}
\] If we Taylor expand the cost we find that for some~\(t \in [0,1]\),
\[
    J(\hat{K})-J_\star = \langle \nabla J(K_\star), \hat{K}-K_\star \rangle + \frac{1}{2} (\hat{K}-K_\star)^T \nabla^2 J(\tilde{K})(\hat{K}-K_\star)\,.
\] where~\(\tilde{K} = (1-t)K_\star + t \hat{K}\). The first term is
equal to zero because~\(K_\star\) is optimal. Since the map from
\((A,B,\Phi,\Psi)\) to~\(K_\star\) is differentiable, there must be
constants~\(L\) and~\(\epsilon_0\) such that
\(\|\hat{K}-K_\star\| \leq L \epsilon\) whenever
\(\epsilon \leq \epsilon_0\). This means that as long as the estimates
for \((A,B,\Phi,\Psi)\) are close enough to the true values, we must
have \[
    J(\hat{K})-J_\star = O(\epsilon^2)\,.
\]

Just how good should the estimates for these quantities be? Let's focus
on the dynamics~\((A,B)\) as the cost matrices~\(\Phi\) and~\(\Psi\) are
typically design parameters, not unknown properties of the system.
Suppose~\(A\) is~\(d\times d\) and~\(B\) is~\(d\times p\). Basic
parameter counting suggests that if we observe~\(T\) sequential states
from the dynamical system, we observe a total of~\(dT\) numbers, one for
each dimension of the state per time step. Hence, a naive statistical
guess would suggest that \[
 \max \left\{ \|\hat{A}-A\|,\, \|\hat{B}-B\|\right\} \leq O\left(\sqrt{\frac{d+p}{T}}\right)\,.
\] Combining this with our Taylor series argument implies that \[
    J(\hat{K})-J_\star = O\left(\frac{d+p}{T}\right)\,.
\] As we already described, this also suggests that certainty equivalent
control accrues a regret of \[
    \mathcal{R}_T = O(\sqrt{T})\,.
\]

This argument can be made completely rigorous.\citep{Mania19} The regret
accrued also turns out to be the optimal.\citep{Simchowitz20} Moreover,
the Taylor series argument here works for any model where the cost
function is twice differentiable. Thus, we'd expect to see similar
behavior in more general SDM problems with continuous state spaces.

\hypertarget{certainty-equivalence-for-tabular-mdps}{%
\subsection{Certainty equivalence for tabular
MDPs}\label{certainty-equivalence-for-tabular-mdps}}

For discounted, tabular MDPs, certainty equivalence also yields an
optimal sample complexity. This result is elementary enough to be proven
in a few pages. We first state an approximation theorem that shows that
if you build a policy with the wrong model, the value of that policy can
be bounded in terms of the inaccuracy of your model. Then, using
Hoeffding's inequality, we can construct a sample complexity bound that
is nearly optimal. The actual optimal rate follows using our main
approximation theorem coupled with slightly more refined concentration
inequalities. We refer readers interested in this more refined analysis
to the excellent reinforcement learning text by Agarwal et
al.\citep{RLTheoryBook}

Let~\(V_\star(x)\) denote the optimal expected reward attainable by some
policy on the discounted problem \[
    \begin{array}{ll}
        \text{maximize} &  (1-\gamma) \E_{W_t}[ \sum_{t=0}^\infty \gamma^t R(X_t,U_t,W_t) ]\\
        \text{subject to} & X_{t+1} = f(X_t, U_t, W_t),~U_t=\pi_t(X_t)\\
        & \text{($x_0=x$).}
    \end{array}
\] Note that~\(V_\star\) is a function of the initial state. This
mapping from initial state to expected rewards is called the \emph{value
function} of the SDM problem. Let~\(V^{\pi}(x)\) denote the expected
reward attained when using some fixed, static policy~\(\pi\). Our aim is
to evaluate the reward of particular policies that arise from certainty
equivalence.

To proceed, first let~\(\hat{Q}\) be any function mapping state-action
pairs to a real value. We can always define a policy
\(\pi_{\hat{Q}}(x) = \arg \max_u \hat{Q}(x,u)\). The following theorem
quantifies the value of~\(\pi_{\hat{Q}}\) when~\(\hat{Q}\) is derived by
solving the Bellman equation with an approximate model of the MDP
dynamics. This theorem has been derived in numerous places in the RL
literature, and yet it does not appear to be particularly well known. As
pointed out by 'Avila Pires and Szepesvari\citep{AvilaPires16}, it
appears as a Corollary to Theorem 3.1 in Whitt\citep{Whitt78} (1978), as
Corollary 2 in Singh and Yee\citep{Singh94} (1994), as a corollary of
Proposition 3.1 in Bertsekas\citep{Bertsekas2012} (2012). We emphasize
it here as it demonstrates immediately why certainty equivalence is such
a powerful tool in sequential decision making
problems.\index{model-error theorem}

\begin{Theorem}

\textbf{Model-error for MDPs.} Consider a~\(\gamma\)-discounted MDP with
dynamics governed by a model~\(p\) and a reward function~\(r\). Let
\(\hat{Q}\) denote the~\(Q\)-function for the MDP with the same rewards
but dynamics~\(\hat{\Pr}\). Then we have \[
 V_\star(x)-V^{\pi_{\hat{Q}}}(x)  \leq \frac{2\gamma}{(1-\gamma)^2} \sup_{x,u} \left| \E_{\hat{\Pr}(\cdot|x,u)}[V_\star] - \E_{\Pr (\cdot|x,u)}[V_\star] \right|\,.
\]

\end{Theorem}

This theorem states that the values associated with the policy that we
derive using the wrong dynamics will be close to optimal if
\(\E_{\hat{\Pr}(\cdot|x,u)}[V_\star]\)
and~\(\E_{\Pr (\cdot|x,u)}[V_\star]\) are close for all state-action
pairs~\((x,u)\). This is a remarkable result as it shows that we can
control our regret by our prediction errors. But the only prediction
that matters is our predictions of the optimal value
vectors~\(V_\star\). Note further that this theorem makes no assumptions
about the size of the state spaces: it holds for discrete, tabular MDPs,
and more general discounted MDPs. A discussion of more general problems
is covered by Bertsekas.\citep{BertsekasAbstractDPBook}

Focusing on the case of finite-state tabular MDPs, suppose the rewards
are in the range~\([0,1]\) and there are~\(S\) states and~\(A\) actions.
Then the values are in the
range~\(V_\star(x) \in [0, (1-\gamma)^{-1}]\). Immediately from this
result, we can derive a sample-complexity bound. Let's suppose that for
each pair~\((x,u)\), we collect~\(n\) samples to estimate the
conditional probabilities~\(\Pr[X'=x'|x,u]\), and define
\(\hat{\Pr}(X'=x'|x,u)\) to be the number of times we observe~\(x'\)
divided by~\(n\). Then by Hoeffding's inequality \[
    \Pr\left[ \left| \E_{\hat{\Pr}[\cdot|x,u]}[V_\star] - \E_{\Pr[\cdot|x,u]}[V_\star] \right|\geq \epsilon \right]
    \leq 2\exp\left(-2n\epsilon^2 (1-\gamma)^2 \right)
\] and therefore, by the union bound \[
    \sup_{x,u} \left| \E_{\hat{\Pr}[\cdot|x,u]}[V_\star] - \E_{\Pr[\cdot|x,u]}[V_\star] \right| \leq \sqrt{\frac{\log\left(\frac{2 SA}{\delta}\right)}{n(1-\gamma)^2}}\,.
\] with probability~\(1-\delta\). If we let~\(N = SAn\) denote the total
number of samples collected, we see that \[
 V_\star(x) -   V^{\pi_{\hat{Q}}}(x) \leq \frac{2\gamma}{(1-\gamma)^3} \sqrt{\frac{SA\log\left(\frac{2 SA}{\delta}\right)}{N}}\,.
\] Our naive bound here is nearly optimal: the dependence on~\(\gamma\)
can be reduced from~\((1-\gamma)^{-3}\) to~\((1-\gamma)^{-3/2}\) using
more refined deviation inequalities, but the dependence on~\(S\),~\(A\),
and~\(N\) is optimal.\citep{RLTheoryBook, Agarwal2020c} That is,
certainty equivalence achieves an optimal sample complexity for the
discounted tabular MDP problem.

\hypertarget{proof-of-the-model-error-theorem}{%
\subsection{Proof of the model-error
theorem}\label{proof-of-the-model-error-theorem}}

\newcommand{\transop}[3]{{\mathcal{T}_{#1}}^{#3}#2}

The proof here combines arguments by Bertsekas \citep{Bertsekas2012} and
Agarwal.\citep{RLTheoryBook} Let us first introduce notation that makes
the proof a bit more elegant. Let~\(Q\) be any function mapping
state-action pairs to real numbers. Given a policy~\(\pi\) and a
state-transition model \(\Pr\), denote~\({\mathcal{T}_{\Pr}}^{}\) to be
the map from functions to functions where \[
    [{\mathcal{T}_{\Pr}}^{}Q ](x,u) = r(x,u)+\gamma \sum_{x'} \max_{u'} Q(x',u') \Pr[X'=x'|x,u]  \,.
\] With this notation, the Bellman equation for the discounted MDP
simply becomes \[
    Q_\star = {\mathcal{T}_{\Pr}}^{}Q_\star\,.
\] If we were to use~\(\hat{\Pr}\) instead of~\(\Pr\), this would yield
an alternative~\(Q\)-function,~\(\hat{Q}\), that satisfies the Bellman
equation \(\hat{Q} ={\mathcal{T}_{\hat{\Pr}}}^{}\hat{Q}\).

The operator~\({\mathcal{T}_{\Pr}}^{}\) is a \emph{contraction mapping}
in the~\(\ell_\infty\) norm. To see this, note that for any
functions~\(Q_1\) and~\(Q_2\), \[
\begin{aligned}
    |[{\mathcal{T}_{\Pr}}^{}Q_1 - {\mathcal{T}_{\Pr}}^{}Q_2 ](x,u)  | &= \left| \gamma \sum_{x'} (\max_{u_1} Q_1(x',u_1) - \max_{u_2} Q_2(x',u_2)) \Pr[X'=x'|x,u]  \right|\\
    & \leq \gamma \sum_{x'} \Pr[X'=x'|x,u]  \left|  \max_{u_1} Q_1(x',u_1) - \max_{x_2} Q_2(x',u_2) \right| \\
    & \leq \gamma\|Q_1-Q_2\|_\infty\,.
    \end{aligned}
\] Since~\({\mathcal{T}_{\Pr}}^{}\) is a contraction, the solution of
the discounted Bellman equations are unique and
\(Q_\star =\lim_{k\rightarrow \infty} {\mathcal{T}_{\Pr}}^{k}Q\) for any
function~\(Q\). Similarly,
\(\hat{Q} =\lim_{k\rightarrow \infty} {\mathcal{T}_{\hat{\Pr}}}^{k}Q\).

Now we can bound \[
\begin{aligned}
    \left\| {\mathcal{T}_{\hat{\Pr}}}^{k}Q_\star - Q_\star\right\|_\infty \leq \sum_{i=1}^k \| {\mathcal{T}_{\hat{\Pr}}}^{i}Q_\star - {\mathcal{T}_{\hat{\Pr}}}^{i-1}Q_\star\|_\infty
    \leq     \sum_{i=1}^k \gamma^k \| {\mathcal{T}_{\hat{\Pr}}}^{}Q_\star -Q_\star\|_\infty\,.
\end{aligned}
\] Taking limits of both sides as~\(k\rightarrow\infty\), we find \[
\begin{aligned}
    \left\| \hat{Q} - Q_\star\right\|_\infty
    \leq    \frac{1}{1-\gamma}\| {\mathcal{T}_{\hat{\Pr}}}^{}Q_\star -Q_\star\|_\infty\,.
\end{aligned}
\] But since~\(Q_\star = {\mathcal{T}_{\Pr}}^{}Q_\star\), and \[
\begin{aligned}
&    [{\mathcal{T}_{\hat{\Pr}}}^{}Q_\star -{\mathcal{T}_{\Pr}}^{}Q_\star](x,u)\\
     =&\gamma  \sum_{x'} \max_{u'} Q_\star(x',u') \left(\hat{\Pr}[X'=x'|x,u] -
      \Pr[X'=x'|x,u]
      \right) \,,
\end{aligned}
\] we have the bound \[
\left\| \hat{Q} - Q_\star\right\|_\infty \leq \frac{\gamma}{1-\gamma} \sup_{x,u} \left| \E_{\hat{\Pr}[\cdot|x,u]}[V_\star] - \E_{\Pr [\cdot|x,u]}[V_\star] \right|\,.
\]

To complete the proof, it suffices to use
\(\left\| \hat{Q} - Q_\star\right\|_\infty\) to upper bound the
difference between the values of the two policies. Let
\(\pi_\star(x) = \arg\max_u Q_\star(x,u)\) and
\(\hat{\pi}(x) = \arg\max_u \hat{Q}_\star(x,u)\) denote the optimal
policies for the models~\(\Pr\) and~\(\hat{\Pr}\) respectively. For any
policy~\(\pi\), we have \[
    V^{\pi}(x) = r(x,\pi(x)) + \gamma \sum_{x'} \Pr[X'=x'|x,\pi(x)] V^\pi(x')\,,
\] and hence we can bound the optimality gap as \[
\begin{aligned}
    V_\star(x) - V^{\hat{\pi}}(x) &= Q_\star(x,\pi_\star(x)) - V^{\hat{\pi}}(x)\\
    &= Q_\star(x,\pi_\star(x)) -Q_\star(x,\hat{\pi}(x))+Q_\star(x,\hat{\pi}(x)) - V^{\hat{\pi}}(x)\\
    &= Q_\star(x,\pi_\star(x)) -Q_\star(x,\hat{\pi}(x))\\
    &\qquad\qquad+ \gamma \sum_{x'} \Pr[X'=x'|x,\hat{\pi}(x)] \left(V_\star(x') - V^{\hat{\pi}}(x') \right)\\
    &\leq Q_\star(x,\pi_\star(x)) - \hat{Q}(x,\pi_\star(x)) + \hat{Q}(x,\hat{\pi}(x)) -Q_\star(x,\hat{\pi}(x))\\
    &\qquad\qquad+ \gamma \sum_{x'} \Pr[X'=x'|x,\hat{\pi}(x)]\left(V_\star(x') - V^{\hat{\pi}}(x') \right)\\
    &\leq 2\|Q_\star- \hat{Q}\|_\infty + \gamma \|V_\star - V^{\hat{\pi}} \|_\infty\,.
\end{aligned}
\] Here, the first inequality holds because
\(\hat{Q}(x,\pi_\star(x)) \leq \max_u \hat{Q}(x,u)=\hat{Q}(x,\hat{\pi}(x))\).
Rearranging terms shows that \[
V_\star(x) - V^{\hat{\pi}}(x) \leq \frac{2}{1-\gamma}\|Q_\star- \hat{Q}\|_\infty\,,
\] which, when combined with our previous bound on
\(\|Q_\star- \hat{Q}\|_\infty\), completes the proof.

\hypertarget{sample-complexity-of-other-rl-algorithms}{%
\subsection{Sample complexity of other RL
algorithms}\label{sample-complexity-of-other-rl-algorithms}}

The sample complexity of reinforcement learning remains an active field,
with many papers honing in on algorithms with optimal complexity.
Researchers have now shown a variety of methods achieve the optimal
complexity of certainty equivalence including those based on ideas from
approximate dynamic programming and Q-learning. For LQR on the other
hand, no other methods are currently competitive.

While sample complexity is important, there are not significant gains to
be made over simple baselines that echo decades of engineering practice.
And, unfortunately, though sample complexity is a well posed problem
which excites many researchers, it does not address many of the
impediments preventing reinforcement learning form being deployed in
more applications. As we will see in a moment, the optimization
framework itself has inherent weaknesses that cannot be fixed by better
sample efficiency, and these weaknesses must be addressed head-on when
designing an SDM system.

\hypertarget{the-limits-of-learning-in-feedback-loops}{%
\section{The limits of learning in feedback
loops}\label{the-limits-of-learning-in-feedback-loops}}

Though we have shown the power of certainty equivalence, it is also a
useful example to guide how reinforcement learning---and optimal
sequential decision making more generally---can go wrong. First, we will
show how optimal decision making problems themselves can be set up to be
very sensitive to model-error. So treating a model as true in these
cases can lead to misguided optimism about performance. Second, we will
adapt this example to the case where the state is partially observed and
demonstrate a more subtle pathology. As we discussed in the last
chapter, when state is not perfectly observed, decision making is
decidedly more difficult. Here we will show an example where improving
your prediction paradoxically increases your sensitivity to model error.

\hypertarget{fragile-instances-of-the-linear-quadratic-regulator}{%
\subsection{Fragile instances of the linear quadratic
regulator}\label{fragile-instances-of-the-linear-quadratic-regulator}}

\index{linear quadratic regulator}

Consider the following innocuous dynamics:

\[
    A = \begin{bmatrix} 0 & 1\\ 0 & 0\end{bmatrix} \,,\quad B = \begin{bmatrix} 0\\1 \end{bmatrix}\,.
\]

This system is a simple, two-state shift register. Write the state out
with indexed components~\(x=[x^{(1)},x^{(2)}]^\top\). New states enter
through the control~\(B\) into the second state. The first
state~\(x^{(1)}\) is simply whatever was in the second register at the
previous time step. The open loop dynamics of this system are as stable
as you could imagine. Both eigenvalues of~\(A\) are zero.

Let's say our control objective aims to try to keep the two components
of the state equal to each other. We can model this with the quadratic
cost matrices

\[
    \Phi = \begin{bmatrix} 1 & -1 \\ -1 & 1 \end{bmatrix} \,, \quad \Psi = 0\,.
\]

Here,~\(\Psi=0\) for simplicity, as the formulae are particularly nice
for this case. But, as we will discuss in a moment, the situation is not
improved simply by having~\(R\) be positive. For the disturbance, assume
that~\(W_t\) is zero mean, has bounded second moment,
\(\Sigma_t = \mathbb{E}[W_t W_t^\top]\), and is uncorrelated
with~\(X_t\) and~\(U_t\).

The cost is asking to minimize \[
    \E\left[\sum_{t=1}^N (X_t^{(1)}-X_t^{(2)})^2\right]
\] When~\(W_t=0\),~\(X_t^{(1)}+X_t^{(2)} = X_{t-1}^{(2)}+U_{t-1}\), so
the intuitive best action would be to set~\(U_{t}=X_t^{(2)}\). This
turns out to be the optimal action, and one can prove this directly
using standard dynamic programming computations or a Discrete Algebraic
Riccati Equation. With this identification, we can write \emph{closed
loop} dynamics by eliminating the control signal: \[
    X_{t+1} = \begin{bmatrix} 0 & 1\\ 0 & 1 \end{bmatrix} X_t + W_t\,.
\] This closed-loop system is \emph{marginally stable}, meaning that
while signals don't blow up, some states will persist forever and not
converge to~\(0\). The second component of the state simply exhibits a
random walk on the real line. We can analytically see that the system is
not stable by computing the eigenvalues of the state-transition matrix,
which are here~\(0\) and~\(1\). The~\(1\) corresponds the state where
the two components are equal, and such a state can persist forever.

If we learned an incorrect model of the dynamics, how would that
influence the closed loop behavior? The simplest scenario is that we
identified~\(B\) from some preliminary experiments. If the true
\(B_\star=\alpha B\), then the closed loop dynamics are \[
    X_{t+1} = \begin{bmatrix} 0 & 1\\ 0 &\alpha \end{bmatrix} X_t + W_t\,.
\] This system is unstable for any~\(\alpha>1\). That is, the system is
arbitrarily sensitive to misidentification of the dynamics. This lack of
robustness has nothing to do with the noise sequence. The structure of
the cost is what drives the system to fragility.

If~\(\Psi>0\), we would get a slightly different policy. Again, using
elementary dynamic programming shows that the optimal control is
\(u_t=\beta_t(\Psi) x_t^{(2)}\) for some~\(\beta_t(\Psi) \in (1/2,1)\).
The closed loop system will be a bit more stable, but this comes at the
price of reduced performance. You can also check that if you add
\(\epsilon\) times the identity to~\(\Phi\), we again get a control
policy proportional to the second component of the state,~\(x_t^{(2)}\).

Similar examples are fairly straightforward to construct. The
state-transition matrix of the closed loop dynamics will always be of
the form~\(A-BK\), and we can first find a~\(K\) such that~\(A-BK\) has
an eigenvalue of magnitude~\(1\). Once this is constructed, it suffices
to find a vector~\(v\) such that~\((v'B)^{-1} v'A = K\). Then the cost
\(\Phi=vv'\) yields the desired pathological example.

One such example that we will use in our discussion of partially
observed systems is the model: \[
    A = \begin{bmatrix} 1 & 1\\ 0 & 1\end{bmatrix} \,,\quad B = \begin{bmatrix} 0\\1 \end{bmatrix}\,,
\] \[
    \Phi = \begin{bmatrix} 1 & 1/2 \\ 1/2 & 1/4 \end{bmatrix} \,, \quad \Psi = 0\,.
\] The dynamics here are our ``Newton's Law'' dynamics studied in our
dynamic programming examples. One can check that the closed loop
dynamics of this system are \[
    X_{t+1} = \begin{bmatrix} 1 & 1\\ -2 &-2 \end{bmatrix} X_t + W_t\,.
\] The transition matrix here has eigenvalues~\(0\) and~\(-1\), and the
state~\(x=[1/2,-1]\) will oscillate in sign and persist forever.

\hypertarget{partially-observed-example}{%
\subsection{Partially observed
example}\label{partially-observed-example}}

Recall the generalization of LQR to the case with imperfect state
observation is called ``Linear Quadratic Gaussian'' control (LQG). This
is the simplest, special case of a POMDP. We again assume linear
dynamics: \[
    X_{t+1} = AX_t + B U_t + W_t\,.
\] where the state is now corrupted by zero-mean Gaussian
noise,~\(W_t\). Instead of measuring the state~\(X_t\) directly, we
instead measure a signal~\(Y_t\) of the form \[
    Y_t = C X_t + V_t\,.
\] Here,~\(V_t\) is also zero-mean Gaussian noise. Suppose we'd still
like to minimize a quadratic cost function \[
\lim_{T\rightarrow \infty} \E\left[\frac{1}{T} \sum_{t=0}^{T} X_t^\top \Phi X_t + U_t^\top \Psi U_t\right] \,.
\] This problem is very similar to our LQR problem except for the fact
that we get an indirect measurement of the state and need to apply some
sort of \emph{filtering} of the~\(Y_t\) signal to estimate~\(X_t\).

The optimal solution for LQG is strikingly elegant. Since the
observation of~\(X_t\) is through a Gaussian process, the maximum
likelihood estimation algorithm has a clean, closed form solution. As we
saw in the previous chapter, our best estimate for~\(X_t\), denoted
\(\hat{x}_t\), given all of the data observed up to time~\(t\) is given
by a Kalman Filter. The estimate obeys a difference equation \[
 \hat{x}_{t+1}  = A\hat{x}_t + B u_t + L(y_t-C\hat{x}_t)\,.
\] The matrix~\(L\) that can be found by solving an discrete algebraic
Riccati equation that depends on the variance of~\(v_t\) and~\(w_t\) and
on the matrices~\(A\) and~\(C\). In particular, it's the DARE with data
\((A^\top,C^\top,\Sigma_w,\Sigma_v)\).

The optimal LQG solution takes the estimate of the Kalman Filter,
\(\hat{x}_t\), and sets the control signal to be \[
    u_t = -K\hat{x}_t\,.
\]

Here,~\(K\) is gain matrix that would be used to solve the LQR problem
with data~\((A,B,\Phi,\Psi)\). That is, LQG performs optimal filtering
to compute the best state estimate, and then computes a feedback policy
as if this estimate was a noiseless measurement of the state. That this
turns out to be optimal is one of the more amazing results in control
theory. It decouples the process of designing an optimal filter from
designing an optimal controller, enabling simplicity and modularity in
control design. This decoupling where we treat the output of our state
estimator as the true state is yet another example of certainty
equivalence, and yet another example of where certainty equivalence
turns out to be optimal. However, as we will now see, LQG highlights a
particular scenario where certainty equivalent control leads to
misplaced optimism about robustness.

Before presenting the example, let's first dive into \emph{why} LQG is
likely less robust than LQR. Let's assume that the true dynamics are
generated as: \[
    X_{t+1} = AX_t + B_\star U_t + W_t \,,
\] though we computed the optimal controller with the matrix~\(B\).
Define an error signal,~\(E_t = X_t - \hat{x}_t\), that measures the
current deviation between the actual state and the estimate. Then, using
the fact that~\(u_t = -K \hat{x}_t\), we get the closed loop dynamics \[
\small
 \begin{bmatrix}
        \hat{X}_{t+1}\\
        E_{t+1}
    \end{bmatrix} = \begin{bmatrix} A-BK & LC\\ (B-B_\star) K & A-LC \end{bmatrix}\begin{bmatrix}
        \hat{X}_t\\
        E_t
    \end{bmatrix} +
    \begin{bmatrix} LV_t\\ W_t-LV_t \end{bmatrix}\,.
\] When~\(B=B_\star\), the bottom left block is equal to zero. The
system is then stable provided~\(A-BK\) and~\(A-LC\) are both stable
matrices (i.e., have eigenvalues with magnitude less than one). However,
small perturbations in the off-diagonal block can make the matrix
unstable. For intuition, consider the matrix \[
 \begin{bmatrix} 0.9 & 1\\ 0 & 0.8 \end{bmatrix}\,.
\] The eigenvalues of this matrix are~\(0.9\) and~\(0.8\), so the matrix
is clearly stable. But the matrix \[
 \begin{bmatrix} 0.9 & 1\\ t & 0.8 \end{bmatrix}
\] has an eigenvalue greater than zero if~\(t>0.02\). So a tiny
perturbation significantly shifts the eigenvalues and makes the matrix
unstable.

Similar things happen in LQG. Let's return to our simple dynamics
inspired by Newton's Laws of Motion \[
    A = \begin{bmatrix} 1 & 1\\ 0 & 1\end{bmatrix} \,,\quad B = \begin{bmatrix} 0\\1 \end{bmatrix}\,, \quad C= \begin{bmatrix} 1 & 0\end{bmatrix}
\] And let's use \emph{any} cost matrices~\(\Phi\) and~\(\Psi\). We
assume that the noise variances are \[
    \mathbb{E}\left[W_t W_t^\top\right]=\begin{bmatrix} 1 & 2 \\ 2 & 4\end{bmatrix} \,,\quad \mathbb{E}\left[V_t^2\right]=\sigma^2
\]

The open loop system here is unstable, having two eigenvalues at~\(1\).
We can stabilize the system only by modifying the second state. The
state disturbance is aligned along the direction of the
vector~\([1/2;1]\), and the state cost only penalizes states aligned
with this disturbance. The SDM goal is simply to remove as much signal
as possible in the~\([1;1]\) direction without using large inputs. We
only are able to measure the first component of the state, and this
measurement is corrupted by Gaussian noise.

What does the optimal policy look like? Perhaps unsurprisingly, it
focuses all of its energy on ensuring that there is little state signal
along the disturbance direction. The optimal~\(L\) matrix is \[
    L=\begin{bmatrix} 3-d_1 \\ 2-d_2 \end{bmatrix}\,.
\] where~\(d_1\) and~\(d_2\) are small positive numbers that go to zero
as \(\sigma\) goes to zero. The optimal~\(K\) will have positive
coefficients whenever we choose for~\(\Phi\) and~\(\Psi\) to be positive
semidefinite: if \(K\) has a negative entry, it will necessarily not
stabilize~\((A,B)\).

Now what happens when we have model mismatch? Let's assume for
simplicity that~\(\sigma=0\). If we set~\(B_\star=tB\) and use the
formula for the closed loop above, we see that closed loop state
transition matrix is \[
A_{cl}=\begin{bmatrix}
1 & 1 & 3 & 0\\
    -k_1 & 1-k_2  & 2 & 0\\
    0 & 0 &-2 &1\\
    k_1(1-t) & k_2(1-t) & -2 & 1
    \end{bmatrix}\,.
\]

It's straight forward to check that when~\(t=1\) (i.e., no model
mismatch), the eigenvalues of~\(A-BK\) and~\(A-LC\) all have real parts
with magnitude less than or equal to~\(1\). For the full closed loop
matrix, analytically computing the eigenvalues themselves is a pain, but
we can prove instability by looking at a characteristic polynomial. For
a matrix to have all of its eigenvalues in the left half plane, its
characteristic polynomial necessarily must have all positive
coefficients. If we look at the linear term in the characteristic
polynomial, of~\(-I-A_{cl}\) we see that if~\(t>1\),~\(A_{cl}\) must
have an eigenvalue with real part less than~\(-1\), and hence the closed
loop is unstable. This is a very conservative condition, and we could
get a tighter bound if we'd like, but it's good enough to reveal some
paradoxical properties of LQG. The most striking is that if we build a
sensor that gives us a better and better measurement, our system becomes
more and more fragile to perturbation and model mismatch. For machine
learning scientists, this seems to go against all of our training. How
can a system become \emph{less} robust if we improve our sensing and
estimation?

Let's look at the example in more detail to get some intuition for
what's happening. When the sensor noise gets small, the optimal Kalman
Filter is more aggressive. The filter rapidly damps any errors in the
disturbance direction~\([1;1/2]\) and, as~\(\sigma\) decreases, it damps
the \([1;1]\) direction less. When~\(t \neq 1\),~\(B-B_\star\) is
aligned in the \([0;1]\) and can be treated as a disturbance signal.
This undamped component of the error is fed errors from the state
estimate, and these errors compound each other. Since we spend so much
time focusing on our control along the direction of the injected state
noise, we become highly susceptible to errors in a different direction
and these are the exact errors that occur when there is a gain mismatch
between the model and reality.

The fragility of LQG has many takeaways. It highlights that noiseless
state measurement can be a dangerous modeling assumption, because it is
then optimal to trust our model too much. Model mismatch must be
explicitly accounted for when designing the decision making policies.

This should be a cautionary tale for modern AI systems. Most papers in
reinforcement learning consider MDPs where we perfectly measure the
system state. Building an entire field around optimal actions with
perfect state observation builds too much optimism. Any realistic
scenario is going to have partial state observation, and such problems
are much thornier.

A second lesson is that it is not enough to just improve the prediction
components in feedback systems that are powered by machine learning.
Improving prediction will increase sensitivity to a modeling errors in
some other part of the engineering pipeline, and these must all be
accounted for together to ensure safe and successful decision making.

\hypertarget{chapter-notes-11}{%
\section{Chapter notes}\label{chapter-notes-11}}

This chapter and the previous chapter overlap significantly with a
survey of reinforcement learning by Recht, which contains additional
connections to continuous control.\citep{RechtRLSurvey}

Bertsekas has written several valuable texts on reinforcement learning
from different perspectives. His seminal book with Tsitsiklis
established the mathematical formalisms of Neurodynamic Programming that
most resemble contemporary reinforcement
learning.\citep{bertsekas1996neuro} The second volume of his Dynamic
Programming Book covers many of the advanced topics in approximate
dynamic programming and infinite horizon dynamic
programming.\citep{BertsekasDPBook2} And his recent book on
reinforcement learning builds ties with his earlier work and recent
advances in reinforcement learning post AlphaGo.\citep{BertsekasRLBook}

For more on bandits from a theoretical perspective, the reader is
invited to consult the comprehensive book by Lattimore and
Szepesvari.\citep{LattimoreBanditBook} Agarwal et al.~provide a thorough
introduction to the theoretical aspects of reinforcement learning from
the perspective of learning theory.\citep{RLTheoryBook}

The control theoretic perspective on reinforcement learning is called
\emph{dual control}. Its originated at a similar time to reinforcement
learning, and many attribute Feldbaum's work as the origin
point.\citep{feldbaum1960dual} Wittenmark surveys the history of this
topic, its limitations, and its comparison to certainty equivalence
methods.\citep{wittenmark1995adaptive} For further exploration of the
limits of classical optimal control and how to think about robustness,
Gunter Stein's ``Respect the Unstable'' remains a classic lecture on the
subject.\citep{stein2003respect}

\chapter{Epilogue}

Unknown outcomes often follow patterns found in past observations. But
when do they not? As powerful as statistical patterns are, they are not
without limitations. Every discipline built on the empirical law also
experiences its failure.

In fact, Halley's contemporaries already bore witness. Seeking to
increase revenue still despite the sale of life annuities, King William
III desired to tax his citizens in proportion to their wealth. An income
tax appeared too controversial and unpopular with his constituents so
that the king's advisors had to come up with something else. In 1696,
the king introduced a property tax based on the number of windows in a
house. It stands to reason that the wealth of a family correlated
strongly with the number of windows in their home. So, the window tax
looked quite reasonable from a statistical perspective.

Although successful on the whole and adopted by many other countries,
the window tax had a peculiar side effect. People adjusted.
Increasingly, houses would have bricked-up window spaces. In Edinburgh
an entire row of houses featured no bedroom windows at all. The
correlation between the number of windows and wealth thus deteriorated.

The problem with the window tax foretold a robust limitation of
prediction. Datasets display a static snapshot of a population.
Predictions on the basis of data are accurate only under an unspoken
stability assumption. Future observations must follow the same
data-generating process. It's the ``more of the same'' principle that we
call generalization in supervised learning.

However, predictions often motivate consequential actions in the real
world that change the populations we observe. Chemist and technology
critic Ursula Franklin summarizes the problem aptly in her 1989 book
called The Real World of Technology:

\begin{quote}
{[}T{]}echnologies are developed and used within a particular social,
economic, and political context. They arise out of a social structure,
they are grafted on to it, and they may reinforce it or destroy it,
often in ways that are neither foreseen nor
foreseeable.\citep{franklin1999real}
\end{quote}

Franklin continues:

\begin{quote}
{[}C{]}ontext is not a passive medium but a dynamic counterpart. The
responses of people, individually, and collectively, and the responses
of nature are often underrated in the formulation of plans and
predictions.
\end{quote}

Franklin understood that predictions are not made in a vacuum. They are
agents of change through the actions they prompt. Decisions are always
part of an evolving environment. It's this dynamic environment that
determines the merit of a decision.

Predictions can fail catastrophically when the underlying population is
subject to unmodeled changes. Even benign changes to a population,
sometimes called distribution shift, can sharply degrade the utility of
statistical models. Numerous results in machine learning are testament
to the fragility of even the best performing models under changing
environments.

Other disciplines have run into the same problem. In his influential
critique from 1976, economist Robert Lucas argued that patterns found in
historical macroeconomic data are an inadequate basis of policy making,
since any policy would inevitably perturb those statistical patterns.
Subsequently, economists sought to ground macroeconomics in the
microeconomic principles of utility theory and rational behavior of the
individual, an intellectual program known as microfoundations dominant
to this day. The hope was that microfoundations would furnish a more
reliable basis of economic policy making.

It is tempting to see dynamic modeling as a possible remedy to the
problem Lucas describes. However, Lucas critique \emph{was} about
dynamic models. Macroeconomists at the time were well aware of dynamic
programming and optimal control. A survey of control-theoretic tools in
macroeconomics from 1976 starts with the lines:

\begin{quote}
In the past decade, a number of engineers and economists have asked the
question: ``If modern control theory can improve the guidance of
airplanes and spacecraft, can it also help in the control of inflation
and unemployment?''\citep{kendrick1976applications}
\end{quote}

If anything, the 60s and 70s had been the heyday of dynamic modeling.
Entire disciplines, such as \emph{system dynamics}, attempted to create
dynamic models of complex social systems, such as, corporations, cities,
and even the western industrial world. Proponents of system dynamics
used simulations of these models to motivate consequential policy
propositions. Reflecting on these times, economist Schumacher wrote in
1973:

\begin{quote}
There have never been so many futurologists, planners, forecasters, and
model-builders as there are today, and the most intriguing product of
technological progress, the computer, seems to offer untold new
possibilities. {[}\ldots{]} Are not such machines just what we have been
waiting for?\citep{schumacher2011small}
\end{quote}

It was not the lack of dynamic models that Lucas criticized, it was the
fact that policy may invalidate the empirical basis of the model. Lucas'
critique puts pressure how we come to \emph{know} a model. Taking action
can invalidate not just a particular model but also disrupt the social
and empirical facts from which we derived the model.

If economics reckoned with this problem decades ago, it's worth taking a
look at how the field has developed since. Oversimplifying greatly, the
ambitious macroeconomic theorizing of the 20th century gave way to a
greater focus on microeconomics and empirical work. Field experiments
and causal inference, in particular, are now at the forefront of
economic research.

Fundamental limitations of dynamic models not only surfaced in
economics, they were also called out by control theorists themselves. In
a widely heralded plenary lecture at the 1989 IEEE Conference on
Decision and Control, Gunter Stein argued against ``the increasing
worship of abstract mathematical results in control at the expense of
more specific examinations of their practical, physical consequences.''
Stein warned that mathematical and algorithmic formulations often elided
fundamental physical limitations and trade-offs that could lead to
catastrophic consequences.

Unstable systems illustrate this point. A stable system has the property
that no matter how you disturb the system, it will always come back to
rest. If you heat water on the stove, it will always eventually return
to room temperature. An unstable system on the other hand can evolve
away from a natural equilibrium exponentially quickly, like a contagious
pathogen. From a computational perspective, however, there is no more
difficulty in mathematically solving a sequential decision making
problem with unstable dynamics than in solving one with stable dynamics.
We can write down and solve decision making problems in both cases, and
they appear to be of equal computational difficulty. But in reality,
unstable systems are dangerous in a way that stable systems are not.
Small errors get rapidly amplified, possibly resulting in catastrophe.
Likely the most famous such catastrophe is the Chernobyl disaster, which
Stein described as the failure to ``respect the unstable'' inherent in
the reactor design.

As the artificial intelligence and machine learning communities
increasingly embrace dynamic modeling, they will inevitably relearn
these cautionary lessons of days past.

\hypertarget{beyond-pattern-classification}{%
\section{Beyond pattern
classification?}\label{beyond-pattern-classification}}

Part of the recent enthusiasm for causality and reinforcement learning
stems from the hope that these formalisms might address some of the
inherent issues with the static pattern classification paradigm. Indeed,
they might. But neither causality nor reinforcement learning are a
panacea. Without hard earned substantive domain knowledge to guide
modeling and mathematical assumptions, there is little that sets these
formalisms apart from pattern classification. The reliance on subject
matter knowledge stands in contrast with the nature of recent advances
in machine learning that largely did without---and that was the point.

Looking ahead, the space of machine learning beyond pattern
classification is full of uncharted territory. In fact, even the basic
premise that there is such a space is not entirely settled.

Some argue that as a practical matter machine learning will proceed in
its current form. Those who think so would see progress coming from
faster hardware, larger datasets, better benchmarks, and increasingly
clever ways of reducing new problems to pattern classification. This
position isn't unreasonable in light of historical or recent
developments. Pattern classification has reemerged several times over
the past 70 years, and each time it has shown increasingly impressive
capabilities.

We can try to imagine what replaces pattern recognition when it falls
out of favor. And perhaps we can find some inspiration by returning one
last time to Edmund Halley. Halley is more well-known for astronomy than
for his life table. Much of astronomy before the 17th century was more
similar to pattern recognition than fundamental physics. Halley himself
had used curve-fitting methods to predict the paths of comets, but found
notable errors in his predictions for the comet Kirch. He discussed his
calculations with Isaac Newton, who solved the problem by establishing a
fundamental description of the laws of gravity and motion. Halley, so
excited by these results, paid to publish Newton's magnum opus
\emph{Philosophiæ Naturalis Principia Mathematica}.

Even if it may not be physics once again or on its own, similarly
disruptive conceptual departures from pattern recognition may be viable
and necessary for machine learning to become a safe and reliable
technology in our lives.

We hope that our story about machine learning was helpful to those who
aspire to write its next chapters.

\chapter{Mathematical background}

The main mathematical tools of machine learning are optimization and
statistics. At their core are concepts from multivariate calculus and
probability. Here, we briefly review some of the concepts from calculus
and probability that we will frequently make use of in the book.

\hypertarget{common-notation}{%
\section{Common notation}\label{common-notation}}

\begin{itemize}
\tightlist
\item
  Lowercase letters \(u, v, w, x, y, z,\) typically denote vectors. We
  use both \(\langle u, v\rangle\) and \(u^T v\) to denote the inner
  product between vectors \(u\) and \(v\).
\item
  Capital letters \(X, Y, Z\) typically denote random variables.
\item
  The conditional probability \(\Pr[A \mid B]\) of an event \(A\)
  conditional on an event \(B\)
\item
  The gradient \(\nabla f(x)\) of a function
  \(f\colon \mathbb{R}^d\to\mathbb{R}\) at a point \(x\in\R^d\) refers
  to the vector of partial derivatives of \(f\) evaluated at \(x\).
\item
  Identity matrix \(\Id\)
\item
  The first \(k\) positive integers \([k]=\{1, 2, \dots, k\}.\)
\end{itemize}

\hypertarget{multivariable-calculus-and-linear-algebra}{%
\section{Multivariable calculus and linear
algebra}\label{multivariable-calculus-and-linear-algebra}}

\hypertarget{positive-definite-matrices}{%
\subsection{Positive definite
matrices}\label{positive-definite-matrices}}

Positive definite matrices are central to both optimization algorithms
and statistics. In this section, we quickly review some of the core
properties that we will use throughout the book.

A matrix~\(M\) is \emph{positive definite} (pd) if it is
symmetric~\(M = M^T\) and~\(z^T M z > 0\) for all nonzero~\(z\in \R^d\).
We denote this as \(M \succ 0\) A matrix~\(M\) is \emph{positive
semidefinite} (psd) if it is symmetric and~\(z^T M z \geq 0\) for all
nonzero~\(z\). We denote this as \(M \succeq 0\). All pd matrices are
psd, but not vice versa.

Some of the main properties of positive semidefinite matrices include.

\begin{enumerate}
\def\labelenumi{\arabic{enumi}.}
\item
  If~\(M_1 \succeq 0\), and~\(M_2 \succeq 0\),
  then~\(M_1 + M_2 \succeq 0\).
\item
  \(a \in \R\),~\(a\geq 0\) implies~\(aM \succeq 0\).
\item
  For any matrix~\(F\),~\(FF^T\) and~\(F^TF\) are both psd. Conversely,
  if \(M\) is psd there exists an~\(F\) such that~\(M = FF^T\).
\end{enumerate}

Note that (1) and (2) still hold if ``psd'' is replaced with ``pd.''''
That is, the sum of two pd matrices is pd. And multiplying a pd matrix
by a positive scalar preserves positive definiteness.

Recall that~\(\lambda\) is a eigenvalue of a square matrix~\(M\) if
there exists a nonzero~\(x\in \R^d\) such that~\(Mx = \lambda x\).
Eigenvalues of psd matrices are all non-negative. Eigenvalues of pd
matrices are all positive. This follows by multiplying the
equation~\(Ax = \lambda x\) on the left by~\(x^T\).

\hypertarget{gradients-taylors-theorem-and-infinitesimal-approximation}{%
\subsection{Gradients, Taylor's Theorem and infinitesimal
approximation}\label{gradients-taylors-theorem-and-infinitesimal-approximation}}

Let~\(\Phi:\R^d\rightarrow \R\). Recall from multivariable calculus that
the \emph{gradient}\index{gradient} of~\(\Phi\) at a point~\(w\) is the
vector of partial derivatives \[
    \nabla \Phi(w) = \begin{bmatrix} \frac{\partial \Phi(w)}{\partial x_1} \\ \frac{\partial \Phi(w)}{\partial x_2} \\ \vdots \\ \frac{\partial \Phi(w)}{\partial x_d} \end{bmatrix}\,.
\] Sometimes we write~\(\nabla_x\Phi(w)\) to make clear which functional
argument we are referring to.

One of the most important theorems in calculus is \emph{Taylor's
Theorem}, which allows us to approximate smooth functions by simple
polynomials. The following simplified version of Taylor's Theorem is
used throughout optimization. This form of Taylor's theorem is sometimes
called the multivariable mean-value theorem. We will use this at
multiple points to analyze algorithms and understand the local
properties of functions.\index{Taylor's theorem}

\begin{Theorem}

\textbf{Taylor's Theorem.}

\begin{itemize}
\item
  If~\(\Phi\) is continuously differentiable, then, for some
  \(t \in [0,1]\,,\)
  \[\Phi(w) = \Phi(w_0) + \nabla \Phi(tw + (1-t)w_0)^T (w - w_0)\,.\]
\item
  If~\(\Phi\) is twice continuously differentiable, then \[
  \nabla \Phi(w) = \nabla \Phi(w_0) + \int_0^1 \nabla^2 \Phi(tw + (1-t)w_0)(w - w_0) \mathrm{d}t
  \] and, for some~\(t \in [0,1]\) \[
  \begin{aligned}
  \Phi(w) &= \Phi(w_0) + \nabla \Phi(w_0)^T(w - w_0) \\
  &\qquad\qquad\quad+ \frac{1}{2} (w-w_0)^T \nabla^2 \Phi(tw + (1-t)w_0)^T (w - w_0)\,.
  \end{aligned}
  \]
\end{itemize}

\end{Theorem}

Taylor's theorem can be used to understand the local properties of
functions. For example, \[
  \Phi(w+ \epsilon v) = \Phi(w) + \epsilon \nabla \Phi(w)^Tv + \frac{\epsilon^2}{2} v^T \nabla^2 \Phi(w + \delta v)^T v
\] for some~\(0\leq \delta \leq \epsilon\). This expression states that
\[
  \Phi(w+ \epsilon v) = \Phi(w) + \epsilon \nabla \Phi(w)^Tv + \Theta(\epsilon^2)\,,
\] So to first order, we can approximate~\(\Phi\) by a linear function.

\hypertarget{jacobians-and-the-multivariate-chain-rule}{%
\subsection{Jacobians and the multivariate chain
rule}\label{jacobians-and-the-multivariate-chain-rule}}

The matrix of first order partial derivatives of a multivariate mapping
\(\Phi\colon\R^n\to\R^m\) is called \emph{Jacobian
matrix}.\index{Jacobian} We define the Jacobian of~\(\Phi\) with respect
to a variable~\(x\) evaluated at a value~\(w\) as the~\(m\times n\)
matrix \[
\jac_x \Phi(w)
= \left[
  \frac{\partial \Phi_i(w)}{\partial x_j}
\right]_{i=1\dots m, j=1\dots n}\,.
\] The~\(i\)-th row of the Jacobian therefore corresponds to the
transpose of the familiar gradient~\(\nabla_x^T \Phi_i(w)\) of
the~\(i\)-th coordinate of~\(\Phi\). In particular, when~\(m=1\) the
Jacobian corresponds to the transpose of the gradient.

The first-order approximation given by Taylor's theorem directly extends
to multivariate functions via the Jacobian matrix. So does the
\emph{chain rule} from calculus for computing the derivatives of
function compositions.\index{chain rule}

Let~\(\Phi\colon\R^n\to\R^m\) and~\(\Psi\colon\R^m\to\R^k\). Then, we
have \[
\jac_x \Psi\circ\Phi(w)
= \jac_{\Phi(w)}\Psi(\Phi(w))\jac_x\Phi(w)\,.
\]

As we did with the gradient notation, when the variable~\(x\) is clear
from context we may drop it from our notation and write~\(\jac\Phi(w)\)

\hypertarget{probability}{%
\section{Probability}\label{probability}}

Contemporary machine learning uses probability as its primary means of
quantifying uncertainty. Here we review some of the basics we will make
use of in this course. This will also allow us to fix
notation.\index{probability}

We note that often times, mathematical rigor gets in the way of
explaining concepts. So we will attempt to only introduce mathematical
machinery when absolutely necessary.

Probability is a function on sets. Let~\(\mathcal{X}\) denote the sample
set. For every~\(A\subset \mathcal{X},\) we have \[
  0 \leq \Pr[A] \leq 1\,, \qquad\qquad \Pr[\mathcal{X}]=1\,, \qquad \qquad \Pr[\emptyset] = 0\,,
\] and \[
\Pr[A\cup B] + \Pr[A\cap B] = \Pr[A] + \Pr[B]\,.
\] This implies that \[
  \Pr[A\cup B] = \Pr[A] + \Pr[B]\,.
\] if and only if~\(\Pr[A\cap B]=0\). We always have the inequality \[
  \Pr[A\cup B] \leq \Pr[A] + \Pr[B]\,.
\] By induction, we get the union bound \[
  \textstyle\Pr\left[\bigcup_{i} A_i\right] \leq \sum_i \Pr[A_i]\,.
\]

\hypertarget{random-variables-and-vectors}{%
\subsection{Random variables and
vectors}\label{random-variables-and-vectors}}

Random variables are a particular way of characterizing outcomes of
random processes. We will use capital letters like~\(X\),~\(Y\),
and~\(Z\) to denote such random variables. The sample space of a random
variable will be the set where a variable can take values. Events are
simply subsets of possible values. Common examples we will encounter in
this book are

\begin{itemize}
\item
  \textbf{Probability that a random variable has a particular value}.
  This will be denoted as~\(\Pr[X=x]\). Note here that we use a lower
  case letter to denote the value that the random variable might take.
\item
  \textbf{Probability that a random variable satisfies some inequality}.
  For example, the probability that~\(X\) is less than a scalar~\(t\)
  will be denoted as~\(\Pr[X\leq t]\).
\end{itemize}

A \emph{random vector} is a random variable whose sample space consists
of \(R^d\). We will not use notation to distinguish between vectors and
scalars in this text.

\hypertarget{densities}{%
\subsection{Densities}\label{densities}}

Random vectors are often characterized by \emph{probability
densities}\index{probability!density} rather than by probabilities. The
density~\(p\) of a random variable~\(X\) is defined by its relation to
probabilities of sets: \[
  \Pr[X \in A] = \int_{x\in A} p(x) \dif x\,.
\]

\hypertarget{expectations}{%
\subsection{Expectations}\label{expectations}}

If~\(f\) is a function on~\(R^d\) and~\(X\) is a random vector, then the
expectation\index{expectation} of~\(f\) is given by \[
  \E[f(X)] = \int f(x) p(x) \dif x
\]

If~\(A\) is a set, the \emph{indicator function of the set} is the
function \[
  I_A(x) = \begin{cases} 1 & \text{if}~x \in A \\ 0 & \text{otherwise} \end{cases}
\] Note that the expectation of an indicator function is a probability:
\[
  \E[I_A(X)] = \int_{x \in A} p(x) \dif x = \Pr[X \in A]\,.
\] This expression links the three concepts of expectation, density, and
probability together.

Note that the expectation operator is linear: \[
  \E[af(X)+bg(X)] = a \E[f(X)] + b \E[g(x)] \,.
\]

Two other important expectations are the mean and covariance. The
\emph{mean} of a random variable is the expected value of the identity
function: \[
  \mu_X := \E[X] = \int x p(x) \dif x\,.
\] The \emph{covariance} of a random variable is the
matrix\index{covariance} \[
  \Sigma_X := \E[(X-\mu_X)(X-\mu_X)^T]\,.
\] Note that covariance matrices are positive semidefinite. To see this,
take a nonzero vector~\(z\) and compute \[
z^T \Sigma_X z := \E[z^T(X-\mu_X)(X-\mu_X)^Tz] = \E[ ((X-\mu_X)^Tz)^2]\,.
\] Since the term inside the expectation is nonnegative, the expectation
is nonnegative as well.

\hypertarget{important-examples-of-probability-distributions}{%
\subsection{Important examples of probability
distributions}\label{important-examples-of-probability-distributions}}

\begin{itemize}
\item
  \textbf{Bernoulli random variables.} A Bernoulli random variable~\(X\)
  can take two values,~\(0\) and~\(1\). In such a
  case~\(\Pr[X=1] = 1-\Pr[X=0]\)
\item
  \textbf{Gaussian random vectors.} Gaussian random vectors are the most
  ubiquitous real valued random vectors. Their densities are
  parameterized only by their mean and covariance: \[
  p(x) = \frac{1}{\operatorname{det}(2\pi \Sigma)^{1/2}} \exp\left( -\tfrac{1}{2} (x-\mu_X)^T \Sigma^{-1} (x-\mu_X) \right)\,.
  \] Gaussian random variables are often called ``normal'' random
  variables. We denote the distribution of a normal random variable with
  mean~\(\mu\) and covariance~\(\Sigma\) as \[
  \mathcal{N}(\mu,\Sigma)\,.
  \] The reason Gaussian random variables are ubiquitous is because of
  the central limit theorem: averages of many independent random
  variables tend to look like Gaussian random variables.
\end{itemize}

\hypertarget{conditional-probability-and-bayes-rule}{%
\section{Conditional probability and Bayes'
Rule}\label{conditional-probability-and-bayes-rule}}

Conditional probability is applied quite cavalierly in machine learning.
It's actually very delicate and should only be applied when we really
know what we're doing.\index{probability!conditional}\index{Bayes' Rule}

\[
  \Pr[A|B] = \frac{\Pr[A\cap B]}{\Pr[B]}
\]

\(A\) and~\(B\) are said to be \emph{independent}
if~\(\Pr[A|B] = \Pr[A]\). Note that from the definition of conditional
probability~\(A\) and~\(B\) are independent if and only if \[
  \Pr[A\cap B] = \Pr[A]\Pr[B]\,.
\]

Bayes' Rule is an immediate corollary of the definition of conditional
probability. In some sense, it's just a restatement of the definition.
\[
\Pr[A|B] = \frac{\Pr[B|A]\Pr[A]}{\Pr[B]}
\] This is commonly applied when~\(A\) is one of a set of several
alternatives. Suppose~\(A_i\) are a collection of disjoint sets such
that \(\cup_i A_i = \mathcal{X}\) then for each~\(i\), Bayes' Rule
states \[
  \Pr[A_i|B] = \frac{\Pr[B|A_i] \Pr[A_i]}{\sum_j \Pr[B|A_j] \Pr[A_j]}\,.
\] This shows that if we have models of the likelihood of~\(B\) under
each alternative~\(A_i\) and if we have beliefs about the probability of
each \(A_i\), we can compute the probability of observing~\(A_i\) under
the condition that~\(B\) has occurred.

\hypertarget{conditional-densities}{%
\subsection{Conditional densities}\label{conditional-densities}}

Suppose~\(X\) and~\(Z\) are random variables whose joint distribution is
continuous. If we try to write down the conditional distribution
for~\(X\) given~\(Z=z\), we find \[
  \Pr[X\in A | Z=z] = \frac{  \Pr[X\in A \cap Z=z] } {\Pr[Z=z]}
\] Both the numerator and denominator are equal to zero. In order to
have a useful formula, we can appeal to densities. \[
\begin{aligned}
  \Pr[x \in A  | z \leq Z \leq z+\epsilon ] &= \frac{  \int_{z}^{z+\epsilon} \int_{x \in A} p(x,z') \dif x \dif z' } {\int_{z}^{z+\epsilon}  p(z') \dif z' }\\
  &\approx \frac{  \epsilon \int_{x\in A} p(x,z) } { \epsilon  p(z) \dif z } \\
  &= \int_{x\in A} \frac{p(x,z)}{p(z)} \dif x
\end{aligned}
\] Letting~\(\epsilon\) go to zero, this calculation shows that we can
use the \emph{conditional density} to compute the conditional
probabilities of \(X\) when~\(Z=z\): \[
  p(x|z) := \frac{p(x,z)}{p(z)}\,.
\]

\hypertarget{conditional-expectation-and-the-law-of-iterated-expectation}{%
\subsection{Conditional expectation and the law of iterated
expectation}\label{conditional-expectation-and-the-law-of-iterated-expectation}}

Conditional expectation is short hand for computing expected values with
respect to conditional probabilities: \[
   \E[f(x,z)|Z=z] = \int f(x,z) p(x|z) \dif x
\] An important formula is the law of iterated expectation: \[
  \E[f(x,z)] = \E[\E[f(x,z)|Z=z]]
\] This formula follows because \[
\begin{aligned}
  \E[f(x,z)] &= \int\int f(x,z) p(x,z) \dif x \dif z \\
  &= \int\int f(x,z) p(x|z) p(z) \dif x \dif z \\
  &= \int \left(\int f(x,z) p(x|z)\dif x\right) p(z) \dif z \,.
  \end{aligned}
\]

\hypertarget{estimation}{%
\section{Estimation}\label{estimation}}

This book devotes much of its attention to probabilistic decision
making. A different but related statistical problem is \emph{parameter
estimation}. Assuming that data~\(X\) is generated by a statistical
model, we'd like to infer some \emph{nonrandom} property about its
distribution. The most canonical examples here would be estimating the
mean or variance of the distribution. Note that estimating these
parameters has a different flavor than decision theory. In particular,
our framework of risk minimization no longer
applies.\index{estimation}\index{parameter estimation}

If we aim to minimize a functional \[
    \text{minimize}_f~\E[\loss(\vartheta,f(x))]
\] then the optimal choice is to set~\(f(x) = \vartheta\). But we don't
know this parameter in the first place. So we end up with an algorithm
that's not implementable.

Instead, what we do in estimation theory is pose a variety of plausible
estimators that might work for a particular parameter and consider the
efficacy of these parameters in different settings. In particular, we'd
like estimators that take a set of observations~\(S=(x_1,\ldots,x_n)\)
and return a guess for the parameter whose value improves as~\(n\)
increases: \[
    \lim_{n\rightarrow \infty} \E_S[\loss(\vartheta,\hat{\vartheta}(S))] = 0
\]

Even though estimators are constructed from data, their design and
implementation require a good deal of knowledge about the underlying
probability distribution. Because of this, estimation is typically
considered to be part of classical statistics and not machine learning.
Estimation theory has a variety of powerful tools that are aimed at
producing high quality estimators, and is certainly worth learning more
about. We need rudimentary elements of estimation to understand popular
baselines and algorithms in causal inference and reinforcement learning.

\hypertarget{plug-in-estimators}{%
\subsection{Plug-in Estimators}\label{plug-in-estimators}}

We will restrict our attention to \emph{plug-in
estimators}.\index{plug-in estimator} Plug-in estimators are functions
of the moments of probability distributions. They are plug-in because we
replace the true distribution with the empirical distribution. To be
precise, suppose there exist vector valued functions \(g\) and~\(\psi\)
such that~\(\vartheta = g(\E[\psi(x)])\). Then, given a
dataset,~\(S=(x_1,\ldots,x_n)\), the associated plug-in estimator of
\(\vartheta\) is \[
\hat{\vartheta}(S) = g\left( \frac{1}{n} \sum_{i=1}^n \psi(x_i)\right)
\] that is, we replace the expectation with the sample average. There
are canonical examples of plugin estimators.

\begin{enumerate}
\def\labelenumi{\arabic{enumi}.}
\item
  \emph{The sample mean.} The sample mean is the plug-in estimator where
  \(g\) and~\(\psi\) are both the identity functions.
\item
  \emph{The sample covariance.} The sample covariance is \[
   \hat{\Sigma}_x = \sum_{i=1}^n x_i x_i^T -\left(\frac{1}{n}\sum_{i=1}^n x_i \right)\left(\sum_{i=1}^n x_i \right)^T\,.
  \] From this formula, we can take \[
   \psi(x) = \begin{bmatrix} 1\\ x \end{bmatrix}\begin{bmatrix} 1\\ x \end{bmatrix}^T ~~~\text{and}~~~
   g\left(\begin{bmatrix} A & B \\ B^T & C\end{bmatrix} \right) = C-BB^T\,.
  \]
\item
  \emph{Least-squares estimator.} Suppose we have three random vectors,
  \(y\),~\(x\), and~\(v\) and we assume that~\(v\) and~\(x\) are
  zero-mean and uncorrelated and that~\(y=Ax+v\) for some matrix~\(A\).
  Let's suppose we'd like to estimate~\(A\) from a set of pairs
  \(S=( (x_1,y_1), \ldots, (x_n, y_n))\). One can check that \[
   A = \Sigma_{yx} \Sigma_{x}^{-1}\,.
  \] And hence the plug-in estimator would use the sample covariances:
  \[
   \hat{A} = \left( \sum_{i=1}^n y_i x_i^T\right) \left( \sum_{i=1}^n x_i x_i^T\right)^{-1}
  \] In this case, we have the formulation \[
   \psi(x) = \begin{bmatrix} x\\ y \end{bmatrix}\begin{bmatrix} x\\ y \end{bmatrix}^T ~~~\text{and}~~~
   g\left(\begin{bmatrix} A & B \\ B^T & C\end{bmatrix} \right) = B A^{-1}\,.
  \]
\end{enumerate}

\hypertarget{convergence-rates}{%
\subsection{Convergence rates}\label{convergence-rates}}

In our study of generalization, we reasoned that the empirical risk
should be close to the true risk because sample averages should be close
to population values. A similar reasoning holds true for plug-in
estimators: smooth functions of sample averages should be close to their
population counterparts.

We covered the case of the sample mean in our discussion of
generalization. To recall, suppose~\(x\) is a Bernoulli random variable
with mean~\(p\). Let~\(x_1,\ldots,x_n\) be independent and identically
distributed as~\(x\). Then Hoeffding's inequality states that \[
    \Pr\left[ \left| \frac{1}{n}\sum_{i=1}^n x_i - p \right| > \epsilon \right] \leq 2 \exp(-2n \epsilon^2)\,.
\] Or, in other words, with probability~\(1-\delta\), \[
     \left| \frac{1}{n}\sum_{i=1}^n x_i - p \right| \leq  \sqrt{\frac{\log(2/\delta)}{2n}}\,.
\]

Let's consider a simple least-squares estimator. Suppose we know that
\(y=w^T x + v\) where~\(w\) and~\(x\) are a vectors,~\(w\) is
deterministic, and \(x\) and~\(v\) are uncorrelated. Consider the
least-squares estimator \(\hat{w}_S\) from~\(n\) data points.. The
estimation error in~\(w\) is the vector~\(e_S = \hat{w}_S-w\). The
expectation of~\(e_S\) is zero and the expected norm of the error is
given by \[
    \E\left[ \Vert e_S \Vert^2 \right] = \operatorname{Trace} \left( \left( \sum_{i=1}^n x_i x_i^T \right)^{-1} \right)\,.
\] This error is small if the sample covariance has large eigenvalues.
Indeed, if~\(\lambda_S\) denotes the minimum eigenvalue of the sample
covariance of~\(x\), then \[
    \E\left[ \Vert e_S \Vert^2 \right] \leq \frac{d}{n} \lambda_S\,.
\] This expression suggests that the distribution of~\(x\) must have
density that covers all directions somewhat equally in order for the
least-squares estimator to have good performance. On top of this, we see
that the squared error decreases roughly as~\(d/n\). Hence, we need far
more measurements than dimensions to find a good estimate of~\(w\). This
is in contrast to what we studied in classification. Most of the
generalization bounds for classification we derived were \emph{dimension
free} and only depended on properties like the margin of the data. In
contrast, in parameter estimation, we tend to get results that scale as
number of parameters over number of data points. This rough rule of
thumb that the error scales as the ratio of number of parameters to
number of data points tends to be a good guiding principle when
attempting to understand convergence rates of estimators.

\backmatter

\footnotesize

\bibliographystyle{unsrt}
\bibliography{references}

\begin{thebibliography}{100}

\bibitem{nytimes1958new}
New navy device learns by doing; psychologist shows embryo of computer designed
  to read and grow wiser.
\newblock {\em The New York Times}, 1958.

\bibitem{bellhouse2011new}
David~R. Bellhouse.
\newblock A new look at {H}alley's life table.
\newblock {\em Journal of the Royal Statistical Society: Series A (Statistics
  in Society)}, 174(3):823--832, 2011.

\bibitem{ciecka2008edmond}
James~E. Ciecka.
\newblock {E}dmond {H}alley's life table and its uses.
\newblock {\em Journal of Legal Economics}, 15:65--74, 2008.

\bibitem{pearson1981history}
Karl Pearson and Egon~S. Pearson.
\newblock The history of statistics in the 17th and 18th centuries against the
  changing background of intellectual, scientific and religious thought.
\newblock {\em British Journal for the Philosophy of Science}, 32(2):177--183,
  1981.

\bibitem{hacking2006emergence}
Ian Hacking.
\newblock {\em The emergence of probability: A philosophical study of early
  ideas about probability, induction and statistical inference}.
\newblock Cambridge University Press, 2006.

\bibitem{mindell2002between}
David~A. Mindell.
\newblock {\em Between human and machine: feedback, control, and computing
  before cybernetics}.
\newblock JHU Press, 2002.

\bibitem{kline2015cybernetics}
Ronald~R. Kline.
\newblock {\em The cybernetics moment: Or why we call our age the information
  age}.
\newblock JHU Press, 2015.

\bibitem{heims1991cybernetics}
Steve~J. Heims.
\newblock {\em The cybernetics group}.
\newblock MIT Press, 1991.

\bibitem{beniger2009control}
James Beniger.
\newblock {\em The control revolution: Technological and economic origins of
  the information society}.
\newblock Harvard University Press, 1986.

\bibitem{Michalski83}
Ryszard~S. Michalski, Jamie~G. Carbonell, and Tom~M. Mitchell, editors.
\newblock {\em Machine learning: An artificial intelligence approach}.
\newblock Springer, 1983.

\bibitem{langley2011changing}
Pat Langley.
\newblock The changing science of machine learning, 2011.

\bibitem{liberman10obituary}
Mark Liberman.
\newblock Obituary: {F}red {J}elinek.
\newblock {\em Computational Linguistics}, 36(4):595--599, 2010.

\bibitem{Church18}
Kenneth~Ward Church.
\newblock Emerging trends: A tribute to {C}harles {W}ayne.
\newblock {\em Natural Language Engineering}, 24(1):155–160, 2018.

\bibitem{Liberman20}
Mark Liberman and Charles Wayne.
\newblock Human language technology.
\newblock {\em {AI} Magazine}, 41(2):22--35, 2020.

\bibitem{mcclelland1986parallel}
James~L. McClelland, David~E. Rumelhart, and PDP~Research Group.
\newblock Parallel distributed processing.
\newblock {\em Explorations in the Microstructure of Cognition}, 2:216--271,
  1986.

\bibitem{jordan2019artificial}
Michael~I. Jordan.
\newblock Artificial intelligence—the revolution hasn’t happened yet.
\newblock {\em Harvard Data Science Review}, 1(1), 2019.

\bibitem{benjamin2019race}
Ruha Benjamin.
\newblock {\em Race after Technology}.
\newblock Polity, 2019.

\bibitem{hutchinson201950}
Ben Hutchinson and Margaret Mitchell.
\newblock 50 years of test (un) fairness: Lessons for machine learning.
\newblock In {\em Conference on Fairness, Accountability, and Transparency},
  pages 49--58, 2019.

\bibitem{barocas-hardt-narayanan}
Solon Barocas, Moritz Hardt, and Arvind Narayanan.
\newblock {\em Fairness and Machine Learning}.
\newblock fairmlbook.org, 2019.
\newblock \url{http://www.fairmlbook.org}.

\bibitem{kleinberg2017inherent}
Jon~M. Kleinberg, Sendhil Mullainathan, and Manish Raghavan.
\newblock Inherent trade-offs in the fair determination of risk scores.
\newblock In {\em Innovations in Theoretical Computer Science}, 2017.

\bibitem{chouldechova2017fair}
Alexandra Chouldechova.
\newblock Fair prediction with disparate impact: A study of bias in recidivism
  prediction instruments.
\newblock {\em Big data}, 5(2):153--163, 2017.

\bibitem{angwin2016machine}
Julia Angwin, Jeff Larson, Surya Mattu, and Lauren Kirchner.
\newblock Machine bias.
\newblock {\em ProPublica}, May 2016.

\bibitem{dieterich16compas}
William Dieterich, Christina Mendoza, and Tim Brennan.
\newblock {COMPAS} risk scales: Demonstrating accuracy equity and predictive
  parity, 2016.

\bibitem{neyman1928use}
Jerzy Neyman and Egon~S. Pearson.
\newblock On the use and interpretation of certain test criteria for purposes
  of statistical inference: Part {I}.
\newblock {\em Biometrika}, pages 175--240, 1928.

\bibitem{neyman1933}
Jerzy Neyman and Egon~S. Pearson.
\newblock On the problem of the most efficient tests of statistical hypotheses.
\newblock {\em Philosophical Transactions of the Royal Society of London.
  Series A}, 231(694-706):289--337, 1933.

\bibitem{wald1939}
Abraham Wald.
\newblock Contributions to the theory of statistical estimation and testing
  hypotheses.
\newblock {\em The Annals of Mathematical Statistics}, 10(4):299--326, 1939.

\bibitem{bt-probability-book}
Dimitri~P. Bertsekas and John~N. Tsitsiklis.
\newblock {\em Introduction to Probability}.
\newblock Athena Scientific, $2$nd edition, 2008.

\bibitem{peterson1954}
W.~Wesley Peterson, Theodore~G. Birdsall, and William~C. Fox.
\newblock The theory of signal detectability.
\newblock {\em Transactions of the {IRE}}, 4(4):171--212, 1954.

\bibitem{tanner1954}
Wilson~P. {Tanner Jr.} and John~A. Swets.
\newblock A decision-making theory of visual detection.
\newblock {\em Psychological Review}, 61(6):401, 1954.

\bibitem{chow1957optimum}
Chao~Kong Chow.
\newblock An optimum character recognition system using decision functions.
\newblock {\em {IRE} Transactions on Electronic Computers}, (4):247--254, 1957.

\bibitem{highleyman1962linear}
Wilbur~H. Highleyman.
\newblock Linear decision functions, with application to pattern recognition.
\newblock {\em Proceedings of the {IRE}}, 50(6):1501--1514, 1962.

\bibitem{broussard2018artificial}
Meredith Broussard.
\newblock {\em Artificial unintelligence: How computers misunderstand the
  world}.
\newblock MIT Press, 2018.

\bibitem{eubanks2018automating}
Virginia Eubanks.
\newblock {\em Automating inequality: How high-tech tools profile, police, and
  punish the poor}.
\newblock St. Martin's Press, 2018.

\bibitem{noble2018algorithms}
Safiya~Umoja Noble.
\newblock {\em Algorithms of oppression: How search engines reinforce racism}.
\newblock {NYU} Press, 2018.

\bibitem{oneil2016weapons}
Cathy O'Neil.
\newblock {\em Weapons of math destruction: How big data increases inequality
  and threatens democracy}.
\newblock Broadway Books, 2016.

\bibitem{rosenblatt58theperceptron}
Frank Rosenblatt.
\newblock The perceptron: A probabilistic model for information storage and
  organization in the brain.
\newblock {\em Psychological Review}, pages 65--386, 1958.

\bibitem{kearns1994toward}
Michael~J. Kearns, Robert~E. Schapire, and Linda~M. Sellie.
\newblock Toward efficient agnostic learning.
\newblock {\em Machine Learning}, 17(2-3):115--141, 1994.

\bibitem{Novikoff1962}
Albert B.~J. Novikoff.
\newblock On convergence proofs on perceptrons.
\newblock In {\em Symposium on the Mathematical Theory of Automata}, pages
  615--622, 1962.

\bibitem{VapnikChervonenkis1974Book}
Vladimir Vapnik and Alexey Chervonenkis.
\newblock {\em Theory of Pattern Recognition: Statistical Learning Problems}.
\newblock Nauka, 1974.
\newblock In Russian.

\bibitem{rosenblatt1958two}
Frank Rosenblatt.
\newblock {\em Two theorems of statistical separability in the perceptron}.
\newblock United States Department of Commerce, 1958.

\bibitem{rosenblatt1962principles}
Frank Rosenblatt.
\newblock {\em Principles of neurodynamics: Perceptions and the theory of brain
  mechanisms}.
\newblock Spartan, 1962.

\bibitem{block1962perceptron}
Hans-Dieter Block.
\newblock The perceptron: A model for brain functioning.
\newblock {\em Reviews of Modern Physics}, 34(1):123, 1962.

\bibitem{papert1961some}
Seymour~A. Papert.
\newblock Some mathematical models of learning.
\newblock In {\em London Symposium on Information Theory}. Academic Press, New
  York, 1961.

\bibitem{minsky2017perceptrons}
Marvin Minsky and Seymour~A. Papert.
\newblock {\em Perceptrons: An introduction to computational geometry}.
\newblock MIT Press, 2017.

\bibitem{mangasarian1965linear}
Olvi~L. Mangasarian.
\newblock Linear and nonlinear separation of patterns by linear programming.
\newblock {\em Operations Research}, 13(3):444--452, 1965.

\bibitem{Aizerman65}
M.~A. Aizerman, E.~M. Braverman, and L.~I. Rozonoer.
\newblock The {R}obbins-{M}onro process and the method of potential functions.
\newblock {\em Automation and Remote Control}, 26:1882--1885, 1965.

\bibitem{hand2010measurement}
David~J. Hand.
\newblock {\em Measurement Theory and Practice: The World Through
  Quantification}.
\newblock Wiley, 2010.

\bibitem{hand2016measurement}
David~J. Hand.
\newblock {\em Measurement: A very short introduction}.
\newblock Oxford University Press, 2016.

\bibitem{bandalos2018measurement}
Deborah~L. Bandalos.
\newblock {\em Measurement theory and applications for the social sciences}.
\newblock Guilford Publications, 2018.

\bibitem{gitelman2013raw}
Lisa Gitelman.
\newblock {\em Raw data is an oxymoron}.
\newblock MIT Press, 2013.

\bibitem{Angelino2017}
Elaine Angelino, Nicholas Larus-Stone, Daniel Alabi, Margo Seltzer, and Cynthia
  Rudin.
\newblock Learning certifiably optimal rule lists for categorical data.
\newblock {\em Journal of Machine Learning Research}, 18(234):1--78, 2018.

\bibitem{cybenko1989approximation}
George Cybenko.
\newblock Approximation by superpositions of a sigmoidal function.
\newblock {\em Mathematics of Control, Signals and Systems}, 2(4):303--314,
  1989.

\bibitem{Barron93}
Andrew~R. Barron.
\newblock Universal approximation bounds for superpositions of a sigmoidal
  function.
\newblock {\em Transactions on Information Theory}, 39(3):930--945, 1993.

\bibitem{Pisier81}
Gilles Pisier.
\newblock Remarques sur un r\'{e}sultat non publi\'{e} de {B}.~{M}aurey.
\newblock In {\em S\'{e}minaire d'analyse fonctionnelle}. Ecole Polytechnique
  Centre de Mathematiques, 1980-1981.

\bibitem{Jones92}
Lee~K. Jones.
\newblock A simple lemma on greedy approximation in {H}ilbert space and
  convergence rates for projection pursuit regression and neural network
  training.
\newblock {\em Annals of Statistics}, 20(1):608--613, 1992.

\bibitem{breiman1993hinging}
Leo Breiman.
\newblock Hinging hyperplanes for regression, classification, and function
  approximation.
\newblock {\em Transactions on Information Theory}, 39(3):999--1013, 1993.

\bibitem{RahimiRecht07}
Ali Rahimi and Benjamin Recht.
\newblock Random features for large-scale kernel machines.
\newblock In {\em Advances in Neural Information Processing Systems}, 2007.

\bibitem{RahimiRechtNIPS08}
Ali Rahimi and Benjamin Recht.
\newblock Weighted sums of random kitchen sinks: Replacing minimization with
  randomization in learning.
\newblock In {\em Advances in Neural Information Processing Systems}, 2008.

\bibitem{Cho09}
Youngmin Cho and Lawrence~K. Saul.
\newblock Kernel methods for deep learning.
\newblock In {\em Advances in Neural Information Processing Systems}, 2009.

\bibitem{willsky1997signals}
Alan~V. Oppenheim, Alan~S. Willsky, and S.~Hamid Nawab.
\newblock {\em Signals and Systems}.
\newblock Prentice-Hall International, 1997.

\bibitem{candes2008introduction}
Emmanuel~J. Cand{\`e}s and Michael~B. Wakin.
\newblock An introduction to compressive sampling.
\newblock {\em {IEEE} Signal Processing Magazine}, 25(2):21--30, 2008.

\bibitem{karhunen1947lineare}
Kari Karhunen.
\newblock {\"U}ber lineare {M}ethoden in der {W}ahrscheinlichkeitsrechnung.
\newblock {\em Annales Academia Scientiarum Fennica Mathematica, Series A},
  (37):1--47, 1947.

\bibitem{loeve1946functions}
Michel Lo{\`e}ve.
\newblock Functions aleatoire de second ordre.
\newblock {\em Revue Science}, 84:195--206, 1946.

\bibitem{Aronszajn50}
N.~Aronszajn.
\newblock Theory of reproducing kernels.
\newblock {\em Transactions of the American Mathematical Society},
  68(3):337--404, 1950.

\bibitem{parzen1961approach}
Emmanuel Parzen.
\newblock An approach to time series analysis.
\newblock {\em The Annals of Mathematical Statistics}, 32(4):951--989, 1961.

\bibitem{Wahba90}
Grace Wahba.
\newblock {\em Spline Models for Observational Data}.
\newblock SIAM, 1990.

\bibitem{SchoelkopfKernelBook}
Bernhard Sch\"{o}lkopf and Alexander~J. Smola.
\newblock {\em Learning with Kernels: Support Vector Machines, Regularization,
  Optimization, and Beyond}.
\newblock MIT Press, 2002.

\bibitem{shawe2004kernel}
John Shawe-Taylor and Nello Cristianini.
\newblock {\em Kernel methods for pattern analysis}.
\newblock Cambridge University Press, 2004.

\bibitem{PinkusBook}
Allan Pinkus.
\newblock {\em N-widths in approximation theory}.
\newblock Springer, 1985.

\bibitem{akst2019machine}
Jef Akst.
\newblock Machine, learning, 1951.
\newblock {\em The Scientist}, May 2019.

\bibitem{RahimiRechtAllerton08}
Ali Rahimi and Benjamin Recht.
\newblock Uniform approximation of functions with random bases.
\newblock In {\em Allerton Conference on Communication, Control, and
  Computing}, 2008.

\bibitem{daniely2016toward}
Amit Daniely, Roy Frostig, and Yoram Singer.
\newblock Toward deeper understanding of neural networks: The power of
  initialization and a dual view on expressivity.
\newblock In {\em Advances in Neural Information Processing Systems}, 2016.

\bibitem{jacot2018neural}
Arthur Jacot, Franck Gabriel, and Cl{\'e}ment Hongler.
\newblock Neural tangent kernel: Convergence and generalization in neural
  networks.
\newblock In {\em Advances in Neural Information Processing Systems}, pages
  8580--8589, 2018.

\bibitem{decoste2002training}
Dennis Decoste and Bernhard Sch{\"o}lkopf.
\newblock Training invariant support vector machines.
\newblock {\em Machine Learning}, 46(1-3):161--190, 2002.

\bibitem{Shankar20a}
Vaishaal Shankar, Alex Fang, Wenshuo Guo, Sara Fridovich-Keil, Jonathan
  Ragan-Kelley, Ludwig Schmidt, and Benjamin Recht.
\newblock Neural kernels without tangents.
\newblock In {\em International Conference on Machine Learning}, 2020.

\bibitem{robbins1951stochastic}
Herbert Robbins and Sutton Monro.
\newblock A stochastic approximation method.
\newblock {\em The Annals of Mathematical Statistics}, pages 400--407, 1951.

\bibitem{gurbuzbalaban2019random}
Mert G{\"u}rb{\"u}zbalaban, Asu Ozdaglar, and Pablo~A Parrilo.
\newblock Why random reshuffling beats stochastic gradient descent.
\newblock {\em Mathematical Programming}, pages 1--36, 2019.

\bibitem{Nemirovski09}
Arkadi Nemirovski, Antoli Juditsky, Guanghui Lan, and Alexander Shapiro.
\newblock Robust stochastic approximation approach to stochastic programming.
\newblock {\em SIAM Journal on Optimization}, 19(4):1574--1609, 2009.

\bibitem{Du18gradient}
Simon~S. Du, Xiyu Zhai, Barnabas Poczos, and Aarti Singh.
\newblock Gradient descent provably optimizes over-parameterized neural
  networks.
\newblock In {\em International Conference on Learning Representations}, 2019.

\bibitem{Wright2021}
Stephen~J. Wright and Benjamin Recht.
\newblock {\em Optimization for Data Analysis}.
\newblock Cambridge University Press, 2021.

\bibitem{WidrowHoffLMS}
Bernard Widrow and Marcian~E. Hoff.
\newblock Adaptive switching circuits.
\newblock In {\em Institute of Radio Engineers, Western Electronic Show and
  Convention, Convention Record}, pages 96--104, 1960.

\bibitem{NemirovskiYudinBook}
A.~Nemirovski and D.~Yudin.
\newblock {\em Problem complexity and method efficiency in optimization}.
\newblock Wiley, 1983.

\bibitem{Pegasos}
Shai Shalev-Shwartz, Yoram Singer, and Nathan Srebro.
\newblock Pegasos: {P}rimal estimated sub-{G}r{A}dient {SO}lver for {SVM}.
\newblock In {\em International Conference on Machine Learning}, 2007.

\bibitem{nesterov-nemirovskii-ip}
Yurii Nesterov and Arkadi Nemirovskii.
\newblock {\em Interior-point polynomial methods in convex programming}.
\newblock SIAM, 1994.

\bibitem{Liang2020}
Tengyuan Liang, Alexander Rakhlin, and Xiyu Zhai.
\newblock On the multiple descent of minimum-norm interpolants and restricted
  lower isometry of kernels.
\newblock In {\em Conference on Learning Theory}, 2020.

\bibitem{resnet}
Kaiming He, Xiangyu Zhang, Shaoqing Ren, and Jian Sun.
\newblock Deep residual learning for image recognition.
\newblock In {\em Computer Vision and Pattern Recognition}, 2016.

\bibitem{huang2019gpipe}
Yanping Huang, Youlong Cheng, Ankur Bapna, Orhan Firat, Dehao Chen, Mia Chen,
  HyoukJoong Lee, Jiquan Ngiam, Quoc~V Le, Yonghui Wu, and Zhifeng Chen.
\newblock Gpipe: Efficient training of giant neural networks using pipeline
  parallelism.
\newblock {\em Advances in Neural Information Processing Systems}, 32:103--112,
  2019.

\bibitem{vapnik1998SLTBook}
Vladimir Vapnik.
\newblock {\em Statistical Larning Theory}.
\newblock Wiley, 1998.

\bibitem{burges1998tutorial}
Christopher J.~C. Burges.
\newblock A tutorial on support vector machines for pattern recognition.
\newblock {\em Data Mining and Knowledge Discovery}, 2(2):121--167, 1998.

\bibitem{shalev2010learnability}
Shai Shalev-Shwartz, Ohad Shamir, Nathan Srebro, and Karthik Sridharan.
\newblock Learnability, stability and uniform convergence.
\newblock {\em Journal of Machine Learning Research}, 11(Oct):2635--2670, 2010.

\bibitem{bousquet2002stability}
Olivier Bousquet and Andr{\'e} Elisseeff.
\newblock Stability and generalization.
\newblock {\em Journal of Machine Learning Research}, 2(Mar):499--526, 2002.

\bibitem{shalev2014understanding}
Shai Shalev-Shwartz and Shai Ben-David.
\newblock {\em Understanding machine learning: From theory to algorithms}.
\newblock Cambridge University Press, 2014.

\bibitem{belkin2019reconciling}
Mikhail Belkin, Daniel Hsu, Siyuan Ma, and Soumik Mandal.
\newblock Reconciling modern machine-learning practice and the classical
  bias{\textendash}variance trade-off.
\newblock {\em Proceedings of the National Academy of Sciences}, 2019.

\bibitem{neyshabur2014search}
Behnam Neyshabur, Ryota Tomioka, and Nathan Srebro.
\newblock In search of the real inductive bias: On the role of implicit
  regularization in deep learning.
\newblock {\em arXiv:1412.6614}, 2014.

\bibitem{zhang2017understanding}
Chiyuan Zhang, Samy Bengio, Moritz Hardt, Benjamin Recht, and Oriol Vinyals.
\newblock Understanding deep learning requires rethinking generalization.
\newblock In {\em International Conference on Learning Representations}, 2017.

\bibitem{schapire1998boosting}
Robert~E Schapire, Yoav Freund, Peter Bartlett, Wee~Sun Lee, et~al.
\newblock Boosting the margin: A new explanation for the effectiveness of
  voting methods.
\newblock {\em The Annals of Statistics}, 26(5):1651--1686, 1998.

\bibitem{zhang2005boosting}
Tong Zhang and Bin Yu.
\newblock Boosting with early stopping: Convergence and consistency.
\newblock {\em The Annals of Statistics}, 33:1538--1579, 2005.

\bibitem{telgarsky2013margins}
Matus Telgarsky.
\newblock Margins, shrinkage, and boosting.
\newblock In {\em International Conference on Machine Learning}, 2013.

\bibitem{kakade2009complexity}
Sham~M. Kakade, Karthik Sridharan, and Ambuj Tewari.
\newblock On the complexity of linear prediction: Risk bounds, margin bounds,
  and regularization.
\newblock In {\em Advances in Neural Information Processing Systems}, pages
  793--800, 2009.

\bibitem{bartlett2002rademacher}
Peter~L. Bartlett and Shahar Mendelson.
\newblock Rademacher and gaussian complexities: Risk bounds and structural
  results.
\newblock {\em Journal of Machine Learning Research}, 3(Nov):463--482, 2002.

\bibitem{koltchinskii2002empirical}
Vladimir Koltchinskii and Dmitry Panchenko.
\newblock Empirical margin distributions and bounding the generalization error
  of combined classifiers.
\newblock {\em The Annals of Statistics}, 30(1):1--50, 2002.

\bibitem{bartlett98}
Peter~L. Bartlett.
\newblock The sample complexity of pattern classification with neural networks:
  the size of the weights is more important than the size of the network.
\newblock {\em Transactions on Information Theory}, 44(2):525--536, 1998.

\bibitem{hardt2016train}
Moritz Hardt, Benjamin Recht, and Yoram Singer.
\newblock Train faster, generalize better: Stability of stochastic gradient
  descent.
\newblock In {\em International Conference on Machine Learning}, 2016.

\bibitem{LiangRecht2021}
Tengyuan Liang and Benjamin Recht.
\newblock Interpolating classifiers make few mistakes.
\newblock {\em arXiv:2101.11815}, 2021.

\bibitem{neyshabur2017exploring}
Behnam Neyshabur, Srinadh Bhojanapalli, David McAllester, and Nati Srebro.
\newblock Exploring generalization in deep learning.
\newblock In {\em Advances in Neural Information Processing Systems}, pages
  5947--5956, 2017.

\bibitem{dziugaite2017computing}
Gintare~Karolina Dziugaite and Daniel~M Roy.
\newblock Computing nonvacuous generalization bounds for deep (stochastic)
  neural networks with many more parameters than training data.
\newblock {\em arXiv:1703.11008}, 2017.

\bibitem{arora2018stronger}
Sanjeev Arora, Rong Ge, Behnam Neyshabur, and Yi~Zhang.
\newblock Stronger generalization bounds for deep nets via a compression
  approach.
\newblock {\em arXiv:1802.05296}, 2018.

\bibitem{zhang2017theory}
Chiyuan Zhang, Qianli Liao, Alexander Rakhlin, Karthik Sridharan, Brando
  Miranda, Noah Golowich, and Tomaso Poggio.
\newblock Theory of deep learning {III}: Generalization properties of {SGD}.
\newblock Technical report, Discussion paper, Center for Brains, Minds and
  Machines (CBMM). Preprint, 2017.

\bibitem{hardt2021generalization}
Moritz Hardt.
\newblock Generalization in overparameterized models.
\newblock In Tim Roughgarden, editor, {\em Beyond the Worst-Case Analysis of
  Algorithms}, page 486–505. Cambridge University Press, 2021.

\bibitem{lowe2004distinctive}
David~G. Lowe.
\newblock Distinctive image features from scale-invariant keypoints.
\newblock {\em International Journal of Computer Vision}, 60(2):91--110, 2004.

\bibitem{dalal2005histograms}
Navneet Dalal and Bill Triggs.
\newblock Histograms of oriented gradients for human detection.
\newblock In {\em Computer Vision and Pattern Recognition}, 2005.

\bibitem{felzenszwalb2009object}
Pedro~F. Felzenszwalb, Ross~B. Girshick, David McAllester, and Deva Ramanan.
\newblock Object detection with discriminatively trained part-based models.
\newblock {\em {IEEE} Transactions on Pattern Analysis and Machine
  Intelligence}, 32(9):1627--1645, 2009.

\bibitem{auer1996exponentially}
Peter Auer, Mark Herbster, and Manfred~K. Warmuth.
\newblock Exponentially many local minima for single neurons.
\newblock In {\em Advances in Neural Information Processing Systems}, 1996.

\bibitem{vu1998infeasibility}
Van~H. Vu.
\newblock On the infeasibility of training neural networks with small
  mean-squared error.
\newblock {\em Transactions on Information Theory}, 44(7):2892--2900, 1998.

\bibitem{goel21hardness}
Surbhi Goel, Adam Klivans, Pasin Manurangsi, and Daniel Reichman.
\newblock Tight hardness results for training depth-2 {ReLU} networks.
\newblock In {\em Innovations in Theoretical Computer Science}, 2021.

\bibitem{ioffe2015batch}
Sergey Ioffe and Christian Szegedy.
\newblock Batch normalization: Accelerating deep network training by reducing
  internal covariate shift.
\newblock In {\em International Conference on Machine Learning}, pages
  448--456. PMLR, 2015.

\bibitem{wu2018group}
Yuxin Wu and Kaiming He.
\newblock Group normalization.
\newblock In {\em European Conference on Computer Vision}, pages 3--19, 2018.

\bibitem{Baum88}
Eric~B. Baum and David Haussler.
\newblock What size net gives valid generalization?
\newblock In {\em Advances in Neural Information Processing Systems}, 1988.

\bibitem{bartlett1998sample}
Peter~L. Bartlett.
\newblock The sample complexity of pattern classification with neural networks:
  the size of the weights is more important than the size of the network.
\newblock {\em IEEE Transactions on Information Theory}, 44(2):525--536, 1998.

\bibitem{heckel2020compressive}
Reinhard Heckel and Mahdi Soltanolkotabi.
\newblock Compressive sensing with un-trained neural networks: Gradient descent
  finds a smooth approximation.
\newblock In {\em International Conference on Machine Learning}, pages
  4149--4158. PMLR, 2020.

\bibitem{page2020how}
David Page.
\newblock \url{https://myrtle.ai/learn/how-to-train-your-resnet/}, 2020.

\bibitem{griewank2008evaluating}
Andreas Griewank and Andrea Walther.
\newblock {\em Evaluating derivatives: principles and techniques of algorithmic
  differentiation}.
\newblock SIAM, 2nd edition, 2008.

\bibitem{jax2018github}
James Bradbury, Roy Frostig, Peter Hawkins, Matthew~James Johnson, Chris Leary,
  Dougal Maclaurin, George Necula, Adam Paszke, Jake Vander{P}las, Skye
  Wanderman-{M}ilne, and Qiao Zhang.
\newblock {JAX}: composable transformations of {P}ython+{N}um{P}y programs,
  2018.

\bibitem{KrizhevskySuHi12}
Alex Krizhevsky, Ilya Sutskever, and Geoffrey Hinton.
\newblock {ImageNet} classification with deep convolutional neural networks.
\newblock In {\em Advances in Neural Information Processing Systems}, 2012.

\bibitem{hardt2017identity}
Moritz Hardt and Tengyu Ma.
\newblock Identity matters in deep learning.
\newblock In {\em International Conference on Learning Representations}, 2017.

\bibitem{bartlett2017spectrally}
Peter~L. Bartlett, Dylan~J. Foster, and Matus~J. Telgarsky.
\newblock Spectrally-normalized margin bounds for neural networks.
\newblock In {\em Advances in Neural Information Processing Systems}, pages
  6240--6249, 2017.

\bibitem{golowich2018size}
Noah Golowich, Alexander Rakhlin, and Ohad Shamir.
\newblock Size-independent sample complexity of neural networks.
\newblock In {\em Conference on Learning Theory}, pages 297--299, 2018.

\bibitem{duda1973pattern}
Richard~O. Duda, Peter~E. Hart, and David~G. Stork.
\newblock {\em Pattern classification and scene analysis}, volume~3.
\newblock Wiley New York, 1973.

\bibitem{hastie2017elements}
Trevor Hastie, Robert Tibshirani, and Jerome Friedman.
\newblock {\em The Elements of Statistical Learning: Data Mining, Inference,
  and Prediction (Corrected 12th printing)}.
\newblock Springer, 2017.

\bibitem{li2019vocal}
Xiaochang Li and Mara Mills.
\newblock Vocal features: From voice identification to speech recognition by
  machine.
\newblock {\em Technology and Culture}, 60(2):S129--S160, 2019.

\bibitem{garofolo1993darpa}
John~S. Garofolo, Lori~F. Lamel, William~M. Fisher, Jonathan~G. Fiscus, and
  David~S. Pallett.
\newblock {DARPA} {TIMIT} acoustic-phonetic continous speech corpus {CD-ROM}.
  {NIST} speech disc 1-1.1.
\newblock {\em STIN}, 93:27403, 1993.

\bibitem{koenecke2020racial}
Allison Koenecke, Andrew Nam, Emily Lake, Joe Nudell, Minnie Quartey, Zion
  Mengesha, Connor Toups, John~R Rickford, Dan Jurafsky, and Sharad Goel.
\newblock Racial disparities in automated speech recognition.
\newblock {\em Proceedings of the National Academy of Sciences},
  117(14):7684--7689, 2020.

\bibitem{aha2020personal}
David Aha.
\newblock personal communication, 2020.

\bibitem{ding2021retiring}
Frances Ding, Moritz Hardt, John Miller, and Ludwig Schmidt.
\newblock Retiring adult: New datasets for fair machine learning.
\newblock In {\em Advances in Neural Information Processing Systems}, 2021.

\bibitem{highleyman1959generalized}
Wilbur~H. Highleyman and Louis~A. Kamentsky.
\newblock A generalized scanner for pattern- and character-recognition studies.
\newblock In {\em Western Joint Computer Conference}, page 291–294, 1959.

\bibitem{highleyman1961character}
Wilbur~H. Highleyman.
\newblock Character recognition system, 1961.
\newblock US Patent 2,978,675.

\bibitem{highleyman1960comments}
Wilbur~H. Highleyman and Louis~A. Kamentsky.
\newblock Comments on a character recognition method of bledsoe and browning.
\newblock {\em IRE Transactions on Electronic Computers}, EC-9(2):263--263,
  1960.

\bibitem{bledsoe1961further}
Woodrow~Wilson Bledsoe.
\newblock Further results on the n-tuple pattern recognition method.
\newblock {\em IRE Transactions on Electronic Computers}, EC-10(1):96--96,
  1961.

\bibitem{chow1962recognition}
Chao~Kong Chow.
\newblock A recognition method using neighbor dependence.
\newblock {\em IRE Transactions on Electronic Computers}, EC-11(5):683--690,
  1962.

\bibitem{highleyman1963data}
Wilbur~H. Highleyman.
\newblock Data for character recognition studies.
\newblock {\em IEEE Transactions on Electronic Computers}, EC-12(2):135--136,
  1963.

\bibitem{highleyman1962design}
Wilbur~H. Highleyman.
\newblock The design and analysis of pattern recognition experiments.
\newblock {\em The Bell System Technical Journal}, 41(2):723--744, 1962.

\bibitem{munson1968experiments}
John~H. Munson, Richard~O. Duda, and Peter~E. Hart.
\newblock Experiments with highleyman's data.
\newblock {\em IEEE Transactions on Computers}, C-17(4):399--401, 1968.

\bibitem{lecun1998gradient}
Yann LeCun, L{\'e}on Bottou, Yoshua Bengio, and Patrick Haffner.
\newblock Gradient-based learning applied to document recognition.
\newblock {\em Proceedings of the IEEE}, 86(11):2278--2324, 1998.

\bibitem{mnisturl}
Yann LeCun.
\newblock \url{http://yann.lecun.com/exdb/mnist/}.
\newblock Accessed 10-31-2021.

\bibitem{grother1995nist}
Patrick~J. Grother.
\newblock {NIST} special database 19 handprinted forms and characters database.
\newblock {\em National Institute of Standards and Technology}, 1995.

\bibitem{yadav2019cold}
Chhavi Yadav and L{\'e}on Bottou.
\newblock Cold case: The lost {MNIST} digits.
\newblock In {\em Advances in Neural Information Processing Systems}, pages
  13443--13452, 2019.

\bibitem{bromley1991neural}
Jane Bromley and Eduard Sackinger.
\newblock Neural-network and k-nearest-neighbor classifiers.
\newblock {\em Rapport Technique}, pages 11359--910819, 1991.

\bibitem{deng2009imagenet}
Jia Deng, Wei Dong, Richard Socher, Li-Jia Li, Kai Li, and Li~Fei-Fei.
\newblock {ImageNet}: A large-scale hierarchical image database.
\newblock In {\em Computer Vision and Pattern Recognition}, pages 248--255,
  2009.

\bibitem{russakovsky2015imagenet}
Olga Russakovsky, Jia Deng, Hao Su, Jonathan Krause, Sanjeev Satheesh, Sean Ma,
  Zhiheng Huang, Andrej Karpathy, Aditya Khosla, Michael Bernstein,
  Alexander~C. Berg, and Fei-Fei Li.
\newblock {ImageNet} large scale visual recognition challenge.
\newblock {\em International Journal of Computer Vision}, 115(3):211--252,
  2015.

\bibitem{MalikCACM}
Jitendra Malik.
\newblock What led computer vision to deep learning?
\newblock {\em Communications of the {ACM}}, 60(6):82--83, 2017.

\bibitem{gray2019ghost}
Mary~L. Gray and Siddharth Suri.
\newblock {\em Ghost work: how to stop Silicon Valley from building a new
  global underclass}.
\newblock Eamon Dolan Books, 2019.

\bibitem{Mania-Model-Sim}
Horia Mania, John Miller, Ludwig Schmidt, Moritz Hardt, and Benjamin Recht.
\newblock Model similarity mitigates test set overuse.
\newblock In {\em Advances in Neural Information Processing Systems}, 2019.

\bibitem{mania-sra}
Horia Mania and Suvrit Sra.
\newblock Why do classifier accuracies show linear trends under distribution
  shift?
\newblock {\em arXiv:2012.15483}, 2020.

\bibitem{buolamwini2018gender}
Joy Buolamwini and Timnit Gebru.
\newblock Gender shades: Intersectional accuracy disparities in commercial
  gender classification.
\newblock In {\em Conference on Fairness, Accountability and Transparency},
  pages 77--91, 2018.

\bibitem{bolukbasi2016man}
Tolga Bolukbasi, Kai-Wei Chang, James~Y. Zou, Venkatesh Saligrama, and Adam~T.
  Kalai.
\newblock Man is to computer programmer as woman is to homemaker? debiasing
  word embeddings.
\newblock {\em Advances in Neural Information Processing Systems},
  29:4349--4357, 2016.

\bibitem{gonen2019lipstick}
Hila Gonen and Yoav Goldberg.
\newblock Lipstick on a pig: Debiasing methods cover up systematic gender
  biases in word embeddings but do not remove them.
\newblock {\em arXiv:1903.03862}, 2019.

\bibitem{narayanan2008robust}
Arvind Narayanan and Vitaly Shmatikov.
\newblock Robust de-anonymization of large sparse datasets.
\newblock In {\em Symposium on Security and Privacy}, pages 111--125. IEEE,
  2008.

\bibitem{dwork2017exposed}
Cynthia Dwork, Adam Smith, Thomas Steinke, and Jonathan Ullman.
\newblock Exposed! {A} survey of attacks on private data.
\newblock {\em Annual Review of Statistics and Its Application}, 4:61--84,
  2017.

\bibitem{dwork2014algorithmic}
Cynthia Dwork and Aaron Roth.
\newblock The algorithmic foundations of differential privacy.
\newblock {\em Foundations and Trends in Theoretical Computer Science},
  9(3-4):211--407, 2014.

\bibitem{levendowski2018copyright}
Amanda Levendowski.
\newblock How copyright law can fix artificial intelligence's implicit bias
  problem.
\newblock {\em Wash. L. Rev.}, 93:579, 2018.

\bibitem{dawes1989clinical}
Robyn~M. Dawes, David Faust, and Paul~E. Meehl.
\newblock Clinical versus actuarial judgment.
\newblock {\em Science}, 243(4899):1668--1674, 1989.

\bibitem{chaaban2007human}
Ibrahim Chaaban and Michael~R. Scheessele.
\newblock Human performance on the {USPS} database.
\newblock {\em Report, Indiana University South Bend}, 2007.

\bibitem{eckersley2017eff}
Peter Eckersley, Yomna Nasser, et~al.
\newblock {EFF} {AI} progress measurement project.
\newblock {\em \url{https://eff.org/ai/metrics}}, 2017.

\bibitem{he2015delving}
Kaiming He, Xiangyu Zhang, Shaoqing Ren, and Jian Sun.
\newblock Delving deep into rectifiers: Surpassing human-level performance on
  {ImageNet} classification.
\newblock In {\em International Conference on Computer Vision}, pages
  1026--1034, 2015.

\bibitem{shankar2020evaluating}
Vaishaal Shankar, Rebecca Roelofs, Horia Mania, Alex Fang, Benjamin Recht, and
  Ludwig Schmidt.
\newblock Evaluating machine accuracy on {ImageNet}.
\newblock In {\em International Conference on Machine Learning}, 2020.

\bibitem{recht2019imagenet}
Benjamin Recht, Rebecca Roelofs, Ludwig Schmidt, and Vaishaal Shankar.
\newblock Do {ImageNet} classifiers generalize to {ImageNet}?
\newblock In {\em International Conference on Machine Learning}, pages
  5389--5400, 2019.

\bibitem{gebru2018datasheets}
Timnit Gebru, Jamie Morgenstern, Briana Vecchione, Jennifer~Wortman Vaughan,
  Hanna Wallach, Hal Daum{\'e}~III, and Kate Crawford.
\newblock Datasheets for datasets.
\newblock {\em arXiv:1803.09010}, 2018.

\bibitem{jo2020lessons}
Eun~Seo Jo and Timnit Gebru.
\newblock Lessons from archives: Strategies for collecting sociocultural data
  in machine learning.
\newblock In {\em Conference on Fairness, Accountability, and Transparency},
  pages 306--316, 2020.

\bibitem{fed2007report}
Board of~Governors of~the Federal Reserve~System.
\newblock Report to the congress on credit scoring and its effects on the
  availability and affordability of credit.
\newblock
  \url{https://www.federalreserve.gov/boarddocs/rptcongress/creditscore/},
  2007.

\bibitem{reichman2001fragile}
Nancy~E. Reichman, Julien~O. Teitler, Irwin Garfinkel, and Sara~S. McLanahan.
\newblock Fragile families: Sample and design.
\newblock {\em Children and Youth Services Review}, 23(4-5):303--326, 2001.

\bibitem{dwork2015preserving}
Cynthia Dwork, Vitaly Feldman, Moritz Hardt, Toniann Pitassi, Omer Reingold,
  and Aaron Roth.
\newblock Preserving statistical validity in adaptive data analysis.
\newblock In {\em Symposium on the Theory of Computing}, pages 117--126, 2015.

\bibitem{dwork2015reusable}
Cynthia Dwork, Vitaly Feldman, Moritz Hardt, Toniann Pitassi, Omer Reingold,
  and Aaron Roth.
\newblock The reusable holdout: Preserving validity in adaptive data analysis.
\newblock {\em Science}, 349(6248):636--638, 2015.

\bibitem{freedman1983note}
David~A. Freedman.
\newblock A note on screening regression equations.
\newblock {\em The American Statistician}, 37(2):152--155, 1983.

\bibitem{blum2015ladder}
Avrim Blum and Moritz Hardt.
\newblock The {L}adder: A reliable leaderboard for machine learning
  competitions.
\newblock In {\em International Conference on Machine Learning}, pages
  1006--1014, 2015.

\bibitem{boyd2012critical}
Danah Boyd and Kate Crawford.
\newblock Critical questions for big data: Provocations for a cultural,
  technological, and scholarly phenomenon.
\newblock {\em Information, Communication \& Society}, 15(5):662--679, 2012.

\bibitem{tufekci2014big}
Zeynep Tufekci.
\newblock Big questions for social media big data: Representativeness, validity
  and other methodological pitfalls.
\newblock In {\em {AAAI} Conference on Weblogs and Social Media}, 2014.

\bibitem{tufekci2014engineering}
Zeynep Tufekci.
\newblock Engineering the public: Big data, surveillance and computational
  politics.
\newblock {\em First Monday}, 2014.

\bibitem{onuoha2016point}
Mimi Onuoha.
\newblock The point of collection.
\newblock {\em Data \& Society: Points}, 2016.

\bibitem{paullada2020data}
Amandalynne Paullada, Inioluwa~Deborah Raji, Emily~M. Bender, Emily Denton, and
  Alex Hanna.
\newblock Data and its (dis)contents: A survey of dataset development and use
  in machine learning research.
\newblock {\em arXiv:2012.05345}, 2020.

\bibitem{olteanu2019social}
Alexandra Olteanu, Carlos Castillo, Fernando Diaz, and Emre Kiciman.
\newblock Social data: Biases, methodological pitfalls, and ethical boundaries.
\newblock {\em Frontiers in Big Data}, 2:13, 2019.

\bibitem{couldry2019data}
Nick Couldry and Ulises~A. Mejias.
\newblock Data colonialism: Rethinking big data's relation to the contemporary
  subject.
\newblock {\em Television \& New Media}, 20(4):336--349, 2019.

\bibitem{bickel1975sex}
Peter~J. Bickel, Eugene~A. Hammel, and J.~William O{'C}onnell.
\newblock Sex bias in graduate admissions: Data from {B}erkeley.
\newblock {\em Science}, 187(4175):398--404, 1975.

\bibitem{humphrey2002postmenopausal}
Linda~L. Humphrey, Benjamin~K.S. Chan, and Harold~C. Sox.
\newblock {Postmenopausal Hormone Replacement Therapy and the Primary
  Prevention of Cardiovascular Disease}.
\newblock {\em Annals of Internal Medicine}, 137(4):273--284, 08 2002.

\bibitem{berkson2014limitations}
Joseph Berkson.
\newblock Limitations of the application of fourfold table analysis to hospital
  data.
\newblock {\em International Journal of Epidemiology}, 43(2):511--515, 2014.
\newblock Reprint.

\bibitem{pearl2009causality}
Judea Pearl.
\newblock {\em Causality}.
\newblock Cambridge University Press, 2009.

\bibitem{peters2017elements}
Jonas Peters, Dominik Janzing, and Bernhard Sch\"olkopf.
\newblock {\em Elements of Causal Inference}.
\newblock MIT Press, 2017.

\bibitem{pearl2016causal}
Judea Pearl, Madelyn Glymour, and Nicholas~P. Jewell.
\newblock {\em Causal Inference in Statistics: A Primer}.
\newblock Wiley, 2016.

\bibitem{pearl2018book}
Judea Pearl and Dana Mackenzie.
\newblock {\em The Book of Why: The New Science of Cause and Effect}.
\newblock Basic Books, 2018.

\bibitem{simpson1951interpretation}
Edward~H. Simpson.
\newblock The interpretation of interaction in contingency tables.
\newblock {\em Journal of the Royal Statistical Society: Series B
  (Methodological)}, 13(2):238--241, 1951.

\bibitem{hernan2011simpson}
Miguel~A. Hern\'{a}n, David Clayton, and Niels Keiding.
\newblock {The Simpson's paradox unraveled}.
\newblock {\em International Journal of Epidemiology}, 40(3):780--785, 03 2011.

\bibitem{spirtes2000causation}
Peter Spirtes, Clark~N. Glymour, Richard Scheines, David Heckerman, Christopher
  Meek, Gregory Cooper, and Thomas Richardson.
\newblock {\em Causation, Prediction, and Search}.
\newblock MIT Press, 2000.

\bibitem{scholkopf2019causality}
Bernhard Sch{\"o}lkopf.
\newblock Causality for machine learning.
\newblock {\em arXiv:1911.10500}, 2019.

\bibitem{neyman1923applications}
Jerzy Neyman.
\newblock Sur les applications de la th{\'e}orie des probabilit{\'e}s aux
  experiences agricoles: Essai des principes.
\newblock {\em Roczniki Nauk Rolniczych}, 10:1--51, 1923.

\bibitem{rubin2005causal}
Donald~B. Rubin.
\newblock Causal inference using potential outcomes: Design, modeling,
  decisions.
\newblock {\em Journal of the American Statistical Association},
  100(469):322--331, 2005.

\bibitem{imbens2015causal}
Guido~W. Imbens and Donald~B. Rubin.
\newblock {\em Causal Inference for Statistics, Social, and Biomedical
  Sciences}.
\newblock Cambridge University Press, 2015.

\bibitem{angrist2008mostly}
Joshua~D. Angrist and J{\"o}rn-Steffen Pischke.
\newblock {\em Mostly harmless econometrics: An empiricist's companion}.
\newblock Princeton University Press, 2008.

\bibitem{hernan2020causal}
Miguel~A. Hern\'{a}n and James Robins.
\newblock {\em Causal Inference: What If}.
\newblock Boca Raton: Chapman \& Hall/CRC, 2020.

\bibitem{morgan2014counterfactuals}
Stephen~L. Morgan and Christopher Winship.
\newblock {\em Counterfactuals and Causal Inference}.
\newblock Cambridge University Press, 2014.

\bibitem{deaton2018understanding}
Angus Deaton and Nancy Cartwright.
\newblock Understanding and misunderstanding randomized controlled trials.
\newblock {\em Social Science \& Medicine}, 210:2--21, 2018.

\bibitem{wager-causal-forests}
Stefan Wager and Susan Athey.
\newblock Estimation and inference of heterogeneous treatment effects using
  random forests.
\newblock {\em Journal of the American Statistical Association},
  113(523):1228--1242, 2018.

\bibitem{almond2010estimating}
Douglas Almond, Joseph~J. Doyle~Jr, Amanda~E. Kowalski, and Heidi Williams.
\newblock Estimating marginal returns to medical care: Evidence from at-risk
  newborns.
\newblock {\em The quarterly journal of economics}, 125(2):591--634, 2010.

\bibitem{bharadwaj2013early}
Prashant Bharadwaj, Katrine~Vellesen L{\o}ken, and Christopher Neilson.
\newblock Early life health interventions and academic achievement.
\newblock {\em American Economic Review}, 103(5):1862--91, 2013.

\bibitem{camacho2011manipulation}
Adriana Camacho and Emily Conover.
\newblock Manipulation of social program eligibility.
\newblock {\em American Economic Journal: Economic Policy}, 3(2):41--65, 2011.

\bibitem{urquiola2009class}
Miguel Urquiola and Eric Verhoogen.
\newblock Class-size caps, sorting, and the regression-discontinuity design.
\newblock {\em American Economic Review}, 99(1):179--215, 2009.

\bibitem{chilton2015challenging}
Adam~S. Chilton and Marin~K. Levy.
\newblock Challenging the randomness of panel assignment in the federal courts
  of appeals.
\newblock {\em Cornell L. Rev.}, 101:1, 2015.

\bibitem{benson2000comparison}
Kjell Benson and Arthur~J. Hartz.
\newblock A comparison of observational studies and randomized, controlled
  trials.
\newblock {\em New England Journal of Medicine}, 342(25):1878--1886, 2000.

\bibitem{concato2000randomized}
John Concato, Nirav Shah, and Ralph~I. Horwitz.
\newblock Randomized, controlled trials, observational studies, and the
  hierarchy of research designs.
\newblock {\em New England journal of medicine}, 342(25):1887--1892, 2000.

\bibitem{freedman1991statistical}
David~A. Freedman.
\newblock Statistical models and shoe leather.
\newblock {\em Sociological Methodology}, pages 291--313, 1991.

\bibitem{hwang2020subprime}
Tim Hwang.
\newblock {\em Subprime Attention Crisis}.
\newblock Farrar, Strauss and Giroux, 2020.

\bibitem{gordon2019comparison}
Brett~R. Gordon, Florian Zettelmeyer, Neha Bhargava, and Dan Chapsky.
\newblock A comparison of approaches to advertising measurement: Evidence from
  big field experiments at facebook.
\newblock {\em Marketing Science}, 38(2):193--225, 2019.

\bibitem{eckles2016design}
Dean Eckles, Brian Karrer, and Johan Ugander.
\newblock Design and analysis of experiments in networks: Reducing bias from
  interference.
\newblock {\em Journal of Causal Inference}, 5(1), 2016.

\bibitem{roemer19how}
John~E. Roemer.
\newblock {\em How We Cooperate: A Theory of Kantian Optimization}.
\newblock Yale University Press, 2019.

\bibitem{athey2017state}
Susan Athey and Guido~W. Imbens.
\newblock The state of applied econometrics: Causality and policy evaluation.
\newblock {\em Journal of Economic Perspectives}, 31(2):3--32, 2017.

\bibitem{marinescu2018quasi}
Ioana~E. Marinescu, Patrick~N. Lawlor, and Konrad~P. Kording.
\newblock Quasi-experimental causality in neuroscience and behavioural
  research.
\newblock {\em Nature Human Behaviour}, 2(12):891--898, 2018.

\bibitem{BertsekasDPBook}
Dimitri~P. Bertsekas.
\newblock {\em Dynamic Programming and Optimal Control}, volume~1.
\newblock Athena Scientific, 4th edition, 2017.

\bibitem{blondel2000survey}
Vincent~D. Blondel and John~N. Tsitsiklis.
\newblock A survey of computational complexity results in systems and control.
\newblock {\em Automatica}, 36(9):1249--1274, 2000.

\bibitem{papadimitriou1987complexity}
Christos~H. Papadimitriou and John~N. Tsitsiklis.
\newblock The complexity of {M}arkov {D}ecision {P}rocesses.
\newblock {\em Mathematics of Operations Research}, 12(3):441--450, 1987.

\bibitem{RechtRLSurvey}
Benjamin Recht.
\newblock A tour of reinforcement learning: The view from continuous control.
\newblock {\em Annual Review of Control, Robotics, and Autonomous Systems}, 2,
  2019.

\bibitem{BorrelliMPCBook}
Francesco Borrelli, Alberto Bemporad, and Manfred Morari.
\newblock {\em Predictive Control for Linear and Hybrid Systems}.
\newblock Cambridge University Press, 2017.

\bibitem{BoydOCNotes}
Stephen Boyd.
\newblock {EE363}: Linear dynamical systems.
\newblock Notes available at \url{https://stanford.edu/class/ee363/}, 2009.

\bibitem{PutermanBook}
Martin~L. Puterman.
\newblock {\em Markov Decision Processes: Discrete Stochastic Dynamic
  Programming}.
\newblock Wiley-Interscience, 1994.

\bibitem{Simon56}
Herbert~A. Simon.
\newblock Dynamic programming under uncertainty with a quadratic criterion
  function.
\newblock {\em Econometrica}, 24(1):74--81, 1956.

\bibitem{Theil57}
Henri Theil.
\newblock A note on certainty equivalence in dynamic planning.
\newblock {\em Econometrica}, 25(2):346--349, 1957.

\bibitem{AuerOrtner10}
Peter Auer and Ronald Ortner.
\newblock {UCB} revisited: Improved regret bounds for the stochastic
  multi-armed bandit problem.
\newblock {\em Periodica Mathematica Hungarica}, 61(1-2):55--65, 2010.

\bibitem{kannan2018smoothed}
Sampath Kannan, Jamie~H. Morgenstern, Aaron Roth, Bo~Waggoner, and
  Zhiwei~Steven Wu.
\newblock A smoothed analysis of the greedy algorithm for the linear contextual
  bandit problem.
\newblock In {\em Advances in Neural Information Processing Systems}, 2018.

\bibitem{hummel2016machine}
Patrick Hummel and R.~Preston McAfee.
\newblock Machine learning in an auction environment.
\newblock {\em Journal of Machine Learning Research}, 17(1):6915--6951, 2016.

\bibitem{bietti2018contextual}
Alberto Bietti, Alekh Agarwal, and John Langford.
\newblock A contextual bandit bake-off.
\newblock {\em arXiv:1802.04064}, 2018.

\bibitem{watkins1992q}
Christopher J. C.~H. Watkins and Peter Dayan.
\newblock Q-learning.
\newblock {\em Machine Learning}, 8(3-4):279--292, 1992.

\bibitem{SARSA}
Gavin~Adrian Rummery and Mahesan Niranjan.
\newblock Online {Q}-learning using connectionist systems.
\newblock Technical report, CUED/F-INFENG/TR 166, Cambridge University
  Engineering Dept., 1994.

\bibitem{williams1992simple}
Ronald~J. Williams.
\newblock Simple statistical gradient-following algorithms for connectionist
  reinforcement learning.
\newblock {\em Machine Learning}, 8(3-4):229--256, 1992.

\bibitem{Rastrigin63}
Leonard~A. Rastrigin.
\newblock About convergence of random search method in extremal control of
  multi-parameter systems.
\newblock {\em Avtomat. i Telemekh.}, 24(11):1467---1473, 1963.

\bibitem{nesterov2017random}
Yurii Nesterov and Vladimir Spokoiny.
\newblock Random gradient-free minimization of convex functions.
\newblock {\em Foundations of Computational Mathematics}, 17(2):527--566, 2017.

\bibitem{Beyer02}
Hans-Georg Beyer and Hans-Paul Schwefel.
\newblock {Evolution Strategies}---a comprehensive introduction.
\newblock {\em Natural Computing}, 1(1):3--52, 2002.

\bibitem{SchwefelThesis}
Hans-Paul Schwefel.
\newblock {\em Evolutionsstrategie und numerische Optimierung}.
\newblock PhD thesis, TU Berlin, 1975.

\bibitem{spall1992multivariate}
James~C. Spall.
\newblock Multivariate stochastic approximation using a simultaneous
  perturbation gradient approximation.
\newblock {\em Transactions on Automatic Control}, 37(3):332--341, 1992.

\bibitem{flaxman2005online}
Abraham~D. Flaxman, Adam~T. Kalai, and H.~Brendan McMahan.
\newblock Online convex optimization in the bandit setting: gradient descent
  without a gradient.
\newblock In {\em Symposium on Discrete Algorithms}, pages 385--394, 2005.

\bibitem{agarwal2010optimal}
Alekh Agarwal, Ofer Dekel, and Lin Xiao.
\newblock Optimal algorithms for online convex optimization with multi-point
  bandit feedback.
\newblock In {\em Conference on Learning Theory}, 2010.

\bibitem{bertsekas1996neuro}
Dimitri~P. Bertsekas and John~N. Tsitsiklis.
\newblock {\em Neuro-Dynamic Programming}.
\newblock Athena Scientific, 1996.

\bibitem{Mania19}
Horia Mania, Stephen Tu, and Benjamin Recht.
\newblock Certainty equivalence is efficient for linear quadratic control.
\newblock In {\em Advances in Neural Information Processing Systems}, 2019.

\bibitem{Simchowitz20}
Max Simchowitz and Dylan Foster.
\newblock Naive exploration is optimal for online {LQR}.
\newblock In {\em International Conference on Machine Learning}, 2020.

\bibitem{RLTheoryBook}
Alekh Agarwal, Nan Jiang, Sham~M. Kakade, and Wen Sun.
\newblock Reinforcement learning: Theory and algorithms.
\newblock Preprint Available at \url{rltheorybook.github.io}, 2020.

\bibitem{AvilaPires16}
Bernardo \'{A}vila Pires and Csaba Szepesv{\'a}ri.
\newblock Policy error bounds for model-based reinforcement learning with
  factored linear models.
\newblock In {\em Conference on Learning Theory}, 2016.

\bibitem{Whitt78}
Ward Whitt.
\newblock Approximations of dynamic programs, {I}.
\newblock {\em Mathematics of Operations Research}, 3(3):231--243, 1978.

\bibitem{Singh94}
Satinder~P. Singh and Richard~C. Yee.
\newblock An upper bound on the loss from approximate optimal-value functions.
\newblock {\em Machine Learning}, 16(3):227--233, 1994.

\bibitem{Bertsekas2012}
Dimitri~P. Bertsekas.
\newblock Weighted sup-norm contractions in dynamic programming: A review and
  some new applications.
\newblock LIDS Tech Report LIDS-P-2884, Department of Electrical Engineering
  and Computer Science, Massachusetts Institute Technology, 2012.

\bibitem{BertsekasAbstractDPBook}
Dimitri~P. Bertsekas.
\newblock {\em Abstract Dynamic Programming}.
\newblock Athena Scientific, 2nd edition, 2018.

\bibitem{Agarwal2020c}
Alekh Agarwal, Sham~M. Kakade, and Lin~F. Yang.
\newblock Model-based reinforcement learning with a generative model is minimax
  optimal.
\newblock In {\em Conference on Learning Theory}, 2020.

\bibitem{BertsekasDPBook2}
Dimitri~P. Bertsekas.
\newblock {\em Dynamic Programming and Optimal Control}, volume~2.
\newblock Athena Scientific, 4th edition, 2012.

\bibitem{BertsekasRLBook}
Dimitri~P. Bertsekas.
\newblock {\em Reinforcement Learning and Optimal Control}.
\newblock Athena Scientific, 2019.

\bibitem{LattimoreBanditBook}
Tor Lattimore and Csaba Szepesv{\'a}ri.
\newblock {\em Bandit Algorithms}.
\newblock Cambridge University Press, 2020.

\bibitem{feldbaum1960dual}
Aleksandr~Aronovich Feldbaum.
\newblock Dual control theory.
\newblock {\em Avtomatika i Telemekhanika}, 21(9):1240--1249, 1960.

\bibitem{wittenmark1995adaptive}
Björn Wittenmark.
\newblock Adaptive dual control methods: An overview.
\newblock In {\em Adaptive Systems in Control and Signal Processing}, pages
  67--72. Elsevier, 1995.

\bibitem{stein2003respect}
Gunter Stein.
\newblock Respect the unstable.
\newblock {\em IEEE Control Systems Magazine}, 23(4):12--25, 2003.

\bibitem{franklin1999real}
Ursula Franklin.
\newblock {\em The real world of technology}.
\newblock House of Anansi, 1999.

\bibitem{kendrick1976applications}
David Kendrick.
\newblock Applications of control theory to macroeconomics.
\newblock In {\em Annals of Economic and Social Measurement, Volume 5, number
  2}, pages 171--190. NBER, 1976.

\bibitem{schumacher2011small}
Ernst~Friedrich Schumacher.
\newblock {\em Small is beautiful: A study of economics as if people mattered}.
\newblock Random House, 2011.

\end{thebibliography}

\listoffigures
\listoftables
\printindex

%
%

\end{document}